\documentclass[runningheads]{llncs}

\usepackage[mobile]{eccv}

\usepackage{eccvabbrv}

\usepackage{graphicx}
\usepackage{booktabs}
\usepackage{url}
\usepackage{svg}

\usepackage[accsupp]{axessibility}  

\usepackage{hyperref}

\usepackage{orcidlink}

\usepackage{epsfig}
\usepackage{graphicx}
\usepackage{comment}
\usepackage{amsmath,amssymb} %
\usepackage{hhline}
\usepackage[export]{adjustbox}
\usepackage{bbm}

\usepackage[font=small,skip=5pt]{caption}
\usepackage{float}

\usepackage{multirow}
\usepackage{tabularx}
\usepackage{soul}
\usepackage{array}
\usepackage{comment}
\usepackage{outlines}
\usepackage{pgfplots}
\pgfplotsset{compat=1.18}

\usepackage{overpic}
\usepackage{wrapfig}
\usepackage{pifont}
\usepackage{tikz}

\newcommand{\method}[0]{NICP}
\newcommand{\pipeline}[0]{NSR}
\newcommand{\pipedef}[0]{Neural Scalable Registration}

\newcommand{\mypar}[1]{\noindent{\textit{\textbf{#1}}}}

\begin{document}

\title{NICP: Neural ICP \\ for 3D Human Registration at Scale} 

\author{Riccardo Marin\inst{1,2}\orcidlink{0000-0003-2392-4612} \and
Enric Corona\inst{3}\orcidlink{0000-0002-4835-1868} \and
Gerard Pons-Moll\inst{1,2,4}\orcidlink{0000-0001-5115-7794}}

\authorrunning{R.~Marin et al.}

\institute{University of Tübingen, Germany \and
Tübingen AI Center, Germany \and
Google Research \and
Max Planck Institute for Informatics, Saarland Informatics Campus, Germany 
}

\maketitle
\noindent\begin{center}
    {\url{https://neural-icp.github.io/}}
\end{center}
\begin{figure}

    \begin{center}
    \captionsetup{type=figure}
    \newcommand{\teaserwidth}{\textwidth}
    \includegraphics[width=\textwidth]{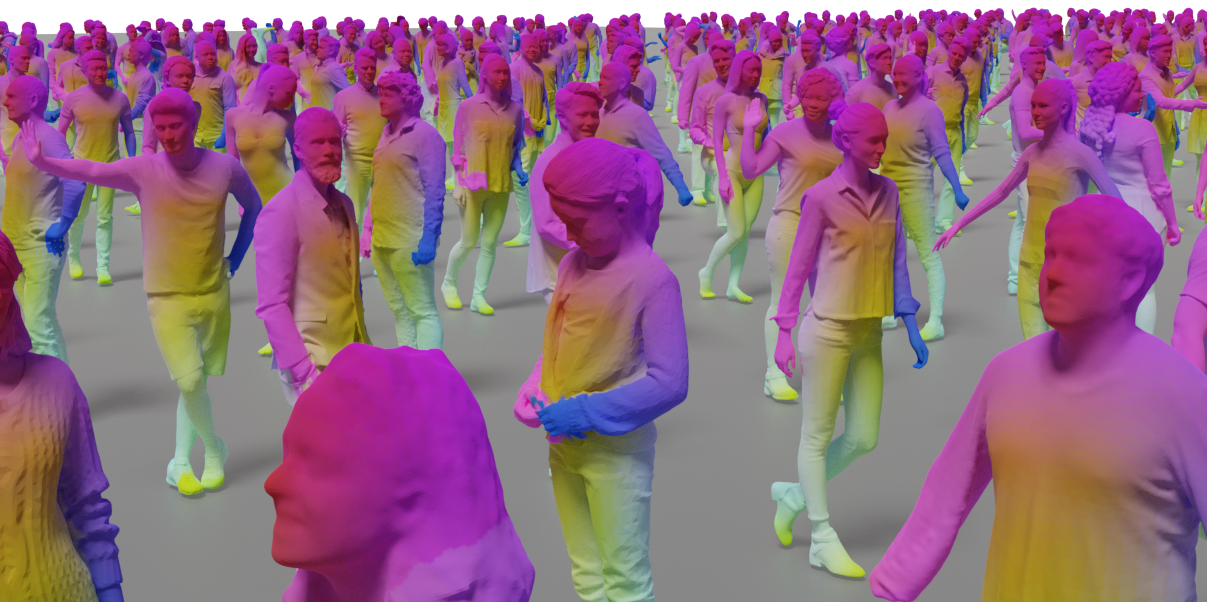}
    \caption{\label{fig:teaser}
    Our method (\pipedef{}, \textbf{\pipeline{}}) provides reliable 3D human point cloud registrations in disparate conditions. The key enabler of such generalization is \textbf{\method{}}, a Neural ICP that improves network predictions at inference time in a self-supervised way. We test \pipeline~on more than 5K shapes from more than 10 data sources, including real scans, partial and cluttered point clouds, clothed humans, loose garments, identities out of distribution, and challenging poses. Colors represent semantic correspondence induced by our registrations. Meshes are depicted for visualization purposes and are not used by our method.}
    \end{center}
\end{figure}
\begin{abstract}
Aligning a template to 3D human point clouds is a long-standing problem crucial for tasks like animation, reconstruction, and enabling supervised learning pipelines. Recent data-driven methods leverage predicted surface correspondences. However, they are not robust to varied poses, identities, or noise. In contrast, industrial solutions often rely on expensive manual annotations or multi-view capturing systems. Recently, neural fields have shown promising results. Still, their purely data-driven and extrinsic nature does not incorporate any guidance toward the target surface, often resulting in a trivial misalignment of the template registration. Currently, no method can be considered the standard for 3D Human registration, limiting the scalability of downstream applications. In this work, we propose a neural scalable registration method, ~\pipeline{}, a pipeline that, for the first time, generalizes and scales across thousands of shapes and more than ten different data sources. Our essential contribution is~\method{}, an ICP-style self-supervised task tailored to neural fields. \method{} takes a few seconds, is self-supervised, and works out of the box on pre-trained neural fields. \pipeline{} combines \method{} with a localized neural field trained on a large MoCap dataset, achieving the state of the art over public benchmarks. The release of our code and checkpoints provides a powerful tool useful for many downstream tasks like dataset alignments, cleaning, or asset animation.

\end{abstract}

\section{Introduction}
\label{sec:intro}
Registration of 3D surfaces is a crucial area of research in Computer Vision, playing a vital role in analyzing shape collections, pattern discovery, and statistical model training. Among all the 3D surfaces, human models are of particular importance. It is worth noting that human registration has enabled the development of today's standard human parametric model (SMPL from \cite{loper2015smpl}) and also facilitates several downstream tasks such as animation, virtual try-on, and is the driving force behind virtual and mixed reality.

However, human registration presents several challenges. Articulations can take various configurations, leading to blind spots and geometry gluing (\eg, due to self-contact), even when hundreds of cameras are involved. Fine-grained details, crucial for capturing human identity and diversity, must be reconstructed accurately. Additionally, the acquisition process is often noisy, particularly in low-resource settings (such as Kinect depth scans) or consumer-level scenarios (such as applying NeRF-based systems like \cite{luma} to smartphone RGB videos).

Scholars approach shape registration intrinsically or extrinsically. Intrinsic approaches like \cite{ovsjanikov2012functional, sun2009concise, melzi2019zoomout, marin2020farm} are favored by Computer Graphics; they obtain pair-wise correspondences between shapes, restricting solutions to the object's surfaces and guaranteeing invariance to rigid deformations. These methods are often impractical for data outside the synthetic case, as clutter and noise ruin their theoretical premises. Extrinsic registration pipelines as \cite{bogo2014faust, ZuffiCVPR2017, flame, romero2022embodied} are popular in Computer Vision research, relying on templates as regularizers. These pipelines require expert annotators, multi-camera views, and priors for specific settings, which makes them inapplicable in many use cases. Also, the final template deformation may fall apart from the target surface even with robust learned priors. A common practice to refine template alignment is to iterate between Euclidean correspondence and registration until convergence, as done in the popular Iterative Closest Point (ICP) of \cite{icp} and its variants (\eg, \cite{li2008global}). However, the pairing step is sensitive to initialization and noise, leading often to local minima. A recent promising direction for iterative alignments that removes the need for Euclidean pairings has come from Neural Fields (\eg, Learned Vertex Descent, LVD \cite{corona2022learned}). The deformation field is trained to aim from any query point in $\mathbb{R}^3$ toward template registration vertices. However, such a data-driven convergence also limits the applicability to data close to training data distribution. 

Currently, no approach shows enough flexibility to be considered a standard or reliable enough for large-scale use. Our work aims to address this gap, proposing a ready-to-go method. Starting from the recent NF approaches, our inspiration is to enrich the learned prior by revisiting ICP principles. We propose Neural ICP (\textbf{\method{}}), the first self-supervised task that, at inference time, iteratively promotes the learned NF to converge toward the target surface. Concretely, \method{} queries the NF directly on \emph{the vertices of the target shape} at inference time. For each point on the target, we pair it with the template vertex  \emph{corresponding to the predicted offset with the minimum norm}, \ie, the closest template point indicated by the NF. Secondly, we sum the retrieved smallest offsets for all the target points, and we use this as a loss to \emph{fine-tune the NF by backpropagation}. Similar to ICP, we iterate between recovering the correspondence suggested by NF deformation and updating the NF deformation to match such correspondence. \method{} is the first refinement specifically designed for NF, takes a few seconds, and improves the backbone registration up to 30\%. Furthermore, \method{} enables an unprecedented generalization out of distribution, including real scans, identities, poses, garments, and even noise from Kinect fusion and partial scans.

Subsequently, we apply \method{} on top of a localized variant of LVD (Localized Vertex Descent \emph{LoVD)} trained on a large MoCap dataset from \cite{mahmood2019amass} (AMASS). Using a \emph{single network}, our Neural Scalable Registration pipeline (\textbf{\pipeline{}}), succeeds on an unprecedented variety of poses, identities, and real-world challenges far from training distribution (\eg, garments, noise point clouds, clutter, partiality).  We experimented \pipeline{} on more than $5$k shapes and $10$ different data sources (Figure~\ref{fig:teaser} shows a subset of these; colors encode the correspondence provided by our registration). We improve state-of-the-art benchmarks for human body registration even surpassing approaches with stronger priors. Our implementation and checkpoint provide a tool that, out of the box, offers robust human registration in \emph{less than a minute}. In summary, our contributions are:
\begin{enumerate}
    \item \emph{\method{}}: The first self-supervised fine-tuning procedure to improve geometric understanding of an NF at inference time, which largely improves registration performance and opens to unprecedented generalization
    \item \emph{\pipeline{} Pipeline}: We combine \method{} with LoVD, a novel localized LVD variant, into a full 3D human registration pipeline trained on a large MoCap dataset, advancing the state of the art on public benchmarks, and capable of addressing large scale real challenges.
    \item \emph{Code and weights release}: We release the code for processing, training, and evaluation, together with network weights; we provide researchers with a way-to-go tool for 3D Human registration that works in many different contexts out of the box.
\end{enumerate}

\section{Related Works}
\label{sec:related}
Shape correspondence has a vast literature, and while we list the works that mainly inspired us, we point to \cite{deng2022survey} for a recent and extensive survey.

\mypar{Shape Matching.} A straightforward idea to solve the correspondence problem is to compute a space for the vertices of the shapes where they are naturally aligned (canonical embedding). Classical approaches in this category are the multi-dimensional scaling from \cite{bronstein2006generalized}, and descriptors proposed by \cite{sun2009concise} (HKS), \cite{aubry2011wave} (WKS), and \cite{salti2014shot} (SHOT). Their dependence on input geometry does not generalize well in the presence of noise. Recently, \cite{marin2020correspondence} proposes a deep learning variation as a baseline. Still, such a unified representation is challenging to learn for non-rigid shapes. Works like \cite{openpose, cao2017realtime, kim2021deep} propose canonicalizing human shapes by inferring a common skeleton from external views, but they suffer from occlusions. Recent works like \cite{marin2020correspondence, huang2022multiway, jiang2023neural} learn a representation where a linear transformation can align shapes. These methods rely on pseudo-inverse computation at training time, which hardly scales and produces numerical instability. Their inspiration can be found in the Functional Maps framework of \cite{ovsjanikov2012functional}, an elegant and theoretically grounded formulation which opened to a set of regularization and variations such as \cite{nogneng2017informative, ren2018continuous, marin2023smoothness, maggioli2024rematching}, and refinements like \cite{huang2020consistent, melzi2019zoomout}. Mainly based on the eigenfunctions of the Laplace-Beltrami Operator, its applications are limited to clean meshes and unrealistic partialities as in \cite{melzi2020intrinsic, rodola2017partial, cosmo2016matching}. 

\mypar{Shape Registrations.} A widely explored approach to recover the correspondence is solving the alignment problem by registering the pair of shapes in $\mathcal{R}^3$. A fundamental algorithm for the rigid case is ICP from \cite{icp}, which aligns two shapes by iterating between the deformation and the correspondence. The simplicity of ICP made it broadly used, but its sensitivity to noise and initialization may limit the convergence toward a local minimum. Countless variations have been proposed; for example, \cite{bouaziz2013sparse} and \cite{chetverikov2002trimmed} address the sparsity and noise, respectively, but they still require careful design choices. Interestingly, \cite{wang2019deep}, \cite{Lu_2019_ICCV}, and \cite{yu2023rotation} propose a deep learning approach to provide a more reliable correspondence. Still, they are limited to rigid deformations. Instead, iterative algorithms to recover non-rigid deformation like \cite{li2008global} often rely on optimizations that trade some rigidity for a more expressive deformation. The Coherent Point Drift proposed in \cite{myronenko2010point} and the subsequent \cite{hirose2020bayesian,hirose2022geodesic} exploits a probabilistic formulation. Deep learning follow-ups from \cite{Li_2022_CVPR} and \cite{wang2019non} tried to learn the deformation from data. Their learning of point-wise features assumes limited non-rigid deformations and the absence of clutter.
Differently from predicting space positions, \cite{sundararaman2022reduced} train a network to predict offsets, using this rich output representation to formulate regularizations. However, these latter are expensive and prohibitive to scale. Finally, domain-specific templates like \cite{Zuffi:CVPR:2017,hesse2018learning} are often available and provide strong regularization, as demonstrated by ~\cite{groueix20183d,bhatnagar2020loopreg,trappolini2021shape,sundararaman2022reduced}), but don't prevent the registration to fall far from the target surface. 

\mypar{3D Human registration.} The classic work of \cite{bogo2014faust} uses expert feedback and controlled setups to lead 3D registration. More flexible, the stitched puppet of \cite{zuffi2015stitched} automatically solves piece-wise optimization of the local body parts, but gluing them leads to major artifacts. \cite{marin2020farm} introduced FARM, an automatic method with 3D landmark detection, which \cite{marin2019high} extended for high resolution. Both rely on the Functional Map of \cite{ovsjanikov2012functional} and struggle with non-watertight meshes.
With the advent of learning, the seminal work of \cite{groueix20183d} proposed a simple yet effective autoencoder architecture. Its global nature motivated follow-up works of \cite{deprelle2019learning} and \cite{trappolini2021shape} to learn local and attention-based relations, but both suffer from clutter. \cite{bhatnagar2020loopreg} proposes Loopreg, a self-supervised schema with a supervised warm-up. Loopreg relies on a diffused SMPL model, which may be inaccurate and discontinuous for points far from the surface; \method{} operates directly on the input surface. \cite{kim2021deep} propose Deep virtual markers (DVM) using synthetic multi-view depth representation. DVM requires a demanding manually annotated training set and suffers from occlusions and self-contact. Many works like~\cite{deng2020nasa, bhatnagar2020combining, mihajlovic2021leap, alldieck2021imghum, mihajlovic2022coap, wang2021locally} incorporate the local rigidity of humans' limbs as an inductive bias which can be used, as in \cite{ArtEq2023}, to learn rotation-invariant features of local parts. The local prior has been proven effective in dealing with complex poses, but it is insufficient to deal with the disruption of local geometry, such as noise, self-contact, and heavy clothes.
\begin{wraptable}[14]{r}{0.45\linewidth}
\scriptsize
  \begin{overpic}[trim=0cm 0cm 0cm 0.0cm,clip, width=\linewidth]{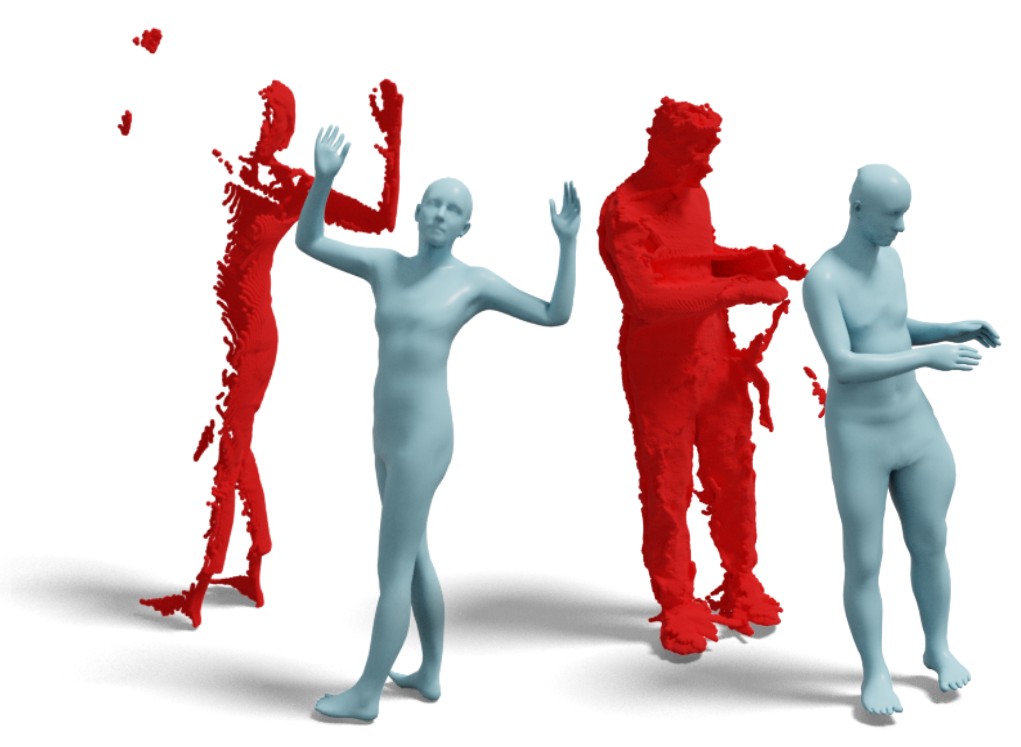}
        \put(19,68){Input}
        \put(43,58){{\textbf{Ours}}}
        \put(63,66){Input}
        \put(83,58){{\textbf{Ours}}}
	\end{overpic}
\captionof{figure}{\small{\label{fig:challenges}Input and results of our method on challenging shapes from \cite{bhatnagar2022behave} and \cite{ma2020learning}.}}
 \vspace{-0.2cm} 
\end{wraptable}
The first use of NF for 3D Human Registration appears in \cite{corona2022learned}, proposing Learned Vertex Descent (LVD), which instead regularizes the training using a richer representation for the template deformation. Their idea is to train a network that, given an input shape, produces an NF defined all over $\mathbb{R}^3$. When querying a 3D point, the NF predicts the ordered offsets towards all the registered template vertices. Supervising the training on every 3D location toward all the template vertices produces a robust prior that benefits from large-scale datasets. However, LVD relies solely on the training distribution, and errors produced at inference time are unrecoverable even after refining the predicted SMPL with Chamfer loss.

\section{Background}
\label{sec:background}
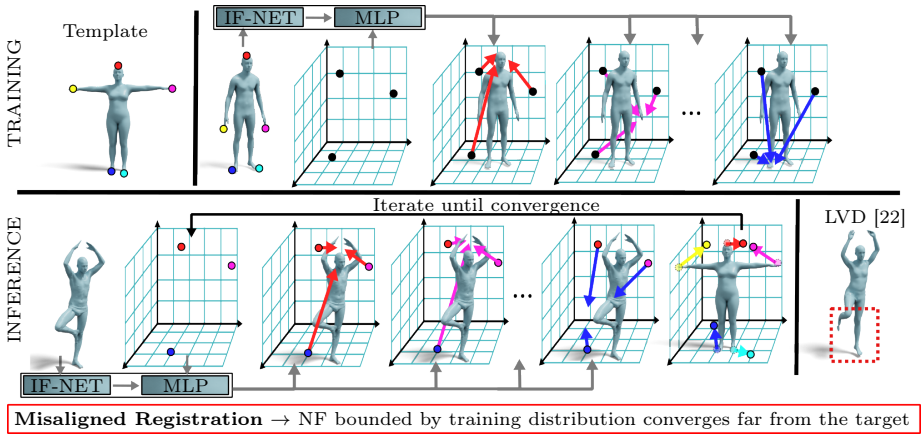
\begin{figure*}[t!]
\scriptsize
 \begin{overpic}[trim=0cm 0cm 0cm 0cm,clip,width=\linewidth]{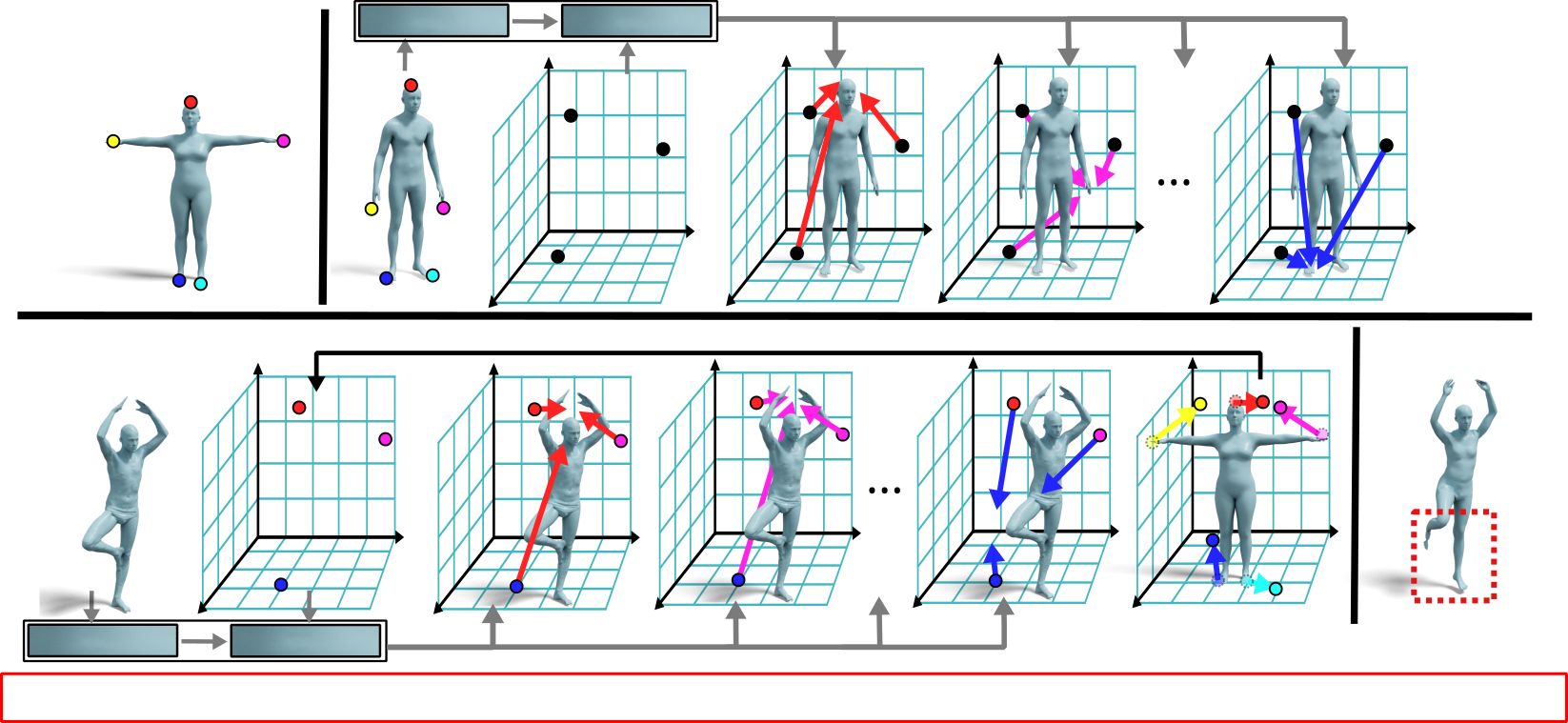}
\put(0.7,0.9){\textbf{Misaligned Registration} $\rightarrow$ NF bounded by training distribution converges far from the target}

\put(6,42.7){Template}
\put(0,30.7){\rotatebox{90}{TRAINING}}
\put(0,9.7){\rotatebox{90}{INFERENCE}}

\put(23.7,44.05){IF-NET}
\put(38, 44.05){MLP}

\put(2.5,4.5){IF-NET}
\put(17,4.5){MLP}

\put(40,24){Iterate until convergence}

\put(89.5,22.7){LVD\cite{corona2022learned}}

\put(91,1){}
        
	\end{overpic}

		\caption{\label{fig:methods}
NF for human registration. During training (top), the network learns to predict, for every query point in space (black dots), the ordered ground-truth offsets toward the template vertices (colored dots). At inference time (bottom), given a target shape, the template vertices locations are queried. The template vertices are then moved by applying the predicted ordered offsets. The process is iterated till convergence. Out-of-distribution data and ruined geometry often result in trivial misalignment.}

\end{figure*}

\subsection{Point Cloud Registration}
3D Point cloud registration is the process of spatially aligning a template with an unsorted target point cloud while respecting its semantics. Given a template with ordered points $\mathbf{X} \in \mathbb{R}^{m \times 3}$ and an unsorted target point cloud $\mathbf{Y} \in \mathbb{R}^{n \times 3}$, the goal is to recover a deformation $F:\mathbf{X} \rightarrow \hat{\mathbf{X}} \in \mathbb{R}^{m \times 3}$ such that $\hat{\mathbf{X}}$ aligns with ${\Pi}(\mathbf{Y})$ under a permutation ${\Pi}: \mathbf{Y} \rightarrow \hat{\mathbf{Y}} \in \mathbf{Y}^{m}$, which encodes the correspondence:
\begin{equation}
    \|F(\mathbf{X}) - {\Pi}(\mathbf{Y})\|_F = 0.
    \label{eq:registration}
\end{equation}

However, the real case is more complex as the correspondence $\Pi$ is not always bijective due to different numbers of points, partial views, noise, and clutter. As a result, often Equation~\ref{eq:registration} does not have a ground-truth solution and is solved by optimization. Figure~\ref{fig:challenges} shows examples of partial and cluttered input (red point clouds), as well as the output of our method (grey meshes).

\mypar{Iterative registration}  Among the possible optimizations to solve Equation~\ref{eq:registration}, a popular solution is to iterate between the correspondence and the deformation such that they refine each other till convergence. A famous seminal example is the Iterative Closest Point (ICP) algorithm from~\cite{besl1992method}. Starting from an initial configuration of the two shapes (\ie, $\widetilde{F}(\mathbf{X}) = \mathbf{X}$), ICP obtains an estimated correspondence $\widetilde{\Pi}$ by Euclidean nearest-neighbor pairing:

\begin{equation}
    \widetilde{\Pi} = \arg\min\limits_{\Pi} \|\widetilde{F}(\mathbf{X}) - \Pi(\mathbf{Y})\|_2^2.
    \label{eq:corr}
\end{equation}

Next, ICP updates the estimated registration $\widetilde{F}$ by minimizing the point-to-point distance with a least-squares objective:

\begin{equation}
    \widetilde{F} = \arg\min\limits_{F} \sum\limits_{i = 1}^{m}\|F(\mathbf{x}_i) - \widetilde{\Pi}(\mathbf{Y})_i\|_2^2.
    \label{eq:def}
\end{equation}
The two steps are iterated multiple times until convergence. To avoid trivial solutions, $\widetilde{F}$ is generally restricted to a specific class (\eg, SE(3)) or regularized (\eg, with an as-rigid-as-possible as in \cite{sorkine2007rigid}). However, the Euclidean nearest-neighbor pairing is undesirable both for computational and stability reasons, requiring mitigation strategies like \cite{gelfand2003geometrically, men2011color, Yew_2020_CVPR} in the presence of outliers. ICP converges at a global minimum only when the shapes are already roughly aligned.  

\mypar{Neural Fields for Registration}
\label{sec:NF}
Neural Fields (NF) have been proven to be a powerful representation of 3D geometry, which parametrizes a quantity defined all over a domain coordinates (a field) using a neural network~\cite{xie2022neural}. \cite{corona2022learned} firstly propose their use for solving the registration problem, calling this procedure Learned Vertex Descent (LVD). The idea is to train a neural network $F^{\theta}: \mathbb{R}^3 \rightarrow \mathbb{R}^{m \times 3}$ that, for any point in $\mathbb{R}^3$, outputs the ordered offsets toward the $n$ vertices of the desired deformed template  $\hat{\mathbf{X}}$ (Figure~\ref{fig:methods}, top). At inference time, the network produces the NF for a target shape (Figure~\ref{fig:methods}, bottom left). The location $\hat{\mathbf{x}}_i$ of the $i$-th registered template vertex is obtained by querying the NF on $\mathbf{x}_i$ and following the $i$-th predicted offset $F^{\theta}(\mathbf{x}_i)_i$ (Figure~\ref{fig:methods},  bottom right):
\begin{equation}
 \hat{\mathbf{x}}_i = \mathbf{x}_i + F^{\theta}( \mathbf{x}_i)_i. %
\end{equation}
The inference procedure is iterated multiple times till convergence of $\hat{\mathbf{x}}_i$.

\mypar{NF Limitations.} As typical of data-driven methods, NF predictions are based on training distribution and are bounded by them. Also, NF does not incorporate any clue or bias to promote solutions on the target geometry. Considering the variety of human identities and poses (further complicated by clutter, \eg, garments), it is common for NF to miss the target (Figure~\ref{fig:methods}, bottom right).

\section{Method}
\label{sec:method}
\begin{figure*}[!t]
\scriptsize
 \begin{overpic}[trim=0cm 0cm 0cm 0cm,clip, width=\linewidth]{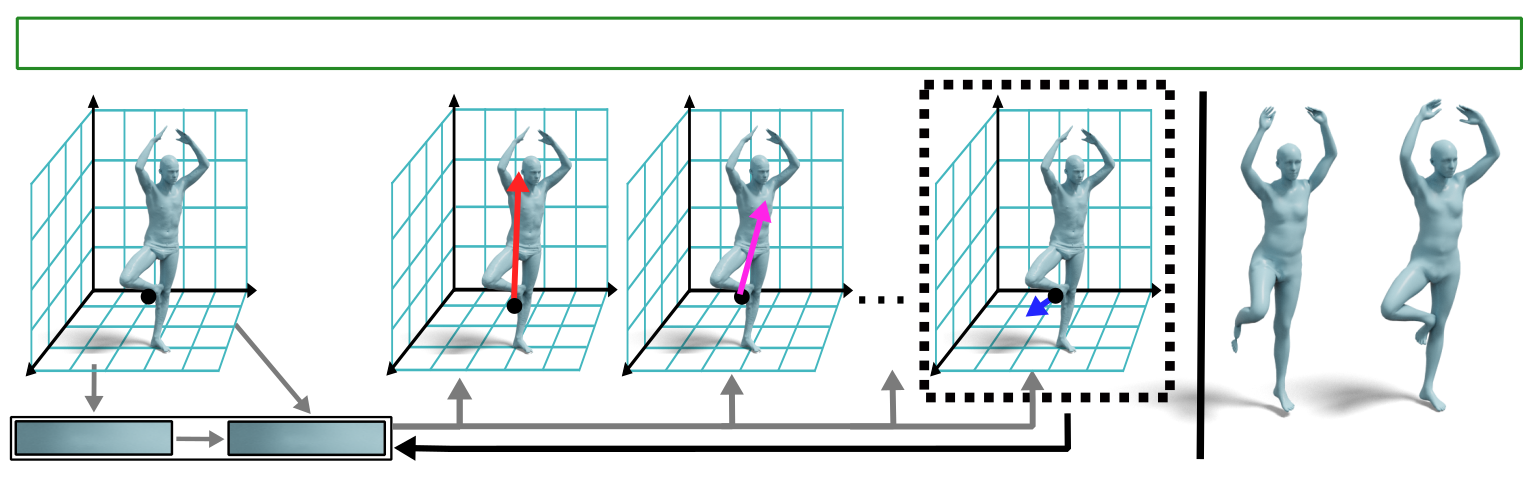}
    \put(17,9){\textbf{(1)}}
    \put(56, 24){\textbf{(2)}}
    \put(37, 0){\textbf{(3)}}
    
    \put(2.5,2.75){IF-Net}
    \put(17.5,2.75){MLP}

    \put(1.8,28.5){\textbf{Neural ICP (NICP)} $\rightarrow$ A self-supervised task, to promote convergence on the target surface}

    \put(40,-0.18){$\sum\limits_{k=1}^{n} \| F^{{\theta}}(\mathbf{y}_k)_{\widetilde{i_k}}\|_2^2$}

    \put(80,2){LVD}
    \put(79.7,-0.3){\cite{corona2022learned}}

    \put(90,2){\textbf{Ours}}
	\end{overpic}

\caption{\label{fig:inlovd} The key enabler of our generalization is \method{}. \textbf{(1)}: At inference time, we query points on the target surface; \textbf{(2)}: For each query point, we select the offset with the minimum norm; \textbf{(3)}: The sum of the retrieved norms for all the vertices is used to tune the backbone NF. \method{} is backbone agnostic and applies to any NF.}

\end{figure*}
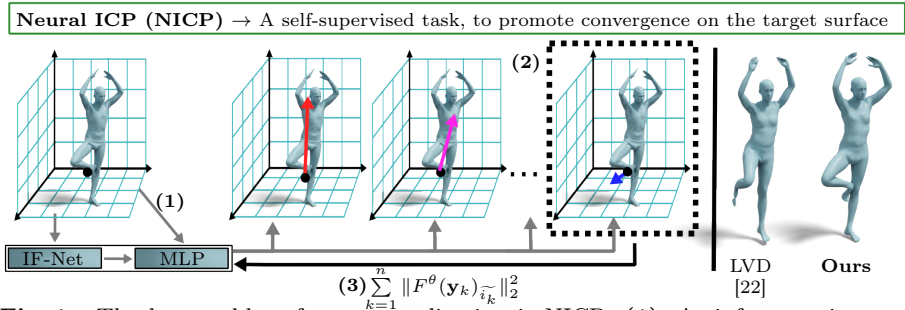

\mypar{Overview of \pipeline{}.} Our 3D Human registration pipeline starts from a 3D point cloud; all the meshes depicted in this paper are for visualization purposes only and are not used. The input is passed to a localized backbone network with multiple heads (LoVD, described in the next paragraph) that obtain NF dedicated to different shape's parts. Before letting the NF converge, we refine it using \method{} (Sec.~\ref{sec:nicp}), a self-supervised task that iteratively improves the backbone. Then, we use the updated NF to register the template to the input point cloud. The registration output by the network is further refined by Chamfer optimization and, if high-frequency details are required, also by vertex displacement. The whole pipeline is explained in Sec.~\ref{sec:NSR}.

\mypar{Local Vertex Descent (LoVD).} We propose a variant of LVD that employs multiple MLPs heads, each specialized in predicting offsets only for vertices in a local region of the template. To define the local regions, a popular choice explored also by~\cite{deng2020nasa, bhatnagar2020combining, mihajlovic2021leap, alldieck2021imghum, mihajlovic2022coap, wang2021locally} is to start from the $24$ skinning weights distribution of SMPL. However, designing different levels of granularity (\eg, considering $10$ or $16$ segments) requires significant manual intervention, and hence this aspect is rarely considered in previous works. 
Instead, we rely on spectral clustering as described in \cite{liu2004segmentation}. Given the SMPL template mesh, we compute its Laplace-Beltrami Operator using the discretization of~\cite{Sharp2020Robust}, we collect the $l$ eigenvectors associated with the $l$ smallest non-zero eigenvalues, and we use them as features for K-Means to obtain $l$ clusters. We empirically found that $l=16$ leads to best results; we report in Sup. Mat. the quantitative analysis. Our localization strategy is general and could be applied to domains that do not enjoy a template rich as SMPL. LoVD follows the same training and inference schema as LVD.

\subsection{Neural ICP (\method{})}
\label{sec:nicp}
\mypar{Insight.} Correspondences should lie exclusively on the target surface. We observe that in the case of NF, the deformation can be queried over the entire $\mathbb{R}^3$, and so on the target shape. Intuitively, we expect that queries of the NF deformation on the target vertices produce one or more offsets with norms close to 0. Our iterative self-supervised task promotes this desirable property. Following the classical ICP procedure, we would update the NF registration to converge toward the target surface. On the other hand, the Euclidean pairing of standard ICP is critical and often leads to local minima.  Our idea is to exploit the data prior learned by the network to pair the points of the target with those of the template. 
Hence, given a target shape, we iterate two steps: solving for the correspondence and solving for the registration. 

\mypar{Correspondence.} We sample points $\mathbf{{y}}_k$ over the target shape and query the NF (Figure~\ref{fig:inlovd}, 1). We pair every query point with a template vertex $\widetilde{i_k}$, corresponding to the one with the smallest predicted offset (Figure~\ref{fig:inlovd}, 2): 
\begin{equation}
 \widetilde{i_k} = \arg \min_i \|F^{\widetilde{\theta}}(\mathbf{y}_k)_i\|_2^2.
 \label{eq:nficp_corr}
\end{equation}
Note that this replaces the Euclidean $\arg\min$ of Equation~\ref{eq:corr} with one suggested by the network data prior and directly operating on the input surface.

\mypar{Registration.} We minimize the sum of minimum offsets $F^{{\theta}}(\mathbf{y}_k)_{\widetilde{i_k}}$ for all the query points $\mathbf{y}_k$, updating the NF parameters by backpropagation (Figure~\ref{fig:inlovd}, 3):
\begin{equation}
    \widetilde{\theta} = \arg\min_{\theta}\sum\limits_{k=1}^{n} \| F^{{\theta}}(\mathbf{y}_k)_{\widetilde{i_k}}\|_2^2.
    \label{eq:nficp_reg}
\end{equation}
Namely, we refine the field such that the deformation converges toward the estimated correspondence (\ie, it predicts a $0$ offset for the associated template vertex). Iterating these two steps refines the backbone network. We stress that our Neural ICP (\method{}) is a self-supervised task performed \emph{at inference time}, does not require data or supervision, and is the first of its kind for neural fields.

\subsection{\pipedef{} (\pipeline{})}
\label{sec:NSR}
\mypar{Input.} Given an input point cloud, we only assume a rough estimation of the Y-axis, while humans can face in any other direction. Such assumption is standard in human body registration pipelines like \cite{groueix20183d} and \cite{trappolini2021shape}, and can be easily approximated, for example, by computing robust landmarks as in \cite{marin2020farm}.

\mypar{\pipeline{} and template fitting.} Given a target point cloud, we use LoVD to obtain the NF, and we fine-tune with \method{} by iterating between Equation~\ref{eq:nficp_corr} and Equation~\ref{eq:nficp_reg}. We update the network weights $20$ times with a learning rate of $1e-5$. The whole procedure takes around $3$ seconds. Then, we use the NF to move the points of the template till convergence and, as in \cite{corona2022learned}, we fit a SMPL model to the prediction, using the same identity and pose prior penalizations \cite{SMPL-X:2019}.

\mypar{Refinement.} A common practice is to refine the template parameters with a Chamfer distance optimization. When we assume bijectivity between the template and target, we use the bidirectional one; otherwise, we rely on a single direction as in \cite{bhatnagar2020loopreg}. Finally, in cases where a high resolution is required, following the spirit of \cite{marin2019high}, we use SMPL+D to catch the finer details of the target, regularizing the displacements with a Laplacian energy \cite{gao2020learning}. We report in Sup. Mat. the technical details of the whole pipeline.

\mypar{Names for Ours Results.} We use \emph{LoVD} for results obtained by fitting SMPL to the LoVD prediction, without further steps; \emph{+\method{}} when we apply our iterative task before NF convergence; \emph{LoVD+\method{}} when we combine the LoVD and \method{}; \emph{\pipeline{}} for the complete registration pipeline, namely when LoVD+\method{} result is further refined. All results with these names are part of our contribution.

\section{Results}
\label{sec:results}
\mypar{Remark.} Apart from Figure~\ref{fig:cape} and Table~\ref{tab:cape}, all shown human registrations here and in Sup. Mat. are obtained starting from \textit{the same network weights}. The quantitative and qualitative experiments stress our method on the FAUST \cite{bogo2014faust} and SHREC19 \cite{melzi2019shrec} challenges; this latter collects shapes from $10$ different datasets. We also consider partial point clouds from CAPE \cite{ma2020learning} and Kinect ones from BEHAVE \cite{bhatnagar2022behave}, NeRF-based 3D human reconstructions from LumaAI \cite{luma} and cases significantly out of distribution from the \cite{scantheworld} Scan the World Project and TOSCA \cite{TOSCA}. We report results for challenging poses from SCAPE \cite{anguelov2005scape}, clothed humans from Twindom \cite{Twindom} and RenderPeople \cite{renderpeople}, and output of acquisition pipeline as Dynamic-Faust \cite{bogo2017dynamic} and the recent HuMMan dataset \cite{cai2022humman}. We are unaware of methods that provide an extensive evaluation and generalization, such as the one in our study.

\mypar{Template.} We consider the human SMPL model sampled at $690$ vertices as a template; hence, our NF will provide $690 \times 3$ values for each point. This subsample speeds up training and inference while we retain enough information to fit a complete SMPL model. 

\mypar{Data.} To train the NF, we leverage the large MoCap AMASS dataset of \cite{mahmood2019amass}, adopting the official splits. The train set comprises roughly 120k SMPL+H~\cite{smplh} shapes animated with motion-captured sequences, and we train the NF for $10$ epochs ($\sim157$k steps). We rely on the FAUST challenge's training set for validation and ablations. We consider as input both the scans (\textbf{Faust$_S$}, $\sim$170K-200K vertices, noisy) and the ground-truth registrations (\textbf{Faust$_R$}, 6890 vertices, clean). We refer to Sup. Mat. for implementation details.

\mypar{Baselines.} We compare against several baselines using the same feature extractor and training set. First, we consider the original LVD formulation (\textbf{LVD}) as described in the previous section. As in its original formulation of \cite{corona2022learned} (and also for our LoVD variant), at inference time, LVD is iterated multiple times till convergence. To stress that \method{} can be applied to different NF, we also propose \textbf{OneShot}, an NF that predicts offsets toward the registered template vertices for every point in $\mathbb{R}^3$ in a single pass. Finally, we also reimplemented the Universal embedding baseline proposed by \cite{marin2020correspondence}, which learns a high-dimensional embedding where shapes are naturally aligned, and the correspondence is obtained by Euclidean nearest neighbor (\ie, forcing the results to lie on the target surface). We consider a $60$ dimensional embedding (\textbf{Uni-60}) as proposed by \cite{marin2020correspondence}, and also a $2070$ one (\textbf{Uni-2070}) to have similar output dimensions of our method ($690$ offset of $3$ dimensions). As described in the next Section, we also compare \pipeline{}, \textbf{PTF}~\cite{wang2021locally}, \textbf{IPNET}~\cite{bhatnagar2020combining}, and LVD on the same CAPE~\cite{ma2020learning} training and test splits.

\mypar{Metrics.} For tests on FAUST, BEHAVE, and CAPE, the error is measured in centimeters as the Euclidean distance from the predicted point position and the ground truth. On SHREC19, the error is the normalized geodesic distance~\cite {kim2011blended}. 

\mypar{Computational Timing.} We conduct our experiments on a computer equipped with a 12-core CPU AMD Ryzen 9 5900X, an NVIDIA GeForce RTX 3080 Ti GPU, and RAM 64GB. The entire \method{} requires 3--4 seconds; \pipeline{} registration for a single shape is around 60 seconds. 
\begin{table}[!t]
\setlength{\tabcolsep}{6pt}
\begin{tabular}{cc}
  \begin{minipage}[t]{0.5\textwidth}
   \setlength{\tabcolsep}{2pt}
\centering
\footnotesize
	\begin{tabular}{@{}lcc@{}}

		    & FAUST$_{R}$ & FAUST$_{S}$                    \\
                \hline
                  OneShot        & 4.45          & 3.34         \\
                  OneShot+\method{}  \textbf{(Ours)}     & \textit{3.64}            & \textit{3.15} \\
                  
                \hline
                LVD            & 4.78          & 3.54                \\ 
                LVD+\method{}     \textbf{(Ours)}    & \textit{3.40}            & \textit{2.81}  \\          \hline
                LoVD \textbf{(Ours)}    & \textbf{4.35} & \textbf{3.11} \\
                LoVD+\method{}    \textbf{(Ours)}    & \textit{\textbf{2.97}}   & \textit{\textbf{2.55}}   \\                  
                \hline \\
          \end{tabular}
  	\captionof{table}{\label{tab:backbones} Comparisons after NF convergence. LoVD provides the best results, and \method{} improves all the backbones.}
    \end{minipage}
    &
      \begin{minipage}[t]{0.45\textwidth}
       \setlength{\tabcolsep}{2pt}
      \centering
\footnotesize   	
\begin{tabular}{lcc}
	          & FAUST$_{R}$ & FAUST$_{S}$          \\
		\hline
  Uni-60 \cite{marin2020correspondence}    &  3.52 &2.66   \\
  Uni-2070 \cite{marin2020correspondence} 
  &  3.49 & 2.62 \\
  LVD pre-Chamfer             & 4.78 & 3.54 \\
  LVD \cite{corona2022learned} & 2.15 & 1.93 \\
  \textbf{\pipeline{} (Ours)}    & \textbf{1.76} & \textbf{1.85} \\
  \hline \\
	\end{tabular}
 		\caption{\label{tab:comparison}
	Comparison with methods trained on AMASS~\cite{mahmood2019amass} and all using the same feature extractors.}
\end{minipage}
\end{tabular}
\end{table}

\subsection{Validation}

\begin{figure}[!t]
    \setlength{\tabcolsep}{12pt}
\centering
\begin{tabular}{cc}

\begin{minipage}{0.20\linewidth}
\setlength{\tabcolsep}{3pt}
    \footnotesize
	\begin{tabular}{@{}lr@{}}
		          & BEHAVE          \\
		\hline
        LoVD       &      6.78         \\
        LoVD+\method{}   &      \textbf{5.69}\\
        \hline \\
	\end{tabular}
	
\end{minipage}
&

\begin{minipage}{0.75\linewidth}
	\footnotesize
	\begin{overpic}[trim=0cm 0.0cm 0cm 0cm,clip, width=0.90\linewidth]{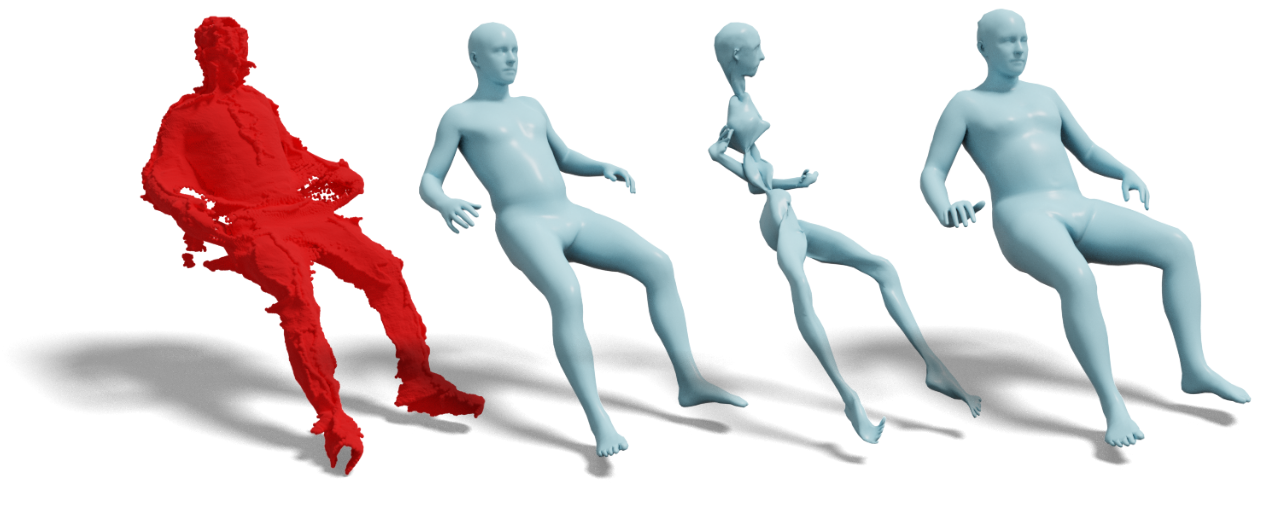}
    \put(13,40){Input}
    \put(37,40){GT}
    
    \put(53,40){LoVD}

    \put(72,40){LoVD+\method{}}
    
	\end{overpic}

\end{minipage}
\end{tabular}
	\captionof{table}{\label{fig:behave} \method{} robustness to noise. On the left, quantitative registration error on  3320 frames of \cite{bhatnagar2022behave}; \method{} improves on average 17,6\% over the backbone and improves results on 80\% of the frames. On the right is a qualitative result.}
\end{figure}

\mypar{LoVD.} In Table~\ref{tab:backbones}, we compare the convergence of our localized NF against other NF backbones. Even before \method{}, LoVD is the NF that provides the best performance. All the NF use the same features and training dataset, proving the effectiveness of our design choice. We include in Sup. Mat. a study on different numbers of segments.
 
\mypar{\method{}.} In Table~\ref{tab:backbones}, we show the effect of \method{} on the convergence of different NF. \method{} solidly improves all the backbones, showing generalization without requiring data or supervision. We also report in Sup. Mat. its application on a pre-trained network for a different domain (\ie, hand registration).

\mypar{LoVD+\method{}.} Table~\ref{tab:backbones} shows that the combination of our two ingredients improves on the two datasets \textbf{30\% and 23\%} w.r.t. the best available NF baseline (OneShot). Considering all these methods have seen the rich AMASS training set, this result further proves that our design choices provide an advantage beyond the already strong data priors.

\mypar{\pipeline{}.}  Table \ref{tab:comparison} reports the comparison with registration and correspondence pipelines trained on AMASS~\cite{mahmood2019amass} and all using the same feature extractor. Uni baselines are challenging competitors, able to surpass the convergence of SoTA NF (LVD pre-chamfer \cite{corona2022learned}). LVD and \pipeline{} outputs can both be refined using chamfer distance, and the two differ only for the backbone (LoVD) and the use of \method{}, which are our original contributions and provide substantial improvement (around \textbf{20\%} on FAUST$_R$).

\mypar{\method{} Generalization and Robustness.}
\method{} updates the network by randomly sampling the target input point cloud. 
We wonder how our self-supervised procedure is sensitive to outliers. Hence, we test \method{} on the BEHAVE dataset proposed in \cite{bhatnagar2022behave}. It includes point clouds from Kinect multi-view depths and presents significant noise (\eg, clutter, missing human parts due to occlusions). We uniformly sample 3320 frames from the sequences, register the fused depth point cloud, and compare the error with the provided ground truth registrations. Results are displayed in Figure~\ref{fig:challenges}, Table~\ref{fig:behave}.
\method{} enhances the backbone by around \textbf{17.6\%, and improves on 80\% of the frames}; it enables robust generalization on clutter and partiality without having seen it at training time, which is often mandatory for other methods \cite{attaiki2021dpfm, predator}. In Sup. Mat. we report many qualitative samples, showing that \method{} is the key enabler of such generalization.

\begin{figure}[!t]
 \footnotesize
 \begin{overpic}[trim=0cm 0cm 0cm 0cm,clip,   width=\linewidth]{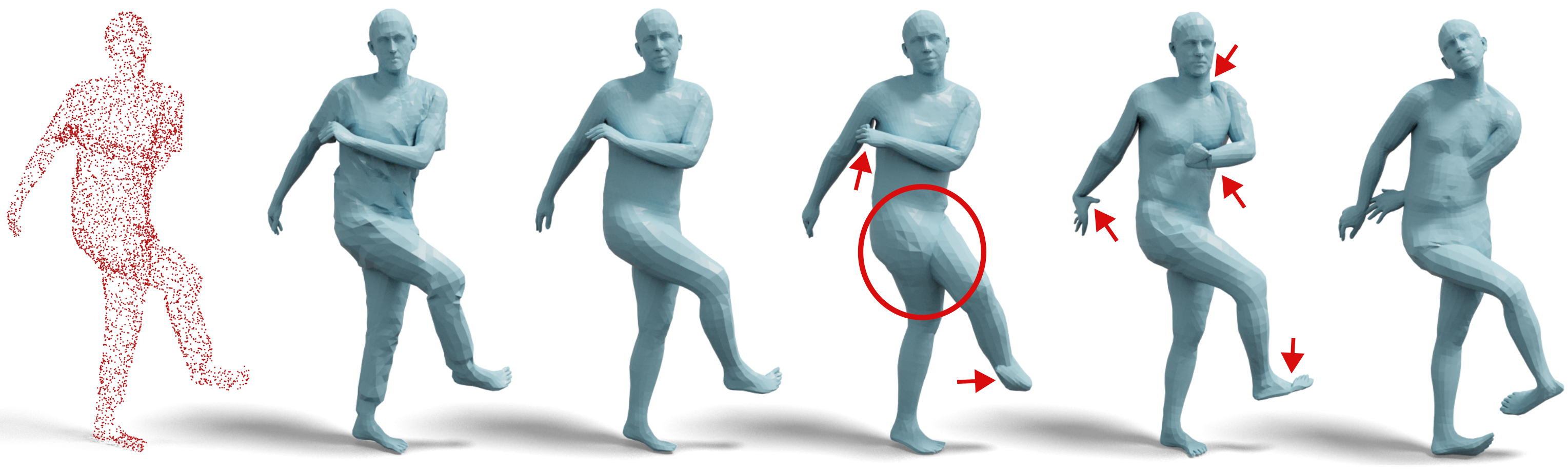}

 \put(6,30.5){Input}
  \put(23,30.5){GT}
  \put(38.8,30.5){\textbf{\pipeline{}}}
  \put(56,30.5){LVD}
  \put(74,30.5){{PTF}}
  \put(90,30.5){{IP-Net}}
\end{overpic}
\footnotesize 
\center
\caption{\label{fig:cape} Comparison between methods trained on the same CAPE training set. LVD swaps the legs due to the twisted pose; self-contact causes PTF artifacts (red arrows).}
\end{figure}

\mypar{SoTA Comparison.}
We compare our method against the pre-trained PTF\cite{wang2021locally} and IP-Net\cite{bhatnagar2020combining} models on the CAPE dataset. 
We trained our method and LVD on the same CAPE training split for $\sim$65k steps (competitors use $\sim$200k steps). We sample the test sequences every 20 frames for 1021 shapes; for each one, we generate the input point clouds as described in PTF's paper.In Tab. \ref{tab:cape}, we report the error w.r.t. the ground truth clothed bodies. Our method improves by \textbf{$\sim15$\%}.\begin{wraptable}[10]{r}{0.35\linewidth}
\footnotesize
\centering
\begin{tabular}{ l r           c c }
Train and test on CAPE  \\
\hline
IP-Net  \cite{bhatnagar2020combining}               & 2.59   \\
PTF \cite{wang2021locally}                    & 2.05  \\  
LVD \cite{corona2022learned}                   & 1.98  \\
\textbf{\pipeline{}} \textbf{(Ours)} & \textbf{1.70}    \\ 
\hline
\end{tabular}
\caption{\label{tab:cape} Quantitative comparison against SoTA methods trained on the same data and tested on 1021 shapes.}
\end{wraptable} Also, PTF and IP-NET take around 3 minutes per shape, while \pipeline{} and LVD require only 1 minute. A qualitative example can be seen in Figure \ref{fig:cape}. LVD fails to attend to local body parts, and even confuses left-right in twisted poses. As mentioned in \cite{wang2021locally}, PTF fails in the presence of self-contact even when the shape comes from the same training distribution. We report in Sup. Mat. further comparison with the general shape matching methods from~\cite{sundararaman2022implicit,cao2023self,jiang2023nonrigid}.

\subsection{3D Human Body Registration}

\begin{table}[!t]
\setlength{\tabcolsep}{12pt}
\begin{tabular}{cc}
  \begin{minipage}[t]{0.5\textwidth}
  \setlength{\tabcolsep}{3pt}
\centering
\footnotesize
	\begin{tabular}{@{}lrr@{}}
		          & INTRA & INTER          \\
		\hline
        UD2E-Net \cite{Chen_2021_ICCV}          & 1.51     & 3.09 \\		
        Loopreg \cite{bhatnagar2020loopreg}    & 1.35     & 2.66 \\
        Elementary \cite{deprelle2019learning}  & 1.74     & 2.58 \\
        \textbf{\pipeline{} (Ours)}   & \textbf{1.06} & \textbf{2.26}\\
        \hline
        \hline 
        DVM (multi-view)$^\dagger$ \cite{kim2021deep}    &  1.19    & 2.37 \\
        \hline
        $^\dagger$: {Uses FAUST training set}\\
        \hline  \\
        
	\end{tabular}
  	\captionof{table}{\label{fig:faust} Results from the public leaderboard on INTRA- and INTER-subject FAUST challenges.}
    \end{minipage}

    &
      \begin{minipage}[t]{0.4\textwidth}
      \centering
      \footnotesize
	\begin{tabular}{@{}lr@{}}
		          & SHREC19          \\
		\hline
        LIE \cite{marin2020correspondence}   & 15.1   \\
        GeoFMAP     \cite{donati2020deep} & 11.2\\
        3DCoded    \cite{groueix20183d} & 10.3\\
        CorrNet3D \cite{zeng2020corrnet3d}   & 9.6   \\
        Transmatch  \cite{trappolini2021shape}& 6.1 \\
        ReduceBasis \cite{sundararaman2022reduced}& 4.8 \\ 
        \textbf{\pipeline{} \textbf{(Ours)}}         & \textbf{2.3} \\
        \hline  \\
	\end{tabular}
	\captionof{table}{\label{fig:shrec19} Results on SHREC19 challenge \cite{melzi2019shrec}. We improve the state of the art by 53\%.}
\end{minipage}
\end{tabular}
\end{table}
\mypar{FAUST Challenge.}
We test our pipeline on real data from the two FAUST challenges\footnote{\url{https://faust-leaderboard.is.tuebingen.mpg.de/leaderboard}} from \cite{bogo2014faust}. The scans are of 10 subjects in 20 poses (200 samples), with missing regions and clutter due to the noise in the acquisition process.
Table~\ref{fig:faust} reports the error results from the leaderboard. We list the methods with similar assumptions to ours in the upper part of the table, on which we achieve an improvement of \textbf{~20\% and ~12\%} on the INTRA-subject and INTER-subject challenges, respectively. We also outperform DVM \cite{kim2021deep}, which requires a manually artist-annotated dataset for training, the render of 72 synthetic views of the target for inference, and \emph{have seen FAUST shapes at training time}, while we do not. In Sup. Mat. we report qualitative results.

\noindent{\textit{\textbf{SHREC19.} }}
We test our method's generalization on the challenging dataset SHREC19~\cite{melzi2019shrec}. 
This database is a collection of 44 shapes from different datasets, facing various challenges: holes, out-of-distribution identities, disconnected components (\eg, earrings), clutter, and different densities (shapes have from 5k to 200k vertices).
The ground truth on this data is provided by a robust registration method that exploits surface information. In Table~\ref{fig:shrec19}, we report our results and the leaderboard from \cite{sundararaman2022reduced}. 
Our method significantly outperforms previous state-of-the-art and improves \textbf{~53\%}. See Sup. Mat. for qualitative results.

\begin{figure*}[!t]

    \centering
 \begin{overpic}[trim=1.0cm 0cm 0cm 0cm,clip, width=\linewidth]{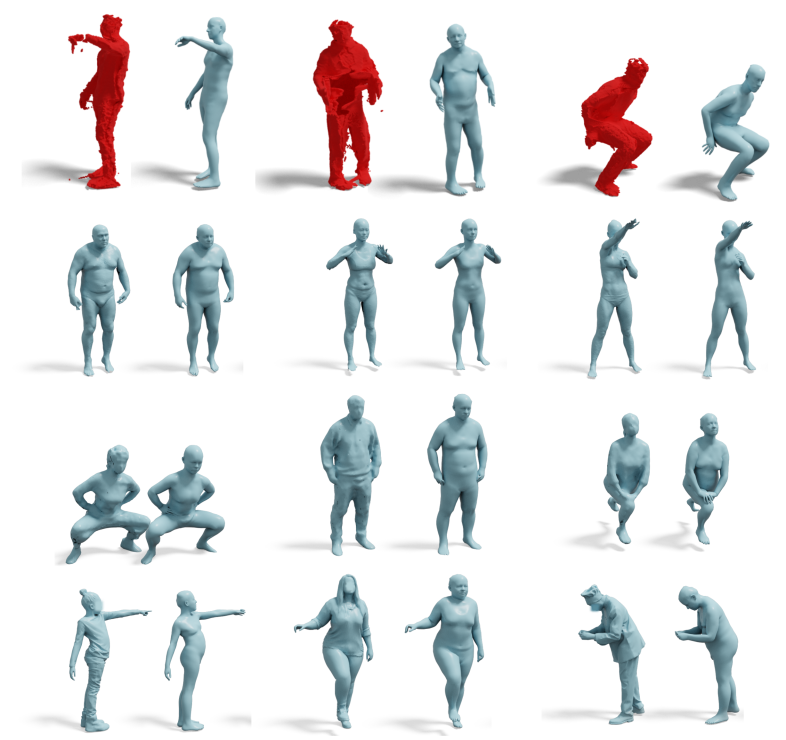}
  \footnotesize
		\put(5,97.8){Input}
            \put(20,97.8){\textbf{\pipeline{}}}
            
		\put(38,97.8){Input}
            \put(52,97.8){\textbf{\pipeline{}}}
            
            \put(76,97.8){Input}
            \put(88,97.8){\textbf{\pipeline{}}}

            \put(-1,4){\rotatebox{90}{Twindom \cite{Twindom}}}
            \put(-1,27){\rotatebox{90}{HuMMan \cite{cai2022humman}}}
            \put(-1,53){\rotatebox{90}{DFAUST \cite{bogo2017dynamic}}}
            \put(-1,78){\rotatebox{90}{BEHAVE \cite{bhatnagar2022behave}}}

	\end{overpic}

    \caption{\label{fig:manyresults} Results on shapes acquired with Kinects (BEHAVE), a 4D Scanner (DFAUST), a Multi-view RGB-D dome (HuMMAN), and a static full-body scanner (Twindom). Our method works out of the box in multiple applicative contexts. We provide in Sup. Mat. further qualitative examples.}
\end{figure*}

\subsection{Applications}

\begin{figure*}[!t]
    \centering
    \footnotesize
 \begin{overpic}[trim=0cm 0cm 0cm 0cm,clip, width=\linewidth]{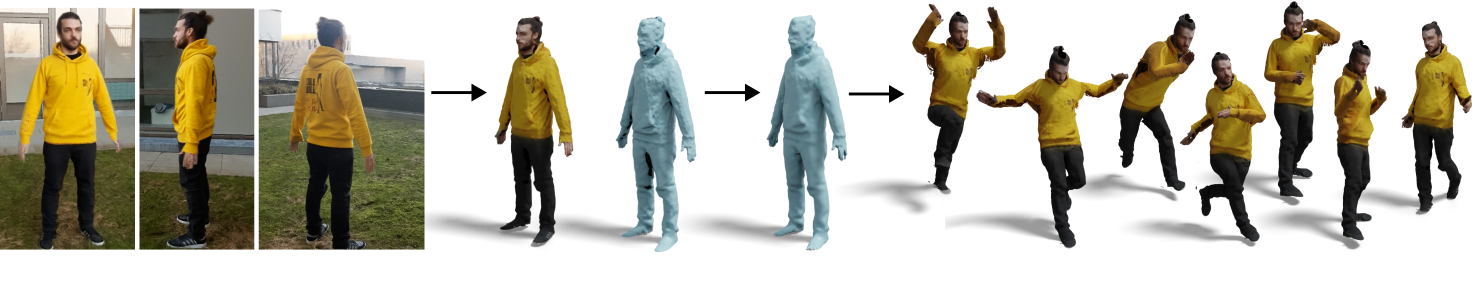}
		\put(6,-0.5){RGB Video}
            \put(31,-0.5){LUMA Output}
            \put(52, -0.5){\textbf{\pipeline{}}}
            \put(65, -0.5){Animation of LUMA Output}
	\end{overpic}
    \caption{\label{fig:LumaAI} Animatable avatar from \cite{luma}. From left to right: images from the smartphone video; the textured geometry obtained by \cite{luma} NeRF; \pipeline{} registration; motion capture animation applied to the input data.}
\end{figure*}

\mypar{Unifying Data Sources.} There are many different procedures to acquire 3D Human models. Our registration method based on point cloud geometry with no further assumption on the input source (\eg, availability of RGB, subject orientation) can provide a standard for research, industry, and customer-level applications. In Figure~\ref{fig:manyresults}, we show registration results on four datasets obtained with respectively with Kinects, a 4D scanner, a multi-view RGB-D dome, and a static body scanner. In Figure~\ref{fig:LumaAI}, we register a geometry extracted by a NeRFs \cite{nerf}, trained on a mobile phone video and using LumaAI website \cite{luma}.

\mypar{Automatic Rigging.} Reliable registrations can be used to transfer information between aligned assets. In animation, for example, transferring the rigging between 3D models is an open problem, and researchers proposed methods like~\cite{seylan20193d,musoni2021functional, yang2018dmat} specifically designed for that. In Figure~\ref{fig:LumaAI}, we show how we can use \pipeline{} to register a noise NeRF acquisition, transfer the SMPL skinning information, and then apply a MoCap animation sequence. The entire procedure, from the NeRF training to the animation, can be automated, opening vast possibilities for content creation applications.

\section{Conclusions}
\label{sec:conclusions}
In this work, we introduce \method{}, a new self-supervised task that at inference time, in a self-supervised way and in a few seconds, improves the registration accuracy and generalization of the backbone Neural Field. Our extensive experiments show how \method{} can be applied out of the box to arbitrary NF, providing a critical advantage on the backbone.
We include it in a full registration pipeline, with a novel localized variant of LVD trained on a large MoCap dataset.  Our method, \pipeline{}, produces a performance boost on public benchmarks and shows robustness to noise and incomplete point clouds, clothing, poses, and all \emph{using the same network weights}. In Supplementary Material, we discuss failure modes and exciting directions for future works. We consider \pipeline{} the first approach showing at-scale results. At the time of publication, code and network weights are released on the project page reported at the beginning of this article, providing a way to go for 3D human registration and a valuable tool for several Computer Vision tasks, users, and researchers.

\section*{Acknowledgements}
Thanks to Garvita Tiwari for the proofreading and feedback, Ilya Petrov for code refactoring, and the whole RVH team for the support. The project was made possible by funding from the Carl Zeiss Foundation. This work is supported by the Deutsche Forschungsgemeinschaft (DFG, German Research Foundation) - 409792180 (Emmy Noether Programme, project:  Real Virtual Humans), German Federal Ministry of Education and Research (BMBF): Tübingen AI Center, FKZ: 01IS18039A. This project has received funding from the European Union’s Horizon 2020 research and innovation program under the Marie Skłodowska-Curie grant agreement No 101109330. Gerard Pons-Moll is a member of the Machine Learning Cluster of Excellence, EXC number 2064/1 - Project number 390727645.

%
%
\bibliographystyle{splncs04}
\bibliography{egbib}

\begin{thebibliography}{10}
\providecommand{\url}[1]{\texttt{#1}}
\providecommand{\urlprefix}{URL }
\providecommand{\doi}[1]{https://doi.org/#1}

\bibitem{alldieck2021imghum}
Alldieck, T., Xu, H., Sminchisescu, C.: imghum: Implicit generative models of 3d human shape and articulated pose. In: Proceedings of the IEEE/CVF International Conference on Computer Vision. pp. 5461--5470 (2021)

\bibitem{anguelov2005scape}
Anguelov, D., Srinivasan, P., Koller, D., Thrun, S., Rodgers, J., Davis, J.: Scape: shape completion and animation of people. In: ACM SIGGRAPH 2005 Papers (2005)

\bibitem{attaiki2021dpfm}
Attaiki, S., Pai, G., Ovsjanikov, M.: Dpfm: Deep partial functional maps. In: 2021 International Conference on 3D Vision (3DV). pp. 175--185. IEEE (2021)

\bibitem{aubry2011wave}
Aubry, M., Schlickewei, U., Cremers, D.: The wave kernel signature: A quantum mechanical approach to shape analysis. In: 2011 IEEE international conference on computer vision workshops (ICCV workshops). pp. 1626--1633. IEEE (2011)

\bibitem{icp}
Besl, P.J., McKay, N.D.: Method for registration of 3-d shapes. In: Sensor fusion IV: control paradigms and data structures. vol.~1611, pp. 586--606. Spie (1992)

\bibitem{besl1992method}
Besl, P.J., McKay, N.D.: Method for registration of 3-d shapes. In: Sensor fusion IV: control paradigms and data structures. vol.~1611, pp. 586--606. Spie (1992)

\bibitem{bhatnagar2020combining}
Bhatnagar, B.L., Sminchisescu, C., Theobalt, C., Pons-Moll, G.: Combining implicit function learning and parametric models for 3d human reconstruction. In: Computer Vision--ECCV 2020: 16th European Conference, Glasgow, UK, August 23--28, 2020, Proceedings, Part II 16. pp. 311--329. Springer (2020)

\bibitem{bhatnagar2020loopreg}
Bhatnagar, B.L., Sminchisescu, C., Theobalt, C., Pons-Moll, G.: Loopreg: Self-supervised learning of implicit surface correspondences, pose and shape for 3d human mesh registration. Advances in Neural Information Processing Systems  \textbf{33},  12909--12922 (2020)

\bibitem{bhatnagar2022behave}
Bhatnagar, B.L., Xie, X., Petrov, I.A., Sminchisescu, C., Theobalt, C., Pons-Moll, G.: Behave: Dataset and method for tracking human object interactions. In: Proceedings of the IEEE/CVF Conference on Computer Vision and Pattern Recognition. pp. 15935--15946 (2022)

\bibitem{bogo2014faust}
Bogo, F., Romero, J., Loper, M., Black, M.J.: Faust: Dataset and evaluation for 3d mesh registration. In: Proceedings of the IEEE conference on computer vision and pattern recognition. pp. 3794--3801 (2014)

\bibitem{bogo2017dynamic}
Bogo, F., Romero, J., Pons-Moll, G., Black, M.J.: Dynamic faust: Registering human bodies in motion. In: Proceedings of the IEEE conference on computer vision and pattern recognition. pp. 6233--6242 (2017)

\bibitem{bouaziz2013sparse}
Bouaziz, S., Tagliasacchi, A., Pauly, M.: Sparse iterative closest point. In: Computer graphics forum. Wiley Online Library (2013)

\bibitem{bronstein2006generalized}
Bronstein, A.M., Bronstein, M.M., Kimmel, R.: Generalized multidimensional scaling: a framework for isometry-invariant partial surface matching. Proceedings of the National Academy of Sciences  \textbf{103}(5),  1168--1172 (2006)

\bibitem{TOSCA}
Bronstein, A.M., Bronstein, M.M., Kimmel, R.: Numerical geometry of non-rigid shapes. Springer Science \& Business Media (2008)

\bibitem{cai2022humman}
Cai, Z., Ren, D., Zeng, A., Lin, Z., Yu, T., Wang, W., Fan, X., Gao, Y., Yu, Y., Pan, L., et~al.: Humman: Multi-modal 4d human dataset for versatile sensing and modeling. In: Computer Vision--ECCV 2022: 17th European Conference, Tel Aviv, Israel, October 23--27, 2022, Proceedings, Part VII. pp. 557--577. Springer (2022)

\bibitem{cao2023self}
Cao, D., Bernard, F.: Self-supervised learning for multimodal non-rigid 3d shape matching. In: Proceedings of the IEEE/CVF Conference on Computer Vision and Pattern Recognition. pp. 17735--17744 (2023)

\bibitem{openpose}
{Cao}, Z., {Hidalgo Martinez}, G., {Simon}, T., {Wei}, S., {Sheikh}, Y.A.: Openpose: Realtime multi-person 2d pose estimation using part affinity fields. IEEE Transactions on Pattern Analysis and Machine Intelligence  (2019)

\bibitem{cao2017realtime}
Cao, Z., Simon, T., Wei, S.E., Sheikh, Y.: Realtime multi-person 2d pose estimation using part affinity fields. In: CVPR (2017)

\bibitem{Chen_2021_ICCV}
Chen, R., Cong, Y., Dong, J.: Unsupervised dense deformation embedding network for template-free shape correspondence. In: Proceedings of the IEEE/CVF International Conference on Computer Vision (ICCV). pp. 8361--8370 (2021)

\bibitem{chetverikov2002trimmed}
Chetverikov, D., Svirko, D., Stepanov, D., Krsek, P.: The trimmed iterative closest point algorithm. In: 2002 International Conference on Pattern Recognition. vol.~3, pp. 545--548. IEEE (2002)

\bibitem{chibane2020implicit}
Chibane, J., Alldieck, T., Pons-Moll, G.: Implicit functions in feature space for 3d shape reconstruction and completion. In: Proceedings of the IEEE/CVF conference on computer vision and pattern recognition. pp. 6970--6981 (2020)

\bibitem{corona2022learned}
Corona, E., Pons-Moll, G., Aleny{\`a}, G., Moreno-Noguer, F.: Learned vertex descent: a new direction for 3d human model fitting. In: Computer Vision--ECCV 2022: 17th European Conference, Tel Aviv, Israel, October 23--27, 2022, Proceedings, Part II. pp. 146--165. Springer (2022)

\bibitem{cosmo2016matching}
Cosmo, L., Rodola, E., Masci, J., Torsello, A., Bronstein, M.M.: Matching deformable objects in clutter. In: 2016 Fourth international conference on 3D vision (3DV). pp. 1--10. IEEE (2016)

\bibitem{deng2022survey}
Deng, B., Yao, Y., Dyke, R.M., Zhang, J.: A survey of non-rigid 3d registration. In: Computer Graphics Forum. pp. 559--589. Wiley Online Library (2022)

\bibitem{deng2020nasa}
Deng, B., Lewis, J.P., Jeruzalski, T., Pons-Moll, G., Hinton, G., Norouzi, M., Tagliasacchi, A.: Nasa neural articulated shape approximation. In: Computer Vision--ECCV 2020: 16th European Conference, Glasgow, UK, August 23--28, 2020, Proceedings, Part VII 16. pp. 612--628. Springer (2020)

\bibitem{deprelle2019learning}
Deprelle, T., Groueix, T., Fisher, M., Kim, V., Russell, B., Aubry, M.: Learning elementary structures for 3d shape generation and matching. Advances in Neural Information Processing Systems  \textbf{32} (2019)

\bibitem{donati2020deep}
Donati, N., Sharma, A., Ovsjanikov, M.: Deep geometric functional maps: Robust feature learning for shape correspondence. In: Proceedings of the IEEE/CVF Conference on Computer Vision and Pattern Recognition. pp. 8592--8601 (2020)

\bibitem{ArtEq2023}
Feng, H., Kulits, P., Liu, S., Black, M.J., Abrevaya, V.F.: Generalizing neural human fitting to unseen poses with articulated se (3) equivariance. In: Proceedings of the IEEE/CVF International Conference on Computer Vision. pp. 7977--7988 (2023)

\bibitem{burov2021dsfn}
Gafni, G., Thies, J., Zollhofer, M., Nie{\ss}ner, M.: Dynamic neural radiance fields for monocular 4d facial avatar reconstruction. In: Proceedings of the IEEE/CVF Conference on Computer Vision and Pattern Recognition. pp. 8649--8658 (2021)

\bibitem{gao2020learning}
Gao, J., Chen, W., Xiang, T., Jacobson, A., McGuire, M., Fidler, S.: Learning deformable tetrahedral meshes for 3d reconstruction. Advances In Neural Information Processing Systems  \textbf{33},  9936--9947 (2020)

\bibitem{gelfand2003geometrically}
Gelfand, N., Ikemoto, L., Rusinkiewicz, S., Levoy, M.: Geometrically stable sampling for the icp algorithm. In: Fourth International Conference on 3-D Digital Imaging and Modeling, 2003. 3DIM 2003. Proceedings. pp. 260--267. IEEE (2003)

\bibitem{groueix20183d}
Groueix, T., Fisher, M., Kim, V.G., Russell, B.C., Aubry, M.: 3d-coded: 3d correspondences by deep deformation. In: Proceedings of the european conference on computer vision (ECCV). pp. 230--246 (2018)

\bibitem{hesse2018learning}
Hesse, N., Pujades, S., Romero, J., Black, M.J., Bodensteiner, C., Arens, M., Hofmann, U.G., Tacke, U., Hadders-Algra, M., Weinberger, R., et~al.: Learning an infant body model from rgb-d data for accurate full body motion analysis. In: Medical Image Computing and Computer Assisted Intervention--MICCAI 2018: 21st International Conference, Granada, Spain, September 16-20, 2018, Proceedings, Part I. pp. 792--800. Springer (2018)

\bibitem{hirose2020bayesian}
Hirose, O.: A bayesian formulation of coherent point drift. IEEE transactions on pattern analysis and machine intelligence  \textbf{43}(7),  2269--2286 (2020)

\bibitem{hirose2022geodesic}
Hirose, O.: Geodesic-based bayesian coherent point drift. IEEE Transactions on Pattern Analysis and Machine Intelligence  (2022)

\bibitem{huang2022multiway}
Huang, J., Birdal, T., Gojcic, Z., Guibas, L.J., Hu, S.M.: Multiway non-rigid point cloud registration via learned functional map synchronization. IEEE Transactions on Pattern Analysis and Machine Intelligence  (2022)

\bibitem{huang2020consistent}
Huang, R., Ren, J., Wonka, P., Ovsjanikov, M.: Consistent zoomout: Efficient spectral map synchronization. In: Computer Graphics Forum. pp. 265--278. Wiley Online Library (2020)

\bibitem{predator}
Huang, S., Gojcic, Z., Usvyatsov, M., Andreas~Wieser, K.S.: Predator: Registration of 3d point clouds with low overlap. In: IEEE Conference on Computer Vision and Pattern Recognition, CVPR (2021)

\bibitem{jiang2023neural}
Jiang, P., Sun, M., Huang, R.: Neural intrinsic embedding for non-rigid point cloud matching. In: Proceedings of the IEEE/CVF Conference on Computer Vision and Pattern Recognition. pp. 21835--21845 (2023)

\bibitem{jiang2023nonrigid}
Jiang, P., Sun, M., Huang, R.: Non-rigid shape registration via deep functional maps prior. In: Oh, A., Naumann, T., Globerson, A., Saenko, K., Hardt, M., Levine, S. (eds.) Advances in Neural Information Processing Systems. vol.~36, pp. 58409--58427. Curran Associates, Inc. (2023)

\bibitem{kim2021deep}
Kim, H., Kim, J., Kam, J., Park, J., Lee, S.: Deep virtual markers for articulated 3d shapes. In: Proceedings of the IEEE/CVF International Conference on Computer Vision. pp. 11615--11625 (2021)

\bibitem{kim2011blended}
Kim, V.G., Lipman, Y., Funkhouser, T.: Blended intrinsic maps. ACM transactions on graphics (TOG)  \textbf{30}(4),  1--12 (2011)

\bibitem{kingma2014adam}
Kingma, D.P., Ba, J.: Adam: A method for stochastic optimization. arXiv preprint arXiv:1412.6980  (2014)

\bibitem{li2008global}
Li, H., Sumner, R.W., Pauly, M.: Global correspondence optimization for non-rigid registration of depth scans. In: Computer graphics forum. pp. 1421--1430. Wiley Online Library (2008)

\bibitem{Li_2022_CVPR}
Li, J., Zhang, J., Wang, Z., Shen, S., Wen, C., Ma, Y., Xu, L., Yu, J., Wang, C.: Lidarcap: Long-range marker-less 3d human motion capture with lidar point clouds. In: Proceedings of the IEEE/CVF Conference on Computer Vision and Pattern Recognition (CVPR). pp. 20502--20512 (2022)

\bibitem{flame}
Li, T., Bolkart, T., Black, M.J., Li, H., Romero, J.: Learning a model of facial shape and expression from {4D} scans. ACM Transactions on Graphics, (Proc. SIGGRAPH Asia)  \textbf{36}(6),  194:1--194:17 (2017)

\bibitem{liu2004segmentation}
Liu, R., Zhang, H.: Segmentation of 3d meshes through spectral clustering. In: 12th Pacific Conference on Computer Graphics and Applications, 2004. PG 2004. Proceedings. pp. 298--305. IEEE (2004)

\bibitem{loper2015smpl}
Loper, M., Mahmood, N., Romero, J., Pons-Moll, G., Black, M.J.: Smpl: A skinned multi-person linear model. ACM transactions on graphics (TOG)  \textbf{34}(6),  1--16 (2015)

\bibitem{Lu_2019_ICCV}
Lu, W., Wan, G., Zhou, Y., Fu, X., Yuan, P., Song, S.: Deepvcp: An end-to-end deep neural network for point cloud registration. In: Proceedings of the IEEE/CVF International Conference on Computer Vision (ICCV) (2019)

\bibitem{luma}
Luma ai. \url{https://lumalabs.ai/}, accessed: 2024-07-10

\bibitem{ma2020learning}
Ma, Q., Yang, J., Ranjan, A., Pujades, S., Pons-Moll, G., Tang, S., Black, M.J.: Learning to dress 3d people in generative clothing. In: Proceedings of the IEEE/CVF Conference on Computer Vision and Pattern Recognition. pp. 6469--6478 (2020)

\bibitem{maggioli2024rematching}
Maggioli, F., Baieri, D., Rodol{\`a}, E., Melzi, S.: Rematching: Low-resolution representations for scalable shape correspondence. In: European Conference on Computer Vision. Springer (2024)

\bibitem{mahmood2019amass}
Mahmood, N., Ghorbani, N., Troje, N.F., Pons-Moll, G., Black, M.J.: Amass: Archive of motion capture as surface shapes. In: Proceedings of the IEEE/CVF international conference on computer vision. pp. 5442--5451 (2019)

\bibitem{marin2023smoothness}
Marin, R., Attaiki, S., Melzi, S., Rodol{\`a}, E., Ovsjanikov, M.: Smoothness and effective regularizations in learned embeddings for shape matching. arXiv  (2023)

\bibitem{marin2019high}
Marin, R., Melzi, S., Rodol{\`a}, E., Castellani, U.: High-resolution augmentation for automatic template-based matching of human models. In: 2019 International Conference on 3D Vision (3DV). pp. 230--239. IEEE (2019)

\bibitem{marin2020farm}
Marin, R., Melzi, S., Rodola, E., Castellani, U.: Farm: Functional automatic registration method for 3d human bodies. In: Computer Graphics Forum. pp. 160--173. Wiley Online Library (2020)

\bibitem{marin2020correspondence}
Marin, R., Rakotosaona, M.J., Melzi, S., Ovsjanikov, M.: Correspondence learning via linearly-invariant embedding. Advances in Neural Information Processing Systems  \textbf{33},  1608--1620 (2020)

\bibitem{melzi2020intrinsic}
Melzi, S., Marin, R., Musoni, P., Bardon, F., Tarini, M., Castellani, U.: Intrinsic/extrinsic embedding for functional remeshing of 3d shapes. Computers \& Graphics  \textbf{88},  1--12 (2020)

\bibitem{melzi2019shrec}
Melzi, S., Marin, R., Rodola, E., Castellani, U., Ren, J., Poulenard, A., Wonka, P., Ovsjanikov, M.: Shrec 2019: Matching humans with different connectivity. In: Eurographics Workshop on 3D Object Retrieval. vol.~7, p.~3. The Eurographics Association (2019)

\bibitem{melzi2019zoomout}
Melzi, S., Ren, J., Rodol\`{a}, E., Sharma, A., Wonka, P., Ovsjanikov, M.: Zoomout: spectral upsampling for efficient shape correspondence. ACM Trans. Graph.  \textbf{38}(6) (2019)

\bibitem{men2011color}
Men, H., Gebre, B., Pochiraju, K.: Color point cloud registration with 4d icp algorithm. In: 2011 IEEE International Conference on Robotics and Automation. pp. 1511--1516. IEEE (2011)

\bibitem{mihajlovic2022coap}
Mihajlovic, M., Saito, S., Bansal, A., Zollhoefer, M., Tang, S.: Coap: Compositional articulated occupancy of people. In: Proceedings of the IEEE/CVF Conference on Computer Vision and Pattern Recognition. pp. 13201--13210 (2022)

\bibitem{mihajlovic2021leap}
Mihajlovic, M., Zhang, Y., Black, M.J., Tang, S.: Leap: Learning articulated occupancy of people. In: Proceedings of the IEEE/CVF Conference on Computer Vision and Pattern Recognition. pp. 10461--10471 (2021)

\bibitem{nerf}
Mildenhall, B., Srinivasan, P.P., Tancik, M., Barron, J.T., Ramamoorthi, R., Ng, R.: Nerf: Representing scenes as neural radiance fields for view synthesis. ECCV  (2020)

\bibitem{musoni2021functional}
Musoni, P., Marin, R., Melzi, S., Castellani, U.: A functional skeleton transfer. Proceedings of the ACM on Computer Graphics and Interactive Techniques  \textbf{4}(3),  1--15 (2021)

\bibitem{myronenko2010point}
Myronenko, A., Song, X.: Point set registration: Coherent point drift. IEEE transactions on pattern analysis and machine intelligence  \textbf{32}(12),  2262--2275 (2010)

\bibitem{nair2010rectified}
Nair, V., Hinton, G.E.: Rectified linear units improve restricted boltzmann machines. In: Proceedings of the 27th international conference on machine learning (ICML-10). pp. 807--814 (2010)

\bibitem{nogneng2017informative}
Nogneng, D., Ovsjanikov, M.: Informative descriptor preservation via commutativity for shape matching. In: Computer Graphics Forum. pp. 259--267. Wiley Online Library (2017)

\bibitem{ovsjanikov2012functional}
Ovsjanikov, M., Ben-Chen, M., Solomon, J., Butscher, A., Guibas, L.: Functional maps: a flexible representation of maps between shapes. ACM Transactions on Graphics (ToG)  \textbf{31}(4),  1--11 (2012)

\bibitem{SMPL-X:2019}
Pavlakos, G., Choutas, V., Ghorbani, N., Bolkart, T., Osman, A.A.A., Tzionas, D., Black, M.J.: Expressive body capture: 3d hands, face, and body from a single image. In: Proceedings IEEE Conf. on Computer Vision and Pattern Recognition (CVPR) (2019)

\bibitem{ren2018continuous}
Ren, J., Poulenard, A., Wonka, P., Ovsjanikov, M.: Continuous and orientation-preserving correspondences via functional maps. ACM Transactions on Graphics (ToG)  \textbf{37}(6),  1--16 (2018)

\bibitem{renderpeople}
renderpeople. \url{https://renderpeople.com/}, accessed: 2024-07-10

\bibitem{rodola2017partial}
Rodol{\`a}, E., Cosmo, L., Bronstein, M.M., Torsello, A., Cremers, D.: Partial functional correspondence. In: Computer graphics forum. pp. 222--236. Wiley Online Library (2017)

\bibitem{romero2022embodied}
Romero, J., Tzionas, D., Black, M.J.: Embodied hands: Modeling and capturing hands and bodies together. ACM Transactions on Graphics, (Proc. SIGGRAPH Asia)  \textbf{36}(6) (2017)

\bibitem{smplh}
Romero, J., Tzionas, D., Black, M.J.: Embodied hands: Modeling and capturing hands and bodies together. ToG  (2017)

\bibitem{salti2014shot}
Salti, S., Tombari, F., Di~Stefano, L.: Shot: Unique signatures of histograms for surface and texture description. Computer Vision and Image Understanding  \textbf{125},  251--264 (2014)

\bibitem{scantheworld}
Scan the world project. \url{https://www.myminifactory.com/scantheworld/full-collection}, accessed: 2024-07-10

\bibitem{seylan20193d}
Seylan, {\c{C}}., Sahillio{\u{g}}lu, Y.: 3d skeleton transfer for meshes and clouds. Graphical Models  \textbf{105},  101041 (2019)

\bibitem{Sharp2020Robust}
Sharp, N., Crane, K.: {A Laplacian for Nonmanifold Triangle Meshes}. Computer Graphics Forum (SGP)  \textbf{39}(5) (2020)

\bibitem{sorkine2007rigid}
Sorkine, O., Alexa, M.: As-rigid-as-possible surface modeling. In: Symposium on Geometry processing. vol.~4, pp. 109--116 (2007)

\bibitem{sun2009concise}
Sun, J., Ovsjanikov, M., Guibas, L.: A concise and provably informative multi-scale signature based on heat diffusion. In: Computer graphics forum. pp. 1383--1392. Wiley Online Library (2009)

\bibitem{sundararaman2022implicit}
Sundararaman, R., Pai, G., Ovsjanikov, M.: Implicit field supervision for robust non-rigid shape matching. In: European Conference on Computer Vision. pp. 344--362. Springer (2022)

\bibitem{sundararaman2022reduced}
Sundararaman, R.S., Marin, R., Rodol\`{a}, E., Ovsjanikov, M.: Reduced representation of deformation fields for effective non-rigid shape matching. In: Advances in Neural Information Processing Systems. vol.~35, pp. 10405--10420. Curran Associates, Inc. (2022)

\bibitem{trappolini2021shape}
Trappolini, G., Cosmo, L., Moschella, L., Marin, R., Melzi, S., Rodol{\`a}, E.: Shape registration in the time of transformers. Advances in Neural Information Processing Systems  \textbf{34},  5731--5744 (2021)

\bibitem{Twindom}
Twindom dataset. \url{https://web.twindom.com/}, accessed: 2024-07-10

\bibitem{wang2019non}
Wang, L., Chen, J., Li, X., Fang, Y.: Non-rigid point set registration networks. arXiv preprint arXiv:1904.01428  (2019)

\bibitem{wang2021locally}
Wang, S., Geiger, A., Tang, S.: Locally aware piecewise transformation fields for 3d human mesh registration. In: Proceedings of the IEEE/CVF Conference on Computer Vision and Pattern Recognition. pp. 7639--7648 (2021)

\bibitem{wang2019deep}
Wang, Y., Solomon, J.M.: Deep closest point: Learning representations for point cloud registration. In: Proceedings of the IEEE/CVF international conference on computer vision. pp. 3523--3532 (2019)

\bibitem{xie2022neural}
Xie, Y., Takikawa, T., Saito, S., Litany, O., Yan, S., Khan, N., Tombari, F., Tompkin, J., Sitzmann, V., Sridhar, S.: Neural fields in visual computing and beyond. In: Computer Graphics Forum. vol.~41, pp. 641--676. Wiley Online Library (2022)

\bibitem{yang2018dmat}
Yang, B., Yao, J., Guo, X.: Dmat: Deformable medial axis transform for animated mesh approximation. In: Computer Graphics Forum. vol.~37, pp. 301--311. Wiley Online Library (2018)

\bibitem{Yew_2020_CVPR}
Yew, Z.J., Lee, G.H.: Rpm-net: Robust point matching using learned features. In: Proceedings of the IEEE/CVF Conference on Computer Vision and Pattern Recognition (CVPR) (2020)

\bibitem{yu2023rotation}
Yu, H., Qin, Z., Hou, J., Saleh, M., Li, D., Busam, B., Ilic, S.: Rotation-invariant transformer for point cloud matching. In: Proceedings of the IEEE/CVF Conference on Computer Vision and Pattern Recognition. pp. 5384--5393 (2023)

\bibitem{zeng2020corrnet3d}
Zeng, Y., Qian, Y., Zhu, Z., Hou, J., Yuan, H., He, Y.: Corrnet3d: Unsupervised end-to-end learning of dense correspondence for 3d point clouds. In: {IEEE/CVF} Conference on Computer Vision and Pattern Recognition (CVPR) (2021)

\bibitem{zuffi2015stitched}
Zuffi, S., Black, M.J.: The stitched puppet: A graphical model of 3d human shape and pose. In: Proceedings of the IEEE Conference on Computer Vision and Pattern Recognition. pp. 3537--3546 (2015)

\bibitem{ZuffiCVPR2017}
Zuffi, S., Kanazawa, A., Jacobs, D., Black, M.J.: {3D} menagerie: Modeling the {3D} shape and pose of animals. In: IEEE Conference on Computer Vision and Pattern Recognition (CVPR). pp. 5524--5532. IEEE Computer Society (2017)

\bibitem{Zuffi:CVPR:2017}
Zuffi, S., Kanazawa, A., Jacobs, D.W., Black, M.J.: 3d menagerie: Modeling the 3d shape and pose of animals. In: Proceedings of the IEEE Conference on Computer Vision and Pattern Recognition (CVPR) (2017)

\end{thebibliography}

\title{SUPPLEMENTARY MATERIALS \\ NICP: Neural ICP \\ for 3D Human Registration at Scale} 

\titlerunning{Sup. Mat. for NICP: Neural ICP for 3D Human Registration at Scale}

\author{Riccardo Marin\inst{1,2}\orcidlink{0000-0003-2392-4612} \and
Enric Corona\inst{3} \and
Gerard Pons-Moll\inst{1,2,4}\orcidlink{0000-0001-5115-7794}}

\authorrunning{R.~Marin et al.}

\institute{University of Tübingen, Germany \and
Tübingen AI Center, Germany \and
Google Research \and
Max Planck Institute for Informatics, Saarland Informatics Campus, Germany 
\\
\email{\{riccardo.marin, gerard.pons-moll\}@uni-tuebingen.de, enriccorona@google.com}}

\maketitle
\noindent\begin{center}
    {\url{https://neural-icp.github.io/}}
\end{center}
\begin{abstract}
In this document, we provide further technical details of our method. We also provide an extensive collection of results from the data presented in the paper and a discussion about failure cases, pointing to interesting challenges for future works. We remark here that our method works only when considering the point cloud and does not consider the mesh of the target. We show the target mesh only for visualization purposes.

\end{abstract}


\section{Technical details}
\subsection{Implementation details}
All the trained NF and Uni- baselines use the same backbone network of \cite{corona2022learned}, which is composed of an IFNet \cite{chibane2020implicit} to extract global features of the target shape, and an MLP network to query $\mathbb{R}^3$ points and output offsets (or the universal embedding, in case of Uni- baselines).

\mypar{IFNet Network.} We start from a $64 \times 64 \times 64$ voxelization of the point cloud. Each voxel is represented as the distance from its center to the nearest point of the input. We also compute the discrete gradient of the functions along the three coordinates, and we apply this information to the input. Then, we use twelve 3D Convolutional layers with the following numbers of filters: $(64,64,128,128,192,192,256,256,256,256,256,256)$. We assign to each point a set of local and global features considering the output of intermediate layers (one every two) for a total of $1152$ features. 

\mypar{MLPs Query Network.} This network takes the coordinates of a point with the global features extracted by IFNet and predicts all the displacements towards $690$ locations, which are the ground truth registered template vertices. In the proposed LoVD, we divided this network into $16$ MLPs, each one composed of $6$ layers of dimensions $(256, 512, 512, 512, 512, n) $ where $n$ is the number of template vertices belonging to that local part. The local region is obtained by performing spectral clustering on the template described in the next paragraph. The activation function for the first five layers is the ReLu \cite{nair2010rectified}.


\mypar{Training.} The backbone network has in total $36776428$ parameters, and it is trained end-to-end. We use a batch size of $8$, a learning rate of $1e-4$, and an Adam optimizer \cite{kingma2014adam}. During training, for each training sample, we query $2200$ points: $400$ uniformly sampled in $\mathbb{R}^3$, and $1800$ near the input point cloud, by perturbing the points with random Gaussian noise of standard-deviation $0.05$. Hence, the network predicts for each of the $2200$ points the offsets towards the ground truth template vertices positions. Before computing the loss, we rescale the offsets with a maximum norm of $0.05$, so the network learns to converge in small steps.

\subsection{\pipeline{} Details}

\mypar{\method{} Iterations} For each iteration of our procedure, we consider the vertices of the target point cloud, pass them toward the query network, compute the loss described in Equation $5$ of the paper, and backpropagate it to update the weights of the backbone network. This is done for $20$ steps using a learning rate of $1e-5$ and an Adam optimizer~\cite{kingma2014adam}. If the input point cloud is particularly dense, we sample $20000$ random points at each iteration.

\mypar{Neural Fields Evaluation.} To evaluate the NF, we initialize $690$ points at $(0,0,0)$ coordinates, and we update their positions $50$ times. For the OneShot baseline, we initialize $690$ points near the target surface and update their position once. After this, we fit a full $6890$ SMPL model to the predicted points, optimizing its parameters for $2000$ steps and a learning rate of $1e-1$, using an Adam optimizer \cite{kingma2014adam}. The loss is a standard L1, plus the statistical pose prior to \cite{SMPL-X:2019} (weighted for $1e-8$) and an L2 regularization on the shape parameters magnitude (weighted for $1e-2$).

\mypar{Chamfer Refinement.} The Chamfer refinement optimizes the SMPL parameters using the bidirectional Chamfer Distance between the obtained SMPL and the target point cloud. We perform $500$ iterations, with a learning rate of $2e-2$ using an Adam optimizer \cite{kingma2014adam}. We generally considered the bidirectional Chamfer loss when the input point cloud is a complete human model. Otherwise, we optimize only for one direction. We keep the regularizers' weights similar to the previous step.

\mypar{SMPL+D.} If the shape contains details that the SMPL model cannot express, we compute the displacements $\mathbf{O} \in \mathbb{R}^{6890 \times 3 }$ to fit the target point cloud. To do so, for each of the SMPL vertices, we optimize the bidirectional Chamfer distance, plus an L2 regularization on the offsets magnitudes and a Laplacian regularization as designed in \cite{gao2020learning}:

\begin{equation}
L_{lap} = \frac{1}{6890} \sum\limits_{i=1}^{6890} (\Delta (\mathbf{v}_i + \mathbf{O}_i) - \Delta \mathbf{v}_i)^2,
\end{equation}

Where $\mathbf{v}_i$ the $i$-th vertex the SMPL template resulting from the previous step, and $\Delta$ the Laplace-Beltrami Operator (in our case, obtained using the Robust Laplacian Python library~\cite{Sharp2020Robust}). This loss promotes the offsets to preserve the smoothness of the input surface. We rely on an Adam optimizer \cite{kingma2014adam}.

\section{Further Validations and Ablations}
\subsection{Number of Segments.}
\begin{wraptable}[12]{r}{0.4\linewidth}
\vspace{-1.2cm}
 \begin{overpic}[trim=5cm 4cm 4cm 3cm,clip,
 width=\linewidth]{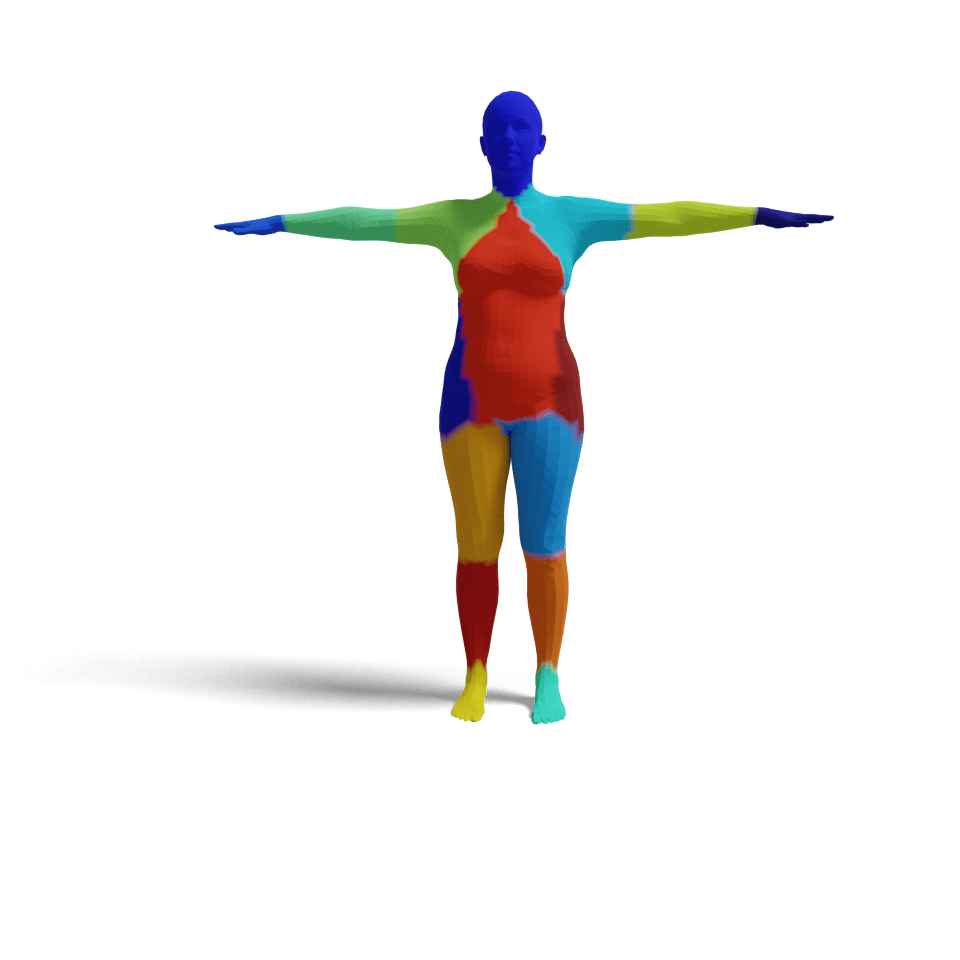}
 \end{overpic}
 \vspace{-1.3cm}
 \captionof{figure}{\small{\label{fig:segms}The $16$ segments considered by our method.}}
 \end{wraptable}
 We trained our backbone network considering different number of segments: $1$ (standard LVD), $10$, $16$, and $24$. In all cases, we resized the query MLPs such that they have a comparable number of parameters. Table~\ref{tab:abl_segments} reports evaluation results. The localization produces a significant gain, while its excess causes performance degradation. Given that using $10$ or $16$ segments performs similarly, we opted for $16$ since it performs better on real scans. We visualize the $16$ segments in Figure~\ref{fig:segms}.

\subsection{\pipeline{} Ablation.}
\begin{figure}
\centering
\noindent
\resizebox{0.8\linewidth}{0.6\linewidth}{

\definecolor{crimson2143940}{RGB}{214,39,40}
\definecolor{darkgray176}{RGB}{176,176,176}
\definecolor{darkorange25512714}{RGB}{255,127,14}
\definecolor{forestgreen4416044}{RGB}{44,160,44}
\definecolor{lightgray204}{RGB}{204,204,204}
\definecolor{steelblue31119180}{RGB}{31,119,180}

\begin{tikzpicture}
\begin{axis}[
legend cell align={left},
legend style={
  fill opacity=0.8,
  draw opacity=1,
  text opacity=1,
  at={(0.97,0.03)},
  anchor=south east,
  draw=lightgray204
},
tick align=outside,
tick pos=left,
xtick={0, 0.025,0.075, 0.1},
xticklabels={0, 0.025,0.075, 0.1},
xmajorgrids,
ymajorgrids,
every x tick label/.append style={font=\color{black}, font=\footnotesize},
every y tick label/.append style={font=\color{black}, font=\footnotesize},
ylabel={ \% Correspondences},
xlabel={ Euclidean Error},
x grid style={darkgray176},
xmin=0, xmax=0.1,
xtick style={color=black},
y grid style={darkgray176},
ymin=0, ymax=100,
ytick style={color=black}
]
\addplot [semithick, steelblue31119180,line width=2pt]
table {%
0 9.60725689404935
0.01 12.7285921625544
0.02 27.4772133526851
0.03 47.3014513788099
0.04 62.6059506531205
0.05 73.4236574746009
0.06 81.1476052249637
0.07 86.6098693759071
0.08 90.5492017416546
0.09 93.2097242380261
0.1 94.9085631349782
0.11 95.9256894049347
0.12 96.5544267053701
0.13 96.9629898403483
0.14 97.2253991291727
0.15 97.3673439767779
0.16 97.4377358490566
0.17 97.4923076923077
0.18 97.5542815674891
0.19 97.6174165457184
0.2 97.6960812772134
0.21 97.7642960812772
0.22 97.8258345428157
0.23 97.8805515239477
0.24 97.9404934687954
0.25 98.0087082728592
0.26 98.0873730043541
0.27 98.1635703918723
0.28 98.2391872278665
0.29 98.2976777939042
0.3 98.3387518142235
0.31 98.3732946298984
0.32 98.4046444121916
0.33 98.4451378809869
0.34 98.4766328011611
0.35 98.511320754717
0.36 98.555732946299
0.37 98.6017416545718
0.38 98.6407837445573
0.39 98.6783744557329
0.4 98.7074020319303
0.41 98.733381712627
0.42 98.7580551523948
0.43 98.7808417997097
0.44 98.8095791001451
0.45 98.8429608127721
0.46 98.8851959361393
0.47 98.9179970972424
0.48 98.9464441219158
0.49 98.9696661828737
};
\addlegendentry{LVD}
\addplot [semithick, darkorange25512714,line width=2pt]
table {%
0 11.6375907111756
0.01 15.2058055152395
0.02 31.755297532656
0.03 53.3939042089985
0.04 68.7629898403483
0.05 78.2635703918723
0.06 84.8195936139332
0.07 89.4622641509434
0.08 92.4403483309144
0.09 94.4988388969521
0.1 95.7791001451379
0.11 96.6005805515239
0.12 97.1306240928882
0.13 97.4930333817126
0.14 97.7023222060958
0.15 97.8095791001451
0.16 97.8650217706821
0.17 97.910885341074
0.18 97.9583454281567
0.19 98.011611030479
0.2 98.077648766328
0.21 98.1300435413643
0.22 98.1718432510885
0.23 98.2159651669086
0.24 98.2619738751814
0.25 98.3050798258345
0.26 98.3519593613933
0.27 98.3943396226415
0.28 98.4320754716981
0.29 98.4586357039187
0.3 98.4917271407837
0.31 98.5306240928882
0.32 98.555587808418
0.33 98.5804063860668
0.34 98.599564586357
0.35 98.6314949201742
0.36 98.6628447024673
0.37 98.6984034833091
0.38 98.7203193033382
0.39 98.7384615384615
0.4 98.7470246734398
0.41 98.755297532656
0.42 98.767053701016
0.43 98.7809869375907
0.44 98.7992743105951
0.45 98.8145137880987
0.46 98.8383164005806
0.47 98.8628447024673
0.48 98.8862119013062
0.49 98.9197387518142
};
\addlegendentry{LoVD}
\addplot [semithick, forestgreen4416044,line width=2pt]
table {%
0 20.9346879535559
0.01 30.6265602322206
0.02 52.4907111756168
0.03 72.6756168359942
0.04 84.3056603773585
0.05 90.4298984034833
0.06 93.9820029027576
0.07 95.9943396226415
0.08 97.0516690856313
0.09 97.6397677793904
0.1 97.9705370101596
0.11 98.1631349782293
0.12 98.2747460087083
0.13 98.344412191582
0.14 98.3970972423803
0.15 98.4319303338171
0.16 98.4626995645864
0.17 98.4966618287373
0.18 98.533381712627
0.19 98.5582002902758
0.2 98.5857764876633
0.21 98.6078374455733
0.22 98.6390420899855
0.23 98.6706821480406
0.24 98.6966618287373
0.25 98.7268505079826
0.26 98.7497822931785
0.27 98.7685050798258
0.28 98.789114658926
0.29 98.811320754717
0.3 98.8287373004354
0.31 98.8483309143686
0.32 98.8634252539913
0.33 98.8751814223512
0.34 98.8880986937591
0.35 98.9018867924528
0.36 98.9174165457184
0.37 98.9365747460087
0.38 98.955297532656
0.39 98.966618287373
0.4 98.9817126269956
0.41 99.0039187227867
0.42 99.0256894049347
0.43 99.0410740203193
0.44 99.0548621190131
0.45 99.0696661828737
0.46 99.0863570391872
0.47 99.1026124818578
0.48 99.1174165457184
0.49 99.132946298984
};
\addlegendentry{LoVD+\method{}}
\addplot [semithick, crimson2143940,line width=2pt]
table {%
0 42.688679245283
0.01 58.4509433962264
0.02 74.0746008708273
0.03 86.9944847605225
0.04 92.9744557329463
0.05 95.7262699564586
0.06 97.1256894049347
0.07 97.8075471698113
0.08 98.200435413643
0.09 98.4818577648766
0.1 98.6978229317852
0.11 98.7979680696662
0.12 98.8657474600871
0.13 98.9193033381713
0.14 98.9637155297533
0.15 98.9812772133527
0.16 99.0104499274311
0.17 99.033091436865
0.18 99.0523947750363
0.19 99.0753265602322
0.2 99.0902757619739
0.21 99.1005805515239
0.22 99.1194484760523
0.23 99.1423802612482
0.24 99.1595065312046
0.25 99.1767779390421
0.26 99.1959361393324
0.27 99.222206095791
0.28 99.2464441219158
0.29 99.2625544267054
0.3 99.2746008708273
0.31 99.2856313497823
0.32 99.2943396226415
0.33 99.3049346879536
0.34 99.3152394775036
0.35 99.322641509434
0.36 99.3362844702467
0.37 99.3455732946299
0.38 99.3535558780842
0.39 99.3619738751814
0.4 99.366908563135
0.41 99.3711175616836
0.42 99.3760522496372
0.43 99.3809869375907
0.44 99.3847605224964
0.45 99.3873730043541
0.46 99.3920174165457
0.47 99.3978229317852
0.48 99.4044992743106
0.49 99.4103047895501
};
\addlegendentry{\pipeline{}}
\end{axis}
\end{tikzpicture}}
 \caption{\label{fig:curves} Error curves for ablation study of the pipeline components. We observe that our \method{} procedure produces a significant impact, doubling the number of exact matches and providing better correspondences.}
\end{figure}
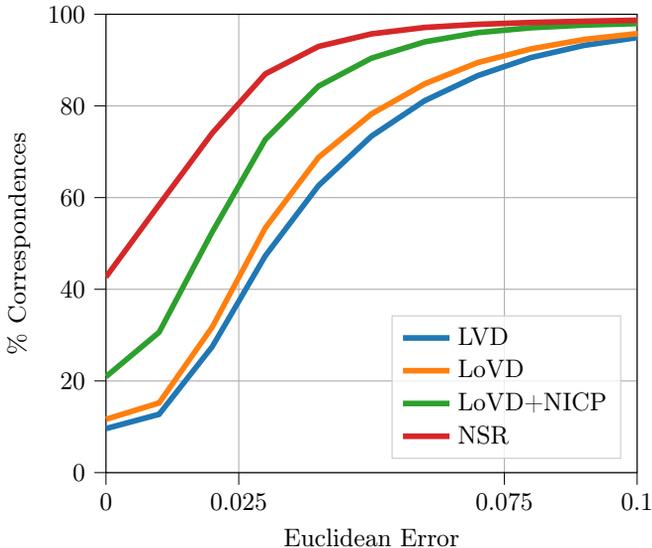

To validate the contribution of our registration method, we report in Table~\ref{tab:ablation} the quantitative results after enabling different components. We appreciate the significant improvement provided by \method{} procedure, upgrading LoVD prediction by a margin of 18\% on real scans. To highlight the nature of the contribution, we report in Figure~\ref{fig:curves} a visualization of the normalized error following the protocol of \cite{kim2011blended} (\textit{i.e.}, error on X-Axis, percentage of correspondences below that error on Y-Axis). We emphasize that \method{} increases the number of correct matches (doubling the points with $0$ error), and is more robust (the curve saturates faster). The similar saturation of curves for our approach before and after the refinement suggests that our method provides a good initialization for the convergence of Chamfer, which uses the geometry to align local features. All our contributions obtain strictly superior performance compared to the LVD baseline.

\begin{table}[!t]
\vspace{-4mm}
\footnotesize
\center
	\begin{tabular}{@{}lrr@{}}
		          & FAUST$_{R}$ & FAUST$_{S}$           \\
		\hline
  LVD         & 4.78          & 3.54 \\
  LoVD - 10 segments   & \textbf{4.27} & 3.13 \\
  LoVD - 16 segments   &   4.35        & \textbf{3.11} \\
  LoVD - 24 segments   & 7.92          & 5.58 \\
        
	\end{tabular}
 \caption{\label{tab:abl_segments}Ablation study on the number of local MLPs. Increasing the number of components helps, but an excess is also detrimental. Using $10$ and $16$ segments lead to a good tradeoff, and our full model relies on the second, which performs better on real data.}
\vspace{-1mm}
 \end{table}

\begin{table}[t!]
\footnotesize
\center
	\begin{tabular}{@{}lrr@{}}
	          & FAUST$_{R}$ & FAUST$_{S}$          \\
		\hline
  LVD               & 4.78  & 3.54 \\
  LoVD {\bf (Ours)}             & 4.35 & 3.11 \\
  LoVD+\method{}  {\bf (Ours)}         & 2.97 & 2.55 \\
  \pipeline{}  {\bf (Ours)}           & \textbf{1.76} & \textbf{1.85} \\
        
	\end{tabular}
 \caption{\label{tab:ablation}Ablation results for different elements of our pipeline. \method{} largely improves the backbone's results, promoting the refinement's better convergence.}
 \end{table}
\begin{figure*}[th!]
\vspace{2mm}
    \centering
    \footnotesize
 \begin{overpic}[trim=0cm 0cm 0cm 0cm,clip, width=\linewidth]{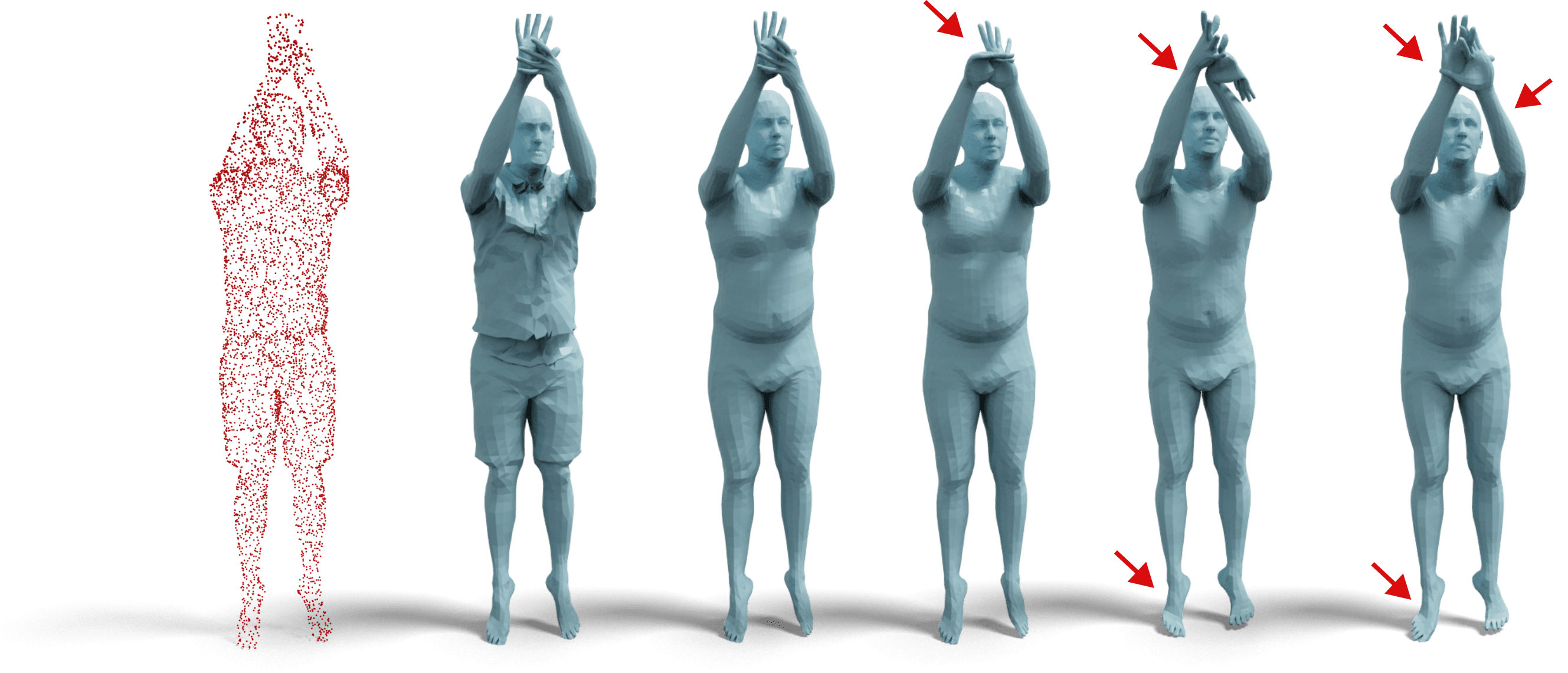}
 \footnotesize
 \put(16,43){Input}
  \put(32, 43){GT}
  \put(46.3,43){\textbf{\pipeline{}}}
  \put(60,43){LVD}
  \put(74,43){{PTF}}
  \put(90,43){{IP-Net}}
\end{overpic}
    \caption{\label{fig:cape2} Comparison between methods trained on the same CAPE training set. \pipeline{} is the only one to obtain the correct hand configuration.}
\end{figure*}

\subsection{\pipeline{} Comparison with SoTA}
In Figure~\ref{fig:cape2}, we emphasize the detailed accuracy of \pipeline{}. We report a further qualitative comparison of methods trained on the same CAPE data split. Despite the generally correct pose, PTF and IP-Net show limitations in feet and head orientations, highlighting a limited capacity to consider details. Also, \pipeline{} is the only method to recover hands perfectly, showing the capability to register even complex configurations. 

\begin{figure}[!t]
\centering
\scriptsize
 \begin{overpic}[trim=0cm 0cm 0cm 0cm,clip, width=\linewidth]{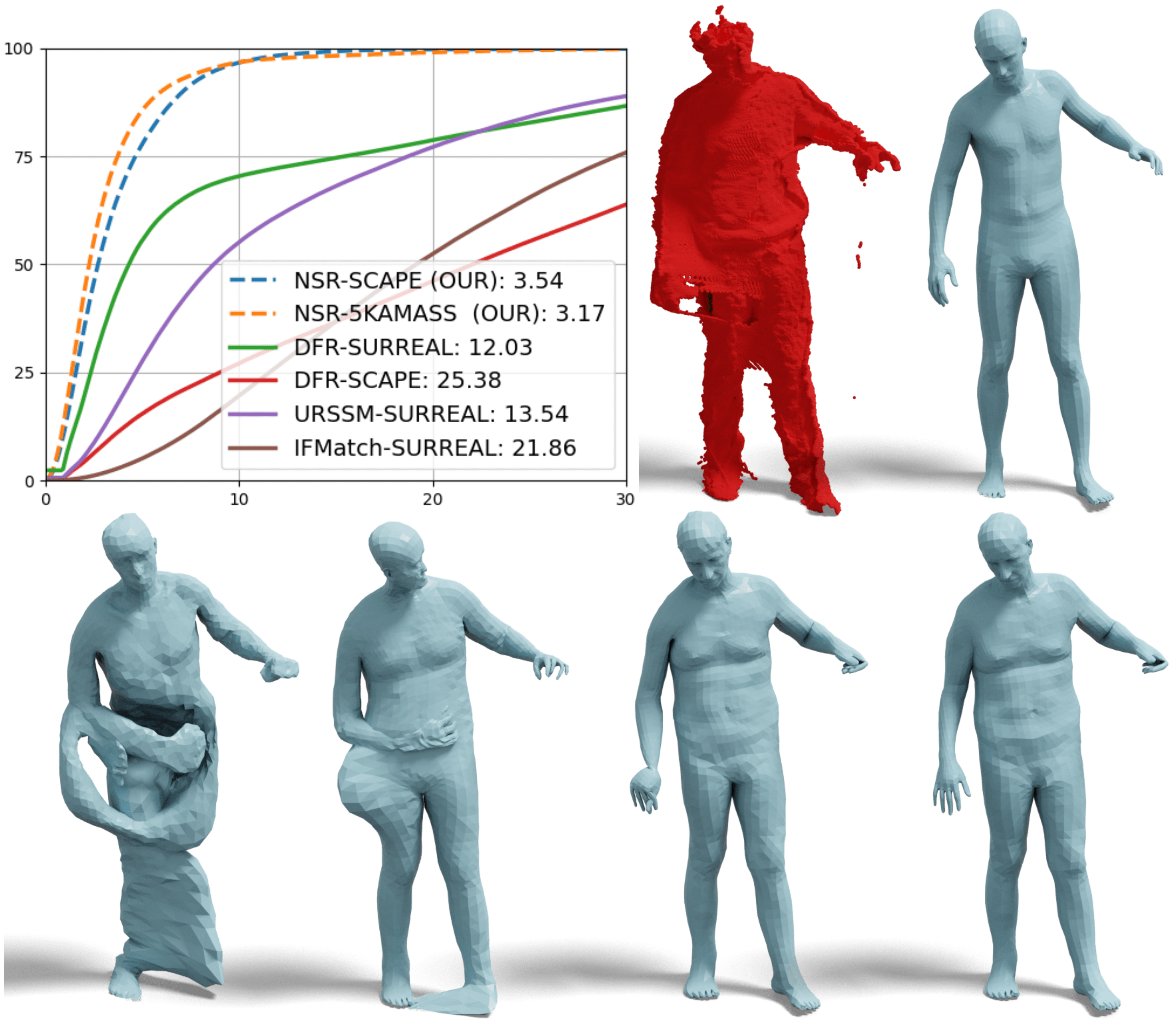}
    \put(19,85){Euclidean Error}
    \put(57,88){Input}
    \put(83,88){GT}

    \put(7,-1.5){DFR-SCAPE}
    \put(30,-1.5){DFR-SURREAL}
    \put(57,-1.5){NSR-SCAPE}
    \put(61,-4.3){\textbf{(OUR)}}
    \put(78,-1.5){NSR-5KAMASS}
    \put(84,-4.3){\textbf{(OUR)}}
	\end{overpic}
    \vspace{0.5cm}
\caption{\small \label{fig:matchsota} Comparison on BEHAVE against shape matching methods. NSR trained on SCAPE surpasses competitors trained on stronger data priors (SURREAL). }
\end{figure}

\mypar{Comparsion with Shape Matching methods.} Although we focus our comparison on SoTA 3D Human Registration pipelines, we also report results against general matching methods for completeness. We collect $20$ BEHAVE Kinect point clouds, obtaining $380$ pairs (similar to FAUST and SCAPE test sets) with ground-truth correspondence. We train our method on the $51$ training shapes of SCAPE (NSR-SCAPE) and $5$K shapes from AMASS (NSR-5KAMASS). We compare with the available checkpoints of \cite{jiang2023nonrigid} (DFR-SCAPE, DFR-SURREAL), \cite{cao2023self} (URSSM-SURREAL), and \cite{sundararaman2022implicit} (IFMatch-SURREAL). Figure~\ref{fig:matchsota} reports the error curves, the mean Euclidean error in the legend, and a qualitative example. Our method trained on little data (SCAPE) largely outperforms baselines trained on stronger priors (SURREAL). NSR handles full resolution point clouds ($\sim$380K points) and takes around 1 minute. Competitors require costly preprocessing; DFR and URSSM require subsampling ($\sim$5-7K points), storing the LBO eigendocomposition (x4 storage memory), and may converge slowly (DFR can exceed 10 minutes).

\subsection{\method{} Robustness and Generalization}
\begin{figure*}
\vspace{-5mm}
    \centering
    \footnotesize
 \begin{overpic}[trim=0cm 0cm 0cm 0cm,clip, width=\linewidth]{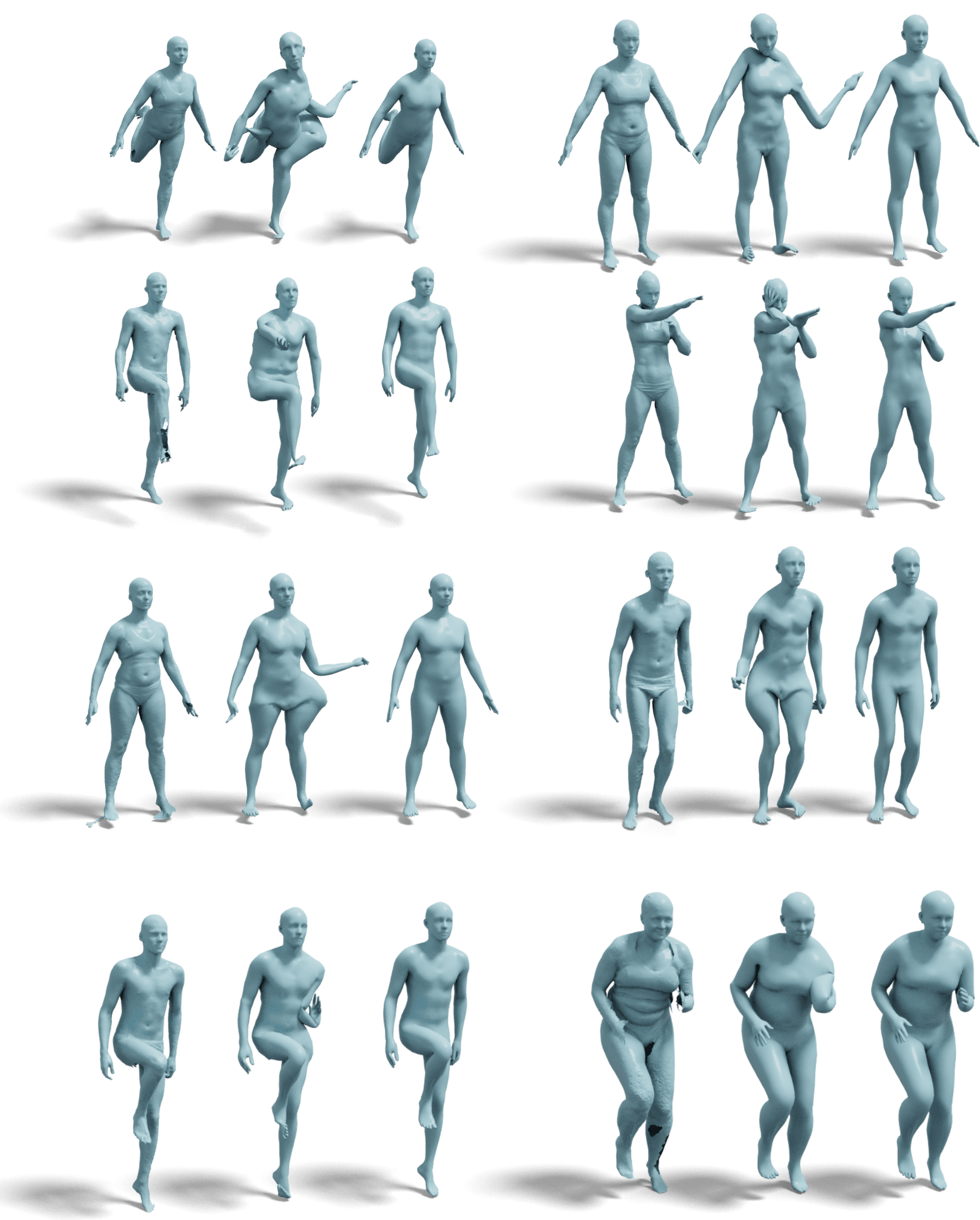}
		\put(12,99){Input}
        \put(22,101){\pipeline{}}
        \put(20,99){w/o \method{}}
        \put(33,99){\textbf{\pipeline{}}}

	\put(50,99){Input}
        \put(60,101){\pipeline{}}
        \put(58,99){w/o \method{}}
        \put(74,99){\textbf{\pipeline{}}}
	\end{overpic}
    \caption{\label{fig:dfaust} Comparison of results from the full pipeline without and with \method{}. DFAUST may contain unreferenced vertices, which significantly change the input implicit representation. \method{} helps to refine these results, bringing the deformation of the backbone towards the majority of points, i.e., the target human.}
\end{figure*}

\begin{table}[th!]
\footnotesize
\center
	\begin{tabular}{@{}lr@{}}
	DFAUST 	            &   v2v      \\ 
		\hline
 \pipeline{}  w/o \method{}                & 3.26 \\
  \pipeline{}                              & \textbf{3.08}         \\ 
        
	\end{tabular}
 \caption{\label{tab:dfaust}Evaluation on $417$ DFAUST shapes compared to the ground-truth registration. \method{} procedure leads to better final registration.}
\vspace{-1mm}
 \end{table}

\mypar{DFAUST.} To further stress our unsupervised \method{} procedure, we validate its impact on the real scans of DFAUST~\cite{bogo2017dynamic}. Such scans contain noise, missing geometry due to occlusion, and identities significantly far from the AMASS distribution~\cite{mahmood2019amass}. We select $11$ sequences of different subjects, and we select one frame every ten for a total of $417$ scans. We compare \pipeline{} with and without \method{} procedure. We evaluate our error using the distance from the provided ground truth registration. We report the results in Table~\ref{tab:dfaust}.
The results confirm that despite the strong initialization provided by the data-driven backbone and the robust SMPL refinement using Chamfer distance, our method provides a further improvement, obtaining a more precise registration. A qualitative comparison is reported in Figure~\ref{fig:dfaust}.

\begin{figure*}
\vspace{-5mm}
    \centering
    \footnotesize
 \begin{overpic}[trim=0cm 0cm 0cm 0cm,clip, width=\linewidth]{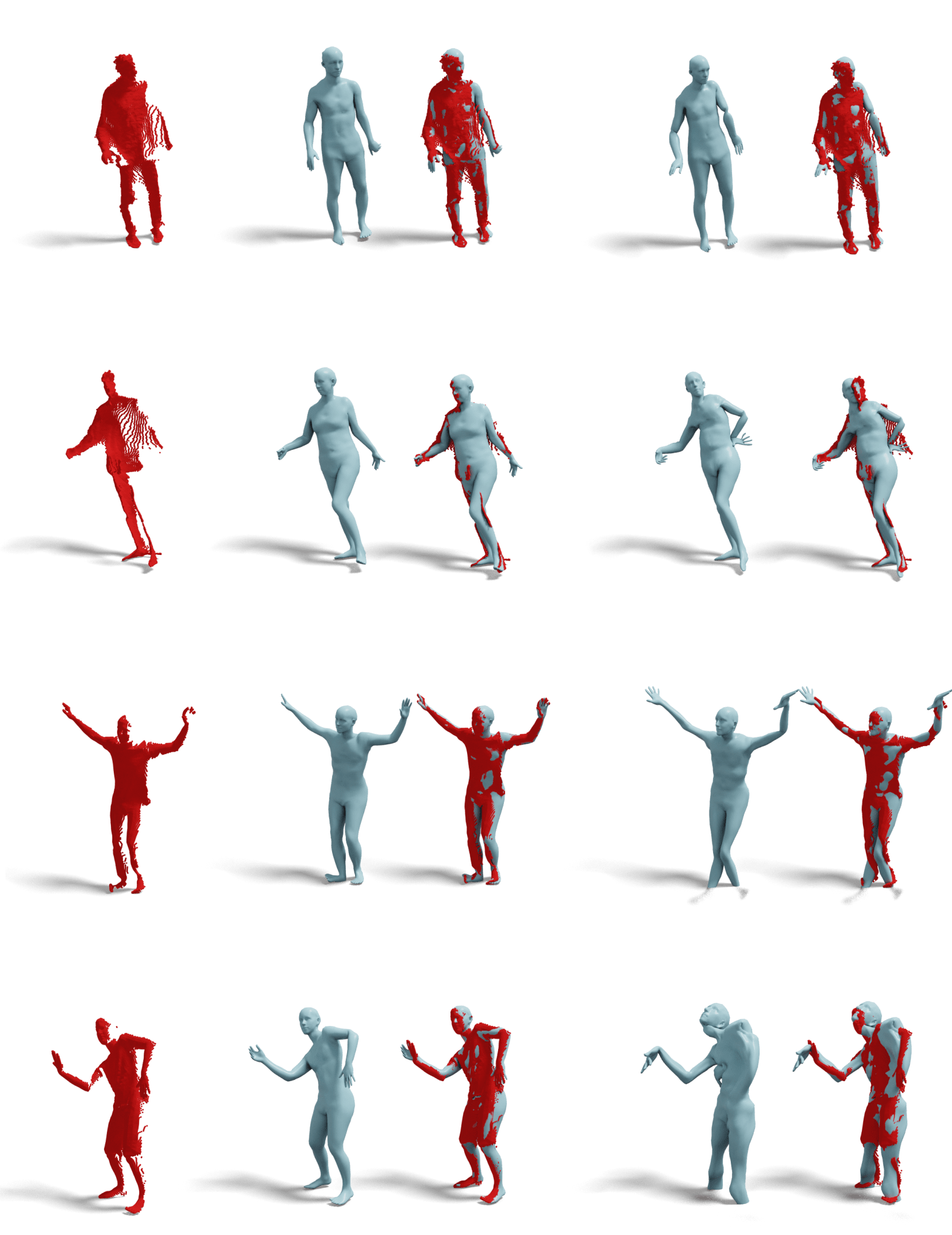}
		\put(8,99){Input}
        \put(56,99){\pipeline{}  w/o \method{}}
        \put(28,99){\textbf{\pipeline{}}}

	\end{overpic}
 \vspace{-0.4cm}
    \caption{\label{fig:pcfails} Comparison of results from \pipeline{} with and without \method{}. Partial point clouds are significantly far from the training distribution seen by the backbone network at training time. \method{} enables this challenging scenario, improving the geometry fitting.}
\end{figure*}

\mypar{Partial Point Clouds}
\method{} enables generalization to shapes with geometry significantly far from the training distribution. Here, we show qualitative results of \pipeline{} when the input is a partial point cloud. In Figure~\ref{fig:pcfails} we report some results of our method on point cloud coming from DSFN~\cite{burov2021dsfn} (first two rows) and CAPE~\cite{ma2020learning} (last two rows), with and without \method{} procedure. \method{} is crucial and lets the network recover the correct pose of the subject. Without it, the prediction of the backbone is not good enough to initialize the Chamfer refinement, leading to catastrophic failures.
We discuss failure cases for our method in the next Section.

\begin{wraptable}[11]{r}{0.45\linewidth}
 \begin{overpic}[trim=0cm 0cm 0cm 0cm,clip,
 width=\linewidth]{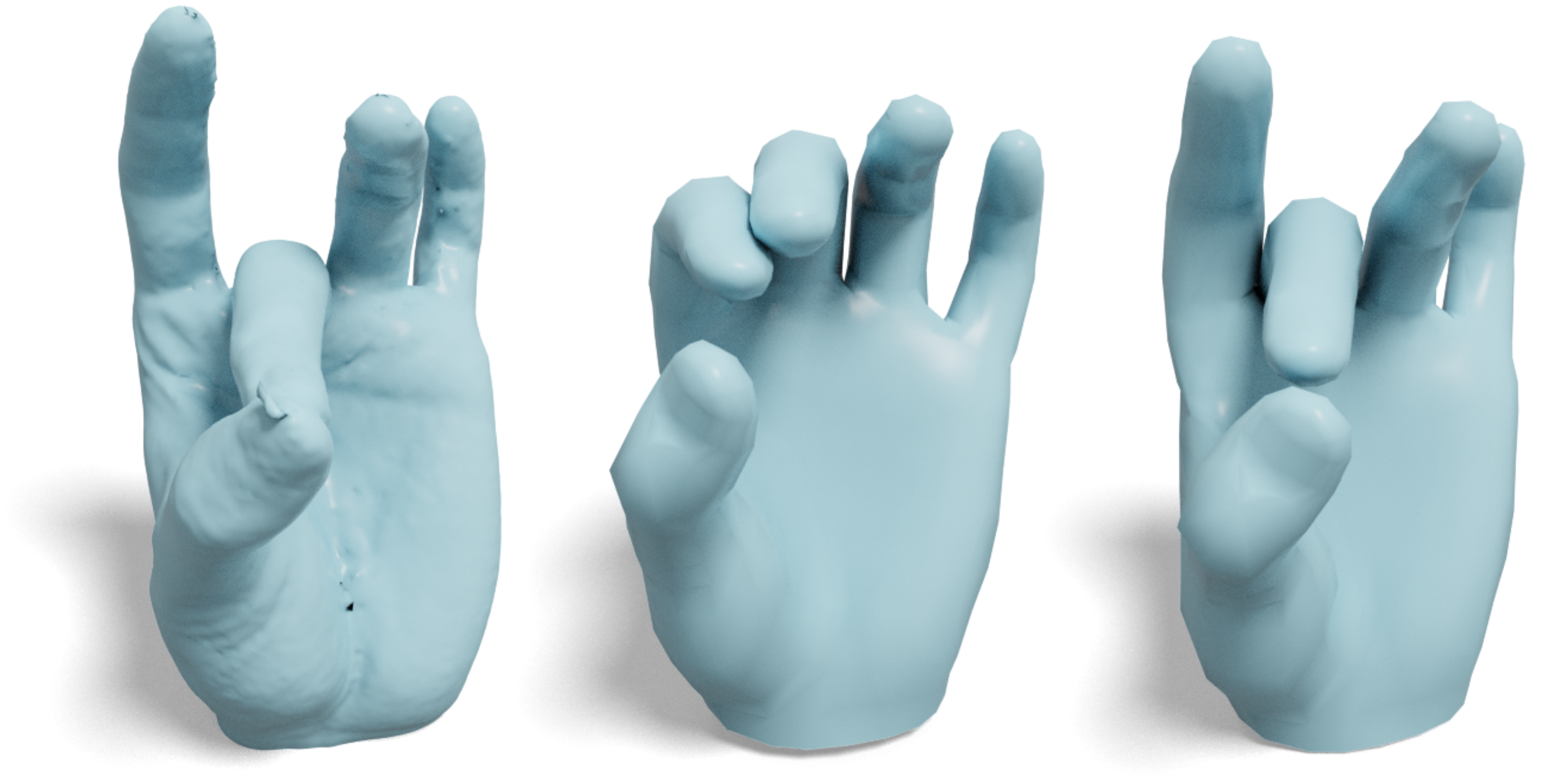}
 \put(13,50.5){Target}
 \put(46,51.5){LVD}
 \put(68,55){LVD+\method{}}
  \put(72,48.6){{\bf (Ours)}}
 \end{overpic}
 \captionof{figure}{\small{\label{fig:mano1}Qualitative results of applying \method{} on a NF trained on hands.}}
 \end{wraptable}
 
\begin{figure*}
    \centering
    \footnotesize
 \begin{overpic}[trim=0cm 0cm 0cm 0cm,clip, width=\linewidth]{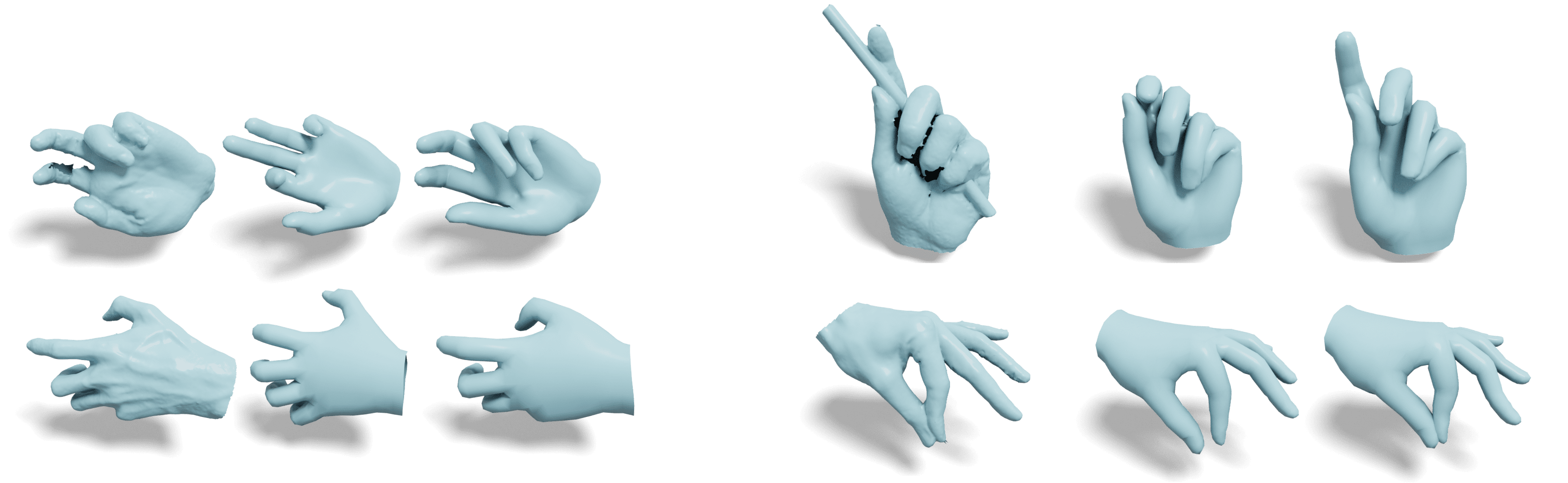}
		\put(7,28){Input}
            \put(18,28){LVD}
            \put(26,30){LVD+\method{}}
            \put(27.5,27){\bf (Ours)}

		\put(60,28){Input}
            \put(73,28){LVD}
            \put(86,30){LVD+\method{}}
            \put(87.5,27){\bf (Ours)}
	\end{overpic}
\caption{\label{fig:supmatmano}Further results on hand fitting.}
\end{figure*}

\mypar{Different Domain.} \method{} runs out of the box on top of any NF. We replicate the experiment on the hands' registration of \cite{corona2022learned}, using publicly released pre-trained weights and evaluating on the same protocol. We report the errors in Table \ref{fig:MANO}. \method{} largely improves both vertices and joint estimations. Qualitative results are reported in Figures \ref{fig:mano1} and \ref{fig:supmatmano}.
\begin{figure*}[!t]
\begin{tabular}{cc}
    \hspace{-0.4cm}
\begin{minipage}{\linewidth}
    \scriptsize
\begin{tabular}{l c c c c c c }
\cmidrule(lr){2-7}
& \multicolumn{2}{c}{MANO Error} & \multicolumn{2}{c}{Reconstruction to Scan} &  \multicolumn{2}{c}{Scan to Reconstruction} \\
\cmidrule(lr){2-3} \cmidrule(lr){4-5} \cmidrule(lr){6-7}
Method & Joint [cm]  & Vertex [cm]  & V2V [mm]  & V2S [mm]& V2V [mm] & V2S [mm] \\
\cmidrule(lr){1-1} \cmidrule(lr){2-7}
LVD        & 9.6 & 12.3 & 5.73 & 5.73 & 8.16 &  6.43 \\
LVD+\method{} {\bf (Ours)} & \textbf{7.9} & \textbf{10.7} & \textbf{4.73} & \textbf{4.70} & \textbf{6.37} & \textbf{4.15} \\
\cmidrule(lr){1-1} \cmidrule(lr){2-7}
\end{tabular}
\end{minipage}




\end{tabular}
	\captionof{table}{\label{fig:MANO} Hands registration. We apply \method{} directly to the pre-trained network released by \cite{corona2022learned}. \method{} works out of the box and improves all the metrics. On the right is a qualitative comparison.}
 \vspace{-0mm}
\end{figure*}

\begin{figure}
    \centering
 \begin{overpic}[trim=0cm 0cm 0cm 0cm,clip, width=\linewidth]{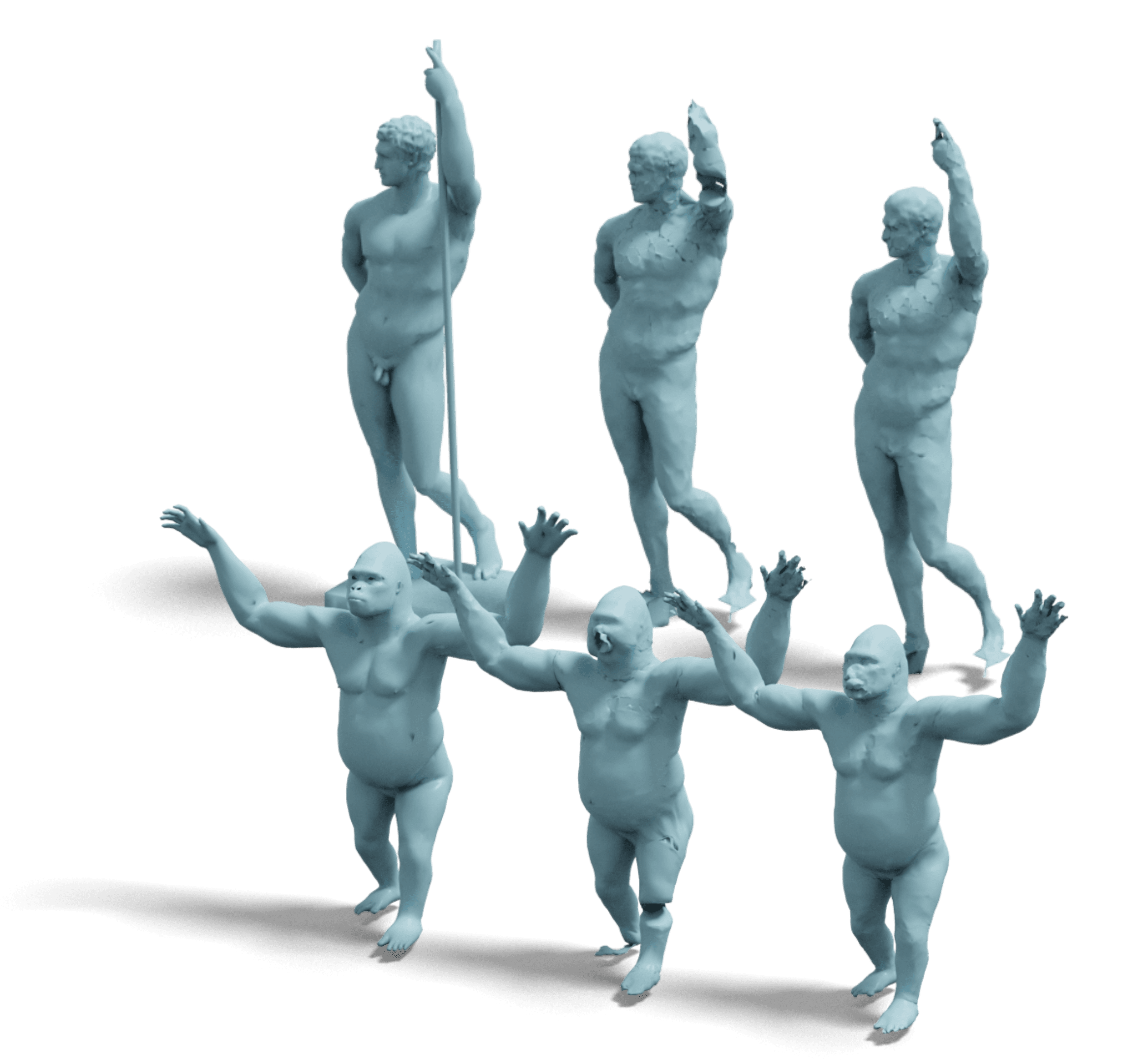}
	\put(30,91){Input}
        \put(53,88){\pipeline{} w/o \method{}}
        \put(80,85){\textbf{\pipeline{}}}
	\end{overpic}
    \caption{\label{fig:statues} Some results on non-humans models. \method{} opens to promising generalization results in the presence of highly non-isometries and heavy clutter. }
\end{figure}
 
\mypar{Non-Humans} 
Finally, we push the limits of \method{} in the presence of out-of-distribution proportions and clutter, showing promising results on animals \cite{TOSCA} or statues\cite{scantheworld} in Figure \ref{fig:statues}. These can still be represented by a human body template, but only when the template is significantly deformed. To emphasize the contribution of \method{} refinement, in the same Figure, we also report the result of our pipeline when the procedure is disabled in \pipeline{}. In these cases, the final output presents significant artifacts, due to a bad initialization of Chamfer refinement that leads to catastrophically local minima.

\section{Further Results}

\subsection{Failure Cases}\label{sec:failures}

\begin{figure*}

    \centering
    \footnotesize
 \begin{overpic}[trim=0cm 0cm 0cm 0cm,clip, width=\linewidth]{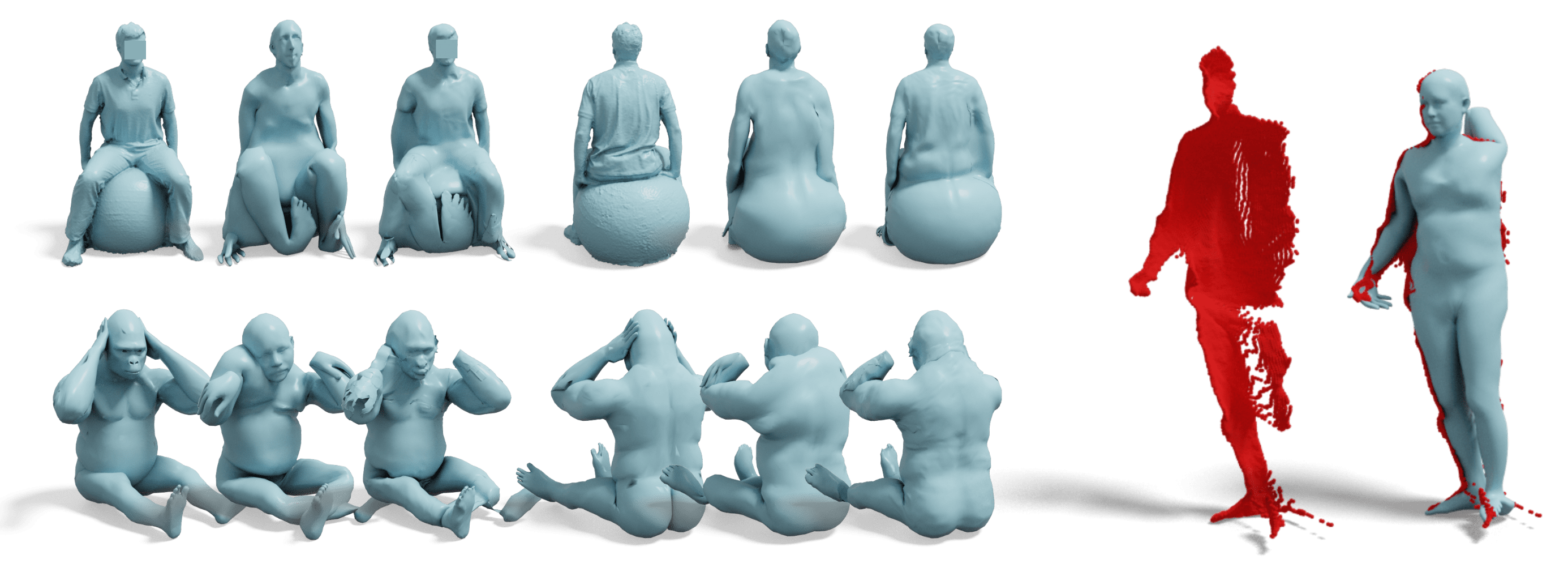}
		\put(5,37){Input}
            \put(21,37){\pipeline{}}

		\put(38,37){Input}
            \put(52,37){\pipeline{}}

       	\put(76,37){Input}
            \put(90,37){\pipeline{}}     
	\end{overpic}
 \caption{\label{fig:failures} Some failure cases for our pipeline. The presence of heavy clutter, highly unnatural poses, and the complete absence of limbs are among the main failure cases we observed.}
\end{figure*}

During our experiments, we observed some recurrent failure modes for our method. While we show robustness to clutter when it constitutes a significant part of the scene (\textit{e.g.}, large objects), our method cannot distinguish between what is human or not. 
Our method performs well with disparate identities even significantly far from the training distribution, even on non-humanoids like the ones in Figure \ref{fig:statues}. Still, when combined with unusual poses, arms and legs might be wrongly located in space.
Finally, our method can also recover the human posture in partial point clouds. However, if the input does not contain enough information to define the position of all the human parts, the registration may not find the correct locations for the missing ones attracted by the input point cloud.
Examples of these failures are reported in Figure~\ref{fig:failures}, where we can appreciate that some of the body parts are correctly located even in failure cases. 
We believe that data augmentation combined with segmentation prediction of the input to remove the clutter or highlight what template parts have an image in the input could be a promising direction to address these challenges. Regularizing the \method{} procedure, considering the initial prediction of the NF could also enhance the procedure's robustness.

\begin{figure*}[th!]

    \centering
    \footnotesize
 \begin{overpic}[trim=0cm 0cm 0cm 0cm,clip, width=\linewidth]{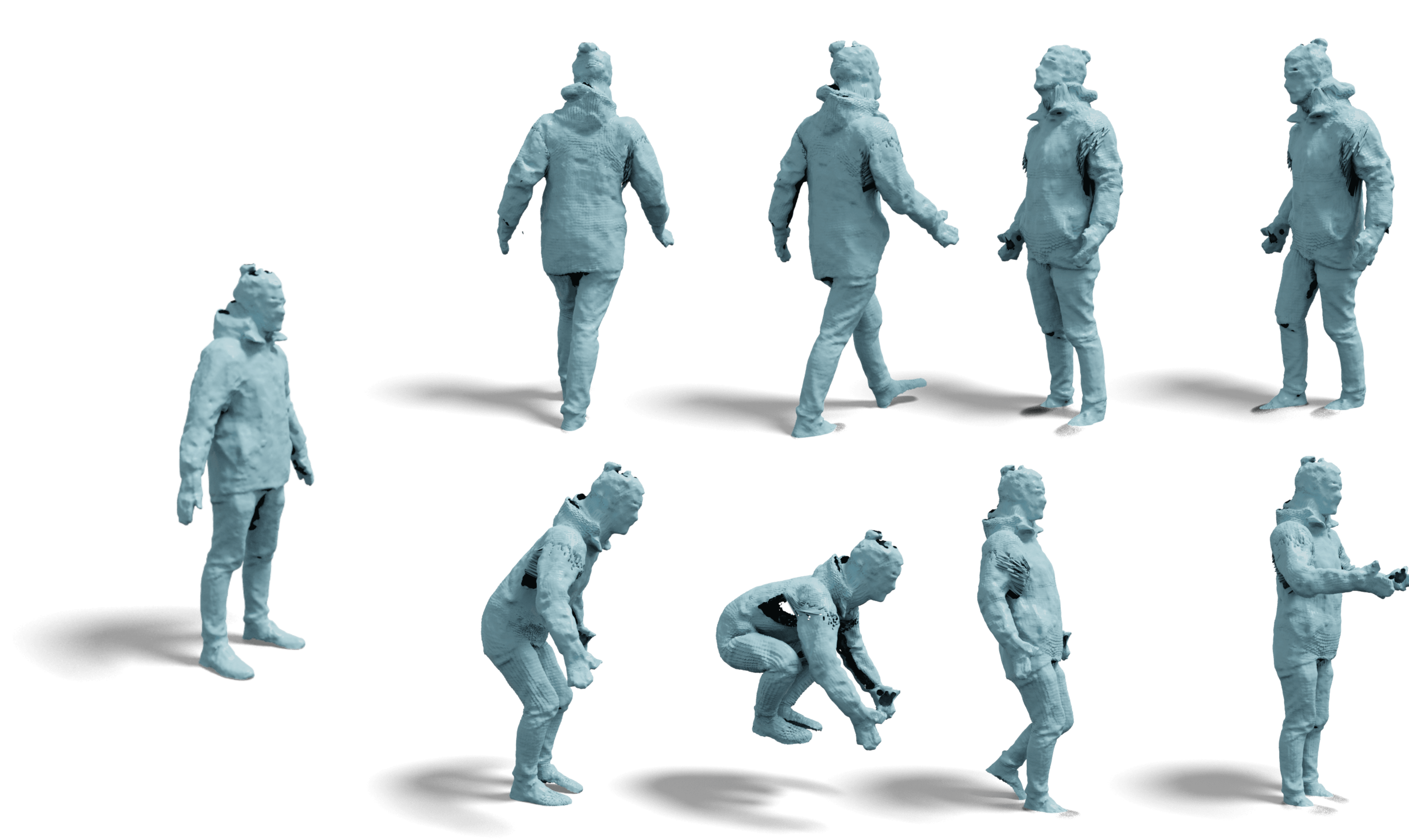}
		\put(11,45){LUMA output}
            \put(53, 60){Animations of LUMA output}
	\end{overpic}
    \caption{\label{fig:LumaAI2} Animatable avatar from \cite{luma}. On the left, the geometry was obtained from LUMA. On the right our animation results. Our animation results show semantically coherent motions despite the heavy clothing and gluing that ruined the geometry extracted by LUMA.}
\end{figure*}


\subsection{Further Qualitative results}

\noindent{\bf Animation} We show a further example of animation with heavier clothes in Figure~\ref{fig:LumaAI2}. Despite the clutter and geometric gluing caused by the garments, our registration is precise enough to provide coherent rigging of the target shape.

\noindent{\bf \pipeline{} results} In the last pages of this document, we show many qualitative results from different datasets. In order:

\begin{itemize}
    \item Pages 33 to 36: RenderPeople \cite{renderpeople}
    \item Pages 37 to 41: D-Faust  \cite{bogo2017dynamic}
    \item Pages 42 to 45: Test shapes from FAUST challenge \cite{bogo2014faust}
    \item Pages 46 to 48: HuMMan \cite{cai2022humman}
    \item Pages 49 to 51: BEHAVE \cite{bhatnagar2022behave}
\end{itemize}
For each row, we will display the input and the result of our \pipeline{} pipeline, both with and without the SMPL+D refinement on the left-hand side. However, for HuMMan and BHEAVE, we will not report the SMPL+D since the shapes in the former do not have significant details, and in the latter, the high-frequency features are primarily noise. It is important to note that we only visualize the meshes for clarity, and our method works solely from the point cloud and does not consider the target mesh in any way. We highlight the variety of poses, clothes, clutter, holes, and identities our method can solve. 

\clearpage
\newpage 

\begin{figure*}
\vspace{2cm}
    \centering
    \footnotesize
 \begin{overpic}[trim=7cm 4cm 12cm 12cm,clip, width=\linewidth]{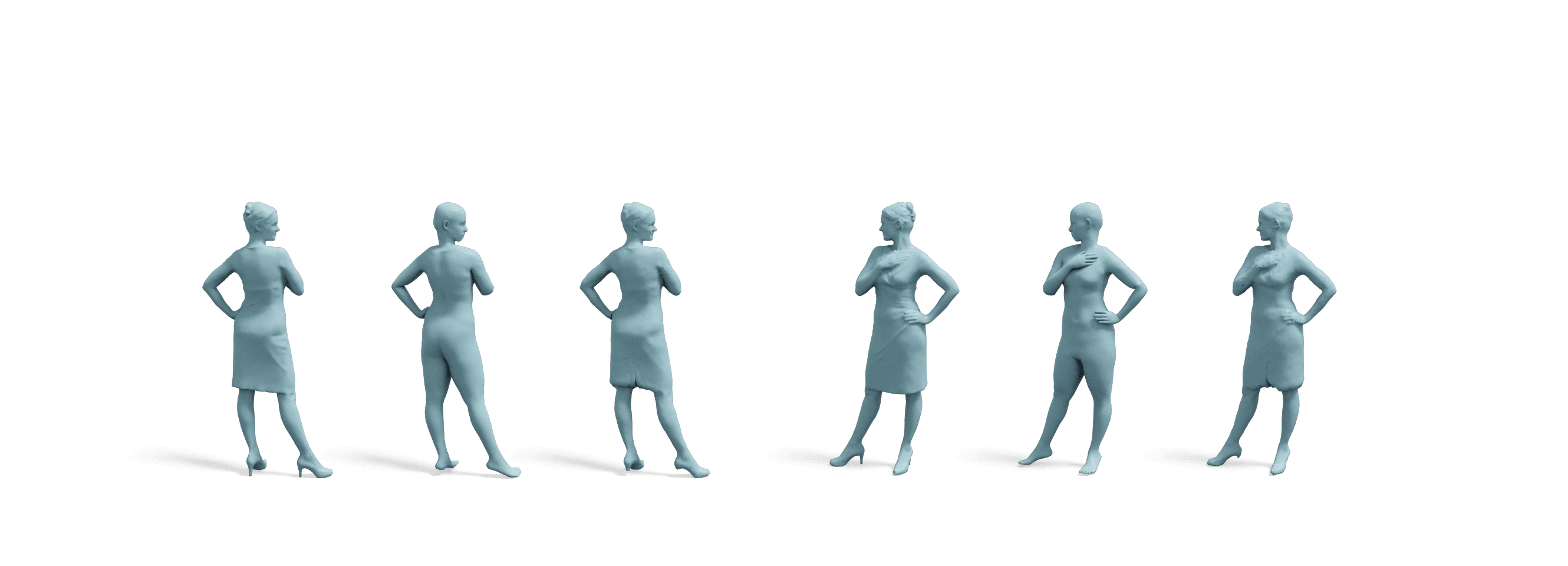}
            \put(40,30){\huge RenderPeople}
		\put(12,25){Input}
        \put(31,25){\pipeline{}}

		\put(60,25){Input}
        \put(81,25){\pipeline{}}
        
	\end{overpic}
 \begin{overpic}[trim=6cm 4cm 12cm 12cm,clip, width=\linewidth]{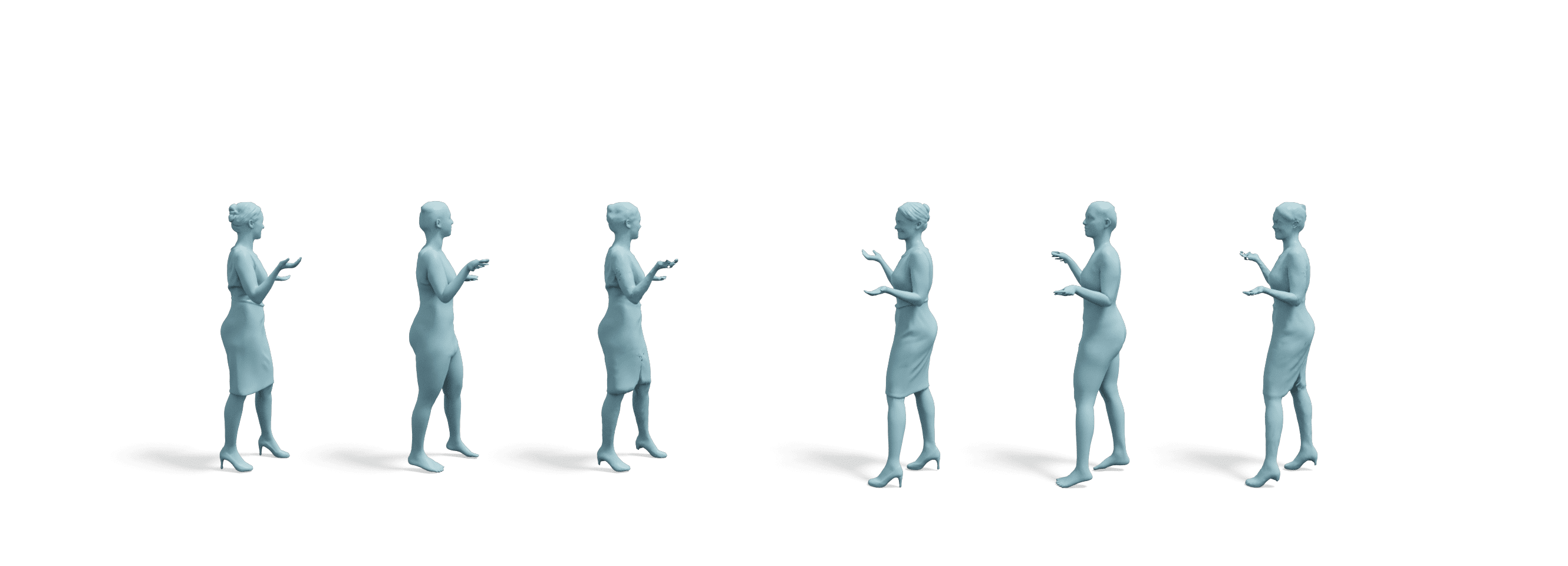}
		\put(12,99){}
	\end{overpic}
  \begin{overpic}[trim=6cm 4cm 12cm 12cm,clip, width=\linewidth]{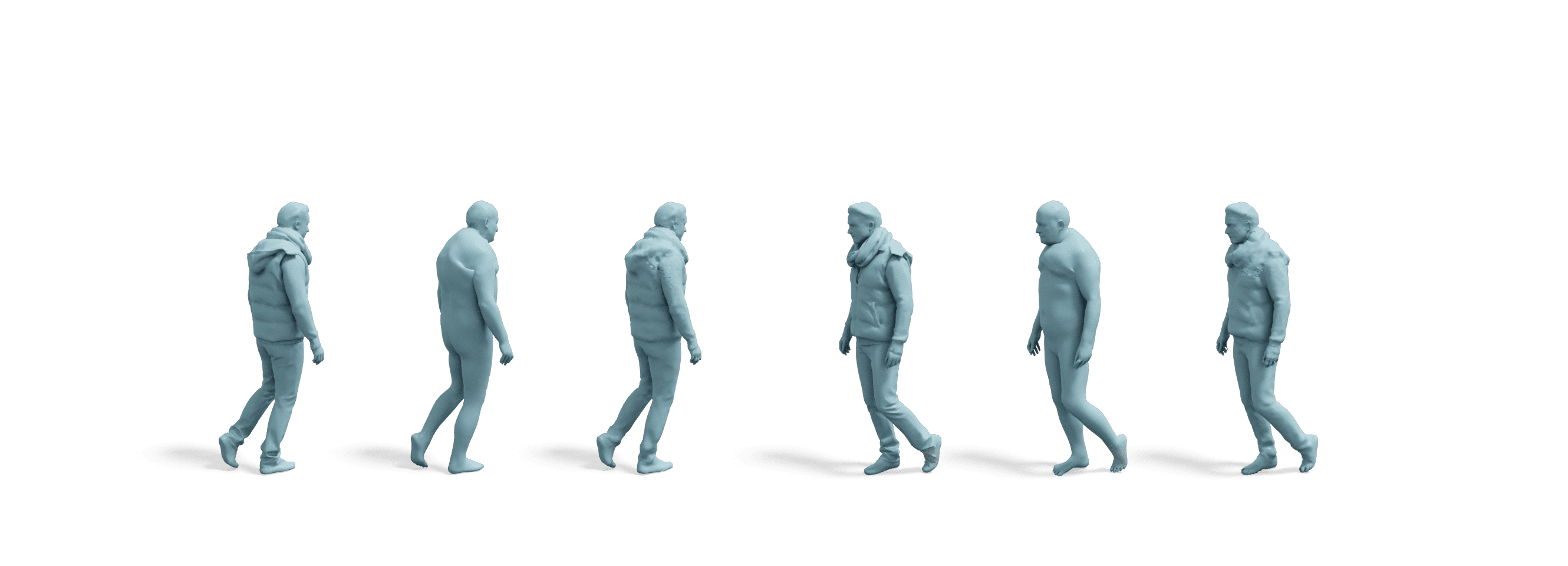}
		\put(12,99){}
	\end{overpic}
  \begin{overpic}[trim=6cm 4cm 12cm 12cm,clip, width=\linewidth]{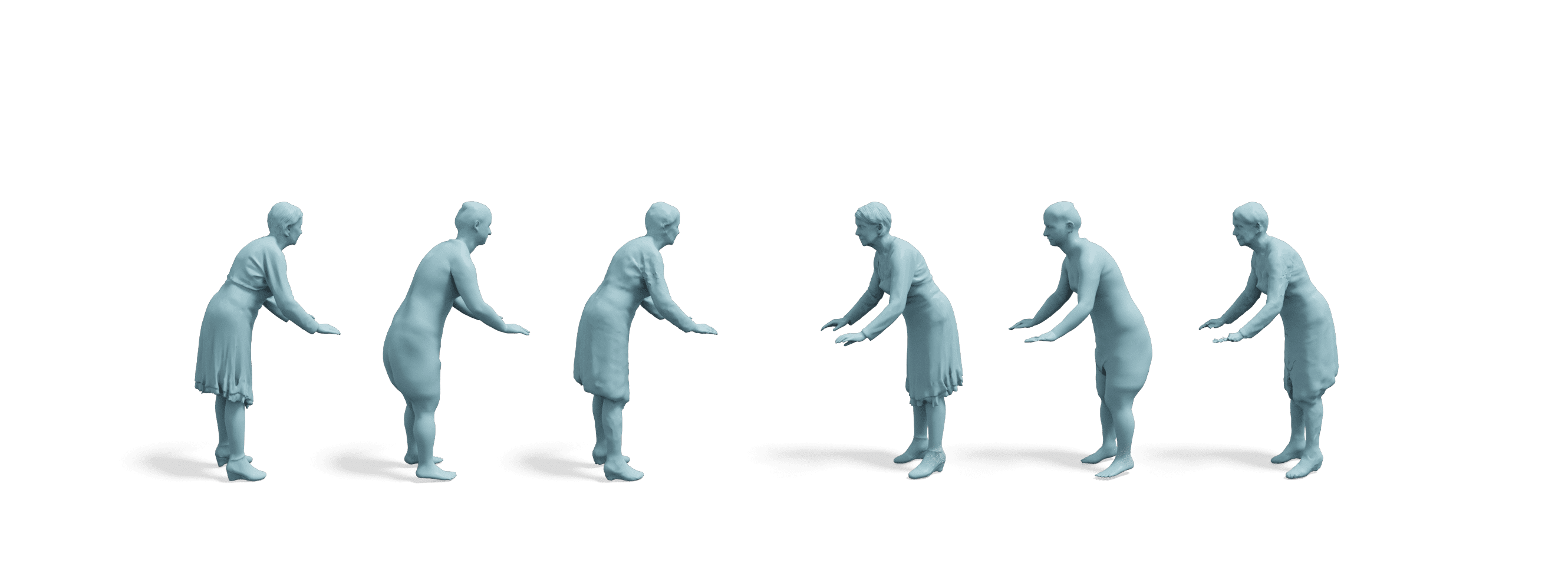}
		\put(12,99){}
	\end{overpic}
  \begin{overpic}[trim=6cm 4cm 12cm 12cm,clip, width=\linewidth]{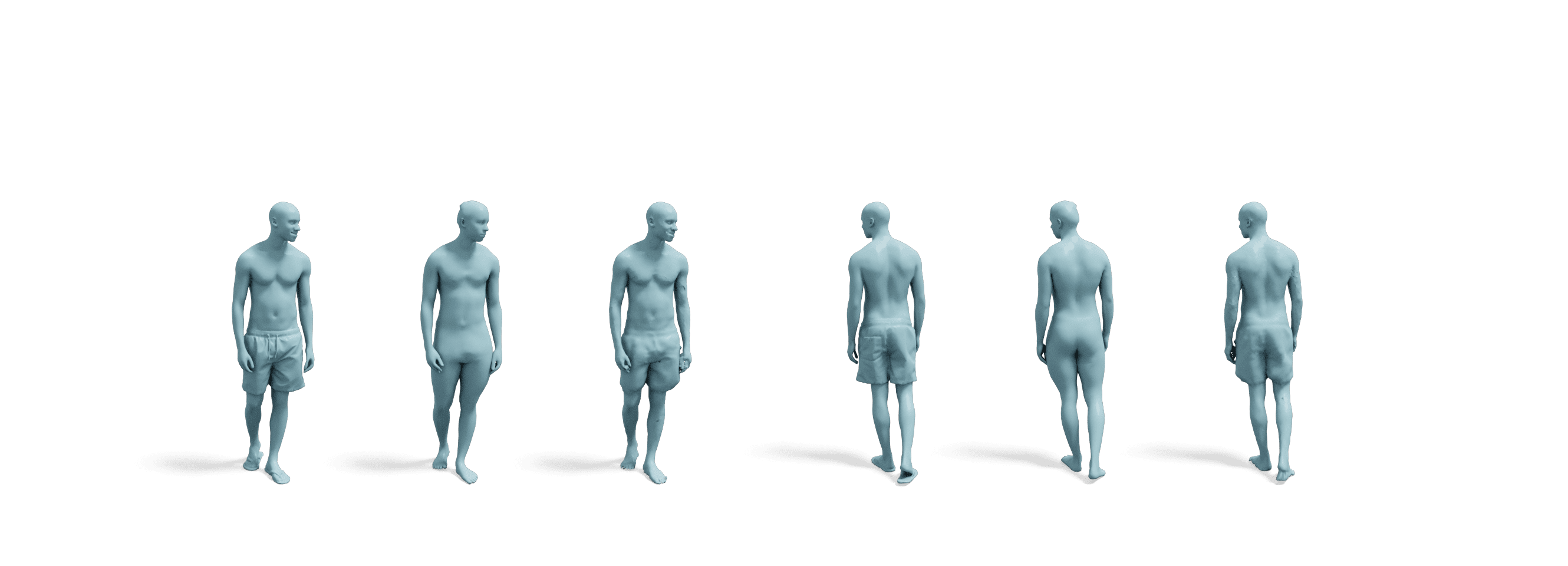}
		\put(12,99){}
	\end{overpic}
\end{figure*}

\begin{figure*}
    \centering
    \footnotesize
 \begin{overpic}[trim=7cm 4cm 12cm 12cm,clip, width=\linewidth]{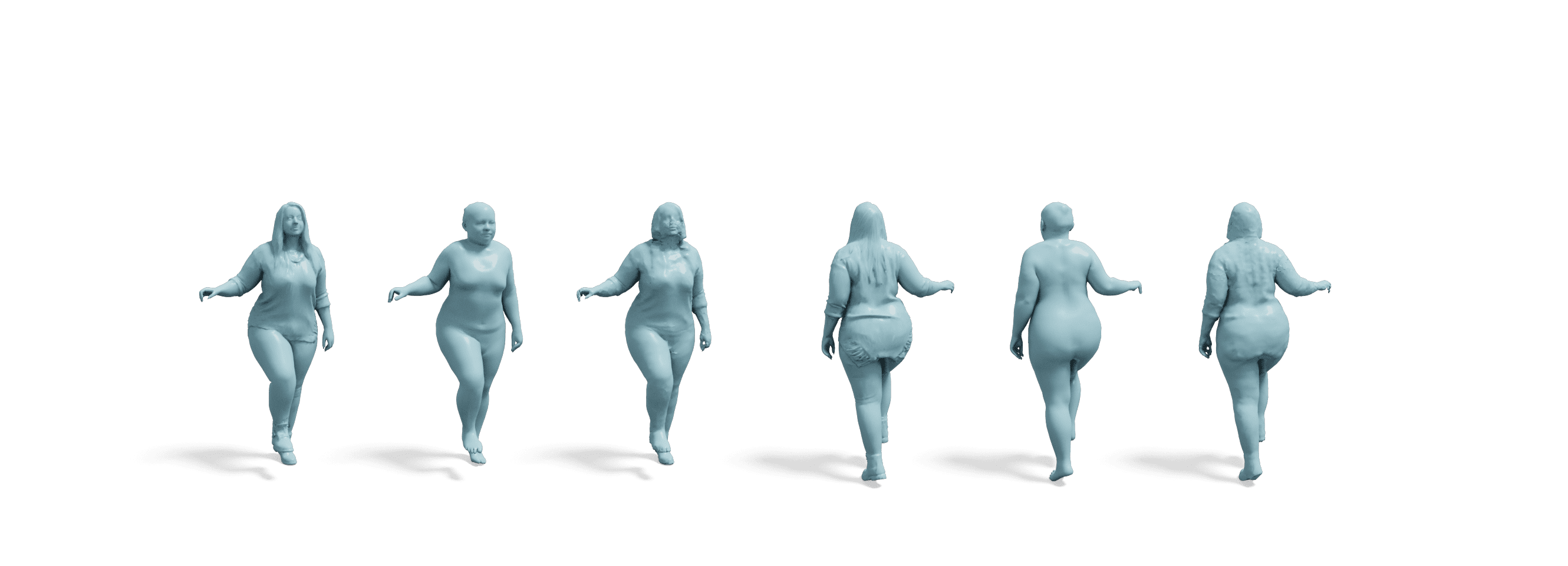}
 \put(40,30){\huge RenderPeople}
		\put(12,25){Input}
        \put(31,25){\pipeline{}}

		\put(60,25){Input}
        \put(81,25){\pipeline{}}
	\end{overpic}
 \begin{overpic}[trim=6cm 4cm 12cm 12cm,clip, width=\linewidth]{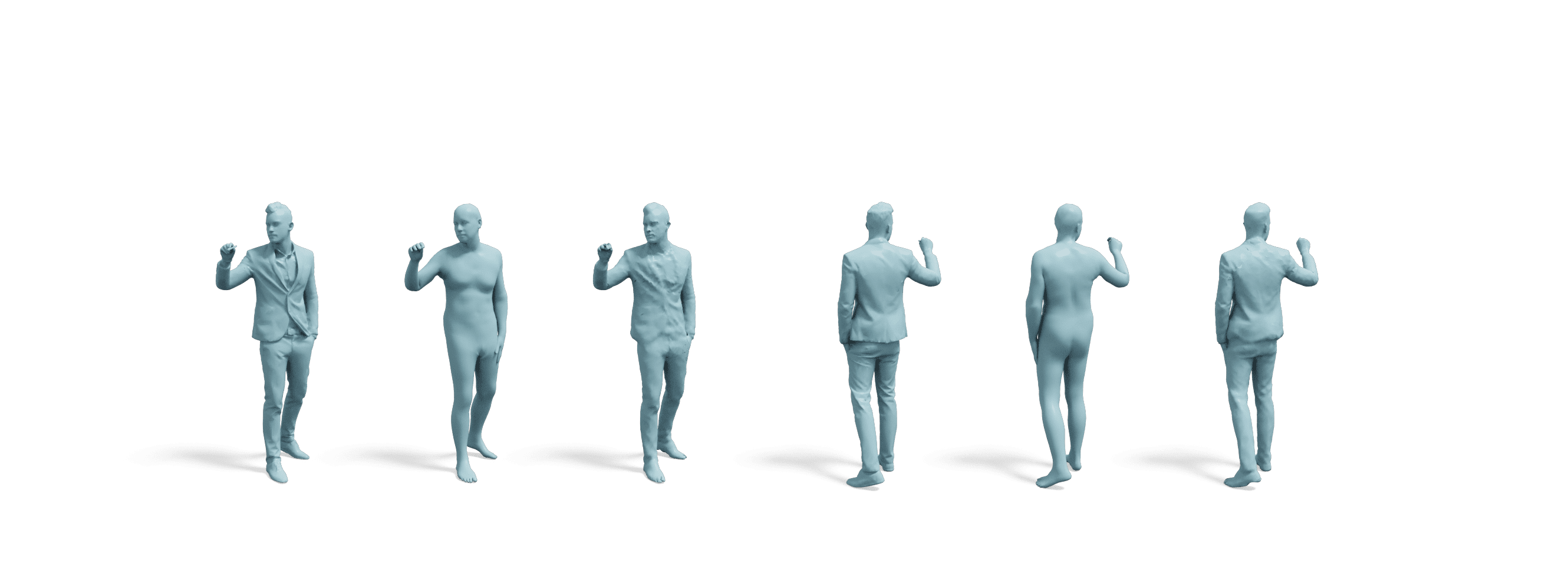}
		\put(12,99){}
	\end{overpic}
  \begin{overpic}[trim=6cm 4cm 12cm 12cm,clip, width=\linewidth]{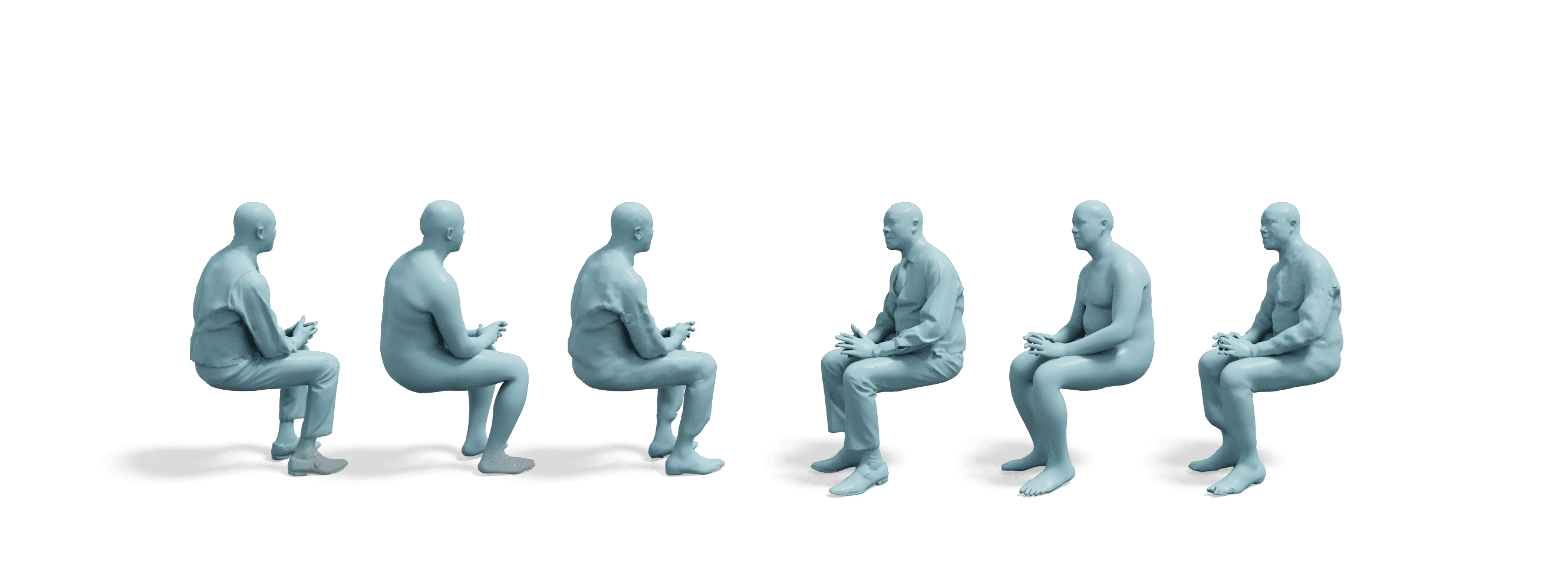}
		\put(12,99){}
	\end{overpic}
  \begin{overpic}[trim=6cm 4cm 12cm 12cm,clip, width=\linewidth]{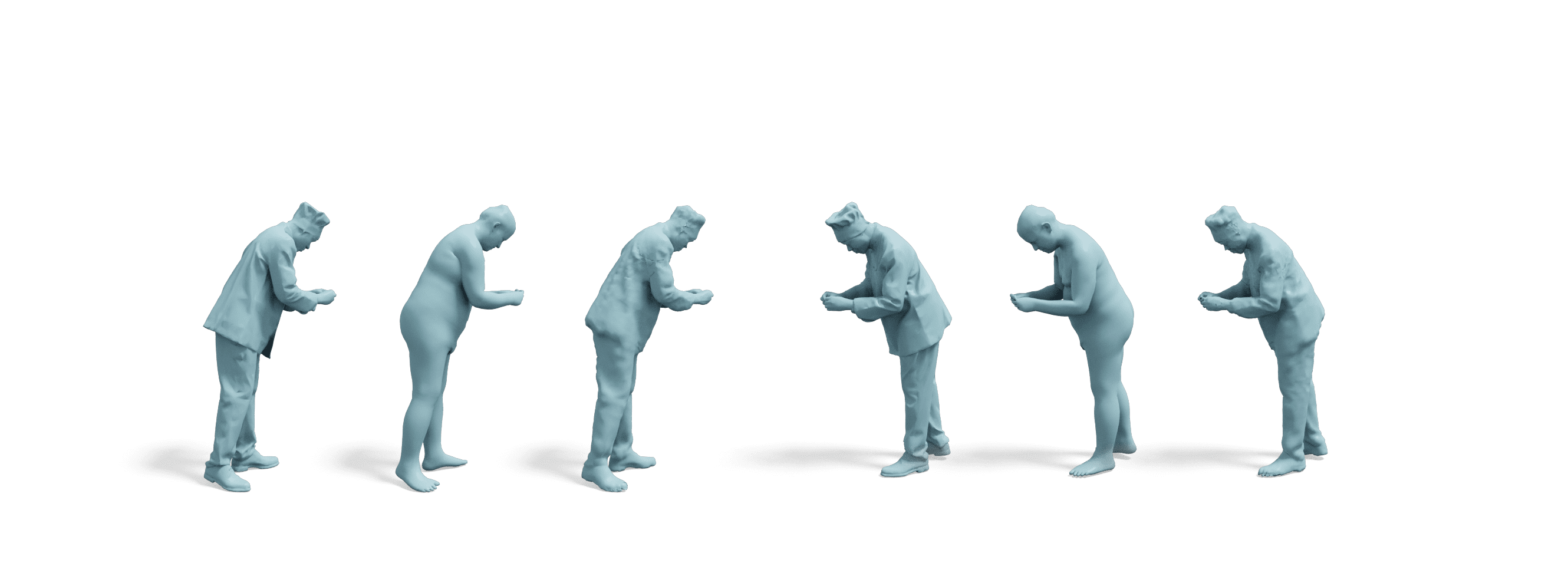}
		\put(12,99){}
	\end{overpic}
  \begin{overpic}[trim=6cm 4cm 12cm 12cm,clip, width=\linewidth]{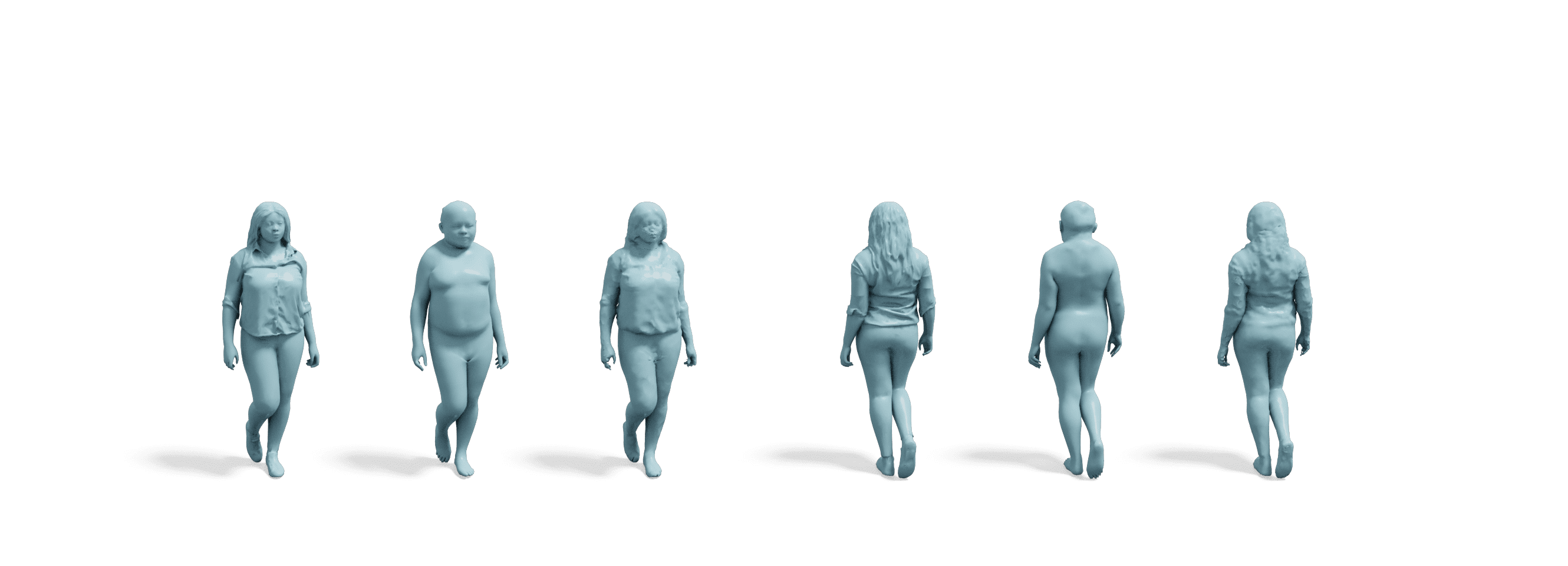}
		\put(12,99){}
	\end{overpic}
\end{figure*}

\begin{figure*}

    \centering
    \footnotesize
 \begin{overpic}[trim=7cm 4cm 12cm 12cm,clip, width=\linewidth]{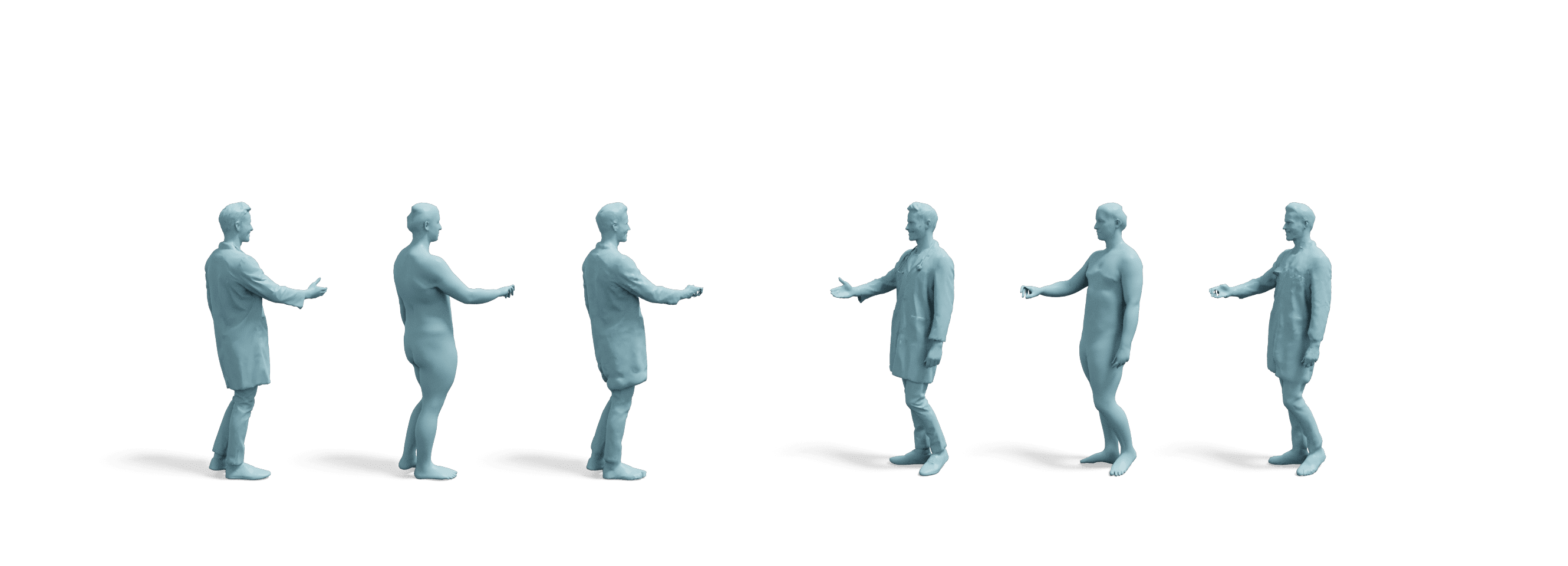}
 \put(40,30){\huge RenderPeople}
		\put(12,25){Input}
        \put(31,25){\pipeline{}}

		\put(60,25){Input}
        \put(81,25){\pipeline{}}
	\end{overpic}
 \begin{overpic}[trim=6cm 4cm 12cm 12cm,clip, width=\linewidth]{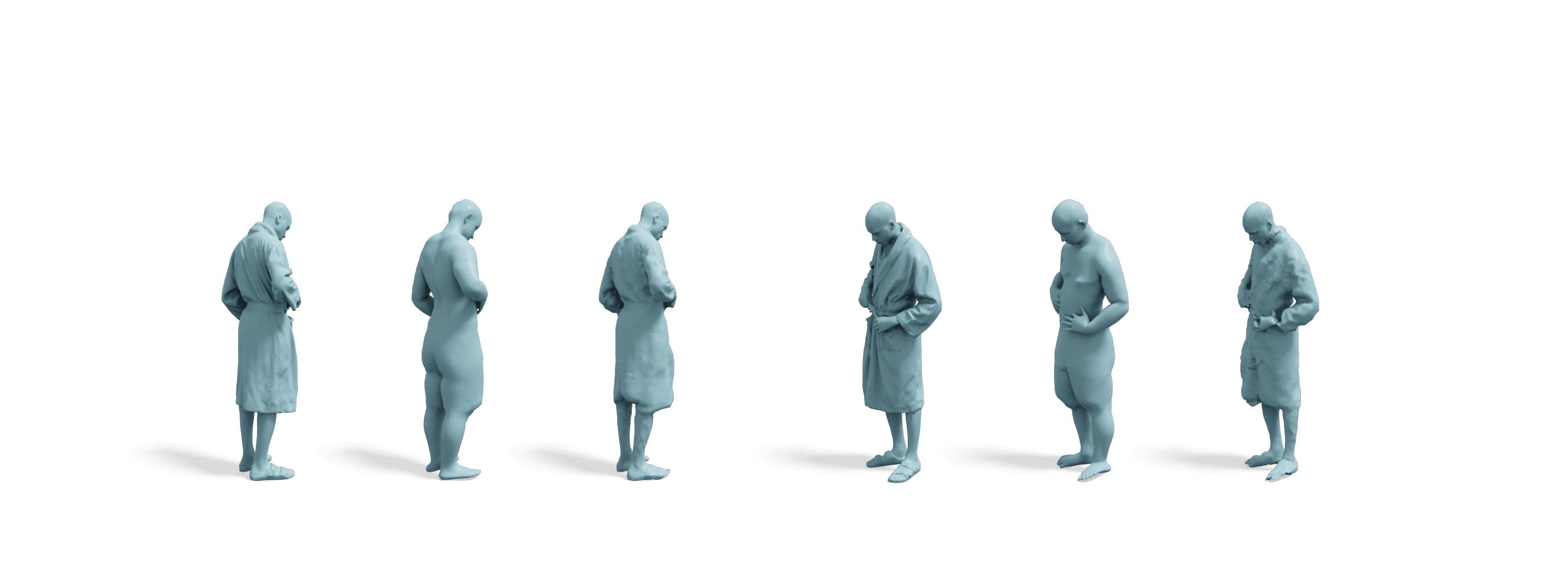}
		\put(12,99){}
	\end{overpic}
  \begin{overpic}[trim=6cm 4cm 12cm 12cm,clip, width=\linewidth]{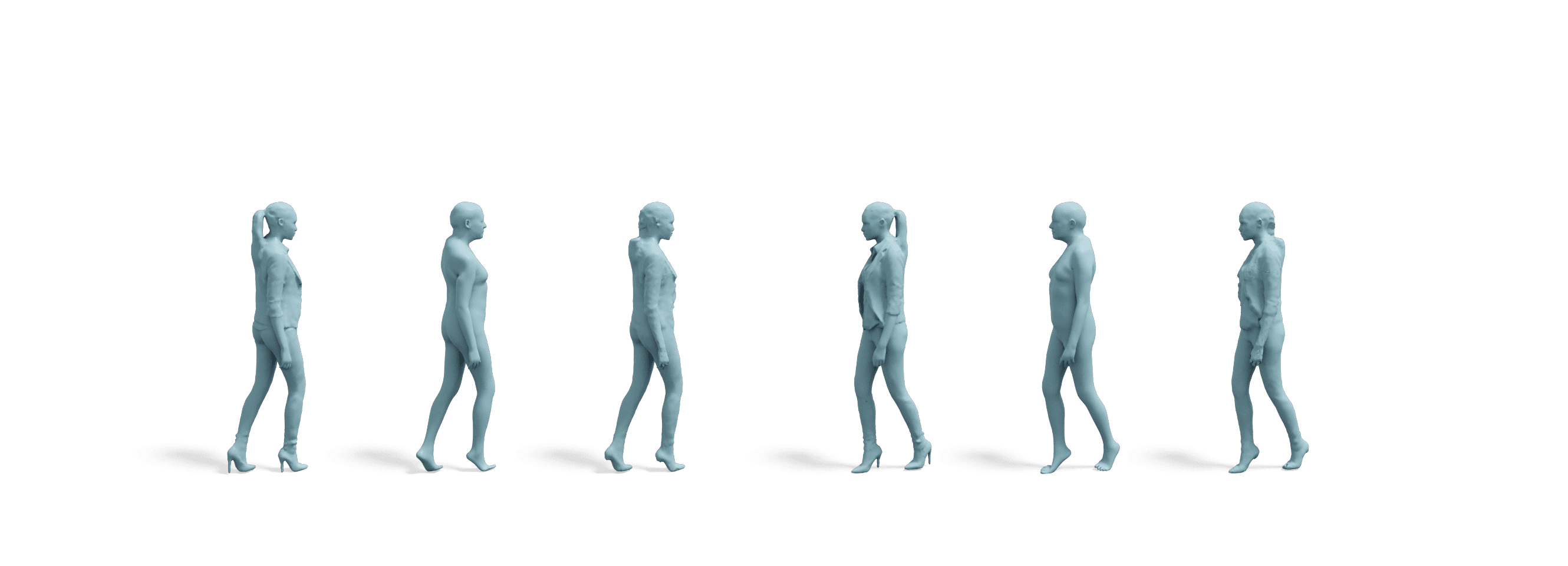}
		\put(12,99){}
	\end{overpic}
  \begin{overpic}[trim=6cm 4cm 12cm 12cm,clip, width=\linewidth]{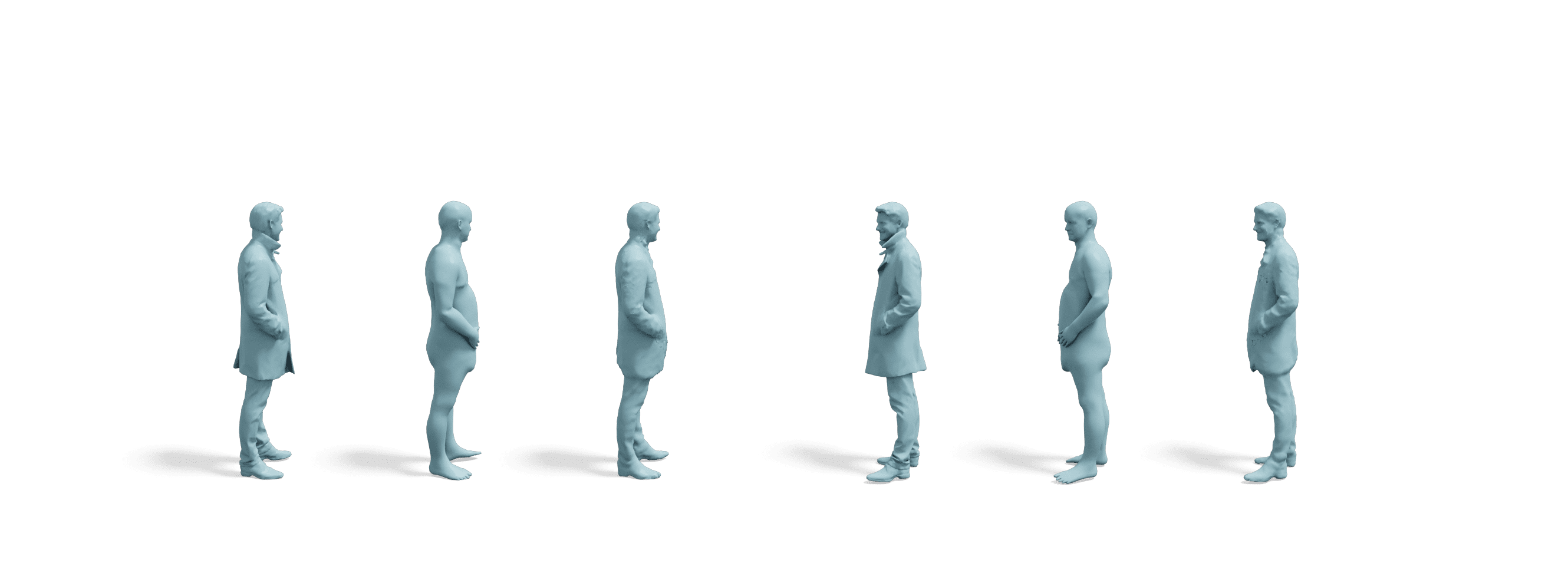}
		\put(12,99){}
	\end{overpic}
  \begin{overpic}[trim=6cm 4cm 12cm 12cm,clip, width=\linewidth]{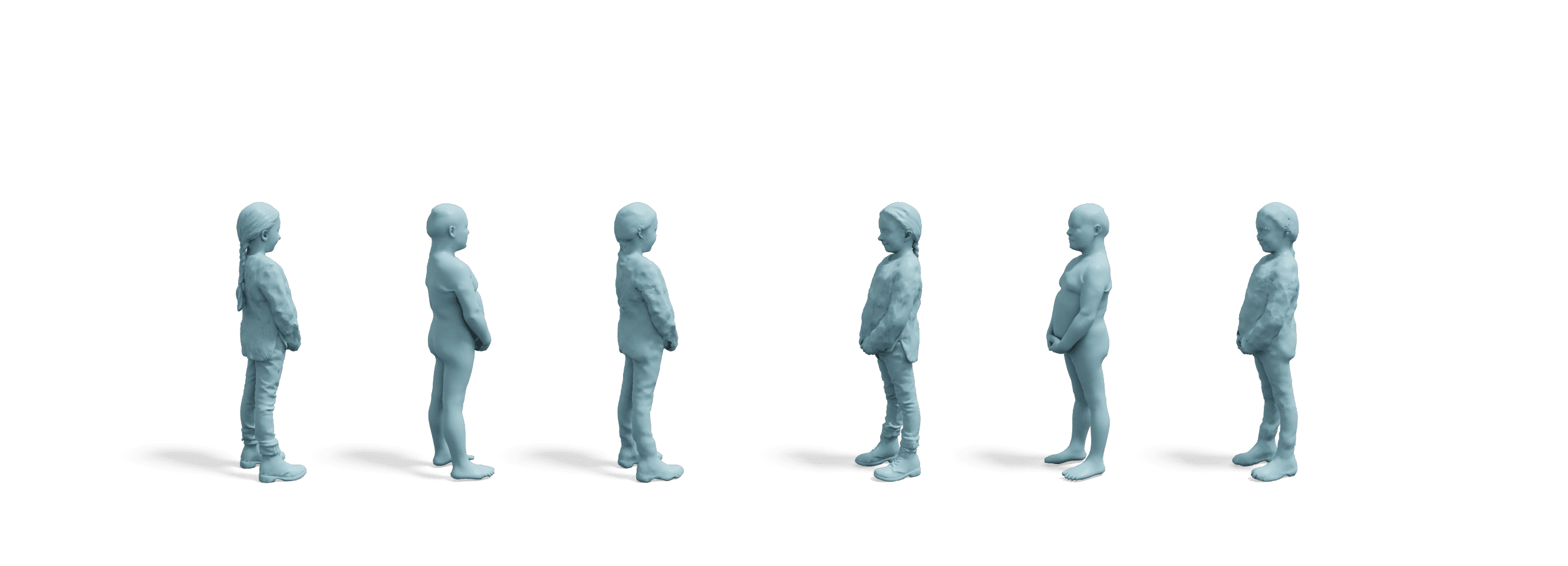}
		\put(12,99){}
	\end{overpic}
\end{figure*}

\begin{figure*}

    \centering
    \footnotesize
 \begin{overpic}[trim=7cm 4cm 12cm 12cm,clip, width=\linewidth]{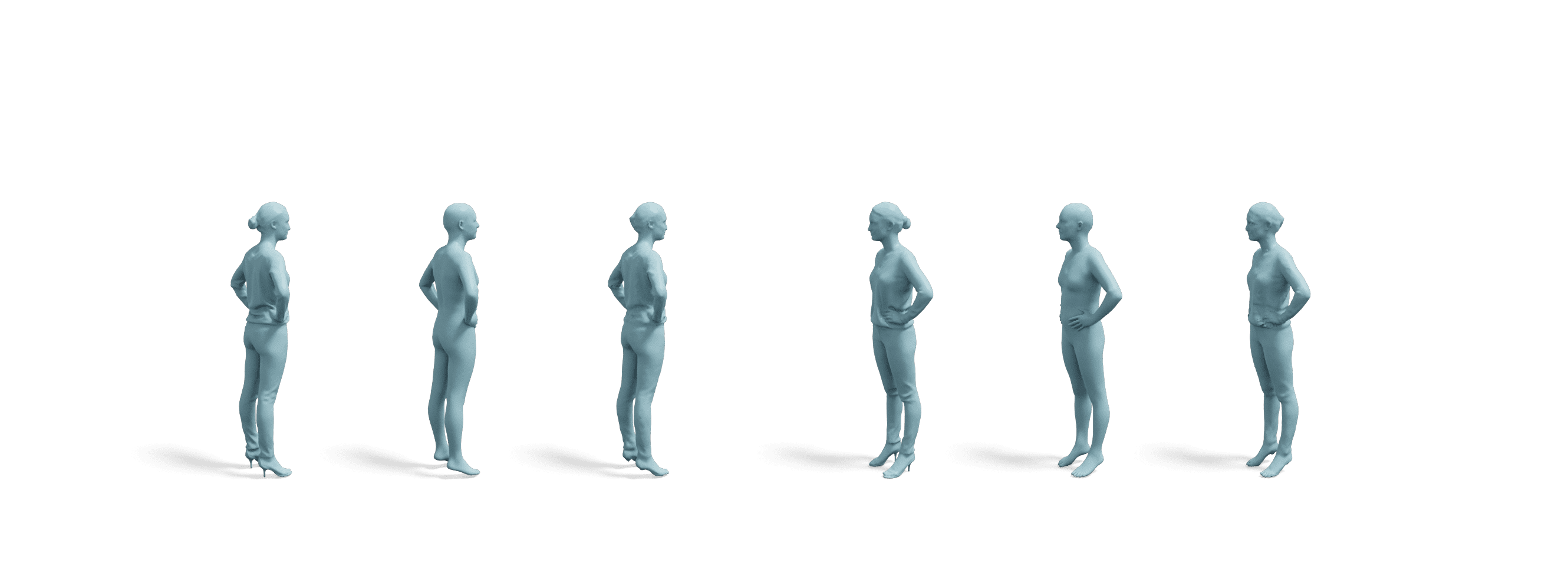}
 \put(40,30){\huge RenderPeople}
		\put(12,25){Input}
        \put(31,25){\pipeline{}}

		\put(60,25){Input}
        \put(81,25){\pipeline{}}
	\end{overpic}
 \begin{overpic}[trim=6cm 4cm 12cm 12cm,clip, width=\linewidth]{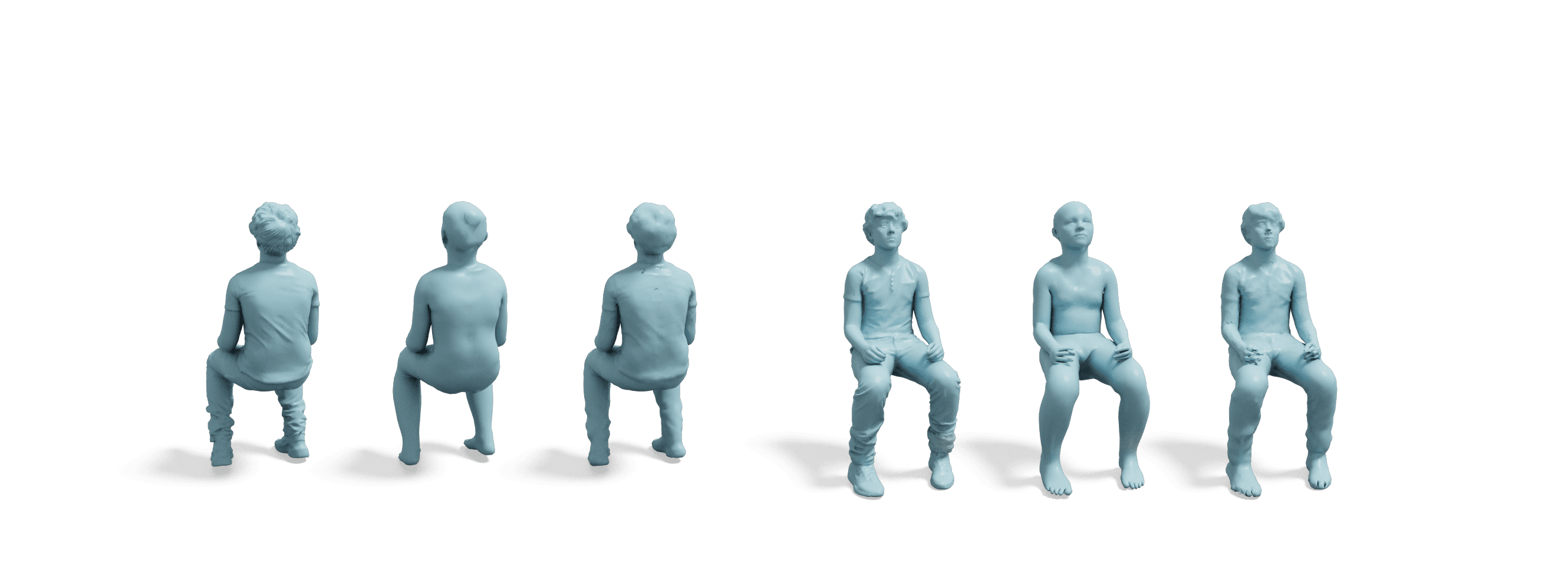}
		\put(12,99){}
	\end{overpic}
  \begin{overpic}[trim=6cm 4cm 12cm 12cm,clip, width=\linewidth]{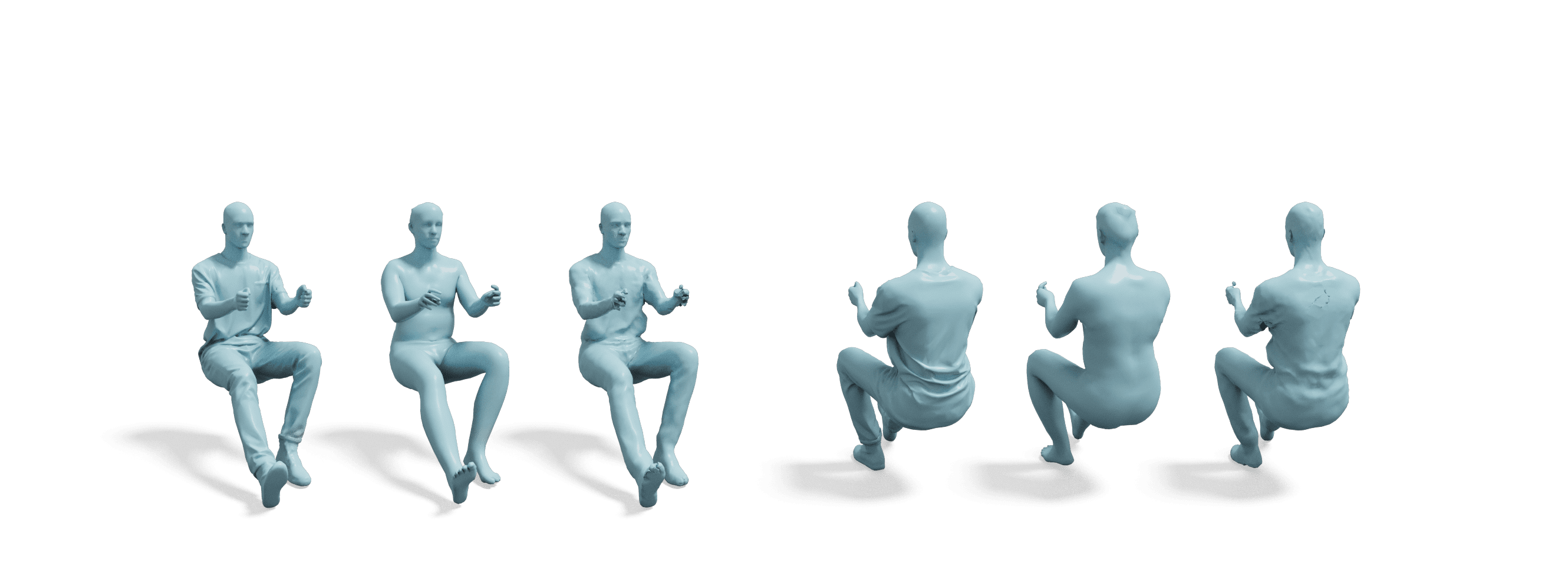}
		\put(12,99){}
	\end{overpic}
  \begin{overpic}[trim=6cm 4cm 12cm 12cm,clip, width=\linewidth]{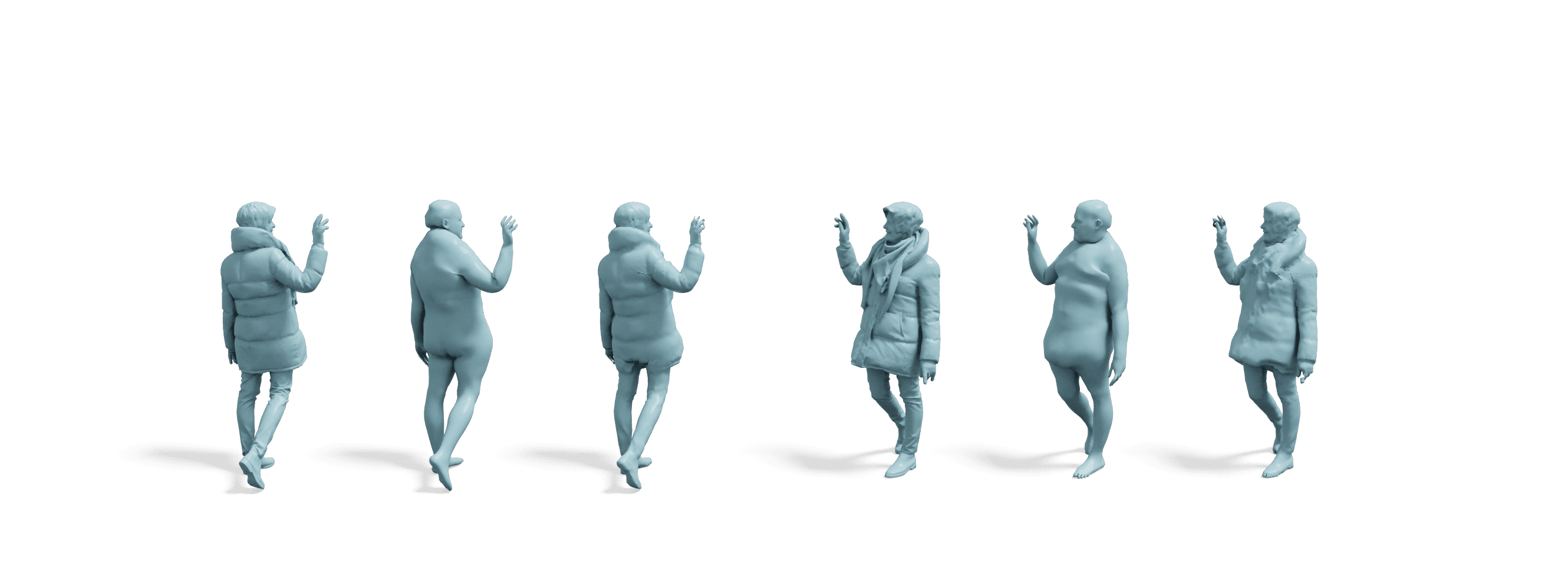}
		\put(12,99){}
	\end{overpic}
  \begin{overpic}[trim=6cm 4cm 12cm 12cm,clip, width=\linewidth]{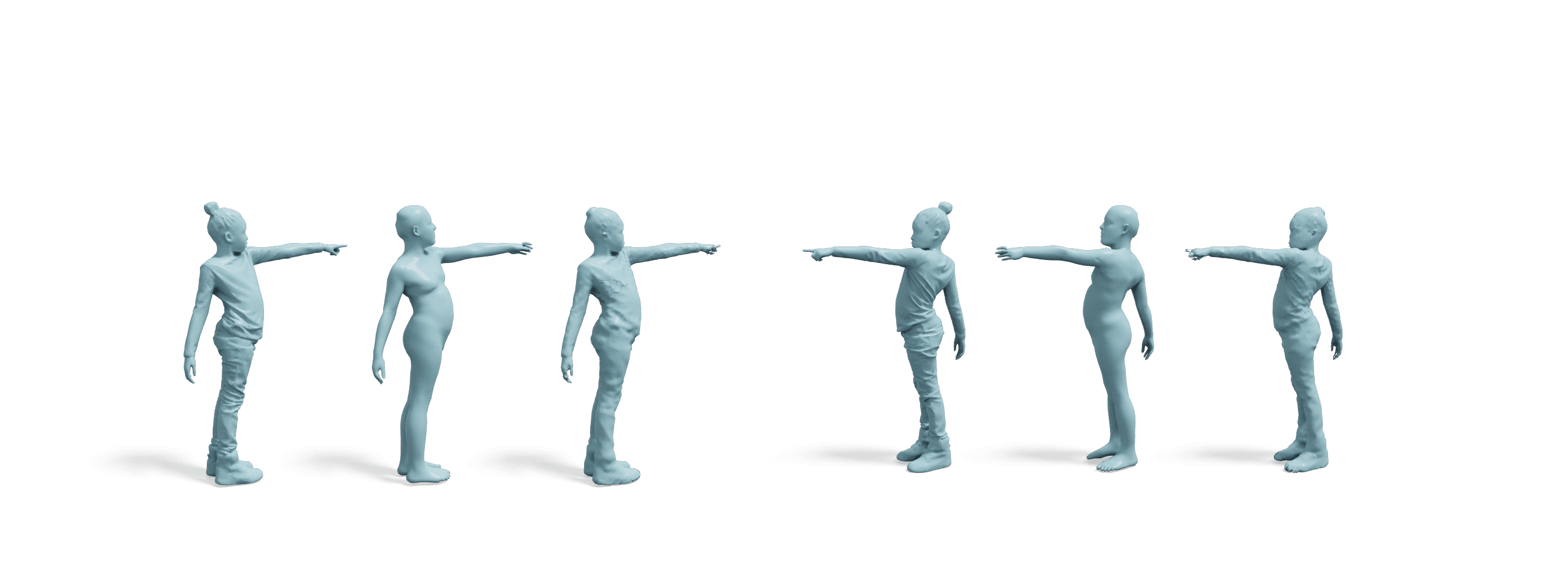}
		\put(12,99){}
	\end{overpic}
\end{figure*}

\begin{figure*}

    \centering
    \footnotesize
 \begin{overpic}[trim=7cm 4cm 12cm 12cm,clip, width=\linewidth]{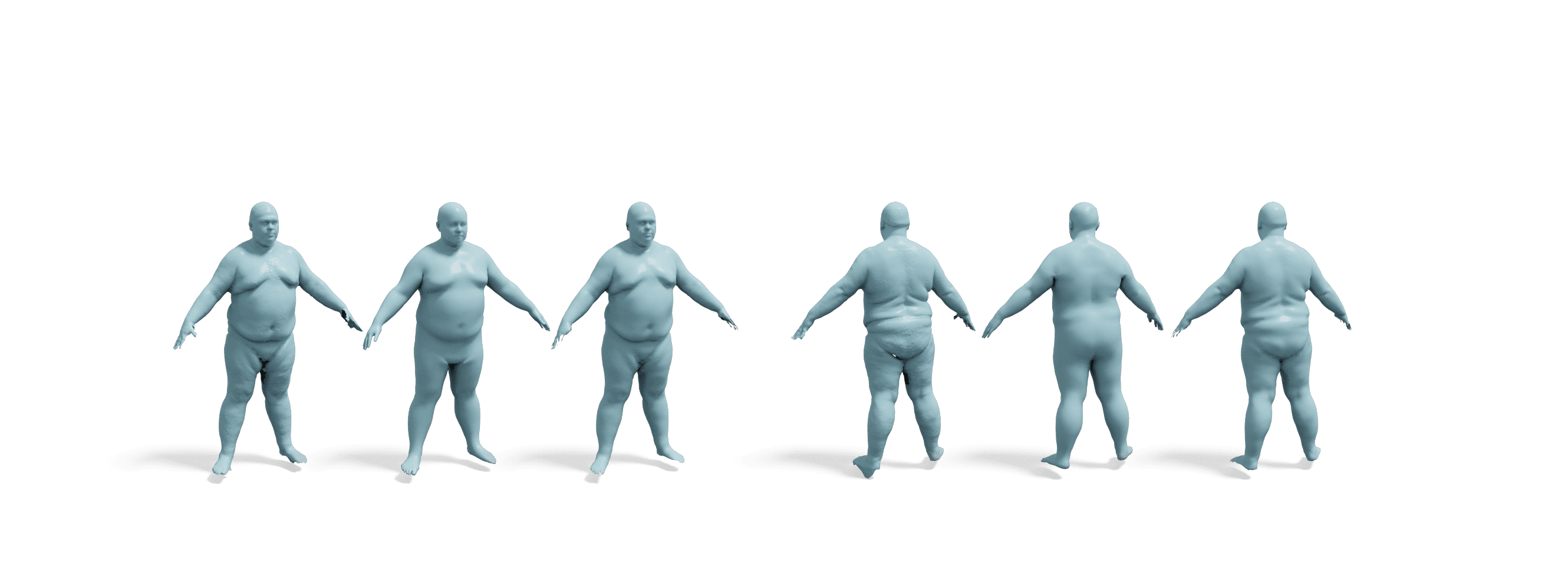}
 \put(40,28){\huge DFAUST}
		\put(12,24){Input}
        \put(31,24){\pipeline{}}

		\put(60,24){Input}
        \put(81,24){\pipeline{}}
	\end{overpic}
 \begin{overpic}[trim=6cm 4cm 12cm 12cm,clip, width=\linewidth]{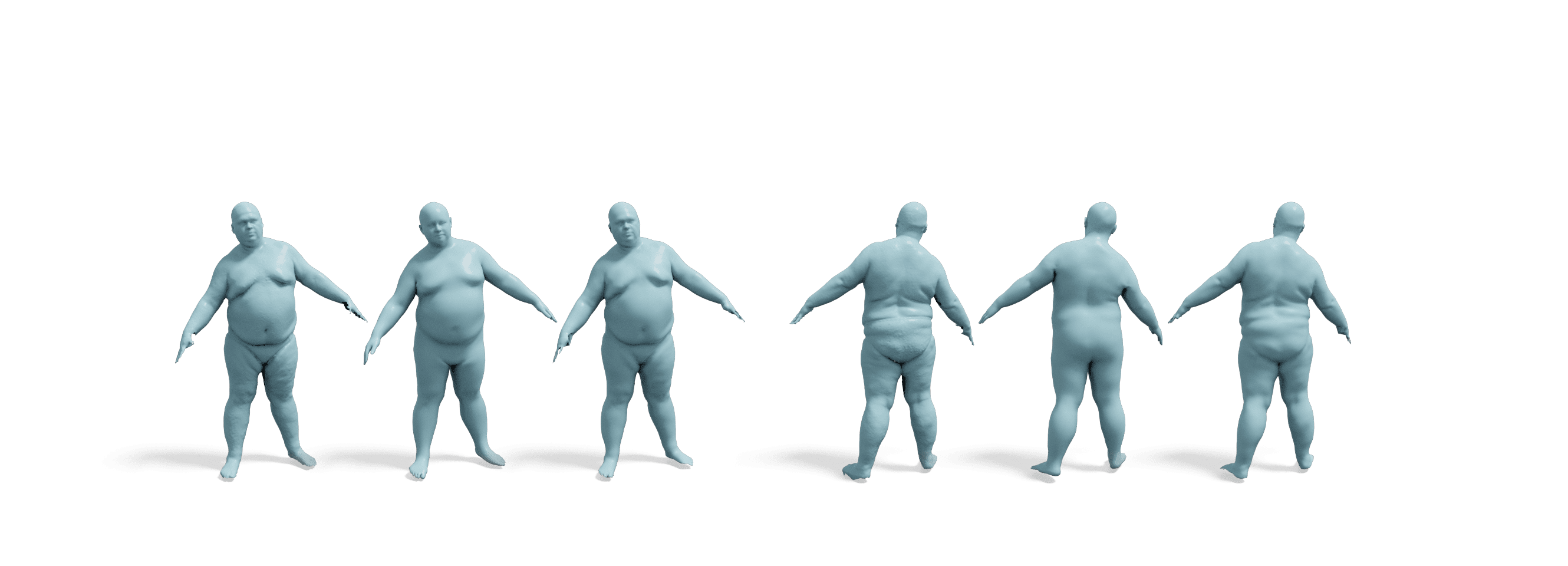}
		\put(12,99){}
	\end{overpic}
  \begin{overpic}[trim=6cm 4cm 12cm 12cm,clip, width=\linewidth]{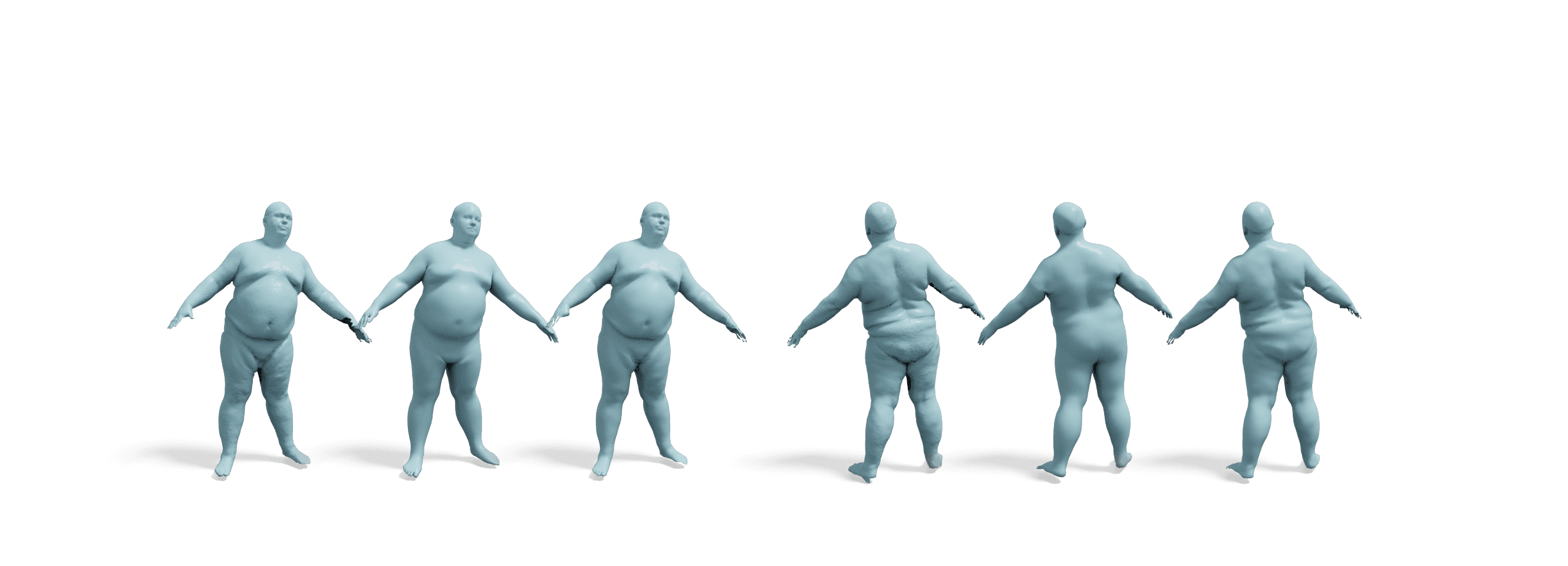}
		\put(12,99){}
	\end{overpic}
  \begin{overpic}[trim=6cm 4cm 12cm 12cm,clip, width=\linewidth]{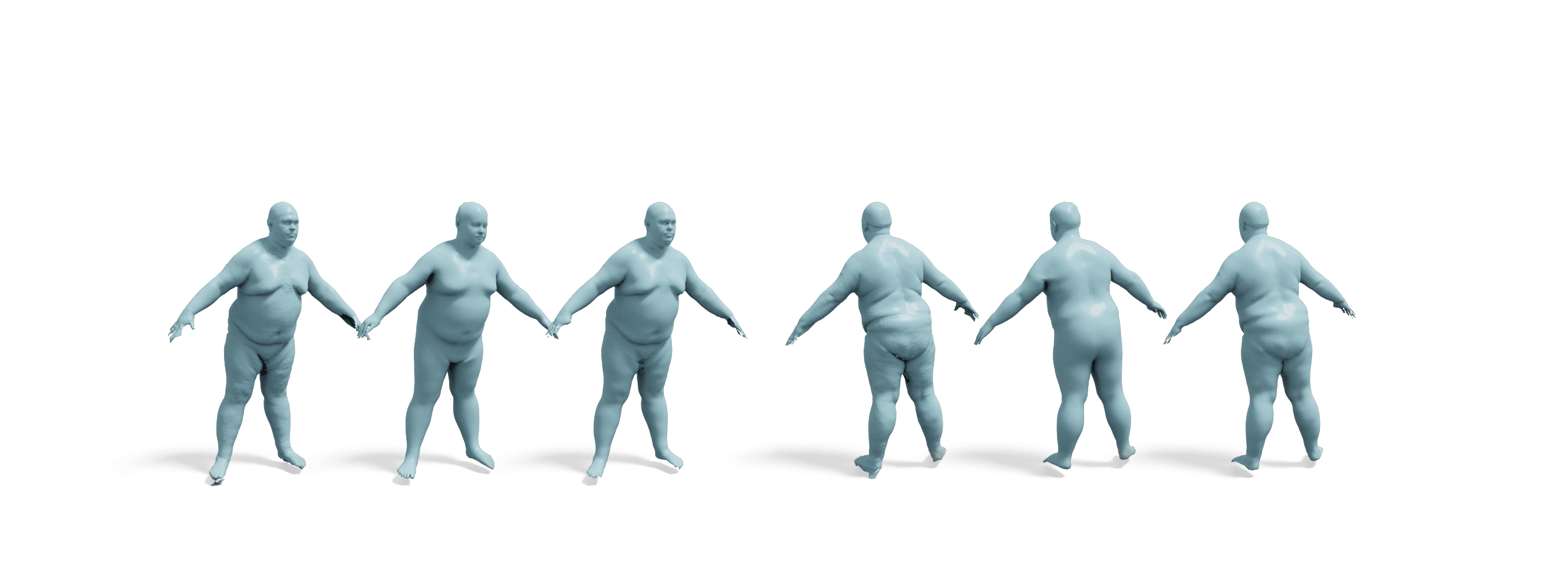}
		\put(12,99){}
	\end{overpic}
  \begin{overpic}[trim=6cm 4cm 12cm 12cm,clip, width=\linewidth]{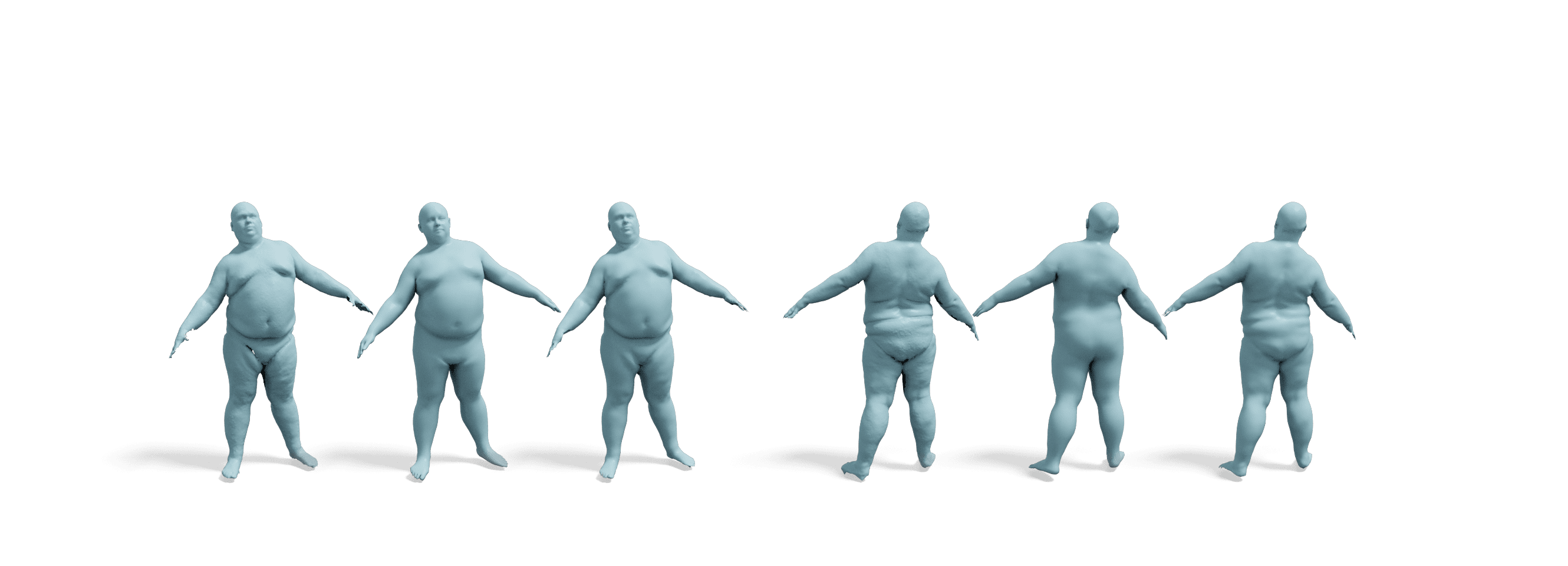}
		\put(12,99){}
	\end{overpic}
   \begin{overpic}[trim=6cm 4cm 12cm 12cm,clip, width=\linewidth]{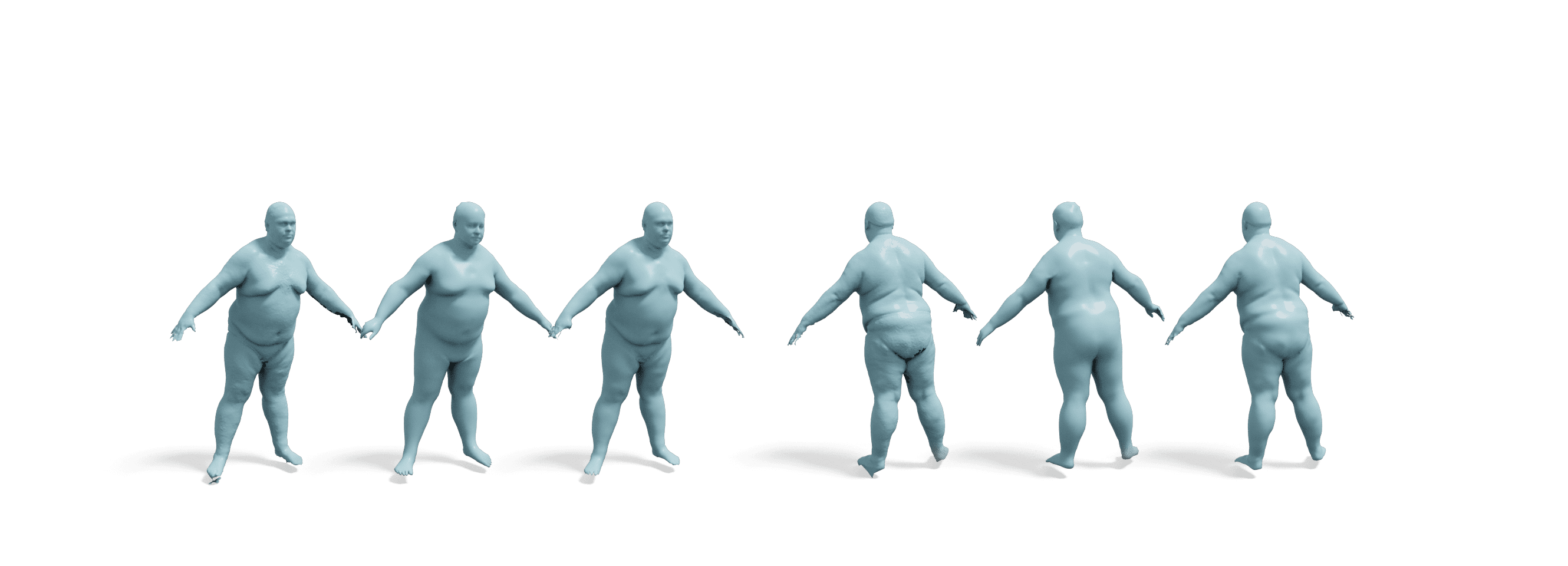}
		\put(12,99){}
	\end{overpic}

\end{figure*}

\begin{figure*}

    \centering
    \footnotesize
 \begin{overpic}[trim=7cm 4cm 12cm 12cm,clip, width=\linewidth]{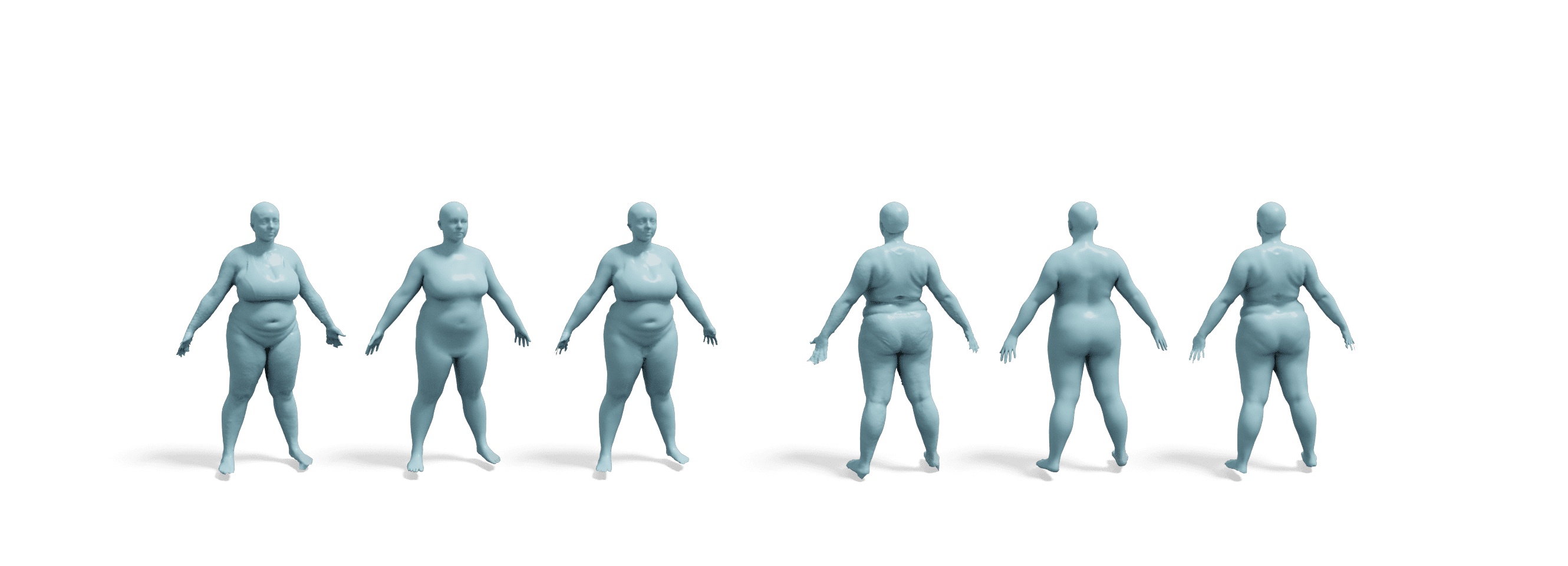}
 \put(40,28){\huge DFAUST}
		\put(12,24){Input}
        \put(31,24){\pipeline{}}

		\put(60,24){Input}
        \put(81,24){\pipeline{}}
	\end{overpic}
 \begin{overpic}[trim=6cm 4cm 12cm 12cm,clip, width=\linewidth]{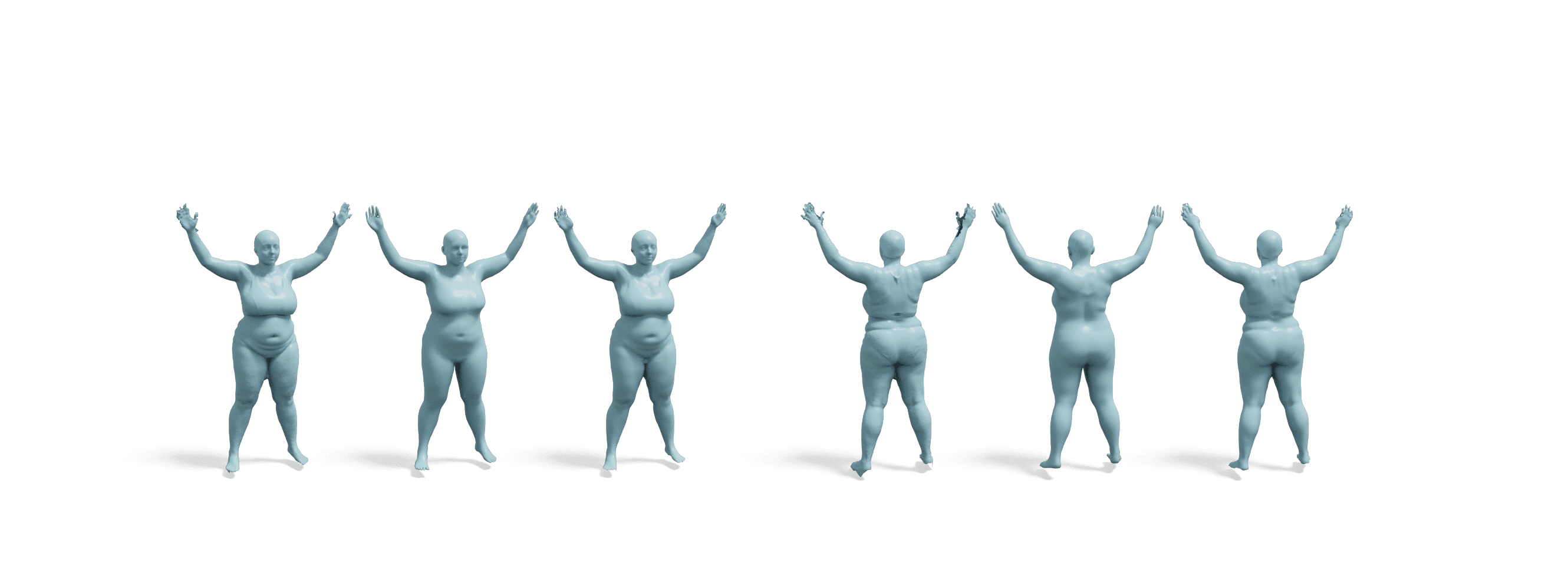}
		\put(12,99){}
	\end{overpic}
  \begin{overpic}[trim=6cm 4cm 12cm 12cm,clip, width=\linewidth]{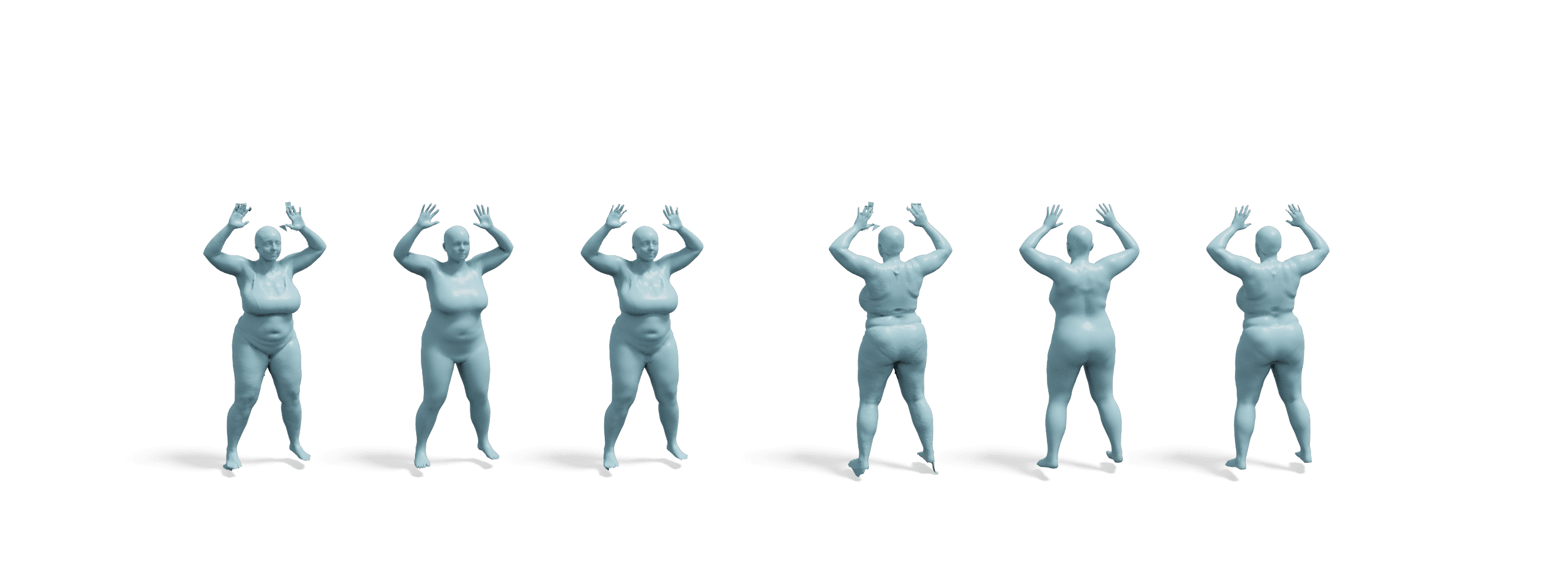}
		\put(12,99){}
	\end{overpic}
  \begin{overpic}[trim=6cm 4cm 12cm 12cm,clip, width=\linewidth]{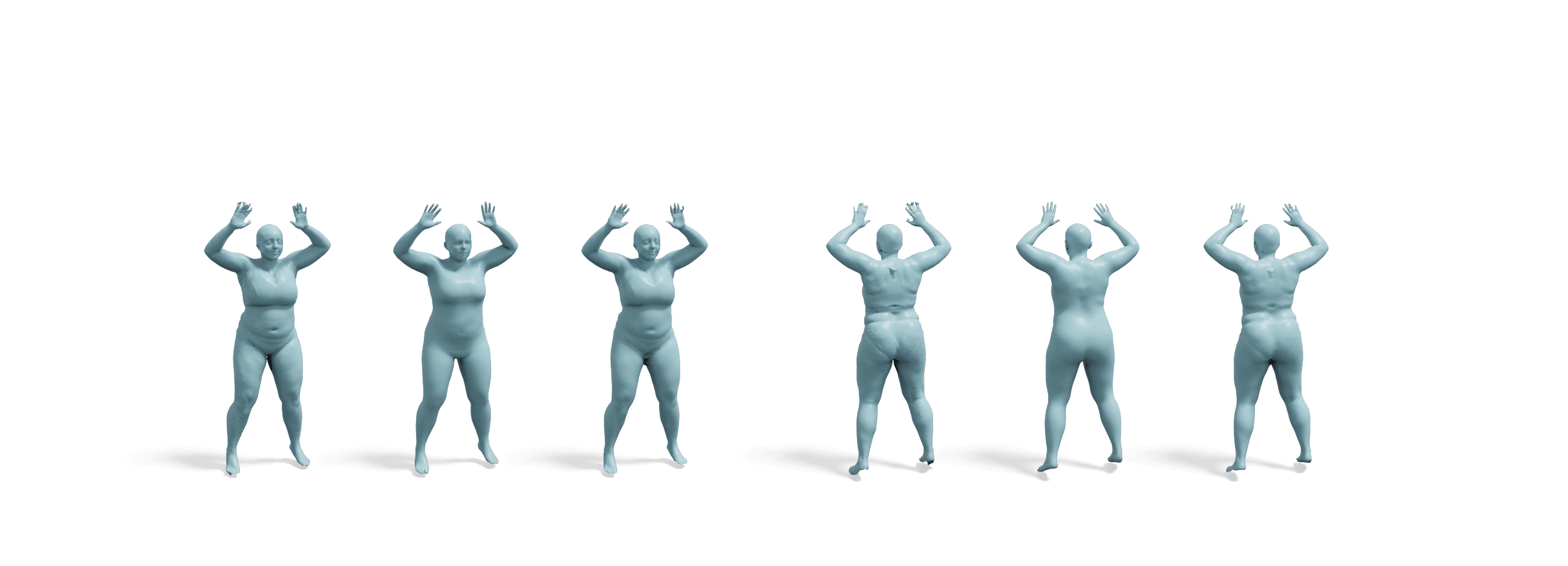}
		\put(12,99){}
	\end{overpic}
  \begin{overpic}[trim=6cm 4cm 12cm 12cm,clip, width=\linewidth]{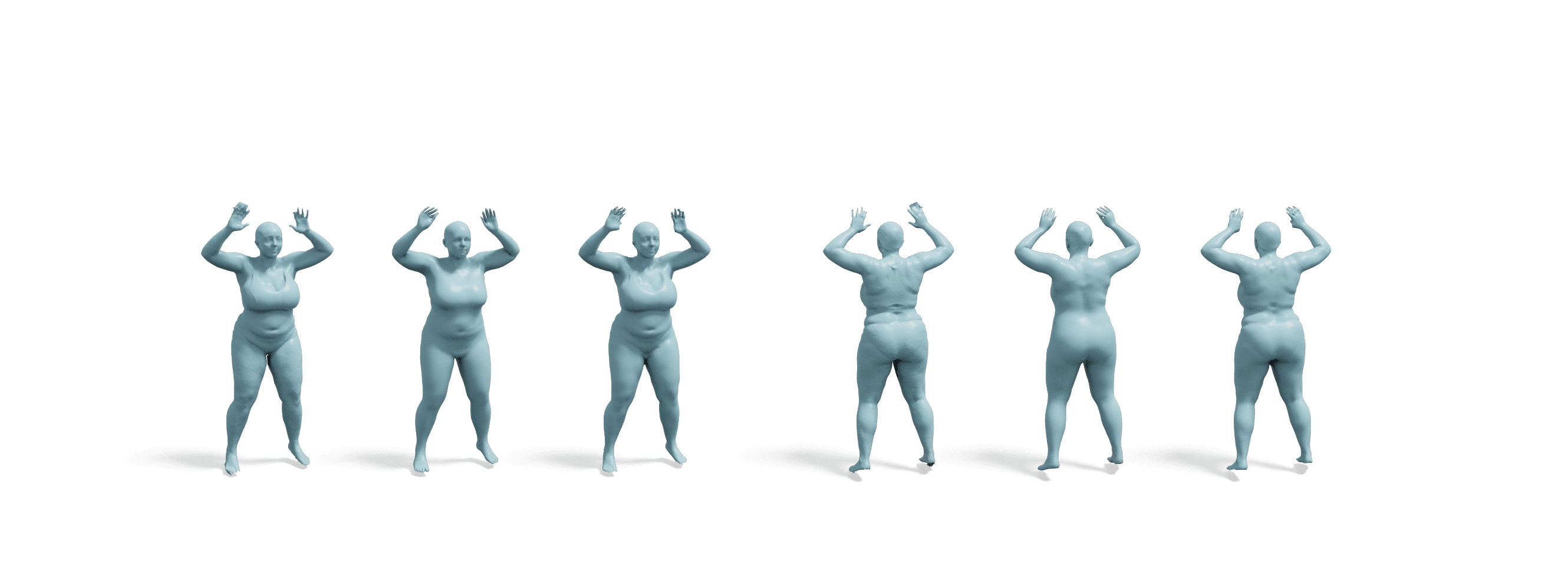}
		\put(12,99){}
	\end{overpic}
   \begin{overpic}[trim=6cm 4cm 12cm 12cm,clip, width=\linewidth]{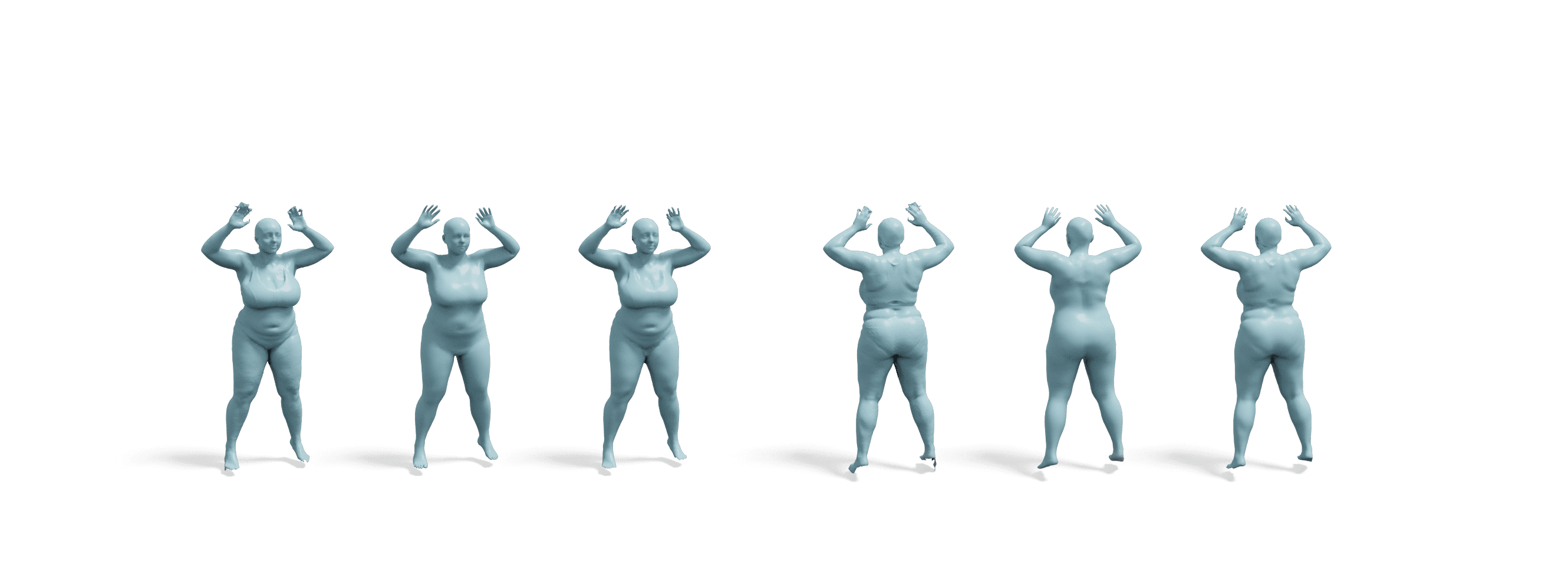}
		\put(12,99){}
	\end{overpic}

\end{figure*}

\begin{figure*}

    \centering
    \footnotesize
 \begin{overpic}[trim=7cm 4cm 12cm 12cm,clip, width=\linewidth]{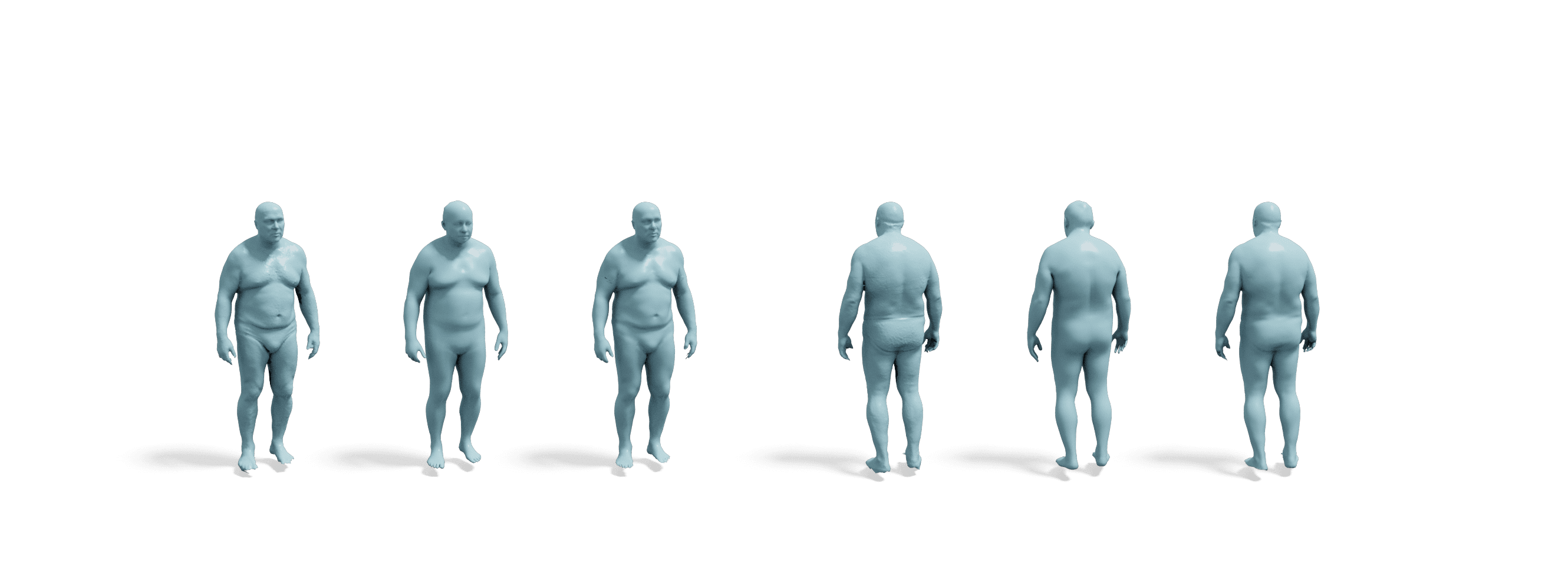}
 \put(40,28){\huge DFAUST}
		\put(12,24){Input}
        \put(31,24){\pipeline{}}

		\put(60,24){Input}
        \put(81,24){\pipeline{}}
	\end{overpic}
 \begin{overpic}[trim=6cm 4cm 12cm 12cm,clip, width=\linewidth]{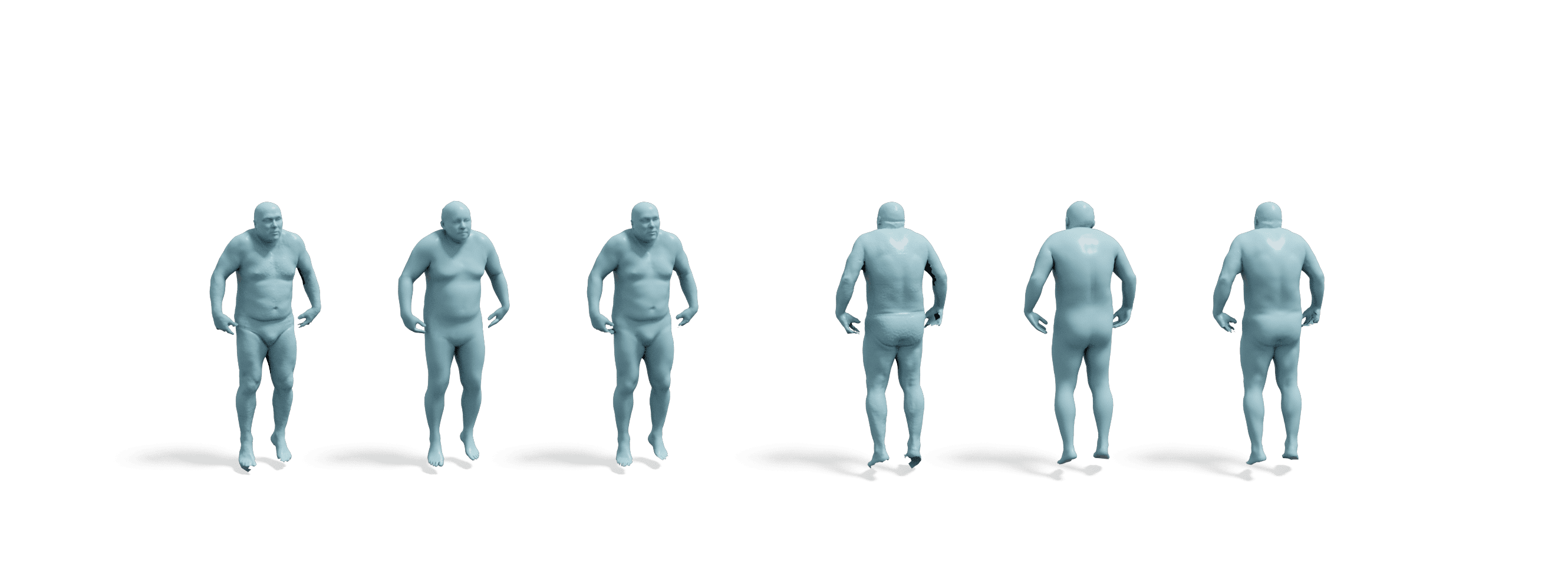}
		\put(12,99){}
	\end{overpic}
  \begin{overpic}[trim=6cm 4cm 12cm 12cm,clip, width=\linewidth]{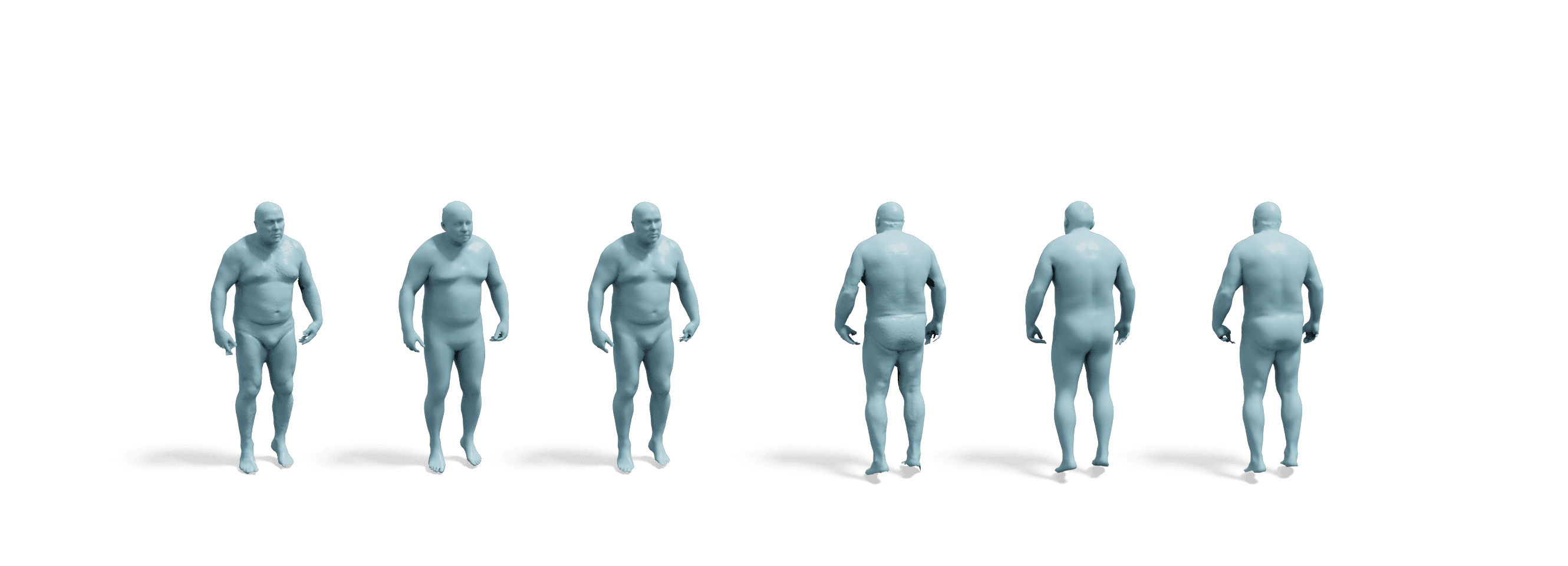}
		\put(12,99){}
	\end{overpic}
  \begin{overpic}[trim=6cm 4cm 12cm 12cm,clip, width=\linewidth]{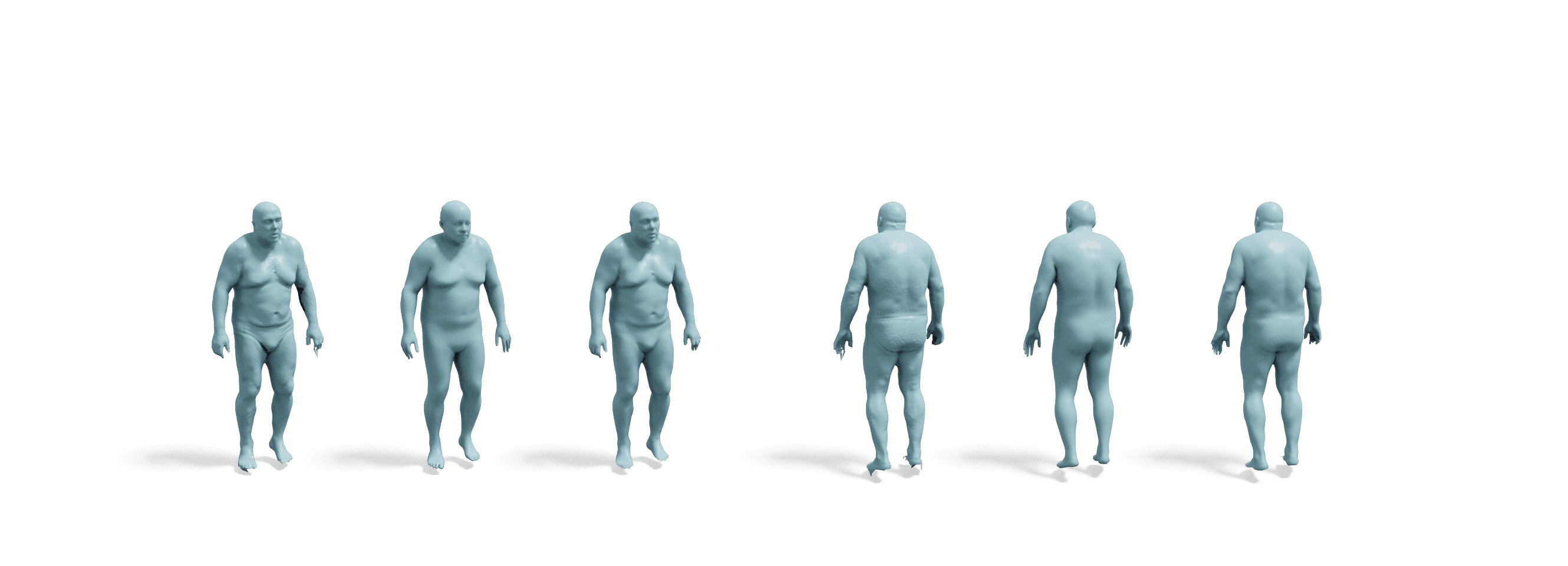}
		\put(12,99){}
	\end{overpic}
  \begin{overpic}[trim=6cm 4cm 12cm 12cm,clip, width=\linewidth]{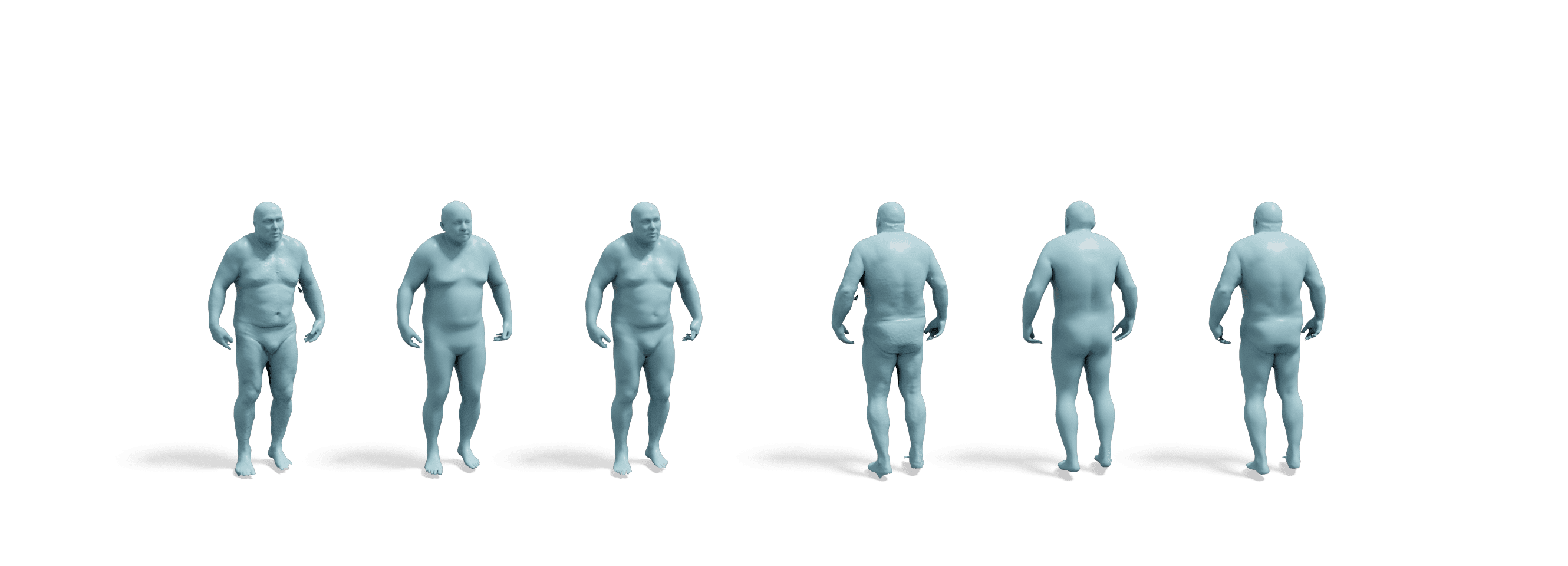}
		\put(12,99){}
	\end{overpic}
   \begin{overpic}[trim=6cm 4cm 12cm 12cm,clip, width=\linewidth]{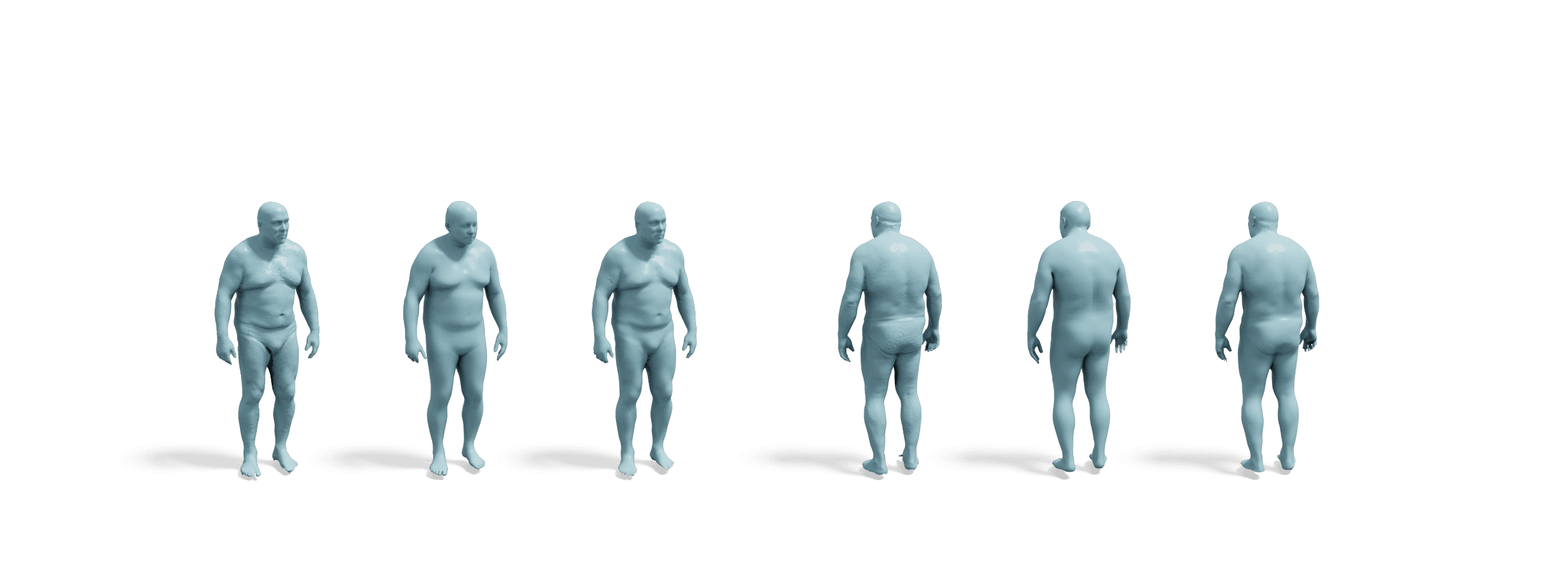}
		\put(12,99){}
	\end{overpic}
\end{figure*}

\begin{figure*}

    \centering
    \footnotesize
 \begin{overpic}[trim=7cm 4cm 12cm 12cm,clip, width=\linewidth]{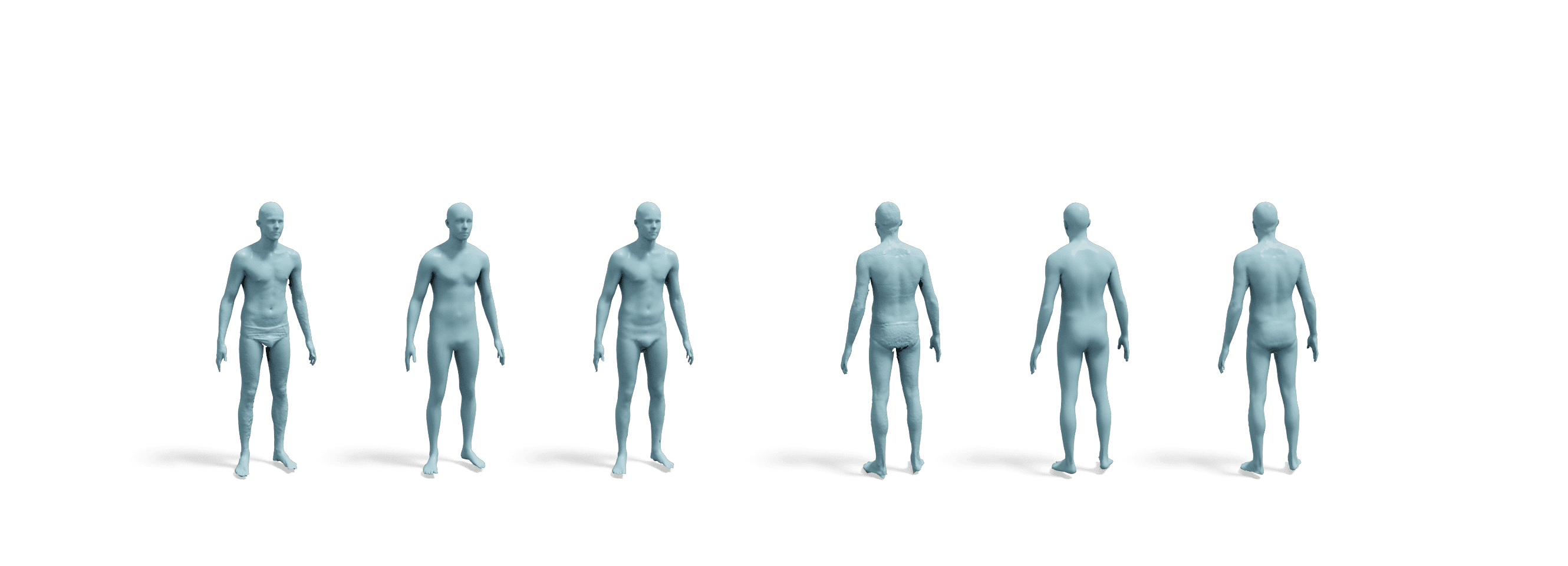}
 \put(40,28){\huge DFAUST}
		\put(12,24){Input}
        \put(31,24){\pipeline{}}

		\put(60,24){Input}
        \put(81,24){\pipeline{}}
	\end{overpic}
 \begin{overpic}[trim=6cm 4cm 12cm 12cm,clip, width=\linewidth]{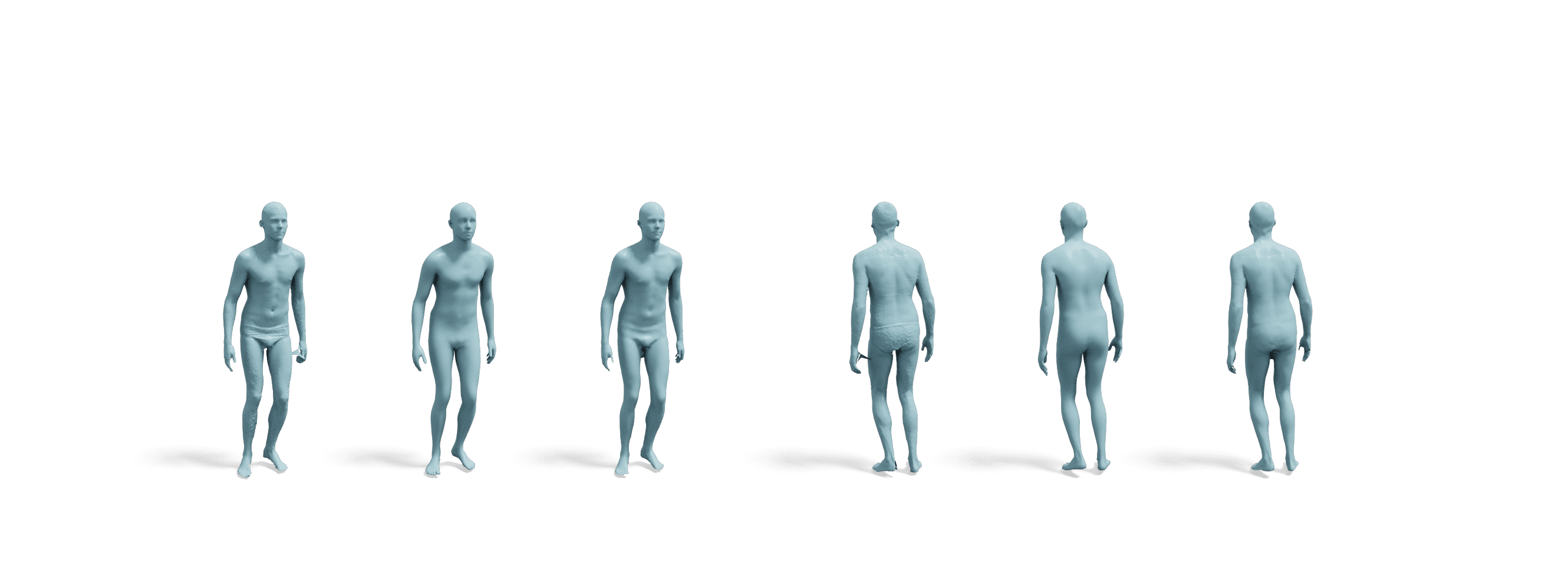}
		\put(12,99){}
	\end{overpic}
  \begin{overpic}[trim=6cm 4cm 12cm 12cm,clip, width=\linewidth]{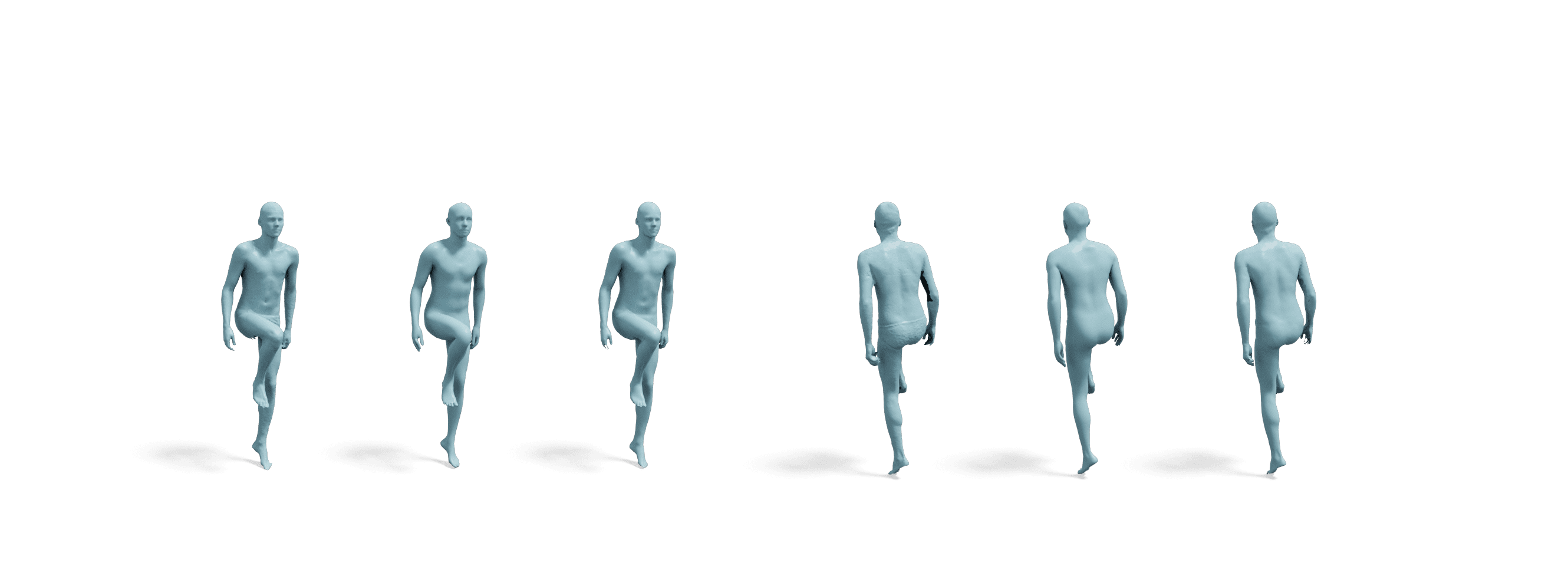}
		\put(12,99){}
	\end{overpic}
  \begin{overpic}[trim=6cm 4cm 12cm 12cm,clip, width=\linewidth]{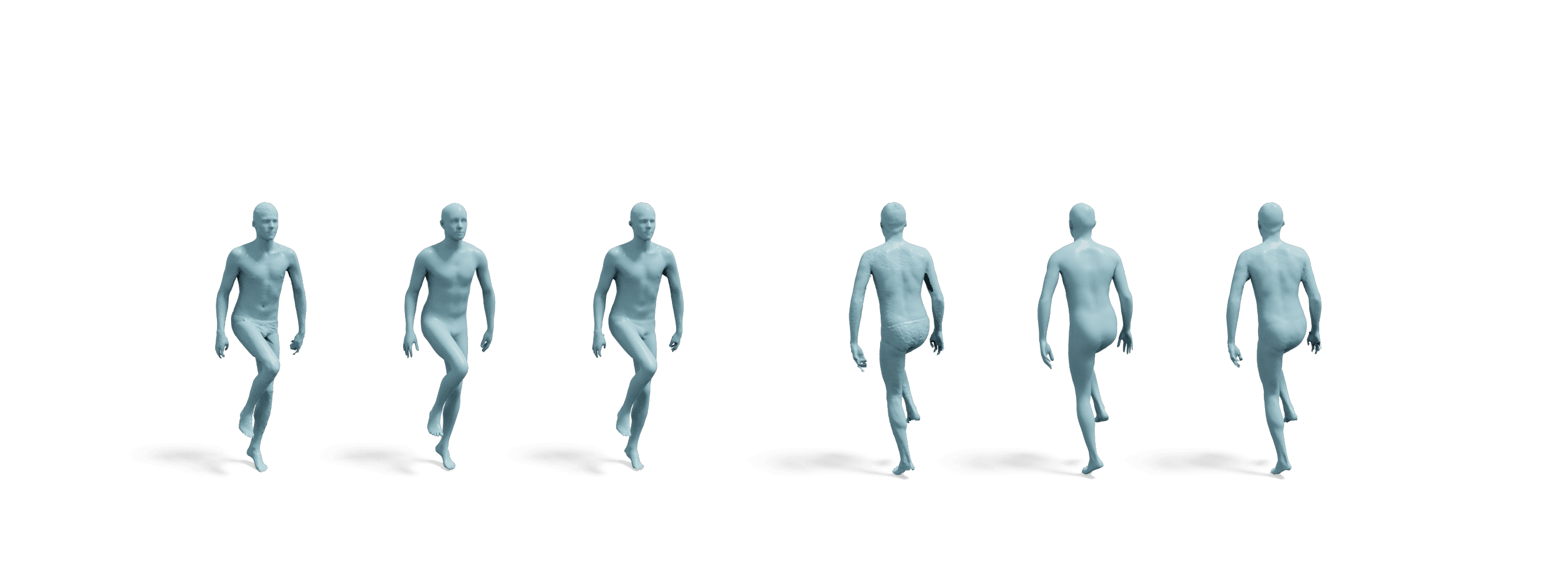}
		\put(12,99){}
	\end{overpic}
  \begin{overpic}[trim=6cm 4cm 12cm 12cm,clip, width=\linewidth]{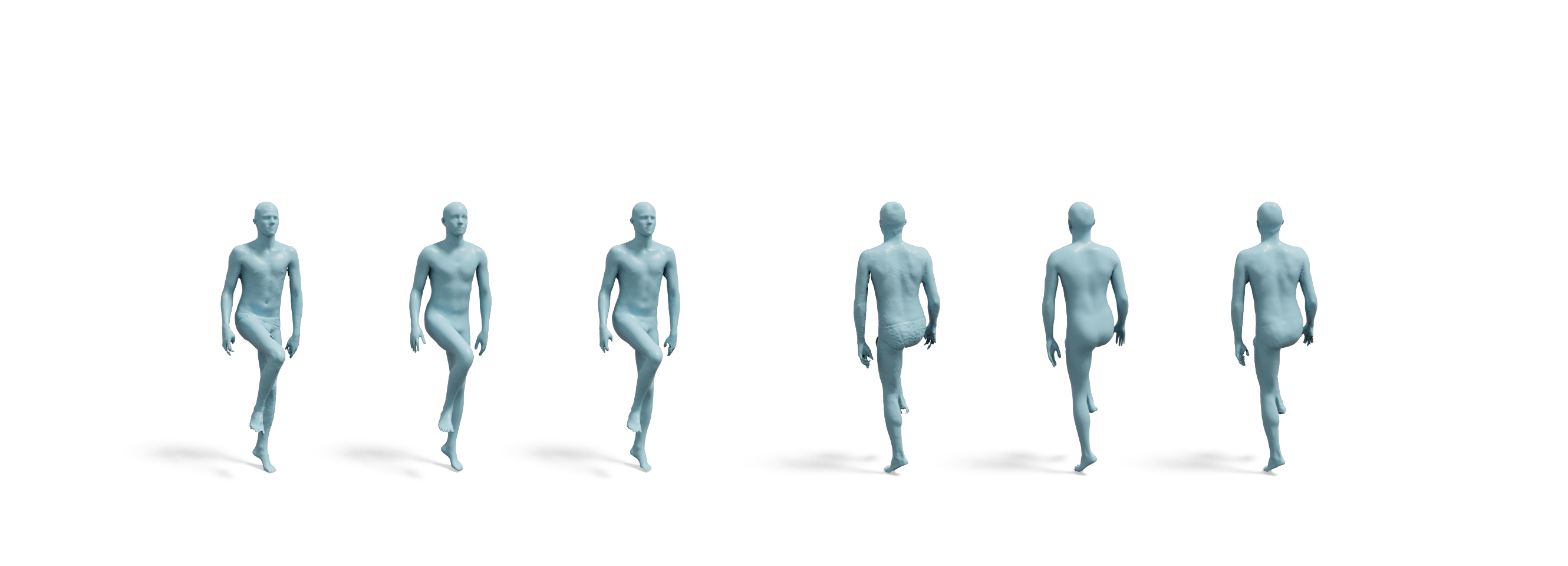}
		\put(12,99){}
	\end{overpic}
   \begin{overpic}[trim=6cm 4cm 12cm 12cm,clip, width=\linewidth]{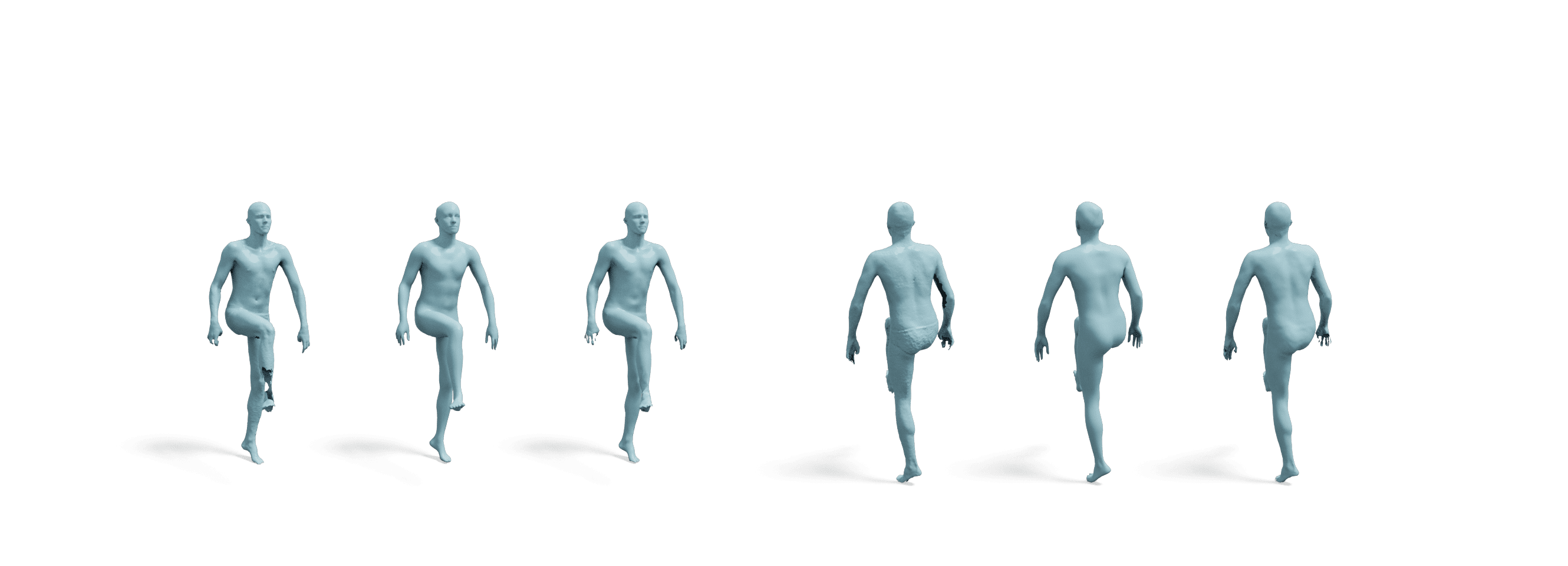}
		\put(12,99){}
	\end{overpic}
\end{figure*}

\begin{figure*}

    \centering
    \footnotesize
 \begin{overpic}[trim=7cm 4cm 12cm 12cm,clip, width=\linewidth]{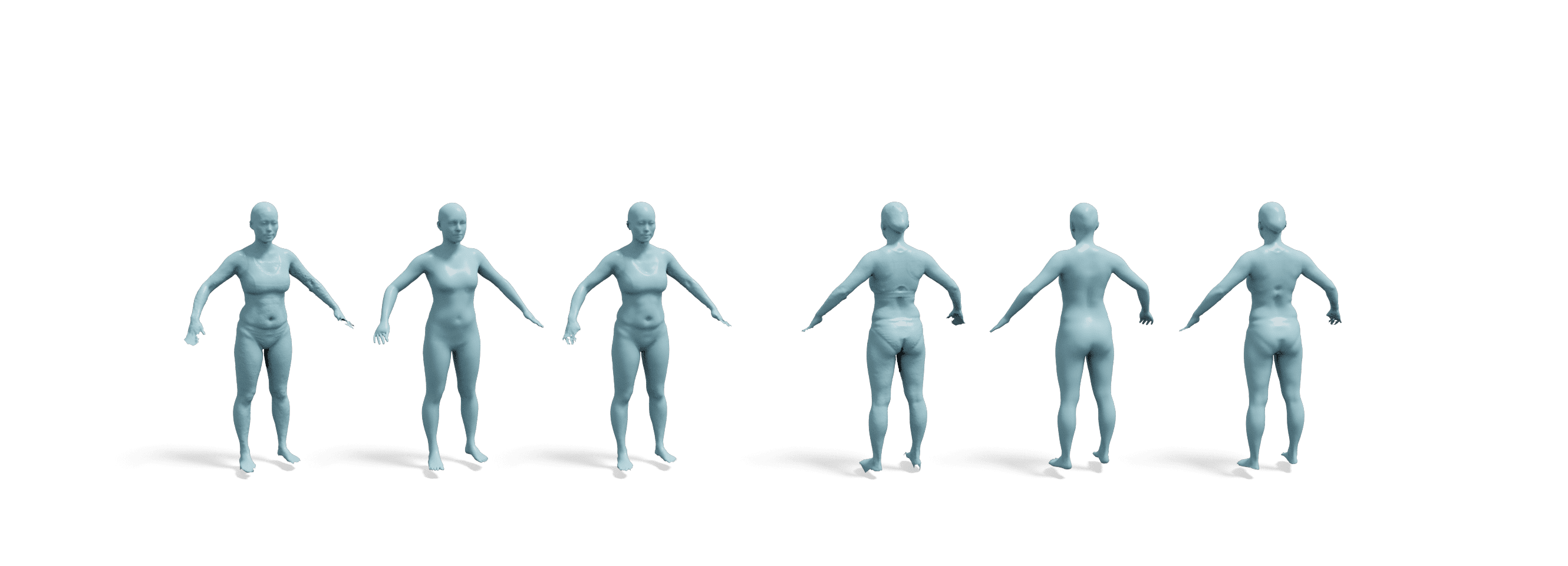}
 \put(40,28){\huge DFAUST}
		\put(12,24){Input}
        \put(31,24){\pipeline{}}

		\put(60,24){Input}
        \put(81,24){\pipeline{}}
	\end{overpic}
 \begin{overpic}[trim=6cm 4cm 12cm 12cm,clip, width=\linewidth]{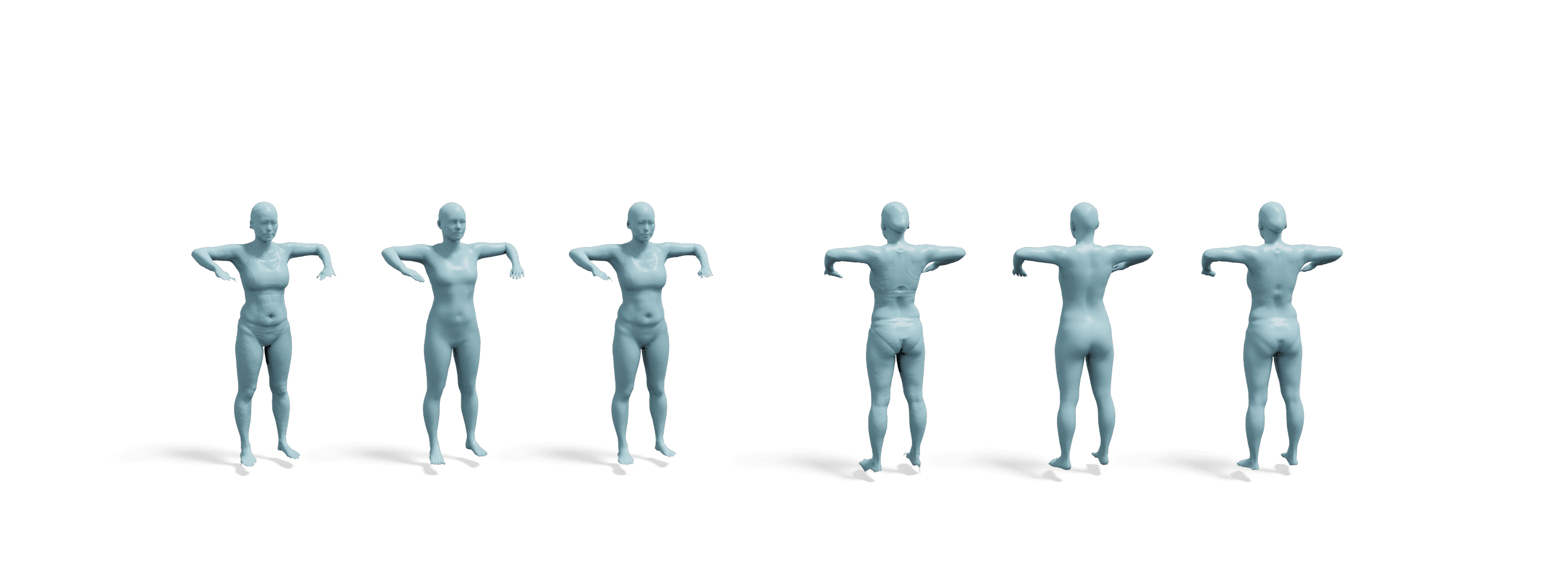}
		\put(12,99){}
	\end{overpic}
  \begin{overpic}[trim=6cm 4cm 12cm 12cm,clip, width=\linewidth]{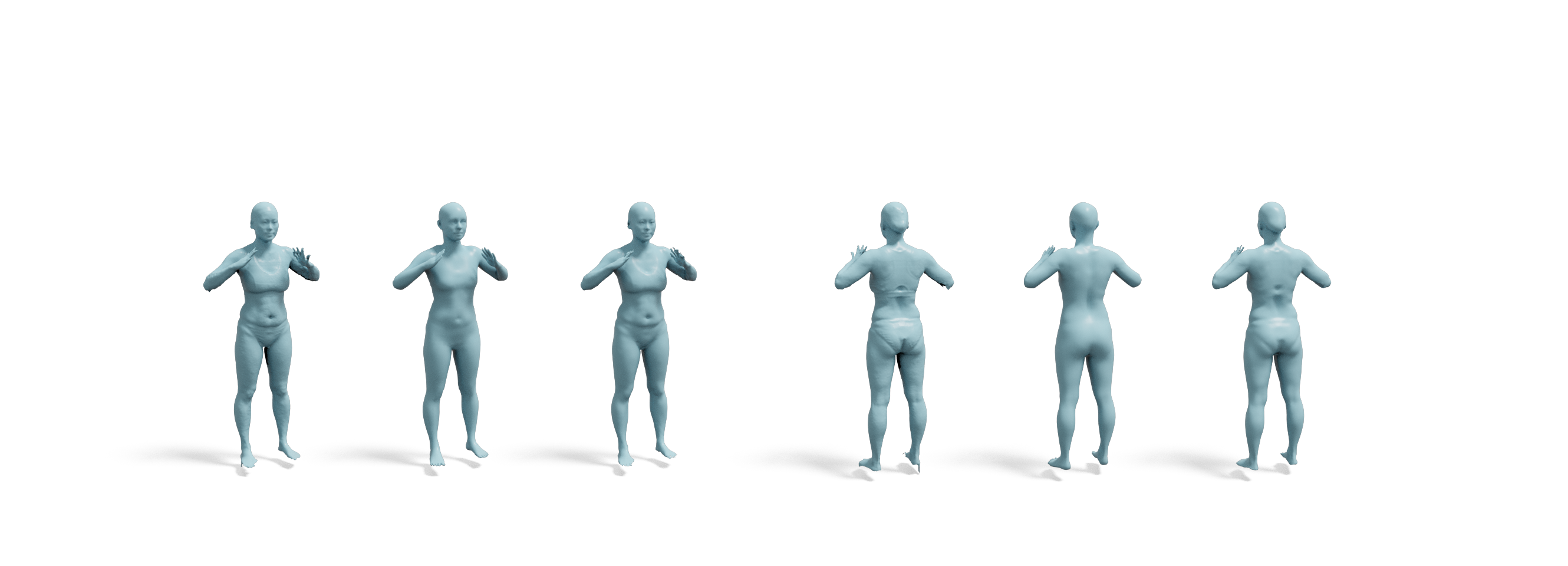}
		\put(12,99){}
	\end{overpic}
  \begin{overpic}[trim=6cm 4cm 12cm 12cm,clip, width=\linewidth]{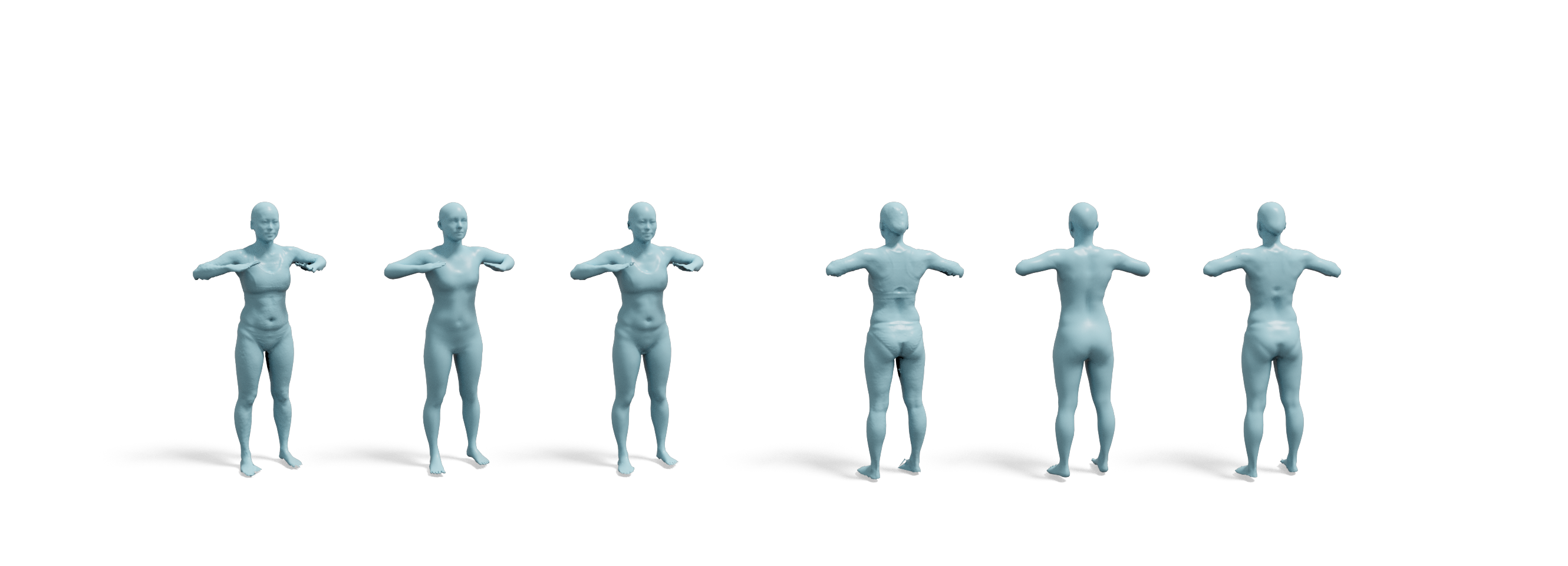}
		\put(12,99){}
	\end{overpic}
  \begin{overpic}[trim=6cm 4cm 12cm 12cm,clip, width=\linewidth]{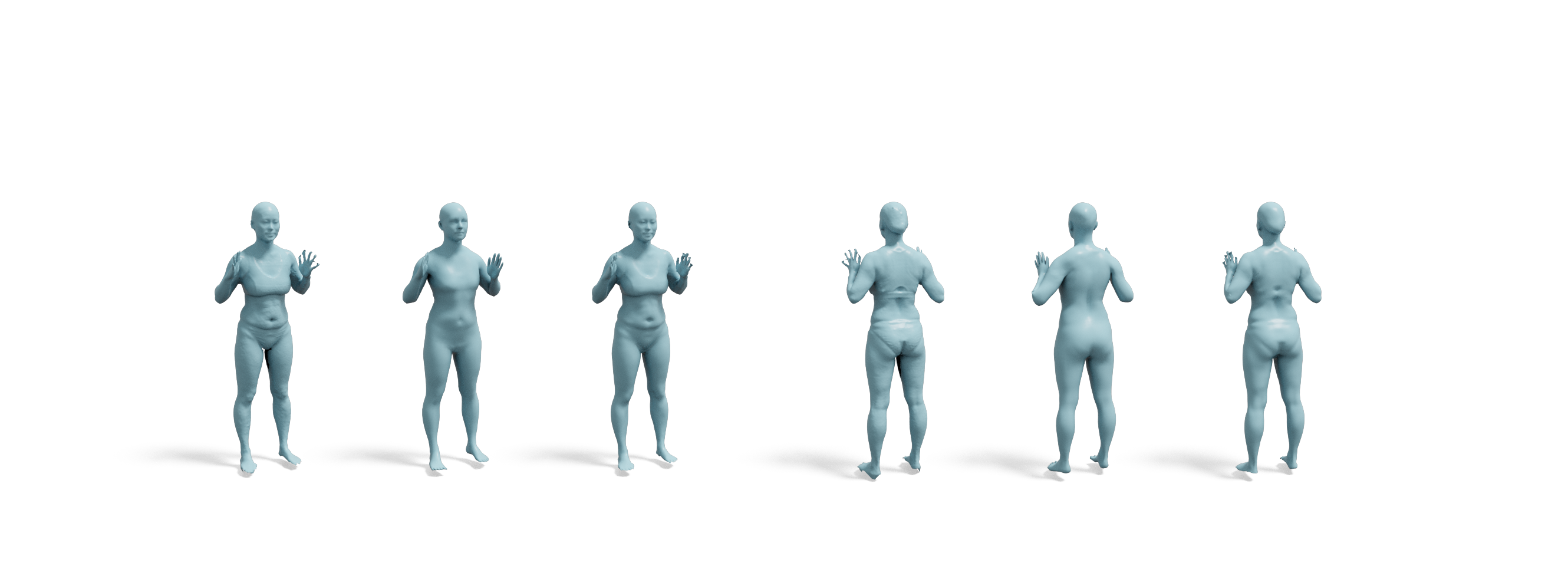}
		\put(12,99){}
	\end{overpic}
\end{figure*}

\begin{figure*}

    \centering
    \footnotesize
 \begin{overpic}[trim=7cm 4cm 12cm 12cm,clip, width=\linewidth]{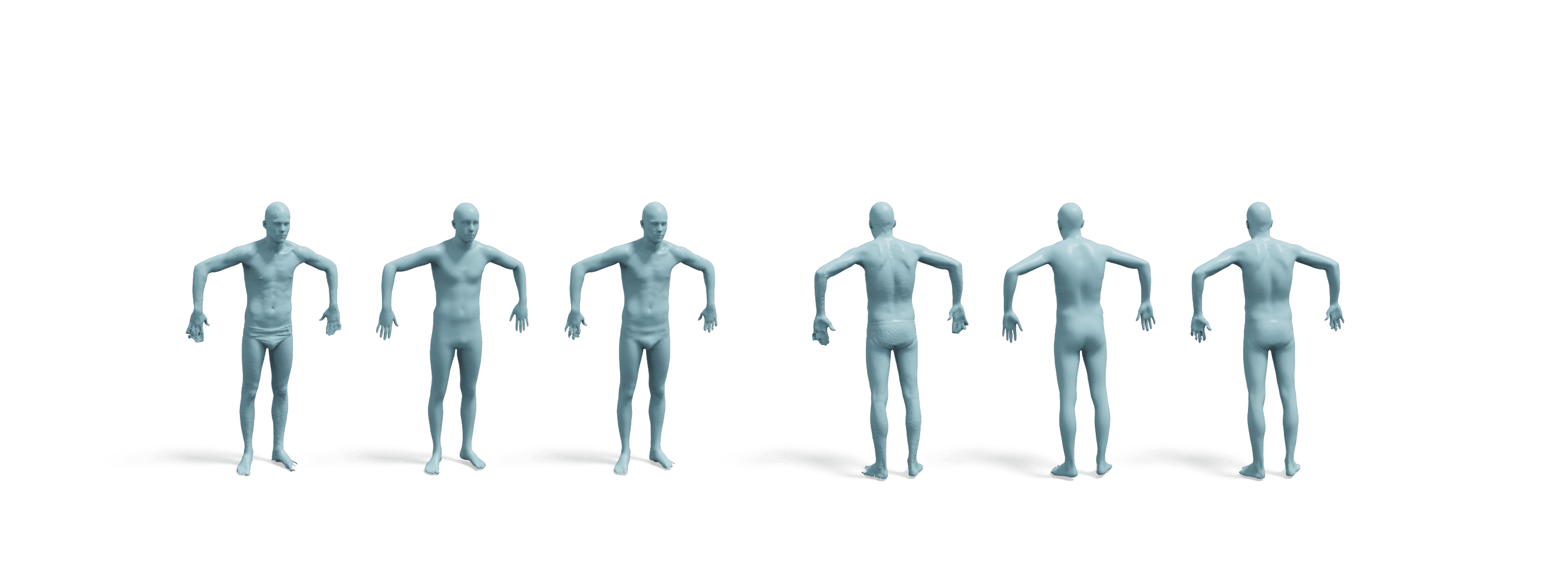}
 \put(37,30){\huge FAUST test set}
		\put(12,25){Input}
        \put(31,25){\pipeline{}}

		\put(60,25){Input}
        \put(81,25){\pipeline{}}
	\end{overpic}
 \begin{overpic}[trim=6cm 4cm 12cm 12cm,clip, width=\linewidth]{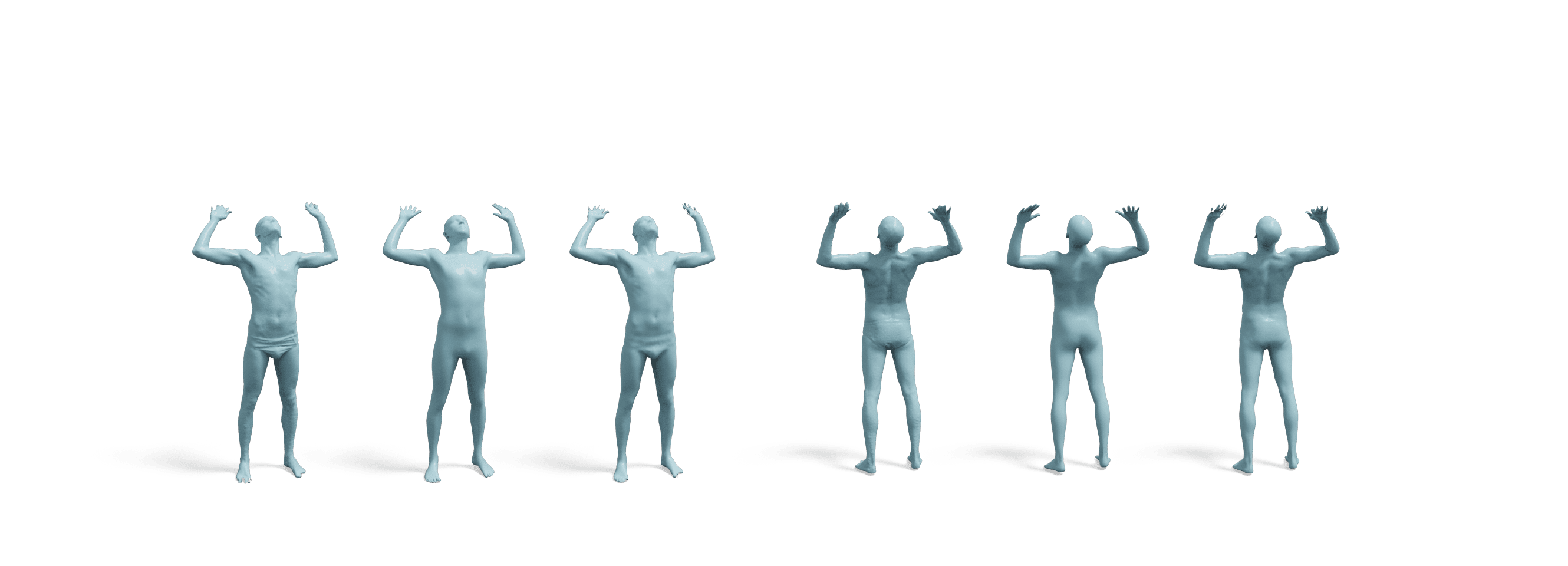}
		\put(12,99){}
	\end{overpic}
  \begin{overpic}[trim=6cm 4cm 12cm 12cm,clip, width=\linewidth]{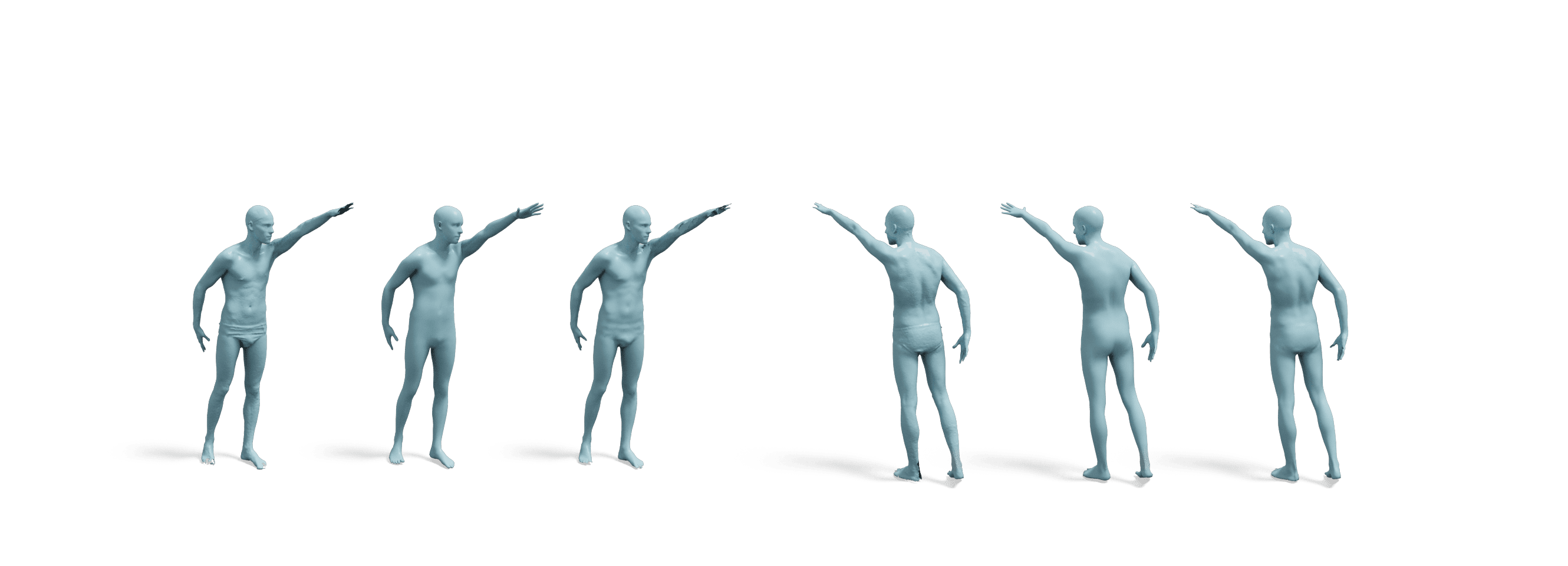}
		\put(12,99){}
	\end{overpic}
  \begin{overpic}[trim=6cm 4cm 12cm 12cm,clip, width=\linewidth]{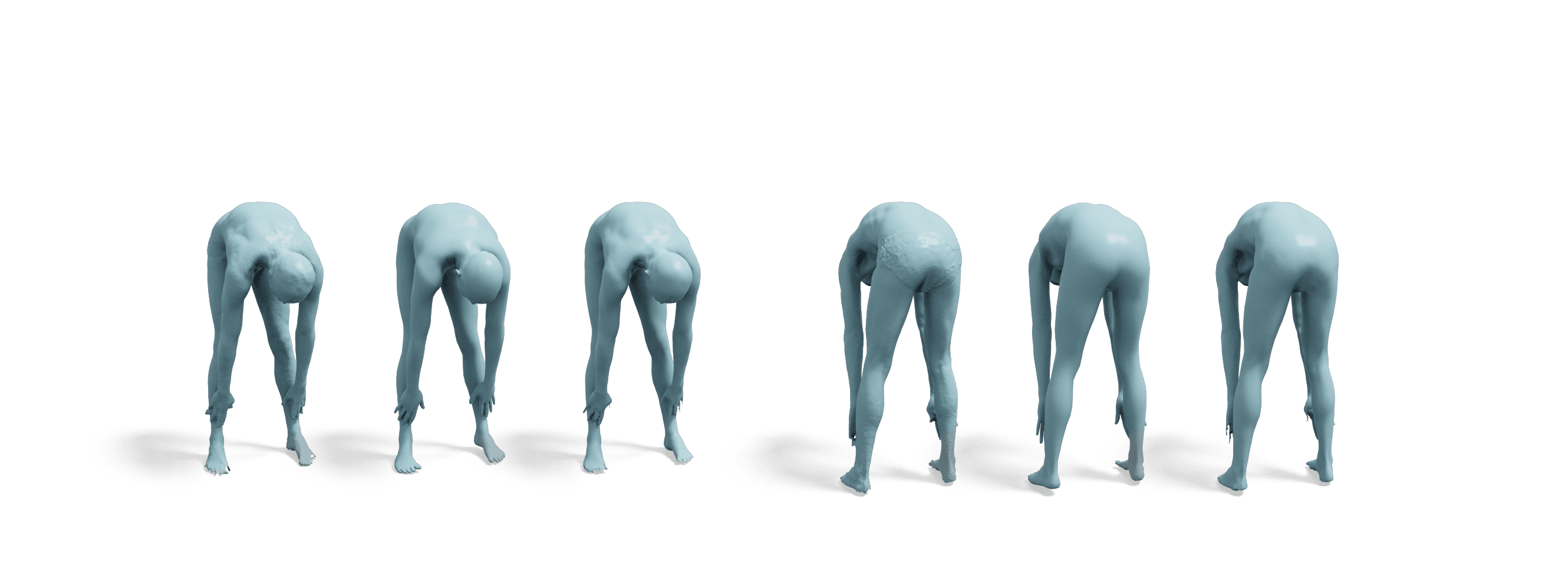}
		\put(12,99){}
	\end{overpic}
  \begin{overpic}[trim=6cm 4cm 12cm 12cm,clip, width=\linewidth]{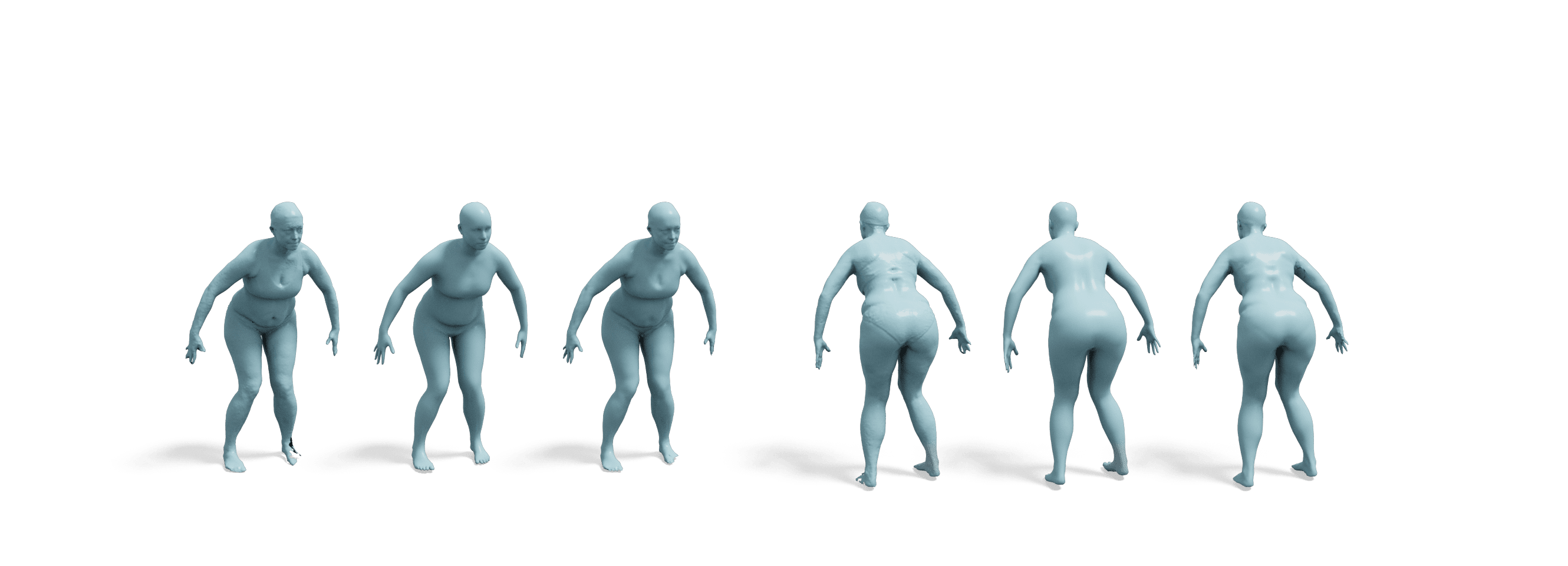}
		\put(12,99){}
	\end{overpic}
\end{figure*}

\begin{figure*}

    \centering
    \footnotesize
 \begin{overpic}[trim=7cm 4cm 12cm 12cm,clip, width=\linewidth]{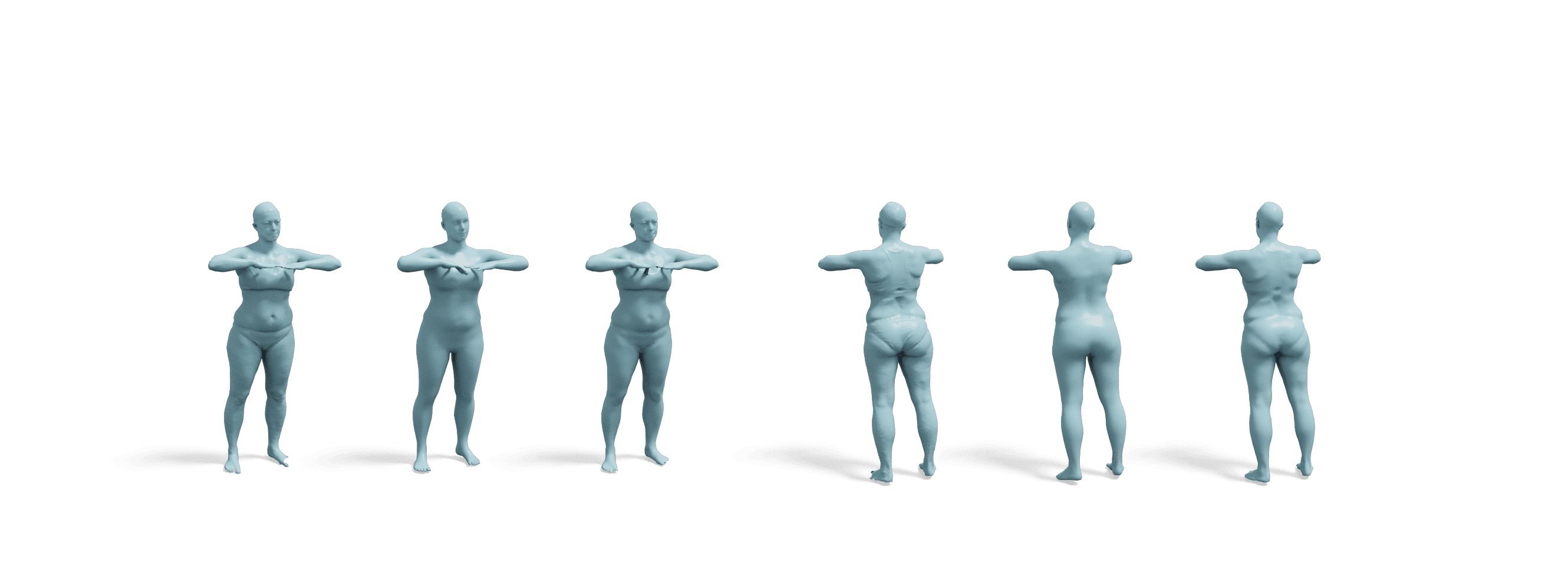}
  \put(37,30){\huge FAUST test set}
		\put(12,25){Input}
        \put(31,25){\pipeline{}}

		\put(60,25){Input}
        \put(81,25){\pipeline{}}
	\end{overpic}
 \begin{overpic}[trim=6cm 4cm 12cm 12cm,clip, width=\linewidth]{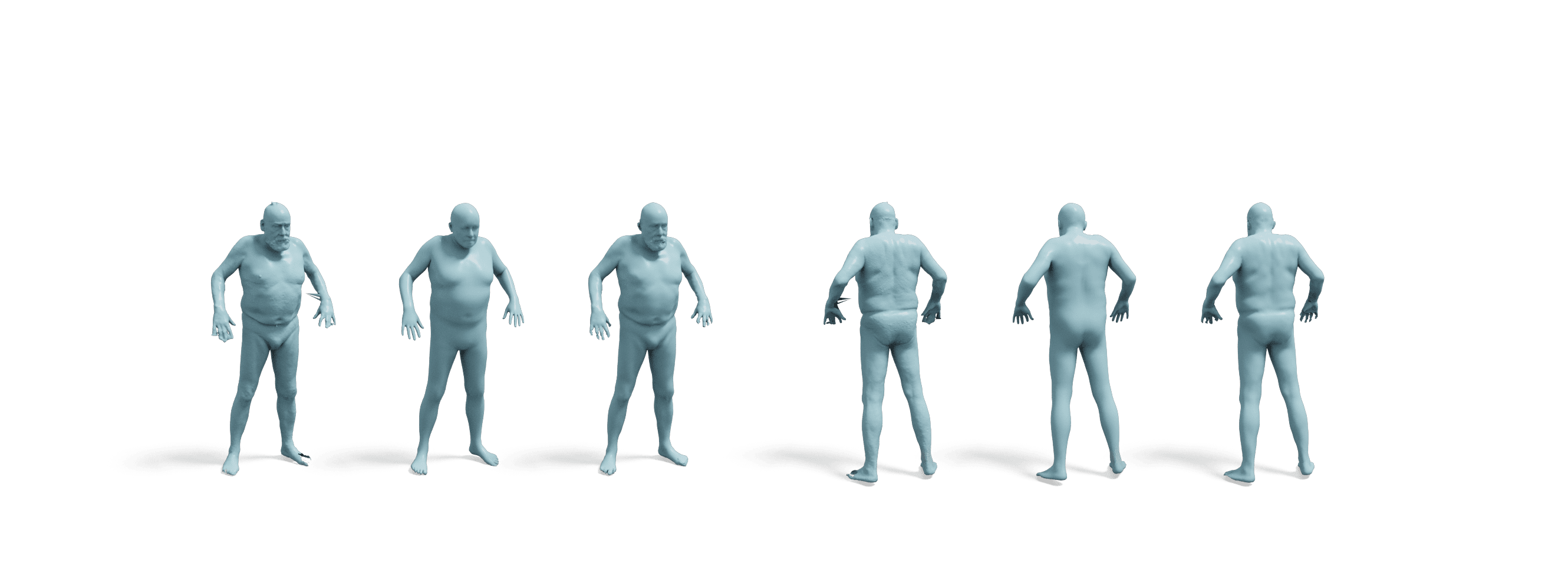}
		\put(12,99){}
	\end{overpic}
  \begin{overpic}[trim=6cm 4cm 12cm 12cm,clip, width=\linewidth]{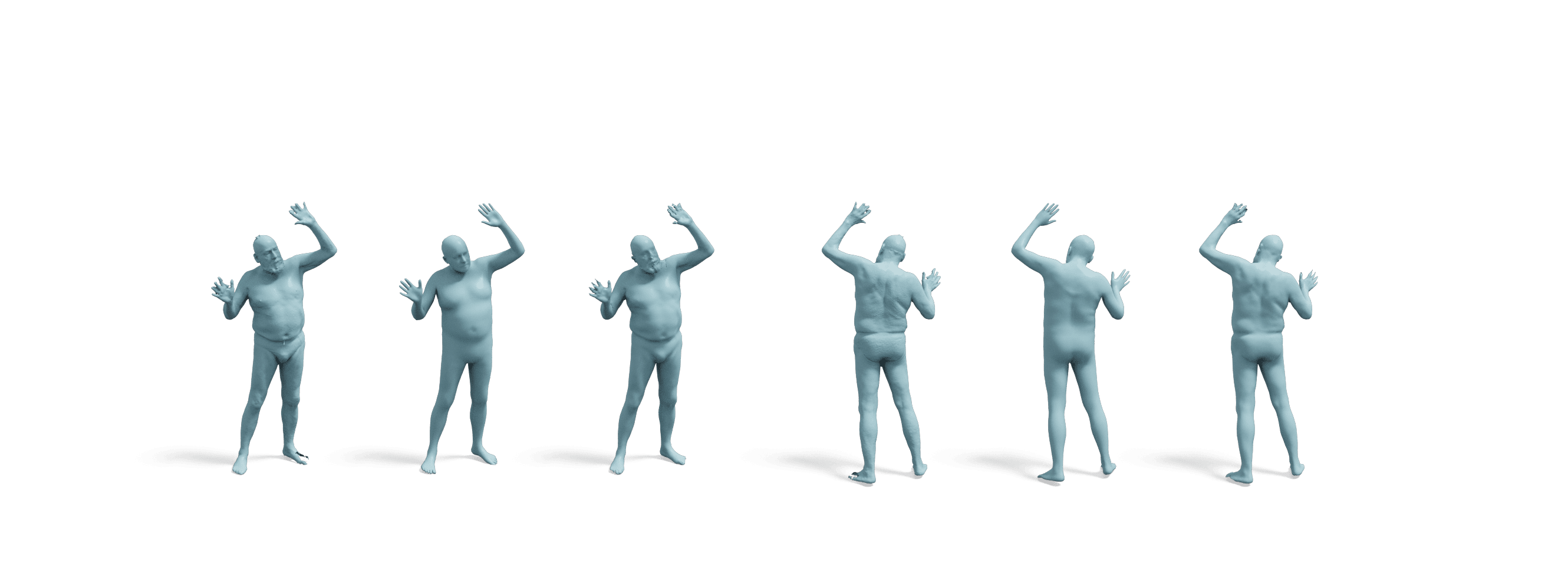}
		\put(12,99){}
	\end{overpic}
  \begin{overpic}[trim=6cm 4cm 12cm 12cm,clip, width=\linewidth]{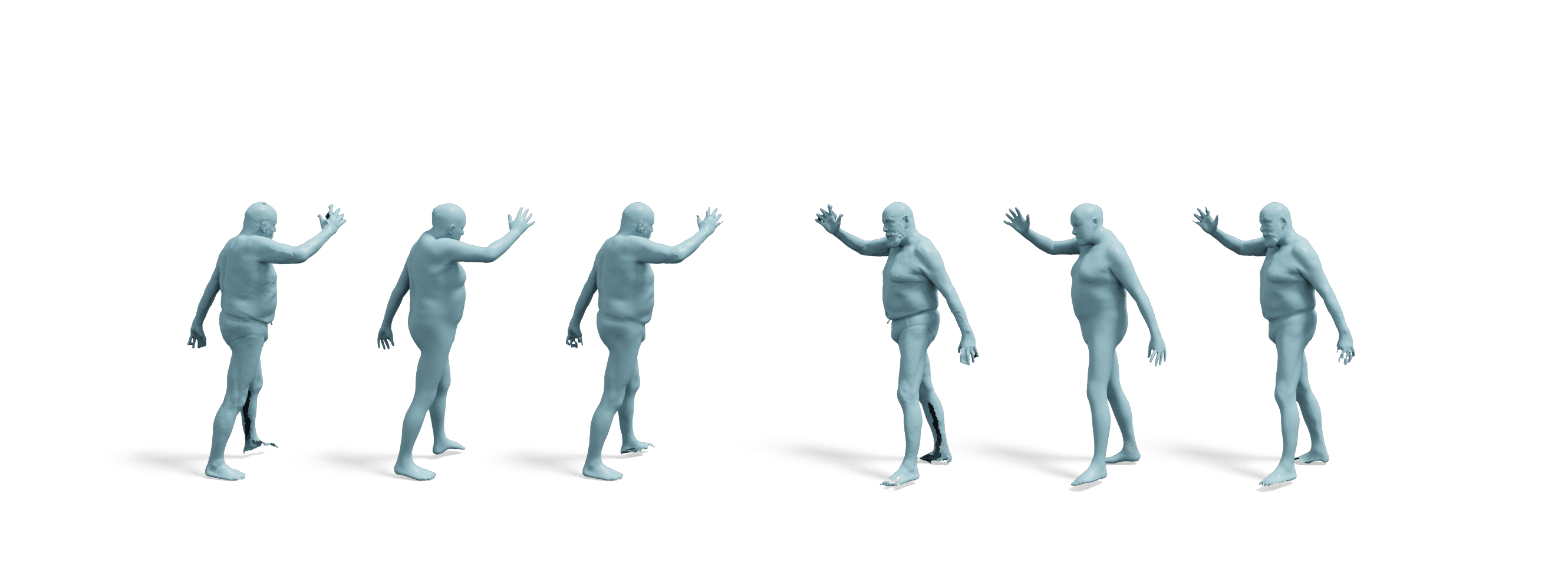}
		\put(12,99){}
	\end{overpic}
  \begin{overpic}[trim=6cm 4cm 12cm 12cm,clip, width=\linewidth]{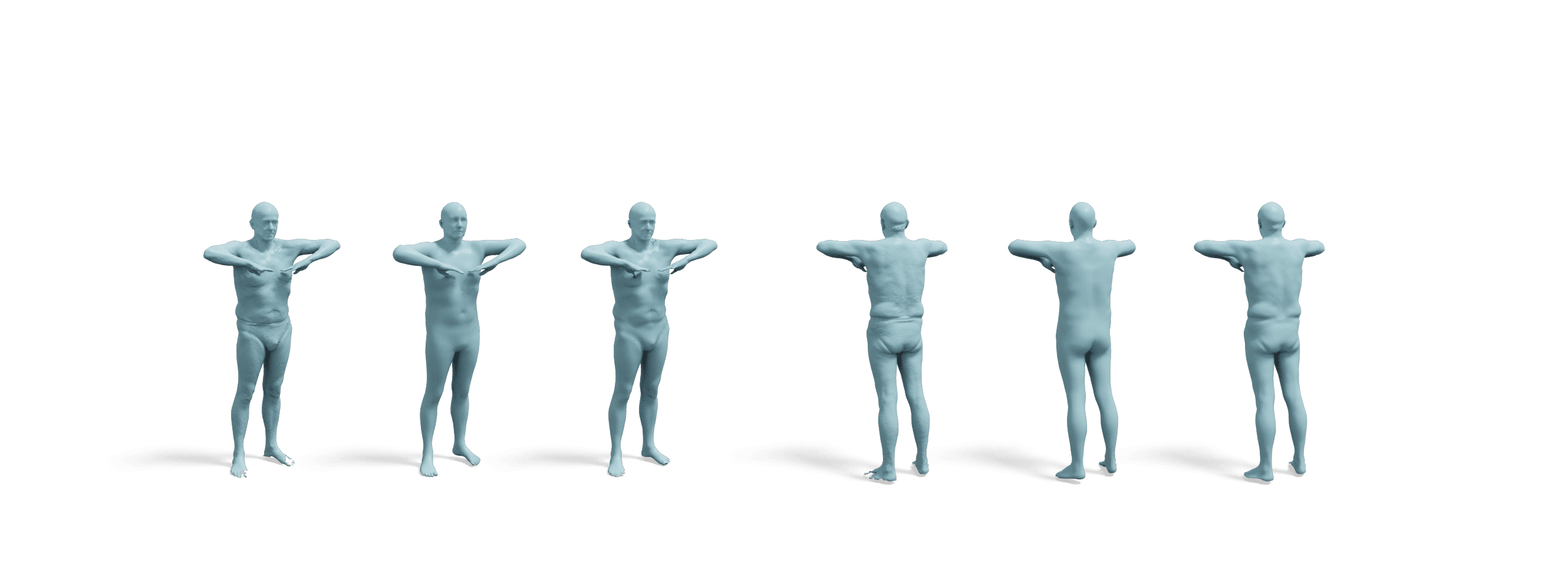}
		\put(12,99){}
	\end{overpic}
\end{figure*}

\begin{figure*}

    \centering
    \footnotesize
 \begin{overpic}[trim=7cm 4cm 12cm 12cm,clip, width=\linewidth]{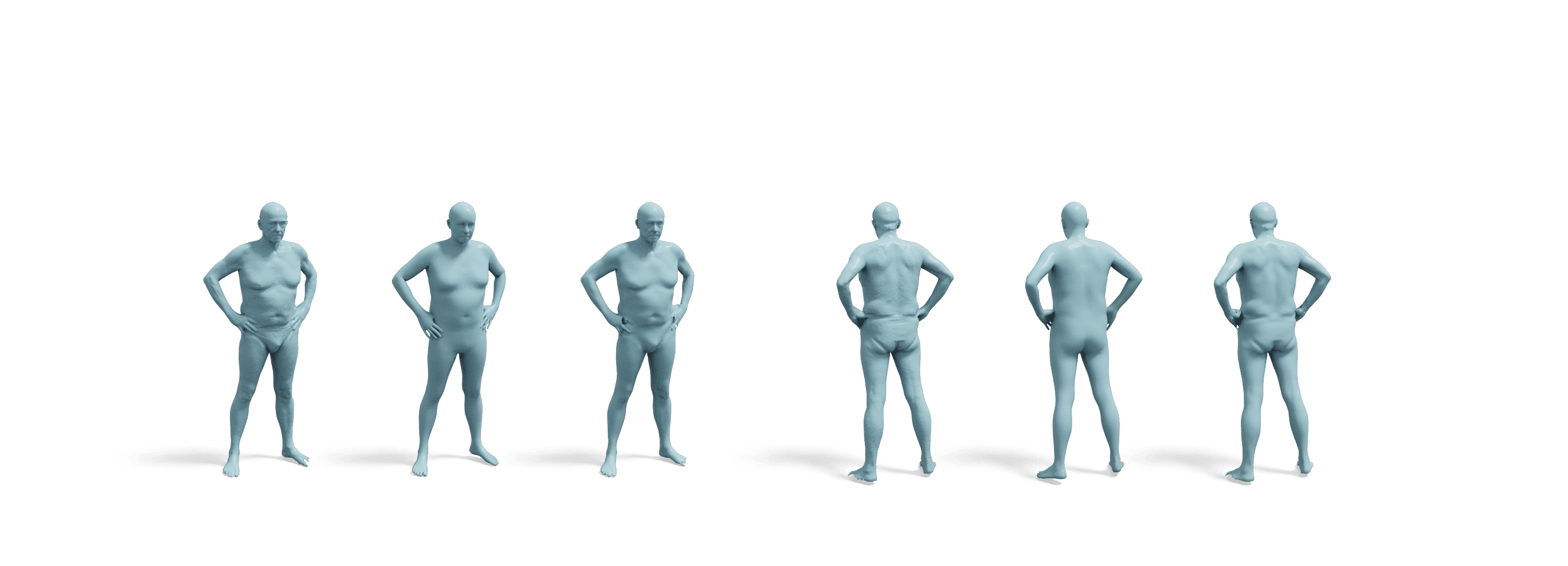}
  \put(37,30){\huge FAUST test set}
		\put(12,25){Input}
        \put(31,25){\pipeline{}}

		\put(60,25){Input}
        \put(81,25){\pipeline{}}
	\end{overpic}
 \begin{overpic}[trim=6cm 4cm 12cm 12cm,clip, width=\linewidth]{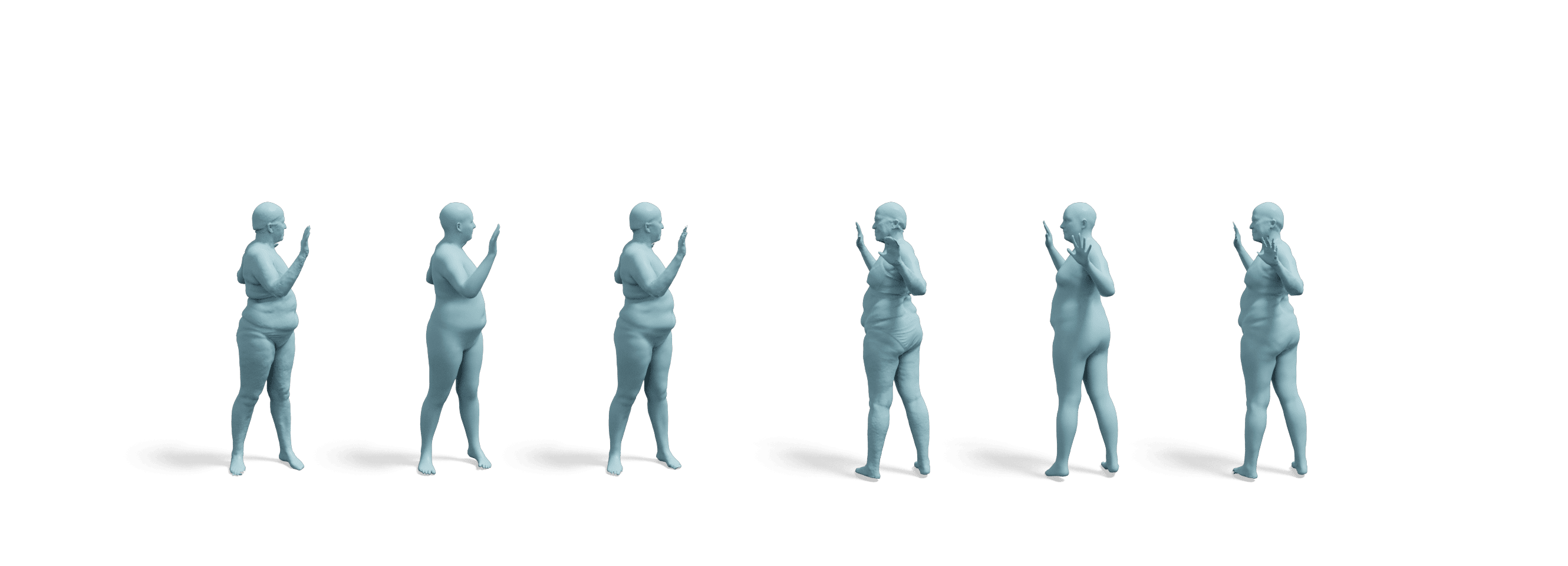}
		\put(12,99){}
	\end{overpic}
  \begin{overpic}[trim=6cm 4cm 12cm 12cm,clip, width=\linewidth]{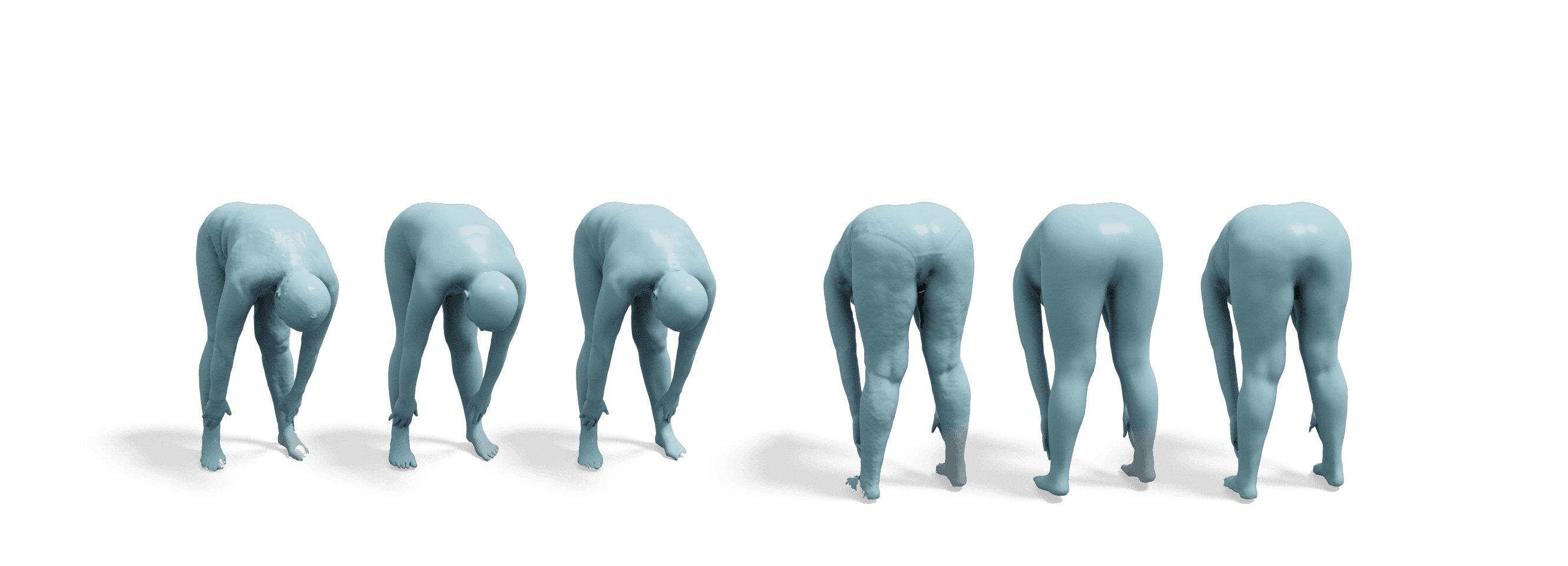}
		\put(12,99){}
	\end{overpic}
  \begin{overpic}[trim=6cm 4cm 12cm 12cm,clip, width=\linewidth]{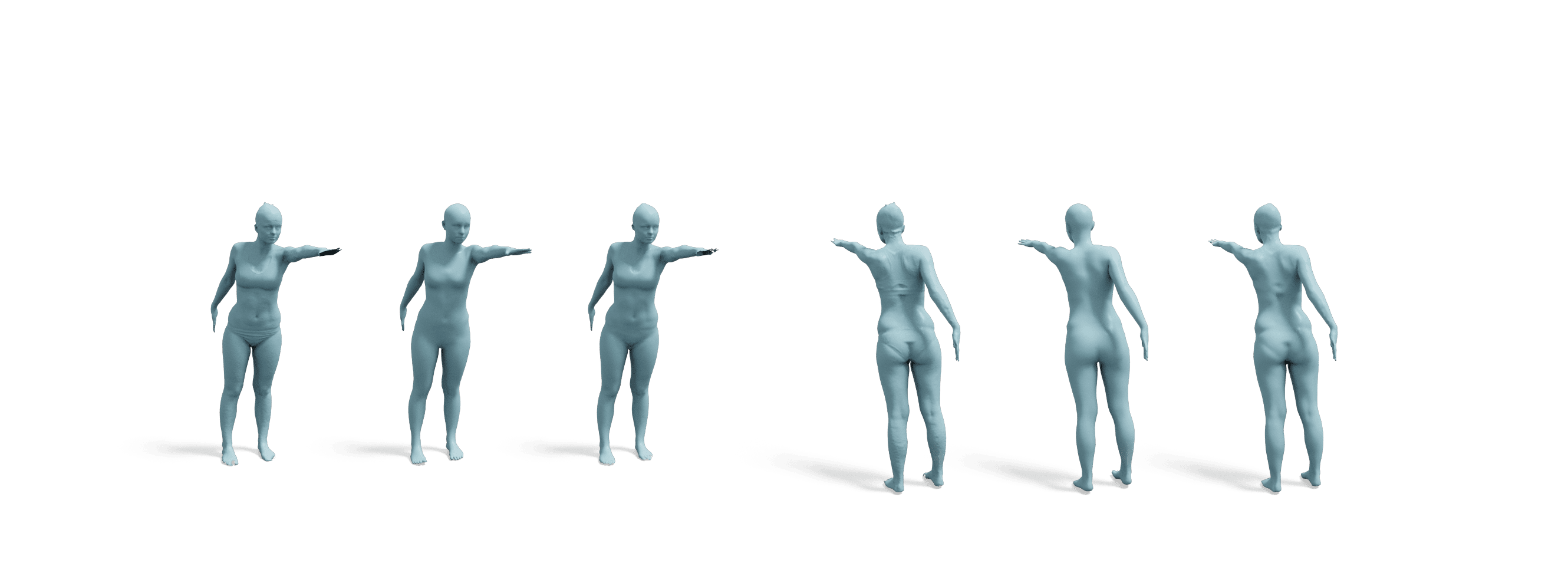}
		\put(12,99){}
	\end{overpic}
  \begin{overpic}[trim=6cm 4cm 12cm 12cm,clip, width=\linewidth]{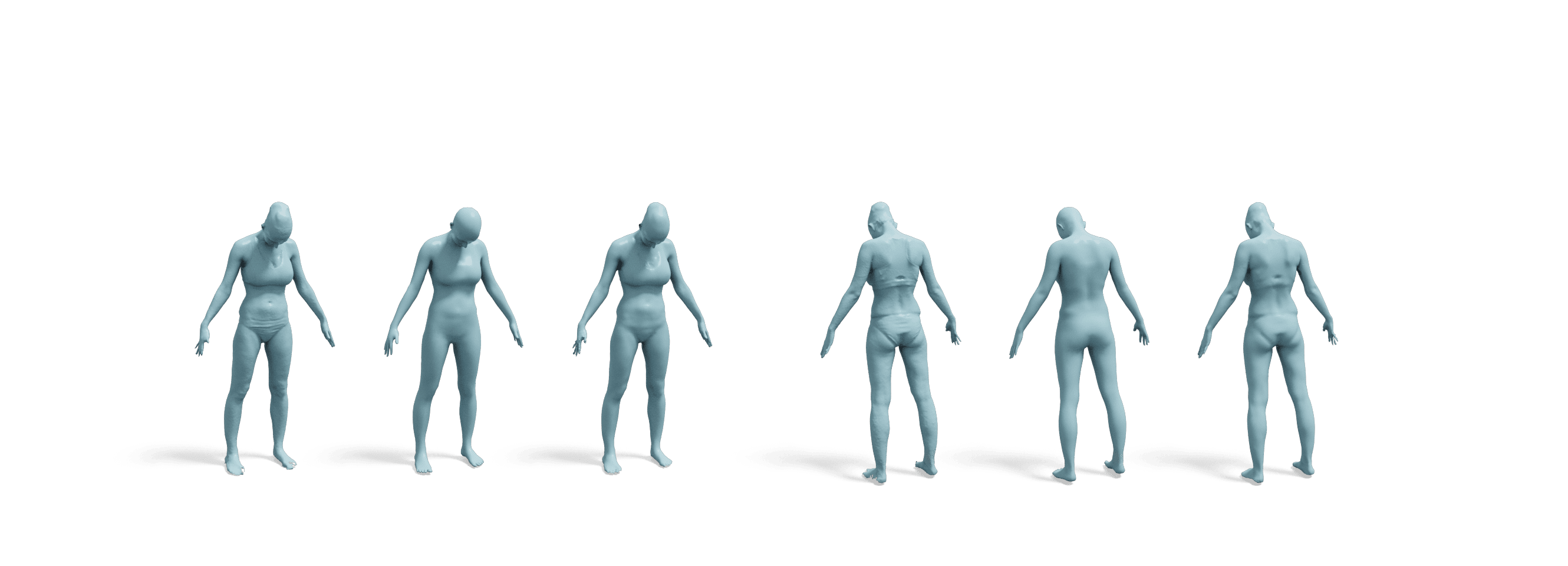}
		\put(12,99){}
	\end{overpic}
\end{figure*}

\begin{figure*}

    \centering
    \footnotesize
 \begin{overpic}[trim=7cm 4cm 12cm 12cm,clip, width=\linewidth]{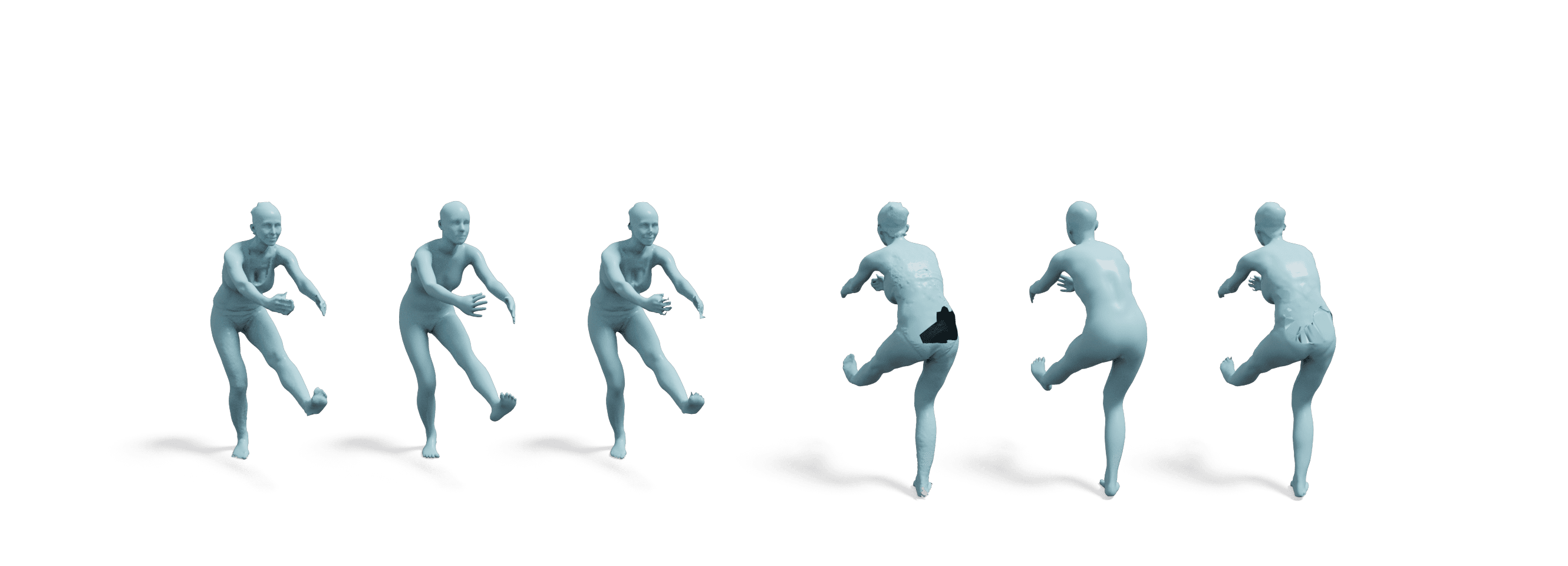}
  \put(37,30){\huge FAUST test set}
		\put(12,25){Input}
        \put(31,25){\pipeline{}}

		\put(60,25){Input}
        \put(81,25){\pipeline{}}
	\end{overpic}
 \begin{overpic}[trim=6cm 4cm 12cm 12cm,clip, width=\linewidth]{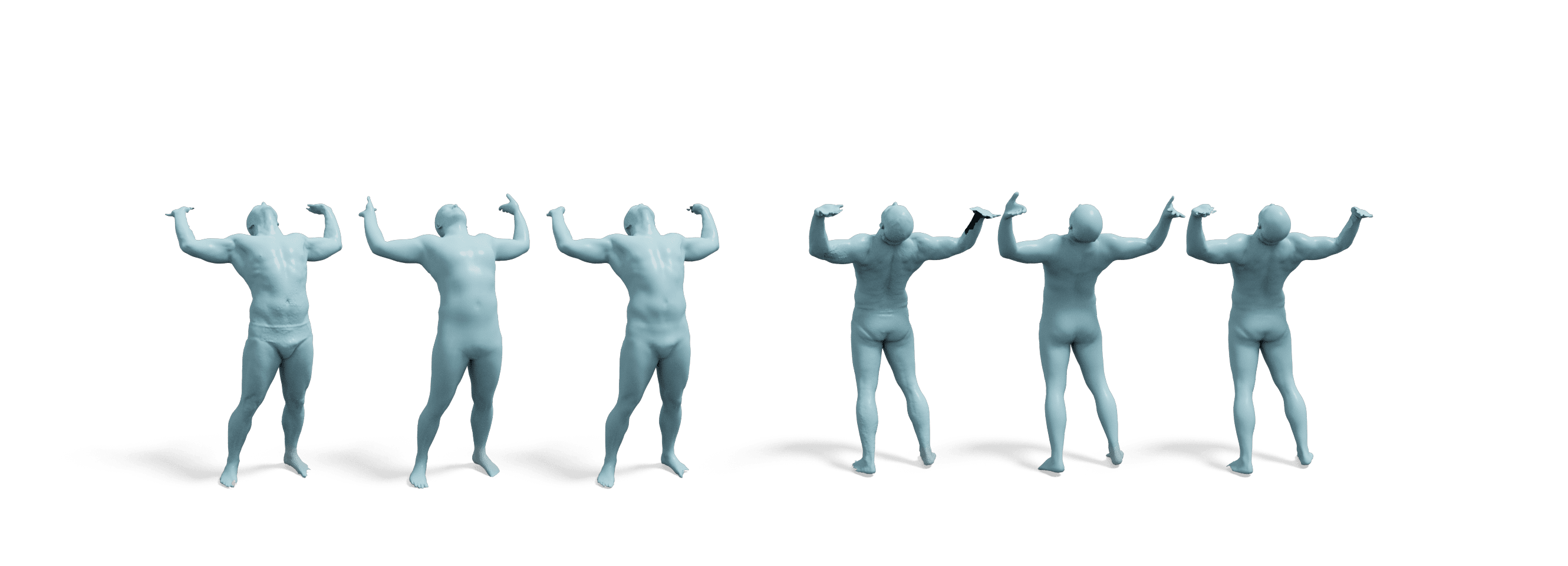}
		\put(12,99){}
	\end{overpic}
  \begin{overpic}[trim=6cm 4cm 12cm 12cm,clip, width=\linewidth]{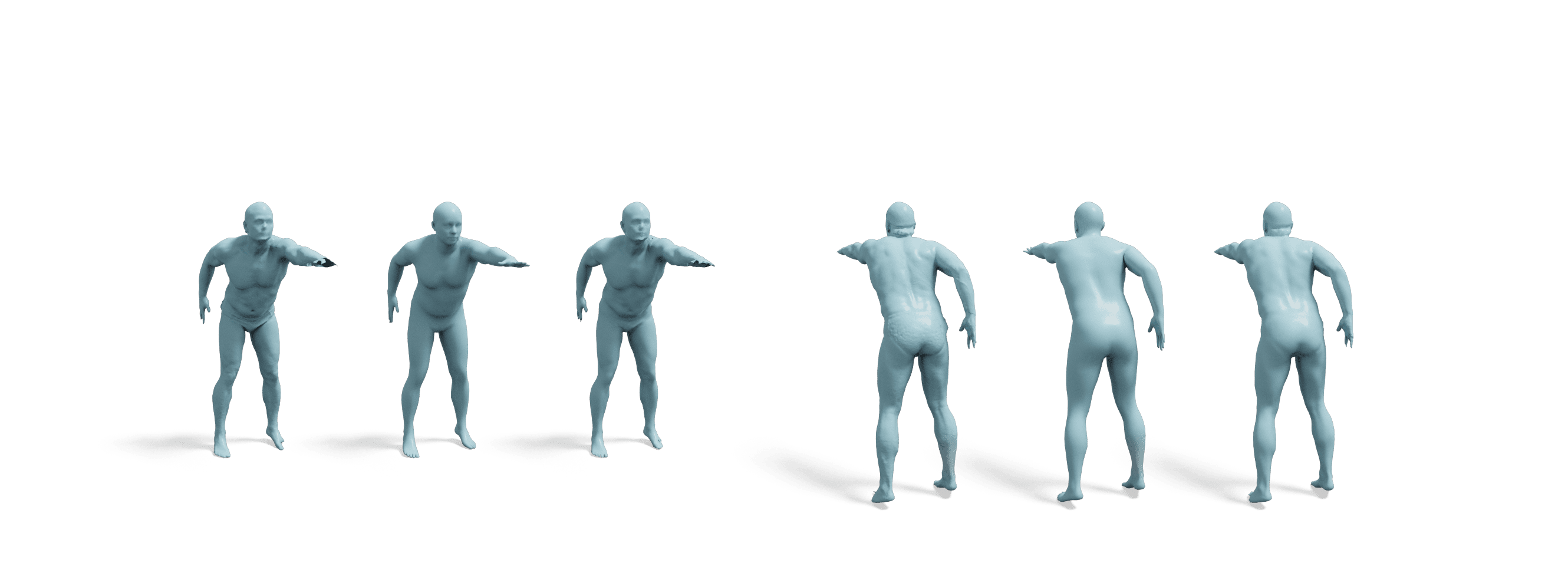}
		\put(12,99){}
	\end{overpic}
  \begin{overpic}[trim=6cm 4cm 12cm 12cm,clip, width=\linewidth]{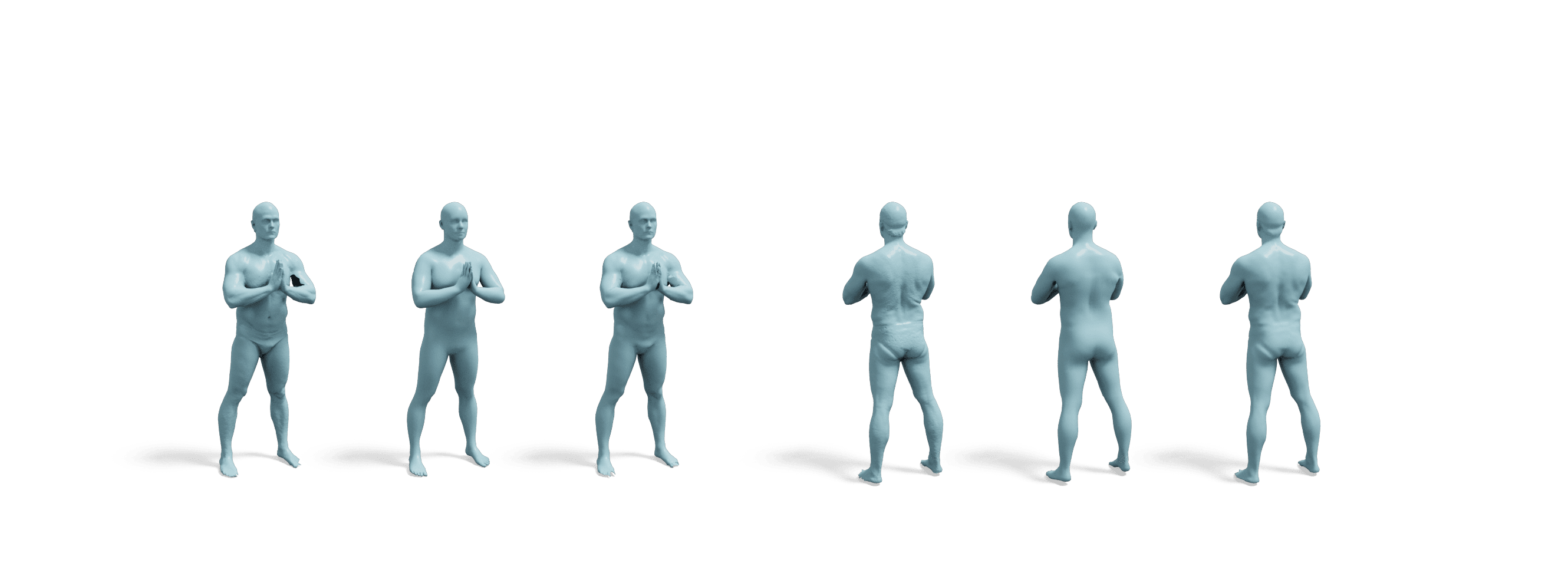}
		\put(12,99){}
	\end{overpic}
  \begin{overpic}[trim=6cm 4cm 12cm 12cm,clip, width=\linewidth]{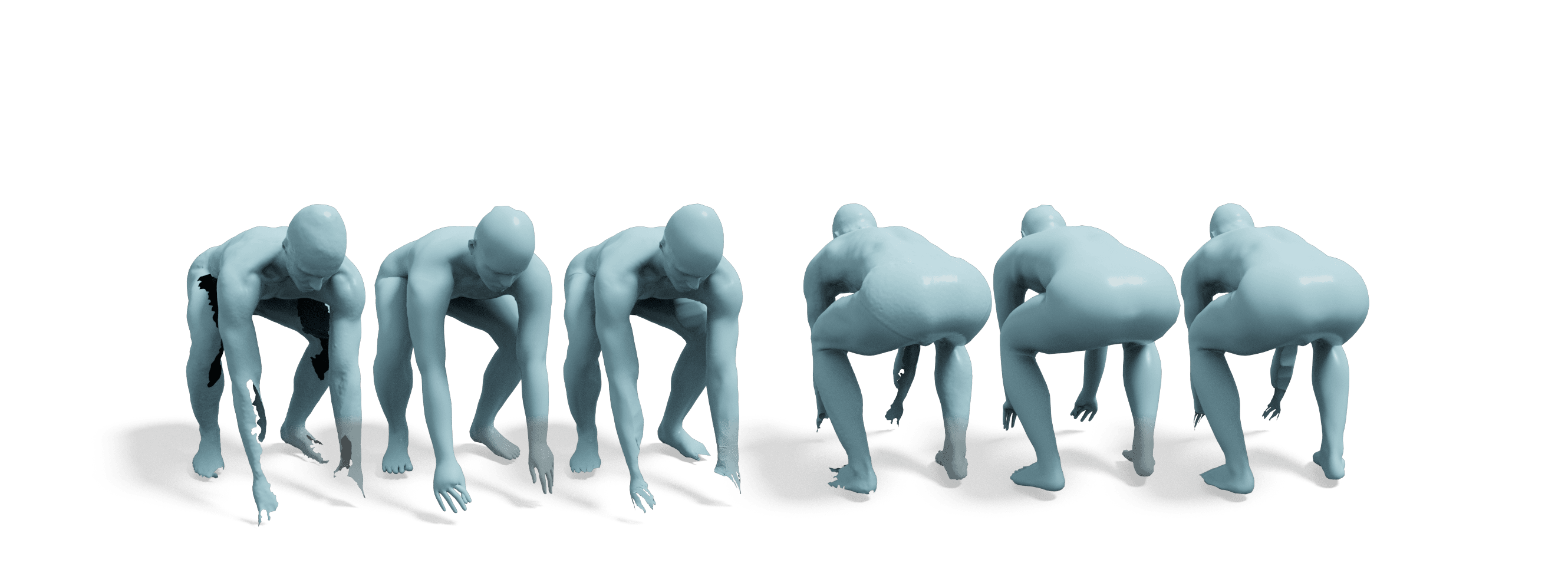}
		\put(12,99){}
	\end{overpic}
\end{figure*}

\begin{figure*}

    \centering
    \footnotesize
 \begin{overpic}[trim=0cm 4cm 0cm 10cm,clip, width=\linewidth]{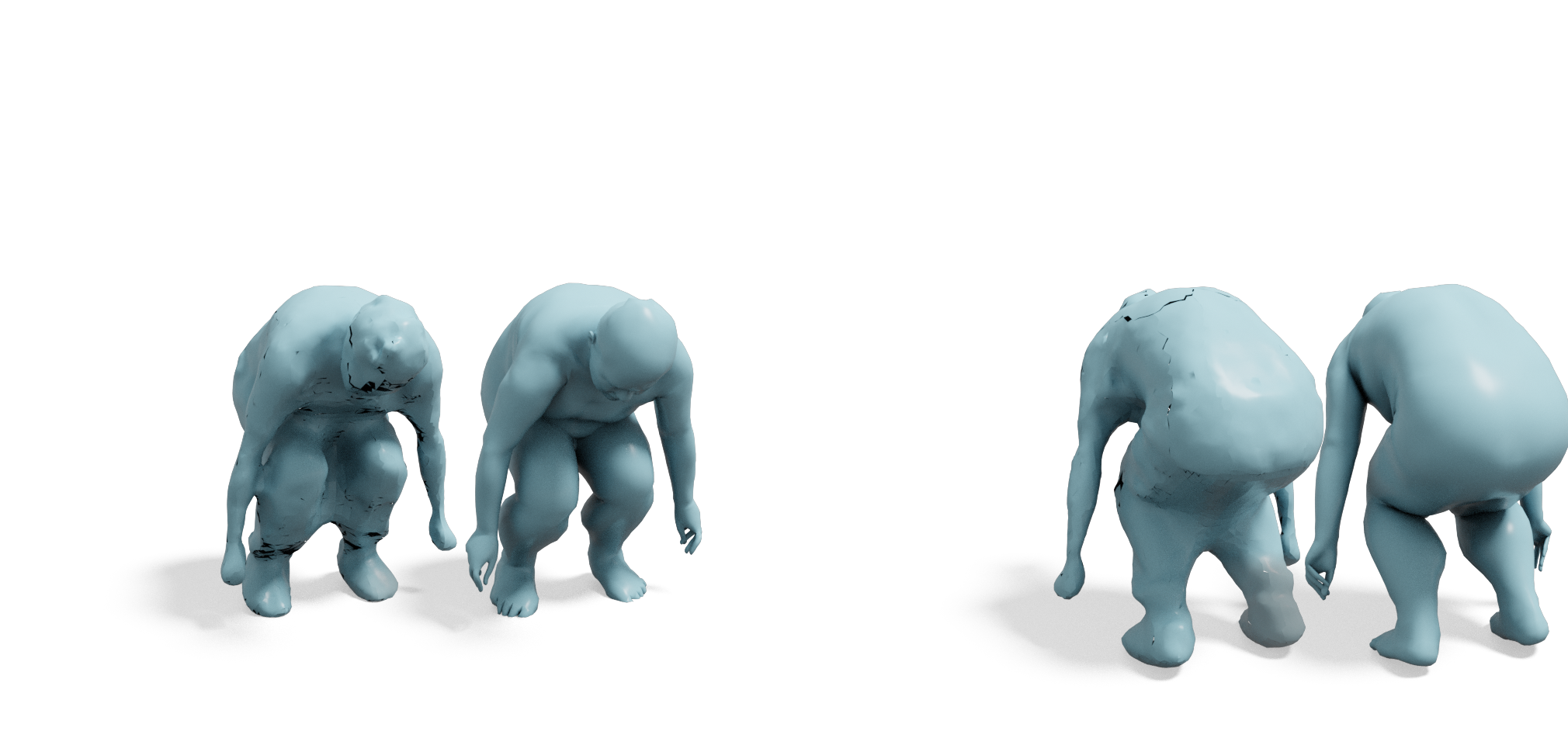}
  \put(43,28){\huge HuMMan}
		\put(12,25){Input}
        \put(31,25){\pipeline{}}

		\put(70,25){Input}
        \put(85,25){\pipeline{}}
	\end{overpic}
 \begin{overpic}[trim=0cm 4cm 0cm 10cm,clip, width=\linewidth]{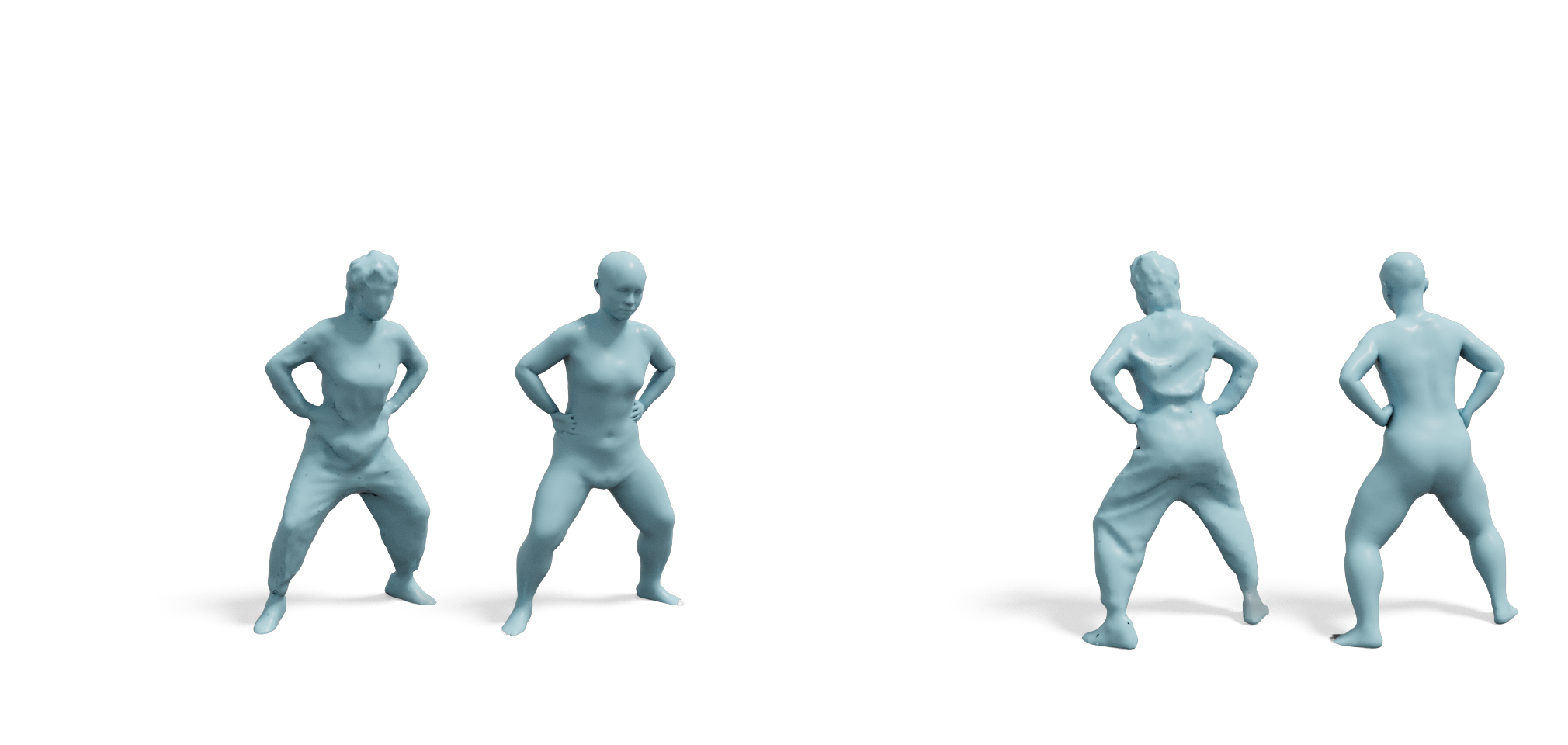}
		\put(12,99){}
	\end{overpic}
  \begin{overpic}[trim=0cm 4cm 0cm 10cm,clip, width=\linewidth]{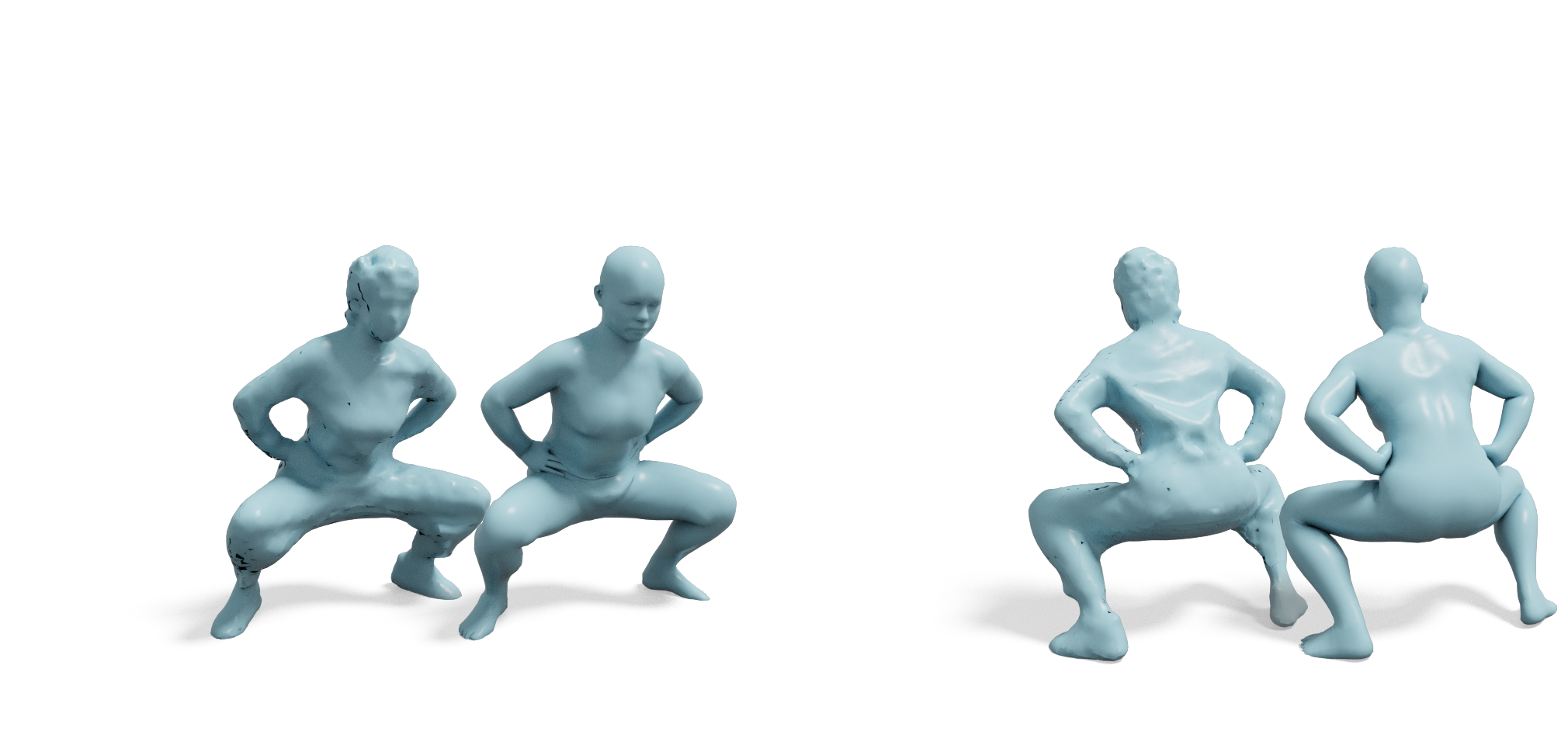}
		\put(12,99){}
	\end{overpic}
  \begin{overpic}[trim=0cm 4cm 0cm 10cm,clip, width=\linewidth]{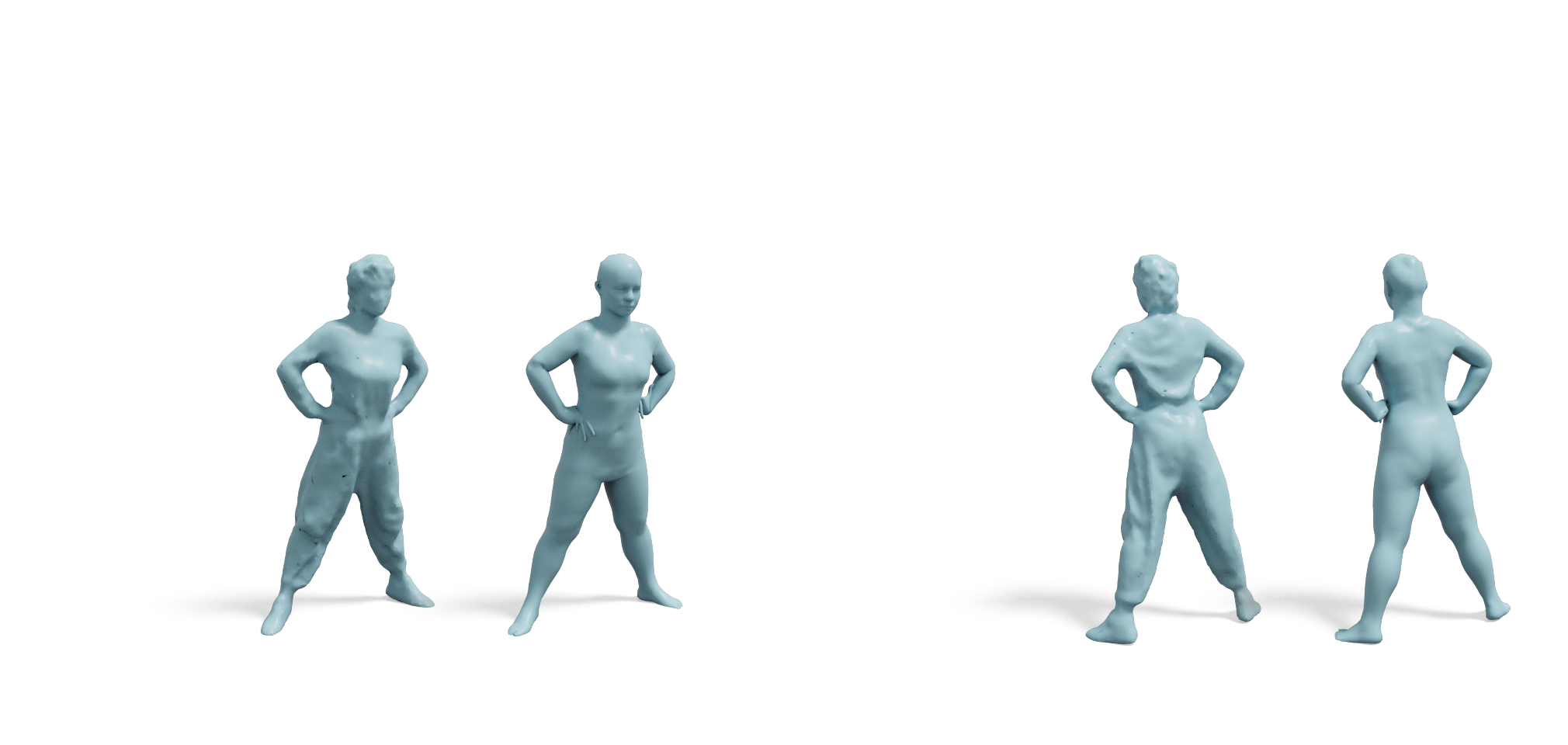}
		\put(12,99){}
	\end{overpic}
  \begin{overpic}[trim=0cm 4cm 0cm 10cm,clip, width=\linewidth]{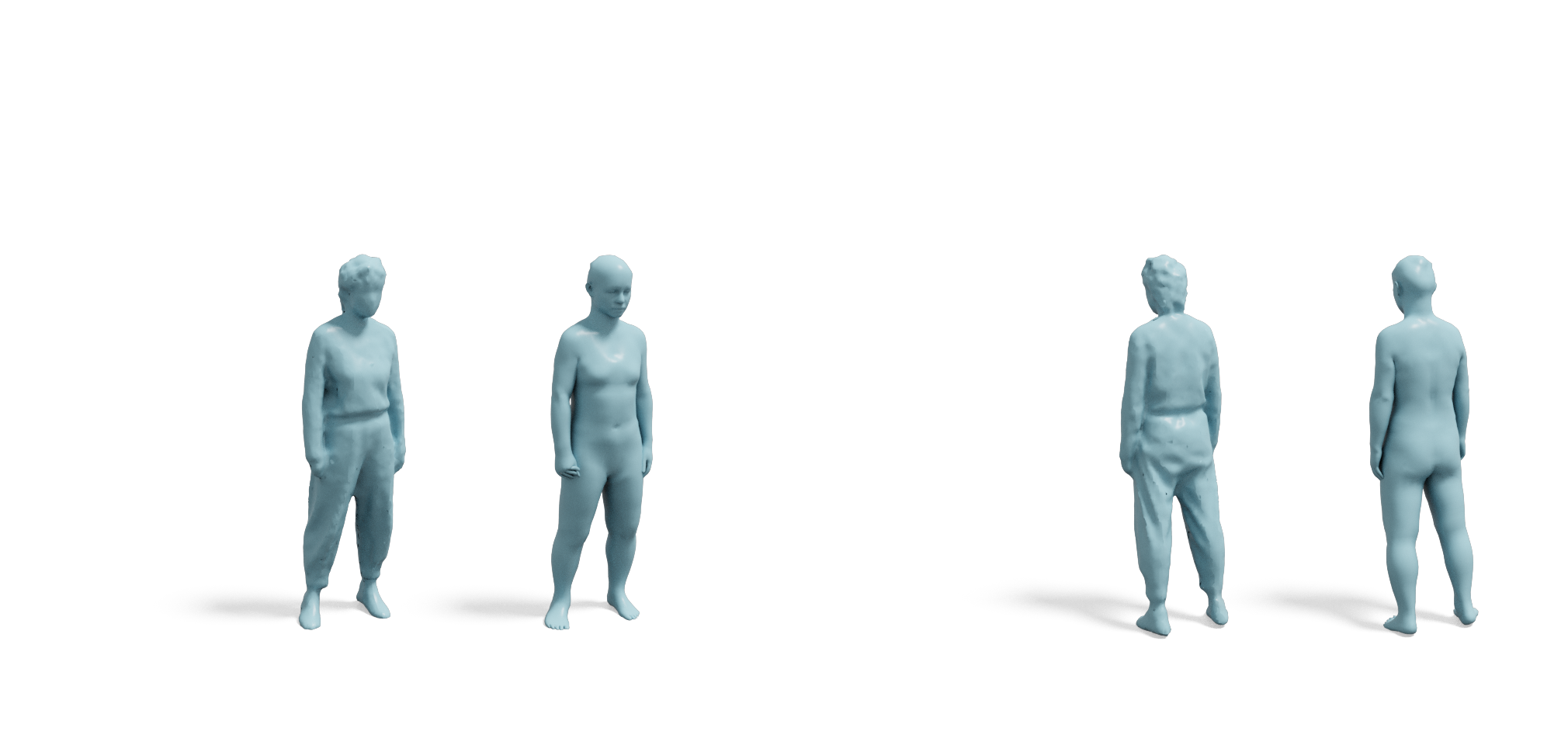}
		\put(12,99){}
	\end{overpic}
\end{figure*}

\begin{figure*}

    \centering
    \footnotesize
 \begin{overpic}[trim=0cm 4cm 0cm 10cm,clip, width=\linewidth]{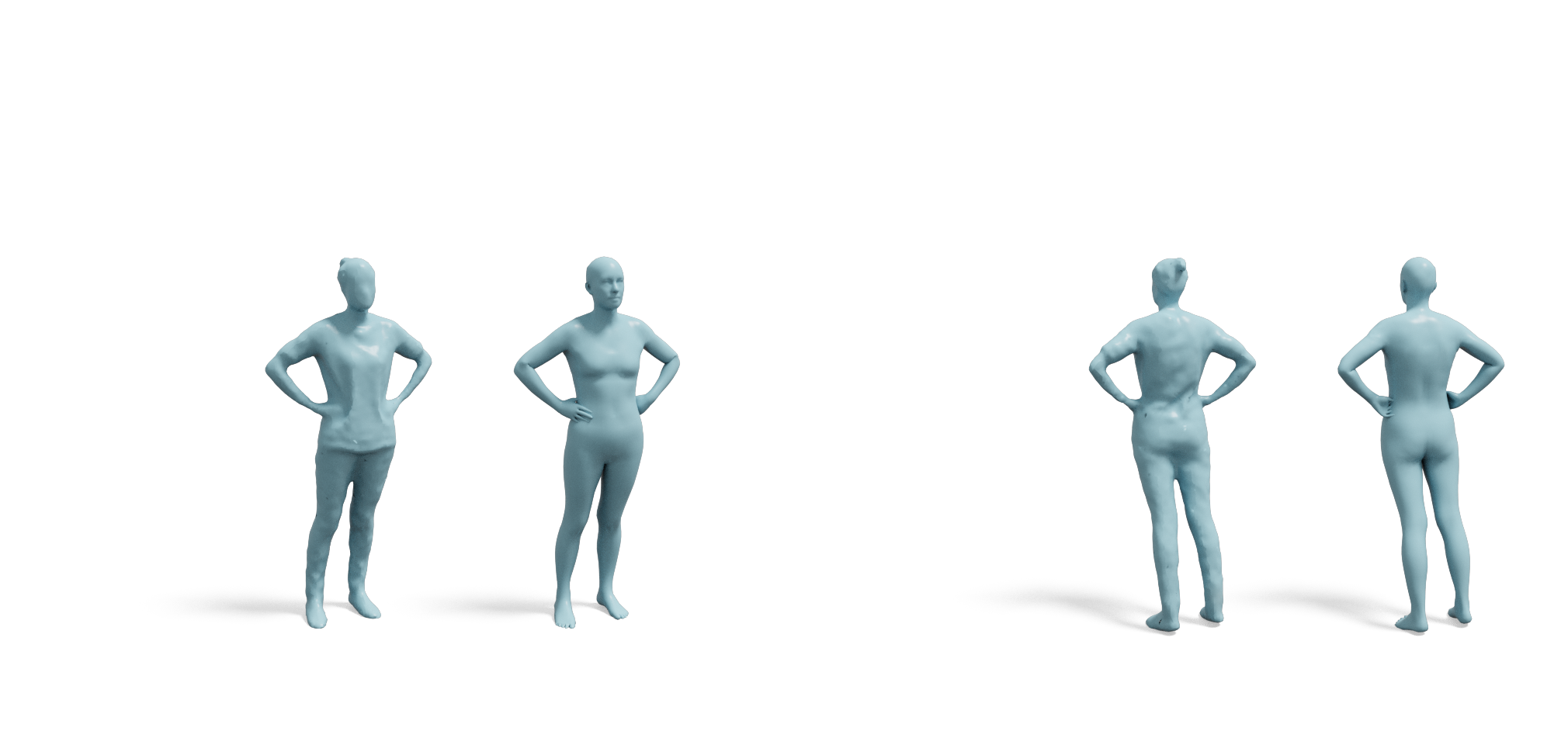}
   \put(43,30){\huge HuMMan}
		\put(12,27){Input}
        \put(31,27){\pipeline{}}

		\put(70,27){Input}
        \put(85,27){\pipeline{}}
	\end{overpic}
 \begin{overpic}[trim=0cm 4cm 0cm 10cm,clip, width=\linewidth]{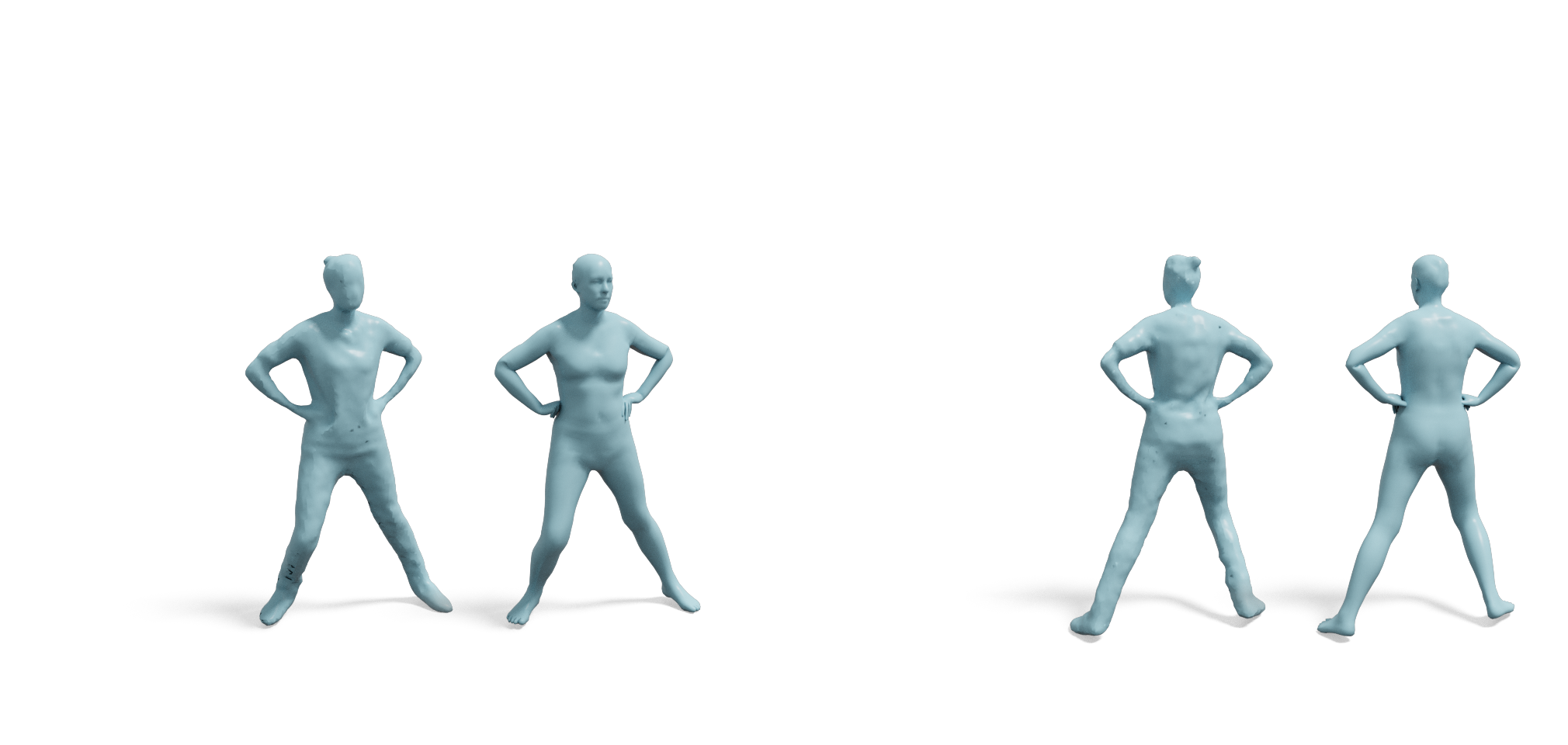}
		\put(12,99){}
	\end{overpic}
  \begin{overpic}[trim=0cm 4cm 0cm 10cm,clip, width=\linewidth]{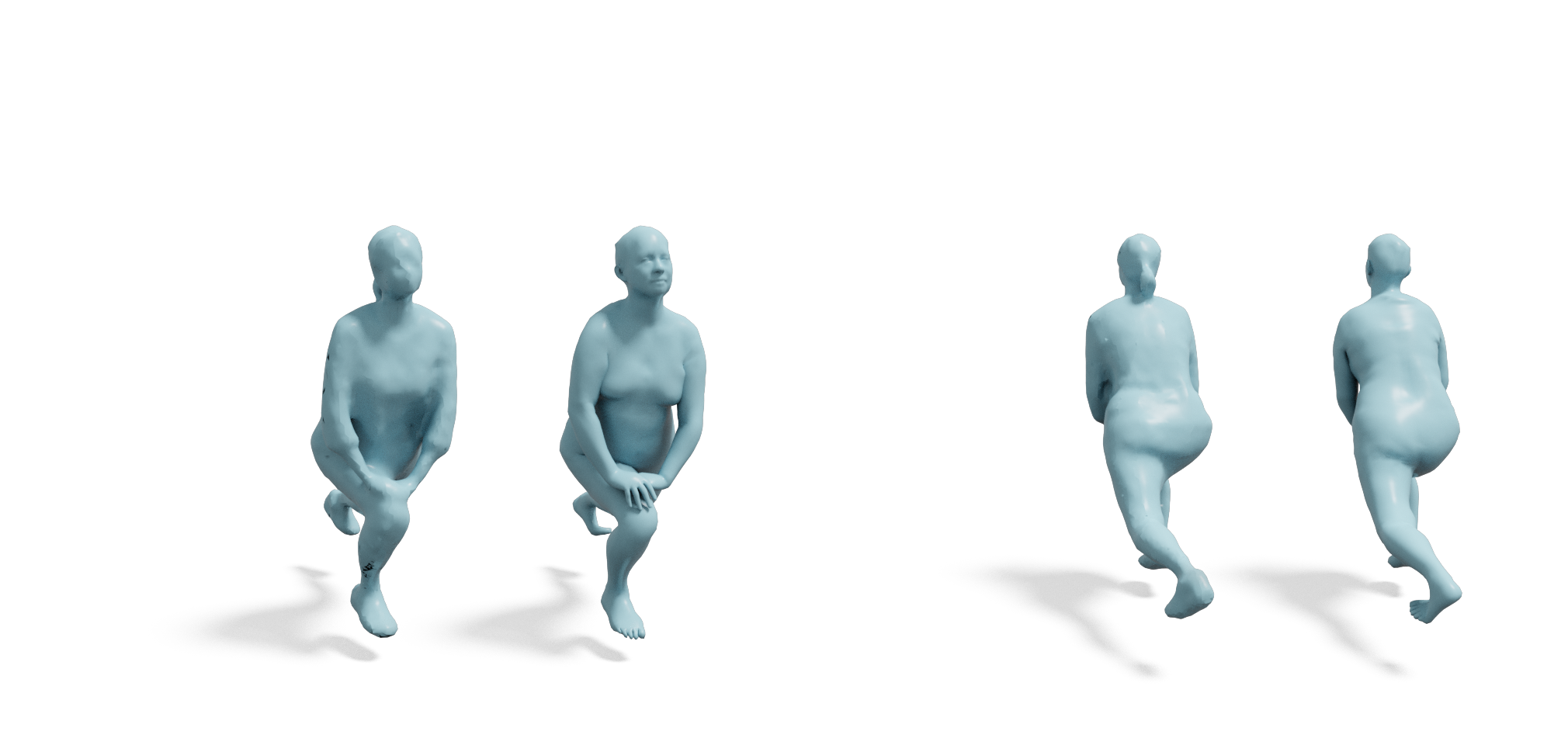}
		\put(12,99){}
	\end{overpic}
  \begin{overpic}[trim=0cm 4cm 0cm 10cm,clip, width=\linewidth]{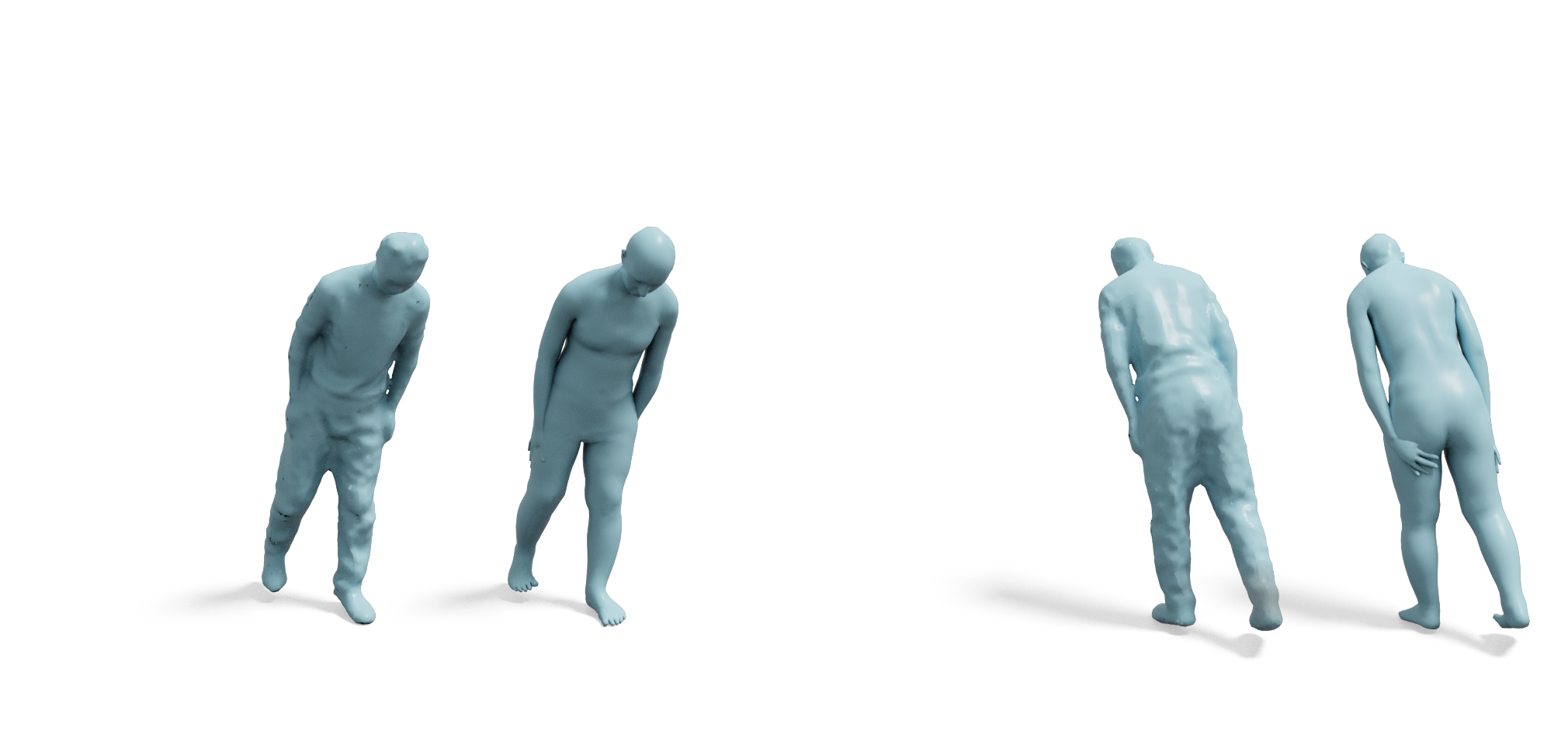}
		\put(12,99){}
	\end{overpic}
  \begin{overpic}[trim=0cm 4cm 0cm 10cm,clip, width=\linewidth]{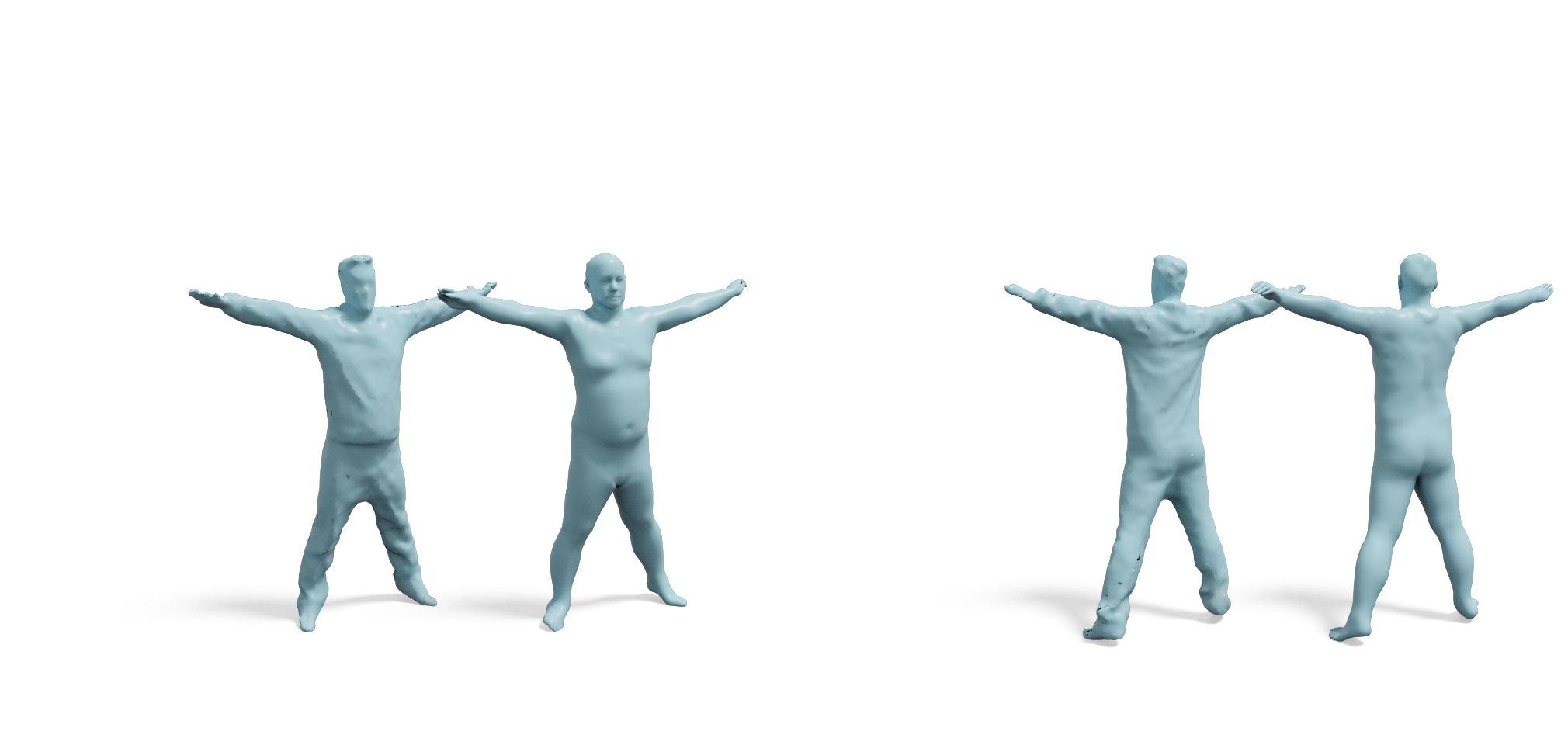}
		\put(12,99){}
	\end{overpic}
\end{figure*}

\begin{figure*}

    \centering
    \footnotesize
 \begin{overpic}[trim=0cm 4cm 0cm 10cm,clip, width=\linewidth]{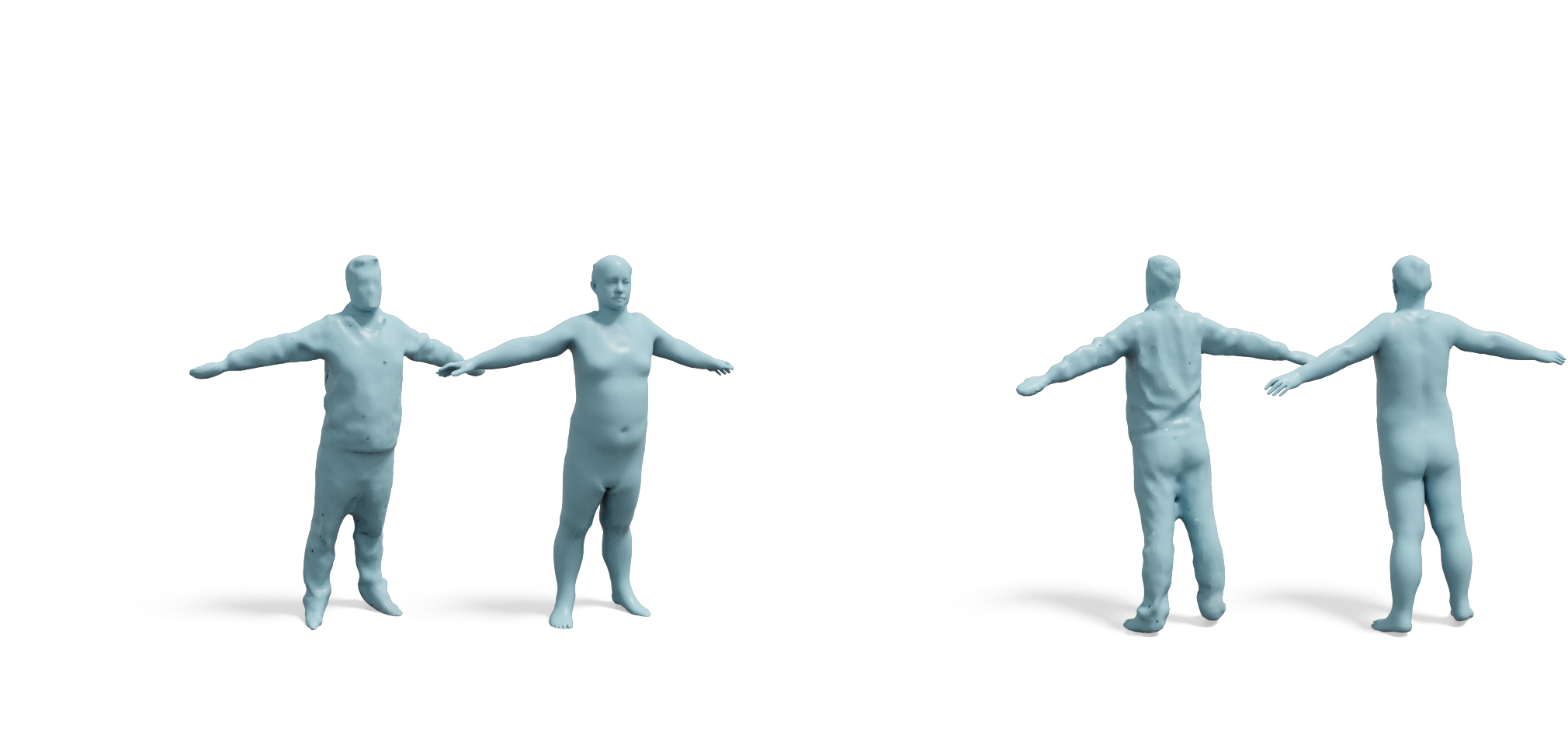}
   \put(43,30){\huge HuMMan}
		\put(12,27){Input}
        \put(31,27){\pipeline{}}

		\put(70,27){Input}
        \put(85,27){\pipeline{}}
	\end{overpic}
 \begin{overpic}[trim=0cm 4cm 0cm 10cm,clip, width=\linewidth]{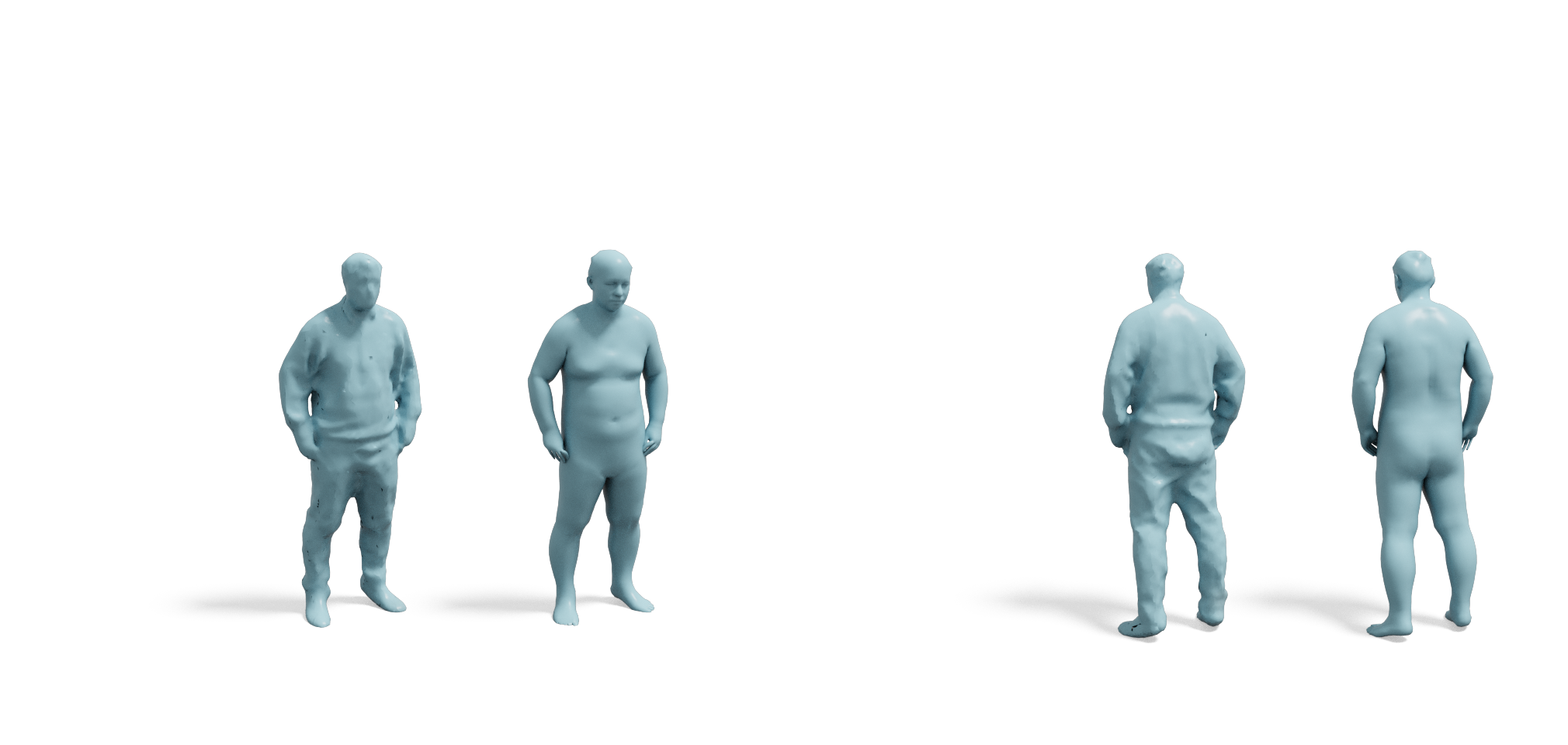}
		\put(12,99){}
	\end{overpic}
  \begin{overpic}[trim=0cm 4cm 0cm 10cm,clip, width=\linewidth]{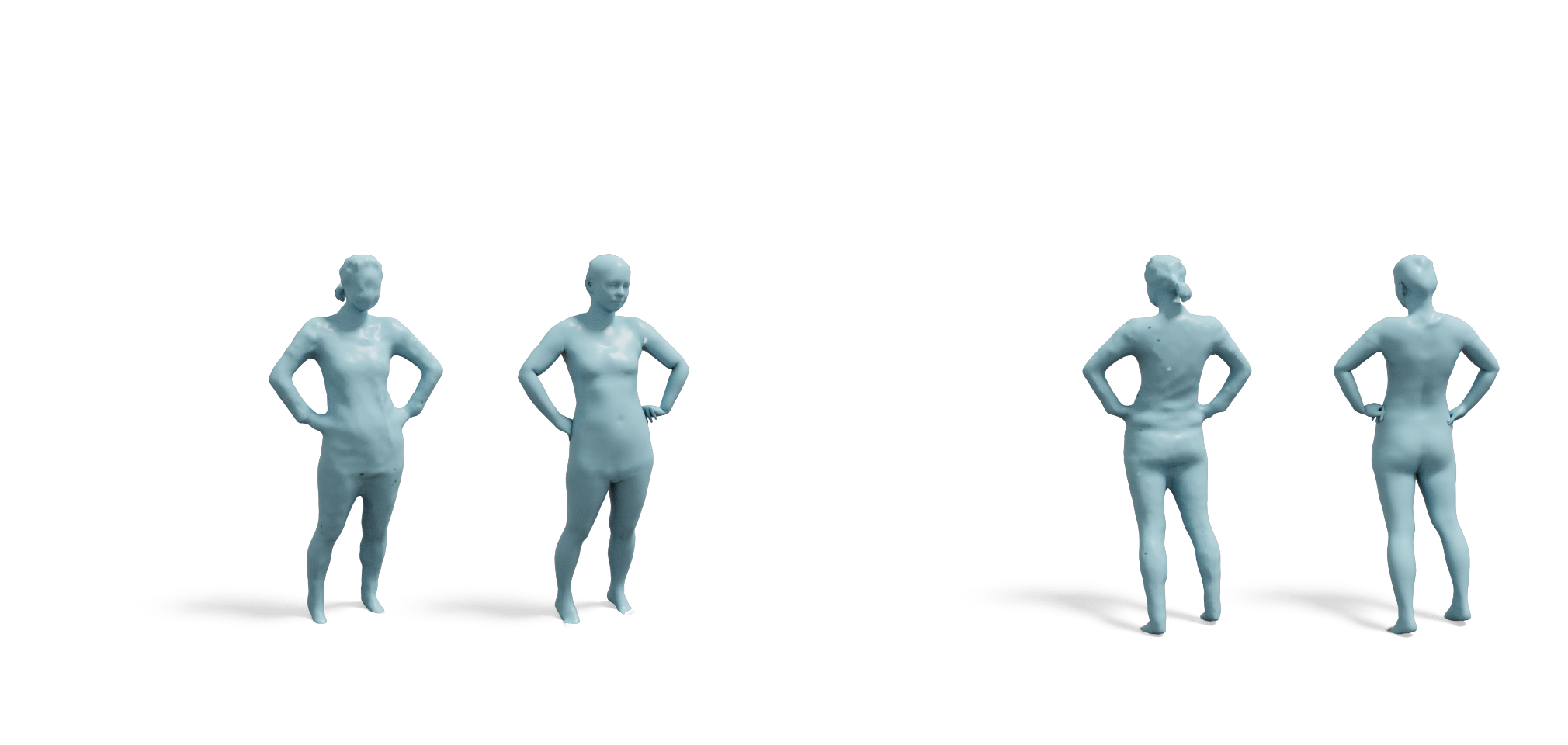}
		\put(12,99){}
	\end{overpic}
  \begin{overpic}[trim=0cm 4cm 0cm 10cm,clip, width=\linewidth]{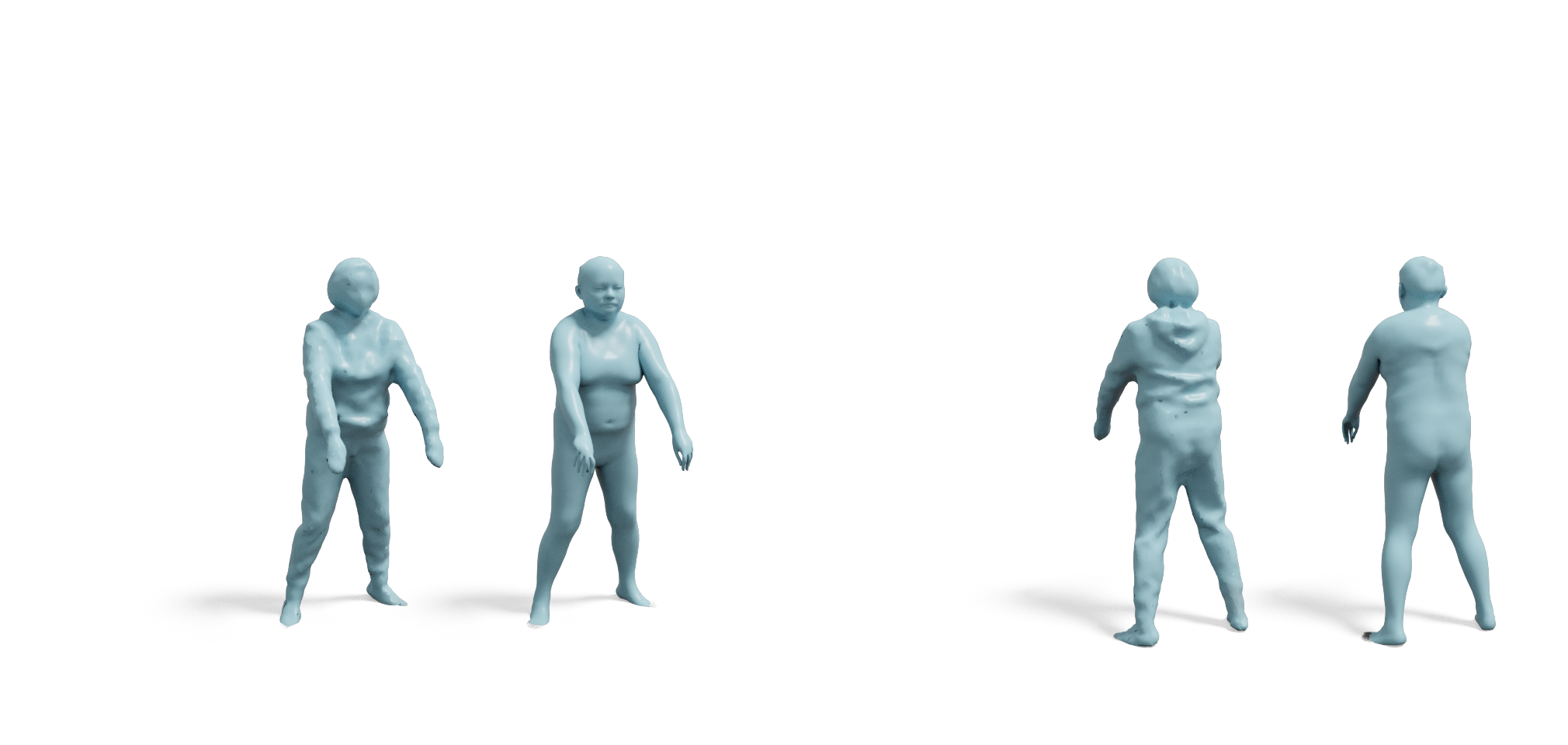}
		\put(12,99){}
	\end{overpic}
  \begin{overpic}[trim=0cm 4cm 0cm 10cm,clip, width=\linewidth]{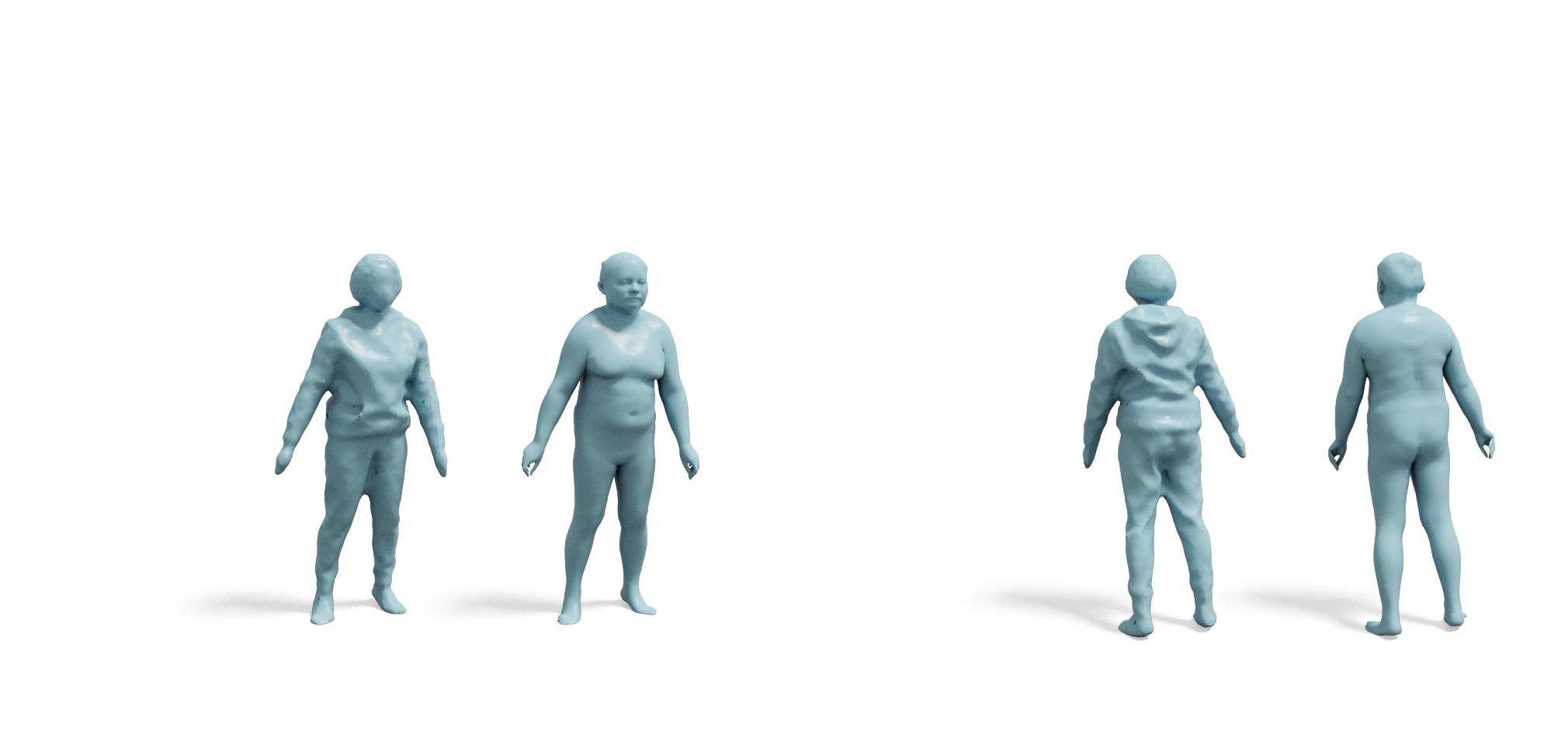}
		\put(12,99){}
	\end{overpic}
\end{figure*}


\newpage

\begin{figure*}

    \centering
    \footnotesize
 \begin{overpic}[trim=0cm 0cm 0cm 0cm,clip, width=\linewidth]{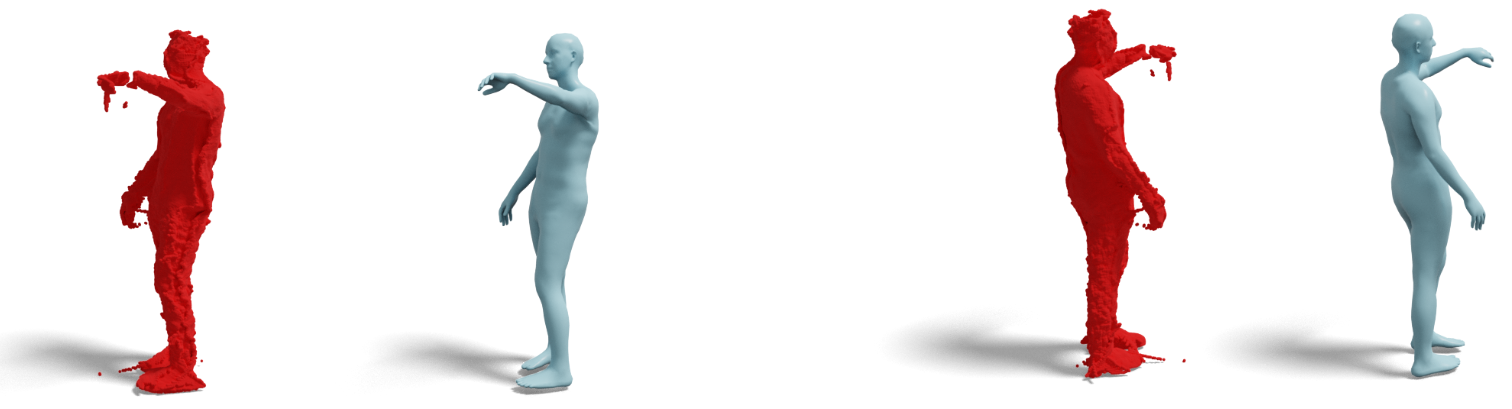}
   \put(43,32){\huge BEHAVE}
		\put(12,30){Input}
        \put(31,30){\pipeline{}}

		\put(70,30){Input}
        \put(85,30){\pipeline{}}
	\end{overpic}
 \begin{overpic}[trim=0cm 0cm 0cm 0cm,clip, width=\linewidth]{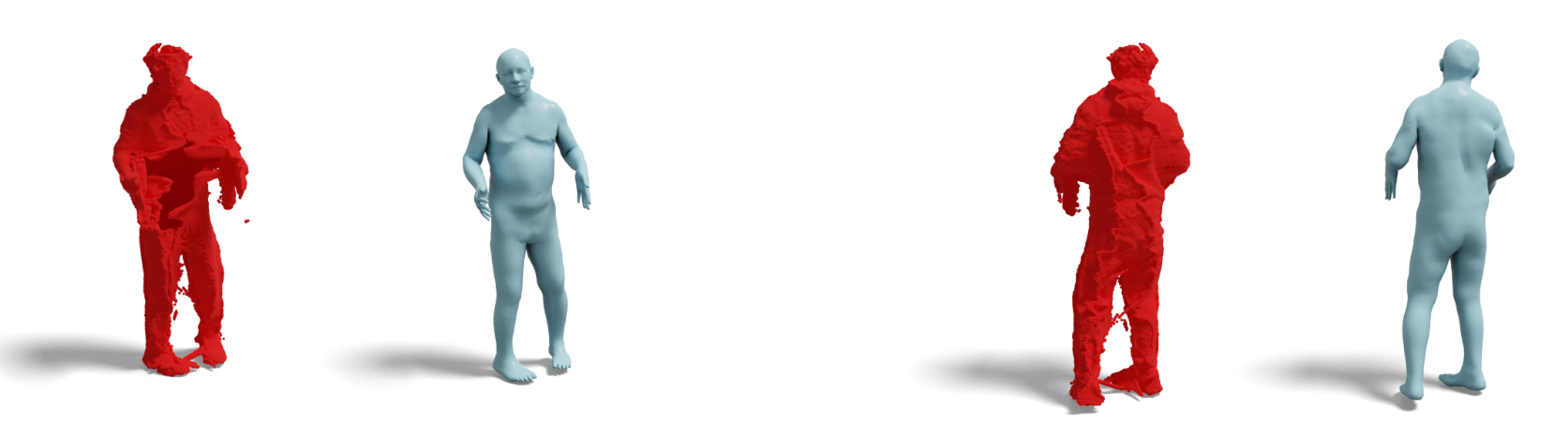}
		\put(12,99){}
	\end{overpic}
  \begin{overpic}[trim=0cm 0cm 0cm 0cm,clip, width=\linewidth]{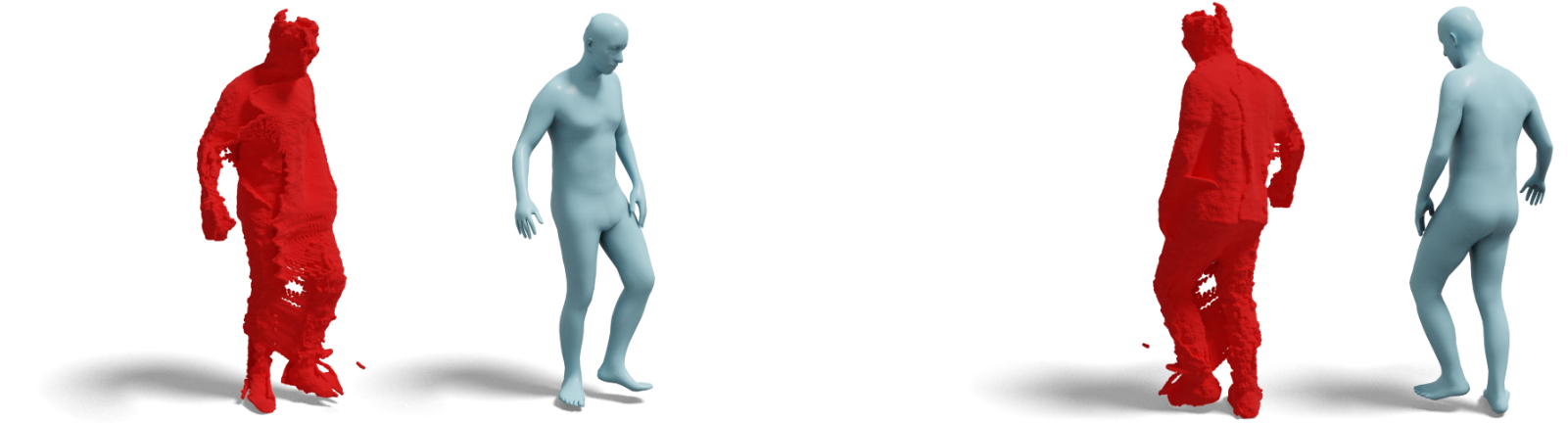}
		\put(12,99){}
	\end{overpic}
  \begin{overpic}[trim=0cm 0cm 0cm 0cm,clip, width=\linewidth]{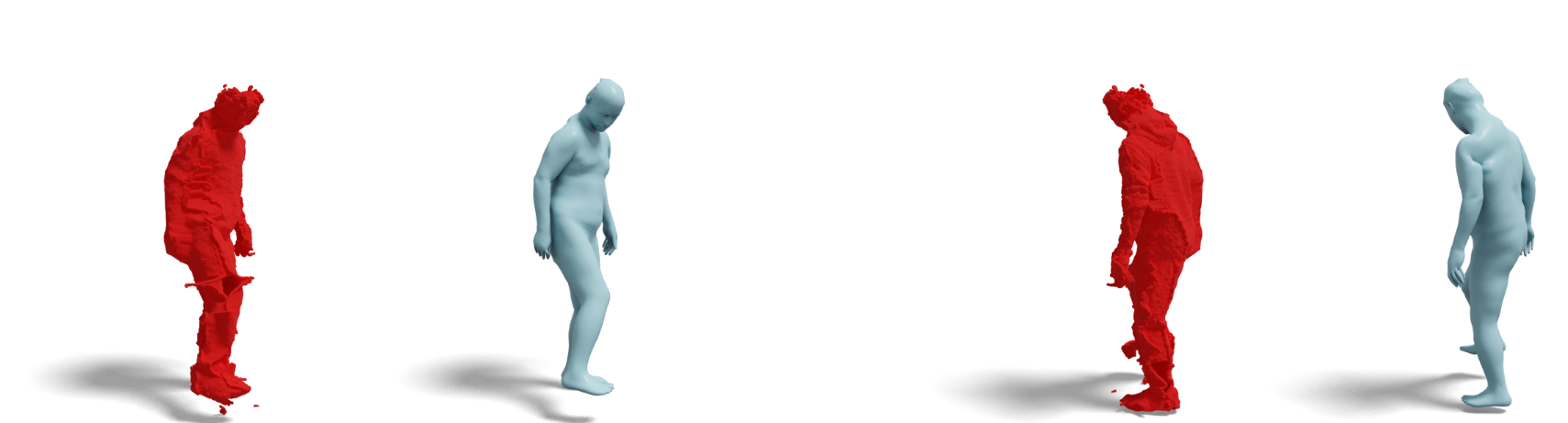}
		\put(12,99){}
	\end{overpic}
  \begin{overpic}[trim=0cm 0cm 0cm 0cm,clip, width=\linewidth]{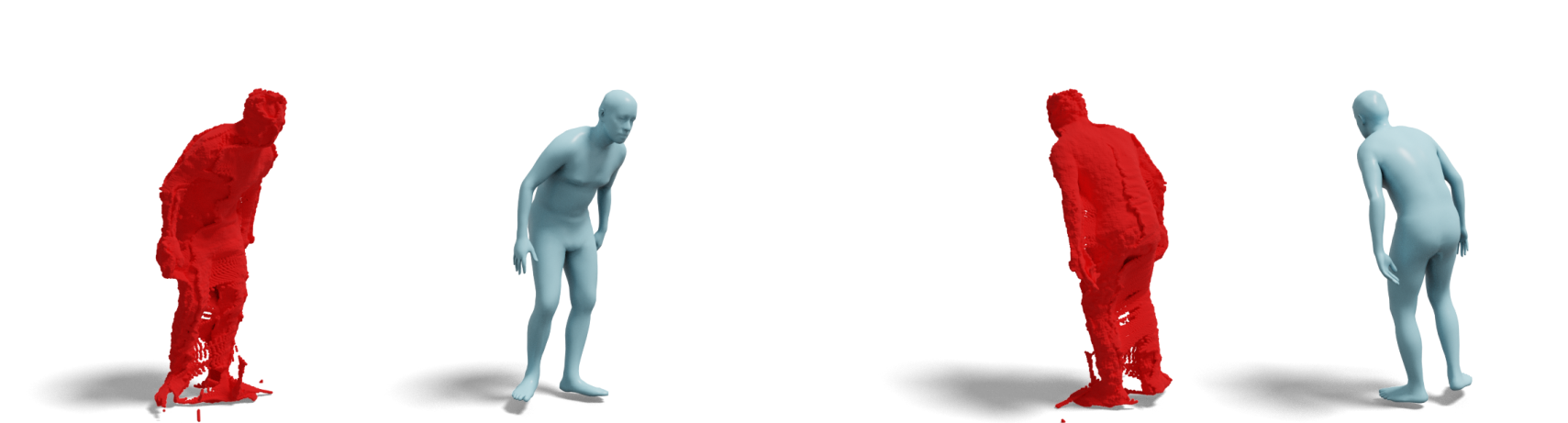}
		\put(12,99){}
	\end{overpic}
\end{figure*}

\newpage

\begin{figure*}

    \centering
    \footnotesize
 \begin{overpic}[trim=0cm 0cm 0cm 0cm,clip, width=\linewidth]{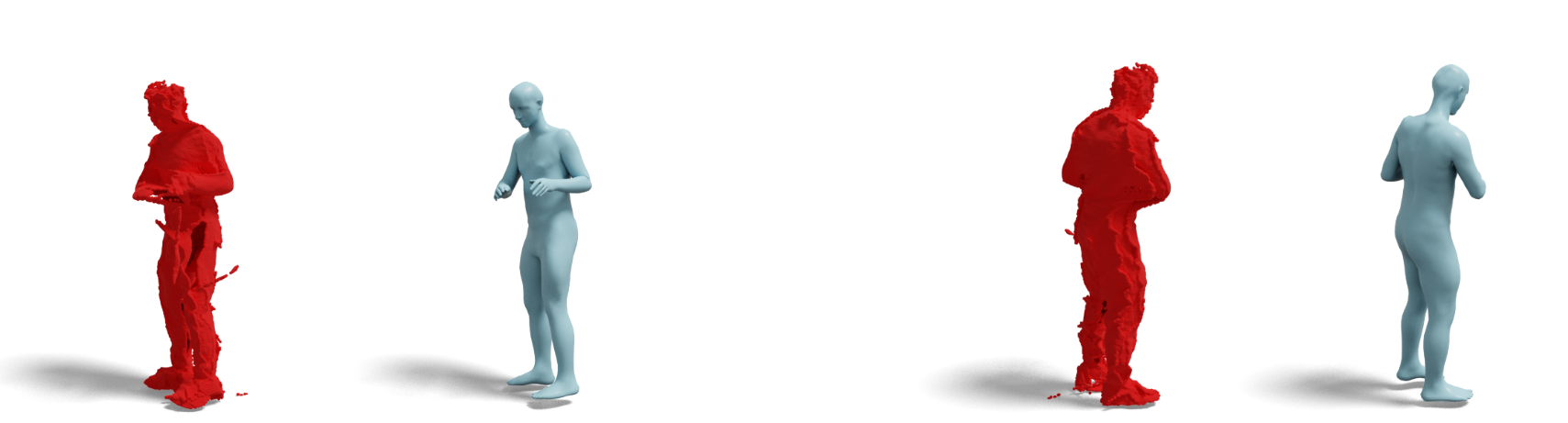}
    \put(43,30){\huge BEHAVE}
		\put(12,25){Input}
        \put(31,25){\pipeline{}}

		\put(70,25){Input}
        \put(85,25){\pipeline{}}
	\end{overpic}
 \begin{overpic}[trim=0cm 0cm 0cm 0cm,clip, width=\linewidth]{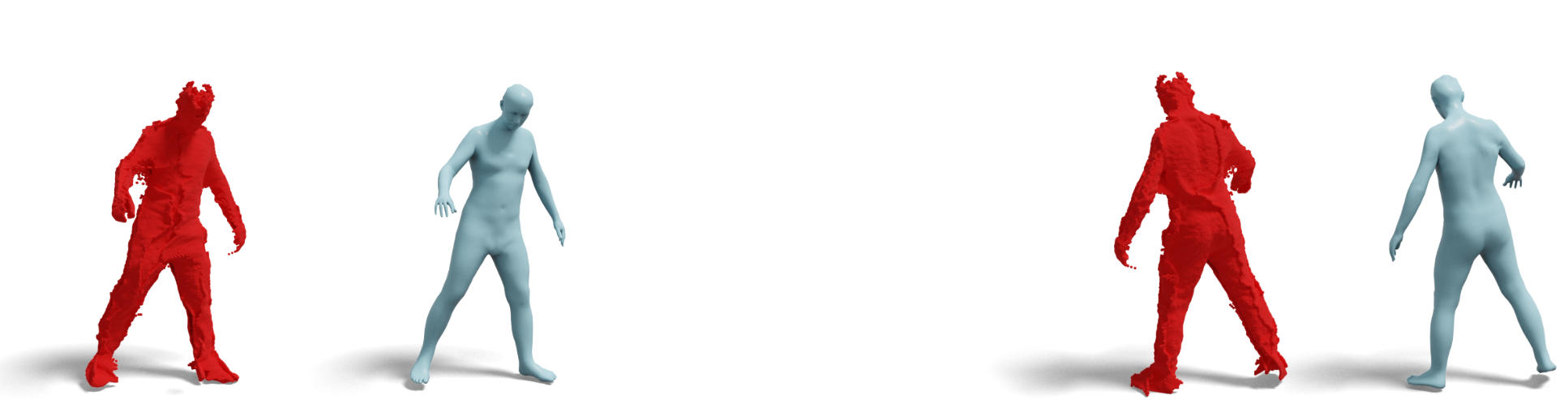}
		\put(12,99){}
	\end{overpic}
  \begin{overpic}[trim=0cm 0cm 0cm 0cm,clip, width=\linewidth]{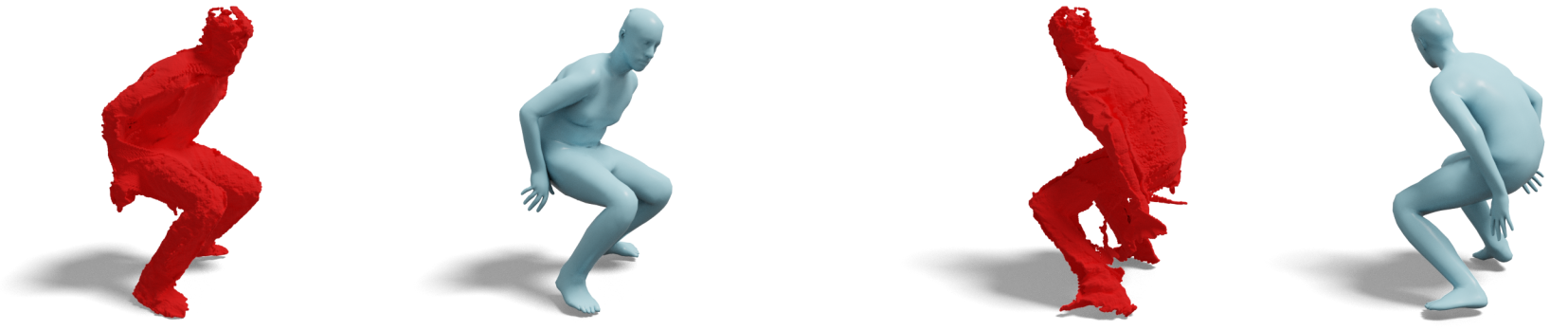}
		\put(12,99){}
	\end{overpic}
  \begin{overpic}[trim=0cm 0cm 0cm 0cm,clip, width=\linewidth]{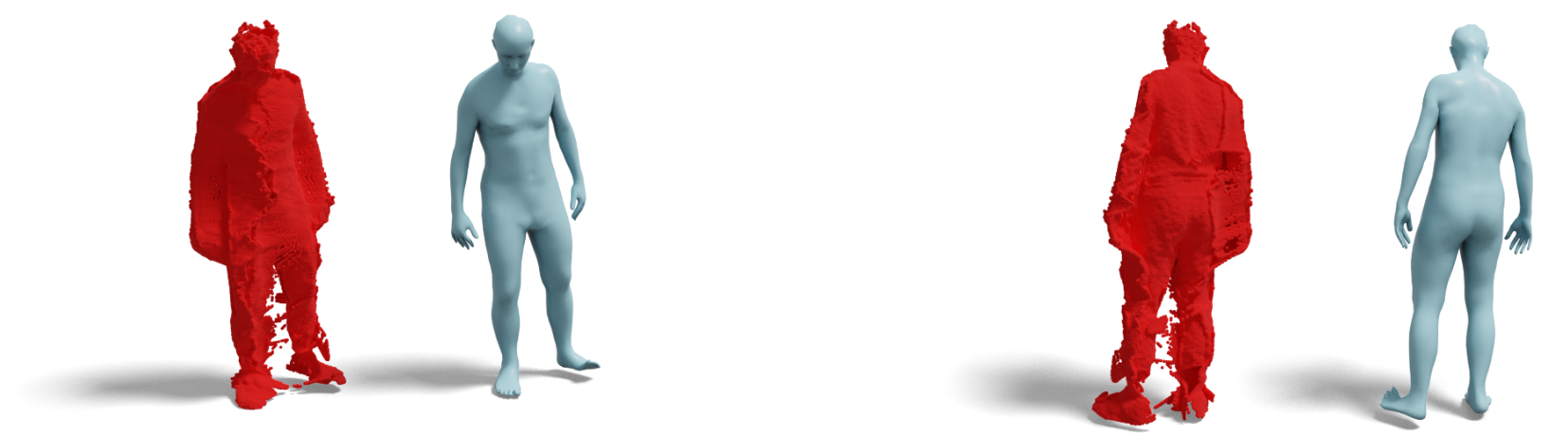}
		\put(12,99){}
	\end{overpic}
  \begin{overpic}[trim=0cm 0cm 0cm 0cm,clip, width=\linewidth]{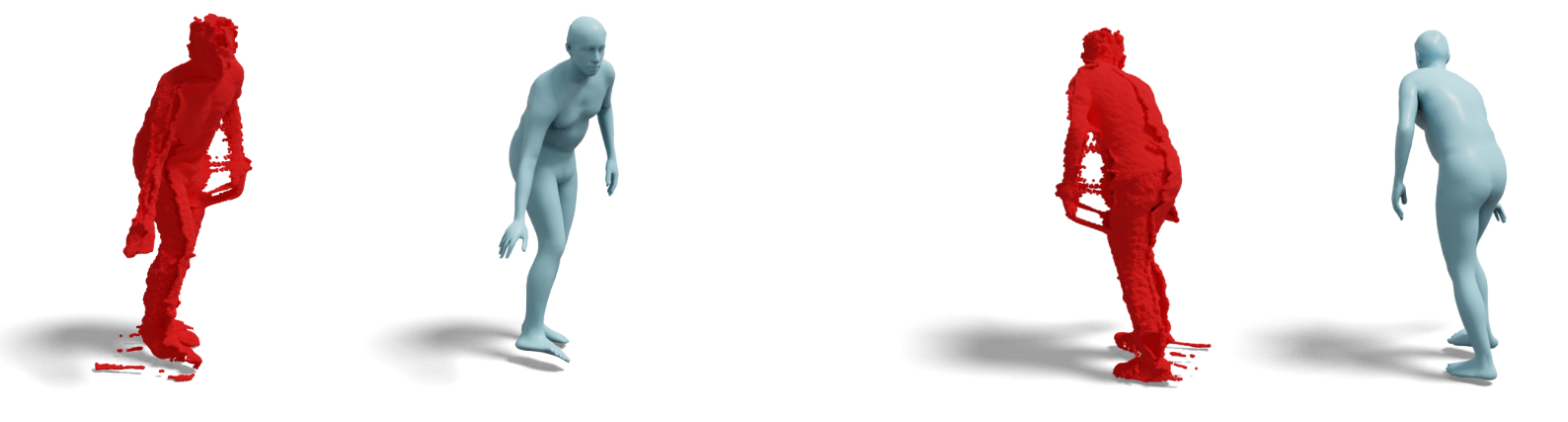}
		\put(12,99){}
	\end{overpic}
\end{figure*}

\clearpage
\newpage

\begin{figure*}

    \centering
    \footnotesize
 \begin{overpic}[trim=0cm 0cm 0cm 0cm,clip, width=\linewidth]{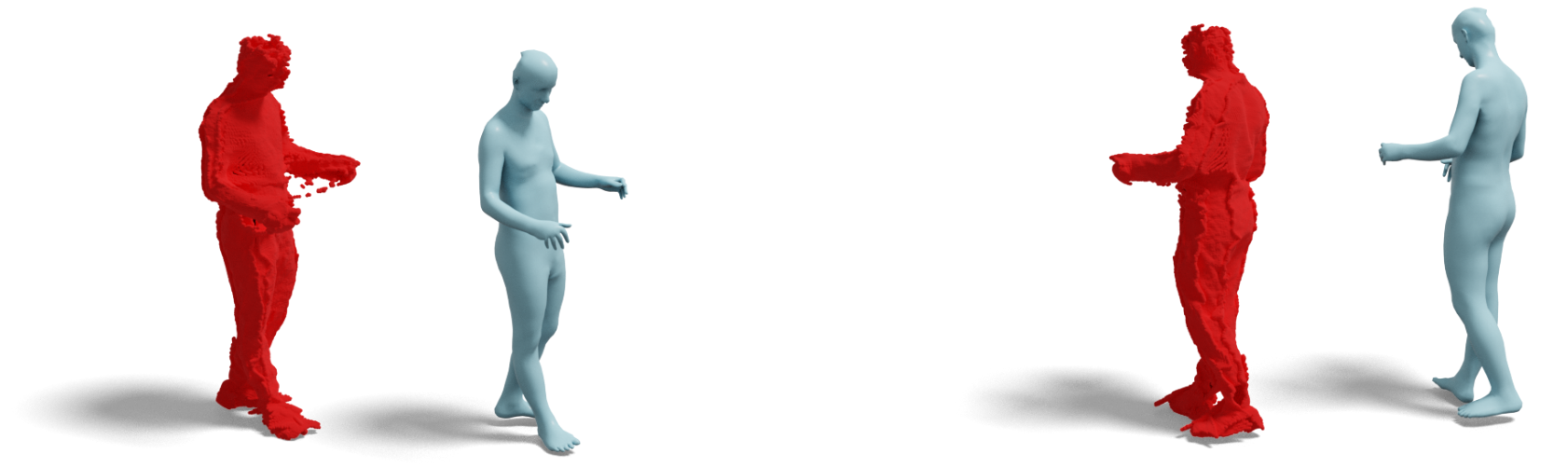}
    \put(43,32){\huge BEHAVE}
		\put(12,30){Input}
        \put(31,30){\pipeline{}}

		\put(70,30){Input}
        \put(87,30){\pipeline{}}
	\end{overpic}
 \begin{overpic}[trim=0cm 0cm 0cm 0cm,clip, width=\linewidth]{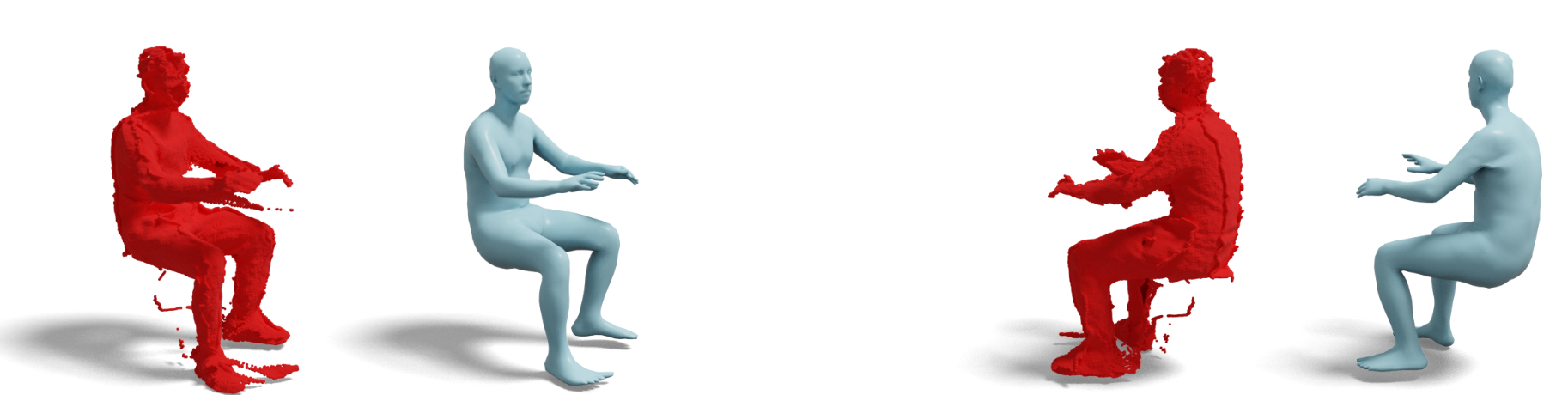}
		\put(12,99){}
	\end{overpic}
  \begin{overpic}[trim=0cm 0cm 0cm 0cm,clip, width=\linewidth]{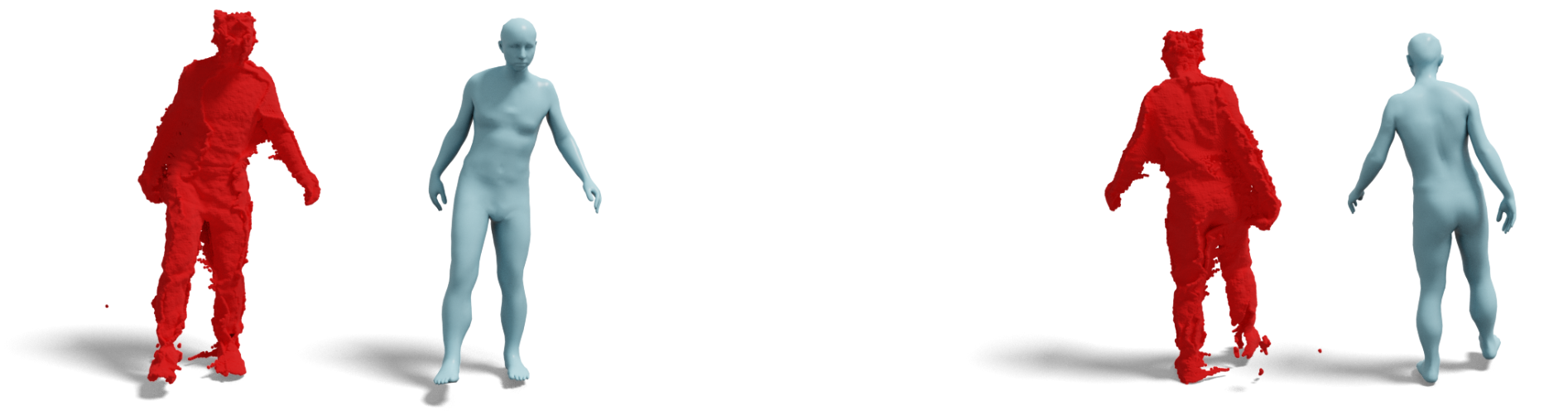}
		\put(12,99){}
	\end{overpic}
  \begin{overpic}[trim=0cm 0cm 0cm 0cm,clip, width=\linewidth]{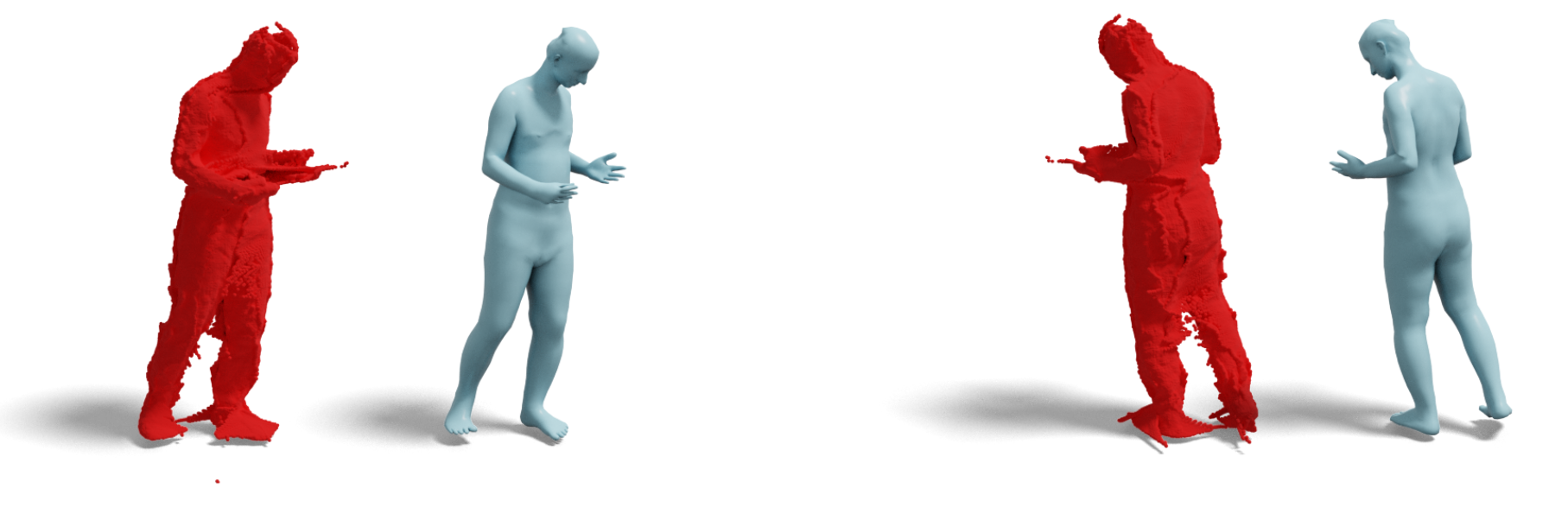}
		\put(12,99){}
	\end{overpic}
  \begin{overpic}[trim=0cm 0cm 0cm 0cm,clip, width=\linewidth]{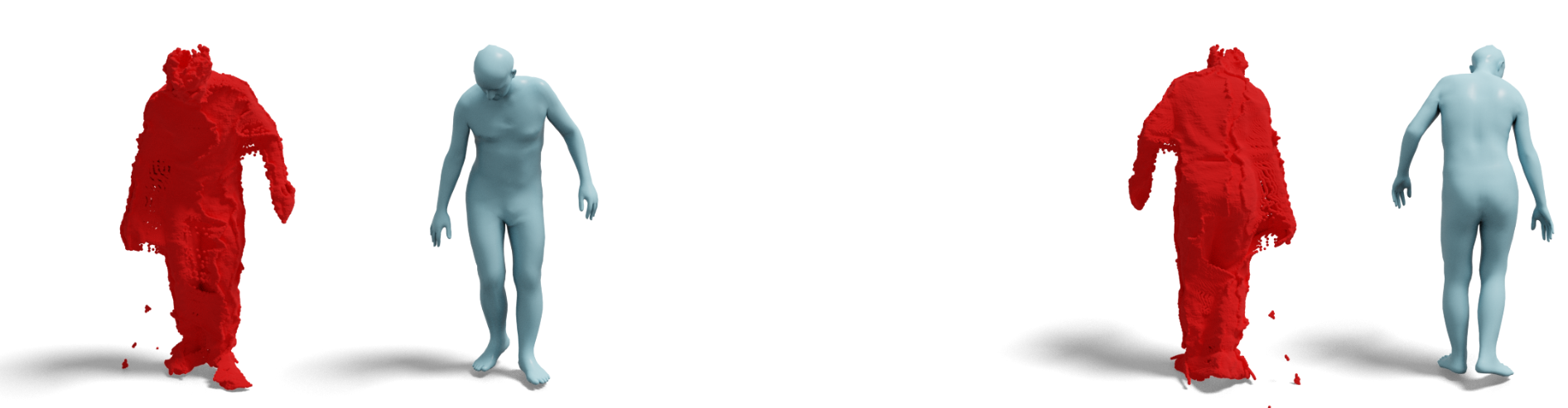}
		\put(12,99){}
	\end{overpic}
\end{figure*}

\clearpage
\newpage






\end{document}